\documentclass{article}
\usepackage[margin=1in]{geometry}
\usepackage[utf8]{inputenc} 
\usepackage[T1]{fontenc}    
\usepackage[colorlinks=true,allcolors=magenta]{hyperref}       
\usepackage{url}            
\usepackage{booktabs}       
\usepackage{amsfonts}       
\usepackage{nicefrac}       
\usepackage{microtype}      
\usepackage[numbers,compress]{natbib}

\usepackage[affil-sl]{authblk}
\title{Stochastic Neighbor Embedding under $f$-divergences}
\author[1,2]{Daniel Jiwoong Im}
\author[1,3]{Nakul Verma}
\author[1]{Kristin Branson}
\affil[1]{Janelia Research Campus, HHMI, Virginia}
\affil[2]{AIFounded Inc., Toronto}
\affil[3]{Columbia University, New York}

\usepackage{amsmath,amsbsy,amsfonts,amssymb,amsthm}
\usepackage{xcolor}
\usepackage{algorithm}
\usepackage[noend]{algpseudocode}
\usepackage{cleveref}
\usepackage{xspace}
\usepackage[shortlabels]{enumitem}

\usepackage[title]{appendix}
\usepackage{subcaption}
\usepackage{graphicx}  
\usepackage{multicol}
\usepackage{multirow}

\def\ddefloop#1{\ifx\ddefloop#1\else\ddef{#1}\expandafter\ddefloop\fi}
\def\ddef#1{\expandafter\def\csname bb#1\endcsname{\ensuremath{\mathbb{#1}}}}
\ddefloop ABCDEFGHIJKLMNOPQRSTUVWXYZ\ddefloop
\def\ddef#1{\expandafter\def\csname c#1\endcsname{\ensuremath{\mathcal{#1}}}}
\ddefloop ABCDEFGHIJKLMNOPQRSTUVWXYZ\ddefloop
\def\ddef#1{\expandafter\def\csname v#1\endcsname{\ensuremath{\boldsymbol{#1}}}}
\ddefloop ABCDEFGHIJKLMNOPQRSTUVWXYZabcdefghijklmnopqrstuvwxyz\ddefloop
\def\ddef#1{\expandafter\def\csname v#1\endcsname{\ensuremath{\boldsymbol{\csname #1\endcsname}}}}
\ddefloop {alpha}{beta}{gamma}{delta}{epsilon}{varepsilon}{veps}{zeta}{eta}{theta}{vartheta}{iota}{kappa}{lambda}{mu}{nu}{xi}{pi}{varpi}{rho}{varrho}{sigma}{varsigma}{tau}{upsilon}{phi}{varphi}{chi}{psi}{omega}{Gamma}{Delta}{Theta}{Lambda}{Xi}{Pi}{Sigma}{varSigma}{Sig}{Upsilon}{Phi}{Psi}{Omega}{ell}\ddefloop

\newtheorem{proposition}{Proposition}

\crefname{proposition}{proposition}{propositions}
\theoremstyle{definition}

\graphicspath{ {./figs/} }

\DeclareMathOperator{\KL}{{\textrm{KL}}}
\DeclareMathOperator{\RKL}{{\textrm{RKL}}}
\DeclareMathOperator{\JS}{{\textrm{JS}}}
\DeclareMathOperator{\HL}{{\textrm{HL}}}
\DeclareMathOperator{\CH}{{\textrm{CH}}}
\DeclareMathOperator{\FP}{{\textrm{FP}}}
\DeclareMathOperator{\FN}{{\textrm{FN}}}
\DeclareMathOperator{\TP}{{\textrm{TP}}}

\begin{document}

\maketitle

\begin{abstract}%
  The $t$-distributed Stochastic Neighbor Embedding ($t$-SNE) is a powerful and
popular method for visualizing high-dimensional data. It minimizes the
Kullback-Leibler (KL) divergence between the original and embedded data
distributions. In this work, we propose extending this method to other
$f$-divergences. We analytically and empirically evaluate the types of latent
structure---manifold, cluster, and hierarchical---that are well-captured
using both the original KL-divergence as well as the proposed $f$-divergence generalization, 
and find that different divergences perform better for different types of
structure.

A common concern with $t$-SNE criterion is that it is optimized using gradient descent, and can
become stuck in poor local minima. We propose optimizing the $f$-divergence based loss 
criteria by minimizing a variational bound. This typically
performs better than optimizing the primal form, and our experiments show that
it can improve upon the embedding results obtained from the original $t$-SNE
criterion as well. 

\end{abstract}

\section{Introduction}

A key aspect of exploratory data analysis is to study two-dimensional
visualizations of the given high-dimensional input data. In order to gain
insights about the data, one hopes that such visualizations faithfully depict
salient structures that may be present in the input.  $t$-distributed Stochastic
Neighbor Embedding ($t$-SNE) introduced by 
\citet{Maaten2008} is a prominent and popular visualization
technique that has been applied successfully in several application domains \citep{chem_app,physics_app,security_app,music_app,cancer_app,bio_app1}.

Arguably, alongside PCA, $t$-SNE has now become the de facto method of choice used by
practitioners for 2D visualizations to study and unravel the structure present
in data.  Despite its immense popularity, very little work has been done to
systematically understand the power and limitations of the $t$-SNE method, and
the quality of visualizations that it produces. Only recently researchers
showed that if the high-dimensional input data does contain prominent clusters
then the 2D $t$-SNE visualization will be able to successfully capture the cluster
structure \citep{Linderman2017,arora2018}. While these results are a promising
start, a more fundamental question remains unanswered: 
\begin{center}
\emph{what kinds of intrinsic
structures can a $t$-SNE visualization reveal?} 
\end{center}

Intrinsic structure in data can take many forms. While clusters are a common structure to 
study, there may be several other important structures such as manifold, sparse or hierarchical structures that are 
present in the data as well. How does the $t$-SNE optimization criterion fare at discovering these other structures?

Here we take a largely experimental approach to answer this question. Perhaps
not surprisingly,  minimizing $t$-SNE's KL-divergence criterion is \emph{not}
sufficient to discover all these important types of structure. We adopt the
neighborhood-centric precision-recall analysis proposed by \citet{Venna2010},
which showed that KL-divergence maximizes recall at the expense of precision.
We show that this is geared specifically towards revealing cluster structure
and performs rather poorly when it comes to finding manifold or hierarchical
structure. In order to discover these other types of structure effectively, one
needs a better balance between precision and recall, and we show that this can
be achieved by minimizing $f$-divergences other than the KL-divergence. 

We prescribe that data scientists create and explore low-dimensional
visualizations of their data corresponding to several different
$f$-divergences, each of which is geared toward different types of structure.
To this end, we provide efficient code for finding $t$-SNE embeddings based on
five different $f$-divergences\footnote{The code is available at $github.com/jiwoongim/ft-SNE$.}. Users can even provide their own specific instantiation of an
$f$-divergence, if needed. Our code can optimize either the standard criterion,
or a variational lower bound based on convex conjugate of the $f$-divergence.
Empirically, we found that minimizing this dual variational
form was computationally more efficient and produced better quality embeddings,
even for the standard case of KL-divergence. To our knowledge, this is the
first work that explicitly compares the
optimization of both the primal and dual form of $f$-divergences, which would be of independent
interest to the reader.

    \begin{table*}[t]
      \centering
      {\small 
	    \caption{A list of commonly used $f$-divergences (along with their generating function) and their corresponsing $t$-SNE objective (which we refer to as $ft$-SNE). The last column describes what kind of distance relationship gets emphasized by different choices of $f$-divergence.
        }
      \label{tab:fSNE}
      \begin{tabular}{lccc}\hline
          $D_f(P\|Q)$           & $f(t)$                            & $ft$-SNE objective                                                      & Emphasis \\
\hline
\hline
          Kullback-Leibler (KL)                    & $t\log t$                             & $\sum p_{ij} \left(\log \frac{p_{ij}}{q_{ij}}\right)$       & Local 
\\
          Chi-square ($\mathcal{X}^2$ or CH)    & $(t-1)^2$                   & $\sum \frac{(p_{ij} - q_{ij})^2}{q_{ij}} $        & Local 
\\ 
          Reverse-KL (RKL)           &  $-\log t$                            & $\sum q_{ij} \left(\log \frac{q_{ij}}{p_{ij}}\right) $      & Global  
\\
          Jensen-Shannon (JS)       & $(t+1)\log\frac{2}{(t+1)} + t\log t$ & $\frac{1}{2}(\textrm{KL}(p_{ij}\|\frac{p_{ij}+q_{ij}}{2}) + \textrm{KL}(q_{ij}|\frac{p_{ij}+q_{ij}}{2}))$ & Both 
\\        
          Hellinger distance (HL)   & $(\sqrt{t} -1)^2$	     & $\sum(\sqrt{p_{ij}} - \sqrt{q_{ij}}  )^2 $ & Both 
\\         
 \hline
      \end{tabular}}
    \end{table*}

\section{Stochastic Neighbor Embedding for Low-Dimensional Visualizations}
\label{sec:sne_background}

Given a set of $m$ high-dimensional datapoints $x_1,\ldots, x_m \in \mathbb{R}^D$,
the goal of Stochastic Neighbor Embedding (SNE) is to represent these
datapoints in one- two- or three-dimensions in a way that faithfully
captures important intrinsic structure that may be present in the given
input. It aims to achieve this by first modelling neighboring pairs of points based on distance in the original, high-dimensional space. Then, SNE
aims to find a low-dimensional representation of the input datapoints whose
pairwise similarities induce a probability distribution that is as \emph{close}
to the original probability distribution as possible. More specifically,
SNE computes $p_{ij}$, the probability of selecting a pair of neighboring points $i$ and $j$, as
$$p_{ij} = \frac{p_{i|j} + p_{j|i}}{2m},$$
where $p_{j|i}$ and $p_{i|j}$ represent the probability that $j$ is $i$'s neighbor and $i$ is $j$'s neighbor, respectively. These are modeled as
$$p_{j|i} := \frac{\exp \left(  - \| x_{i} - x_{j} \|^2 / 2\sigma^2_i \right)}{\sum_{k\neq i} \exp \left(  - \| x_{i} - x_k \|^2 / 2\sigma^2_i \right)}.$$
The parameters $\sigma_i$ control the effective neighborhood size for the
individual datapoints $x_i$. In practical implementations the neighborhood
sizes are controlled by the so-called \emph{perplexity} parameter, which can be
interpreted as the effective number of neighbors for a given datapoint and is
proportional to the neighborhood size \cite{Maaten2008}.
 
The pairwise similarities between the corresponding low-dimensional datapoints $y_1,\ldots,y_m \in \mathbb{R}^d$ (where $d = 1, 2$ or $3$ typically), are modelled as
Student's $t$-distribution
$$
q_{ij} :=  \frac{(1+\|y_i-y_j\|^2)^{-1}}{\sum_{k\neq i} (1+\|y_i-y_k\|^2)^{-1}}. 
$$
The choice of a heavy-tailed $t$-distribution to model the low-D similarities is
deliberate and is key to circumvent the so-called \emph{crowding
problem} \cite{Maaten2008}, hence the name $t$-SNE. 

The locations of the mapped $y_i$'s are determined by minimizing the discrepancy between the original high-D pairwise similarity distribution $P = (p_{ij})$ and the corresponding low-D distribution $Q=(q_{ij})$.
$t$-SNE prescribes minimizing the KL-divergence ($ D_\textrm{KL}$) between distributions $P$ and $Q$ to find an optimal configuration of the mapped points
$$
J_{\textrm{KL}}(y_1,\ldots,y_m) := D_\textrm{KL}(P||Q) = \sum_{i\neq j} p_{ij} \log \frac{p_{ij}}{q_{ij}}.
$$
While it is reasonable to use KL-divergence to compare the pairwise distributions $P$ and $Q$, there is no compelling reason why it should be preferred over other measures. 
In fact we will demonstrate that using KL-divergence is restrictive for some types of structure discovery, and one should explore other divergence-based measures as well to gain a wholistic 
understanding of the input data.

\begin{figure*}[t]
   \begin{minipage}{\textwidth}
       \begin{minipage}{0.195\textwidth}
       \includegraphics[width=\linewidth]{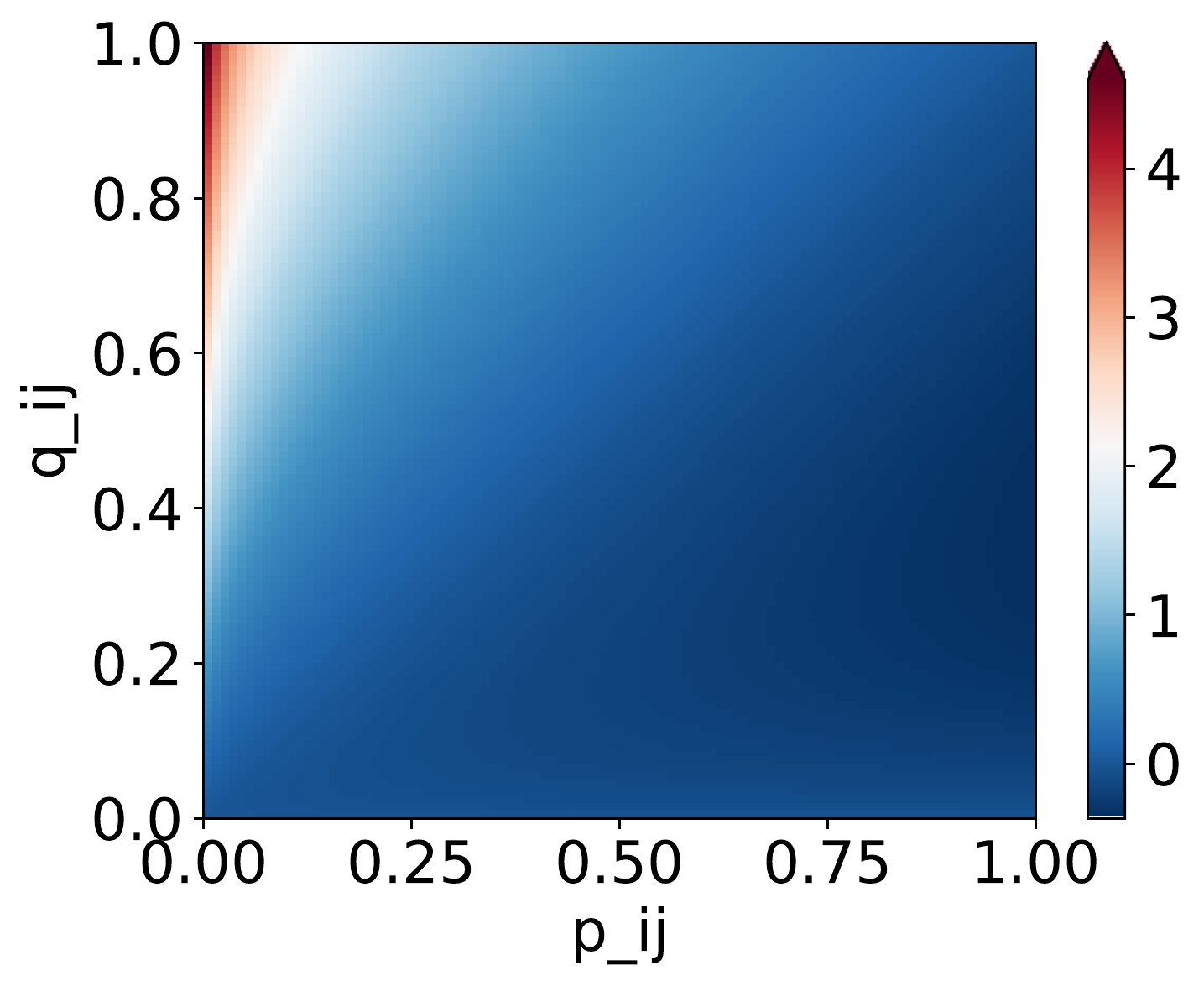}
       \end{minipage}   
       \begin{minipage}{0.195\textwidth}
       \includegraphics[width=\linewidth]{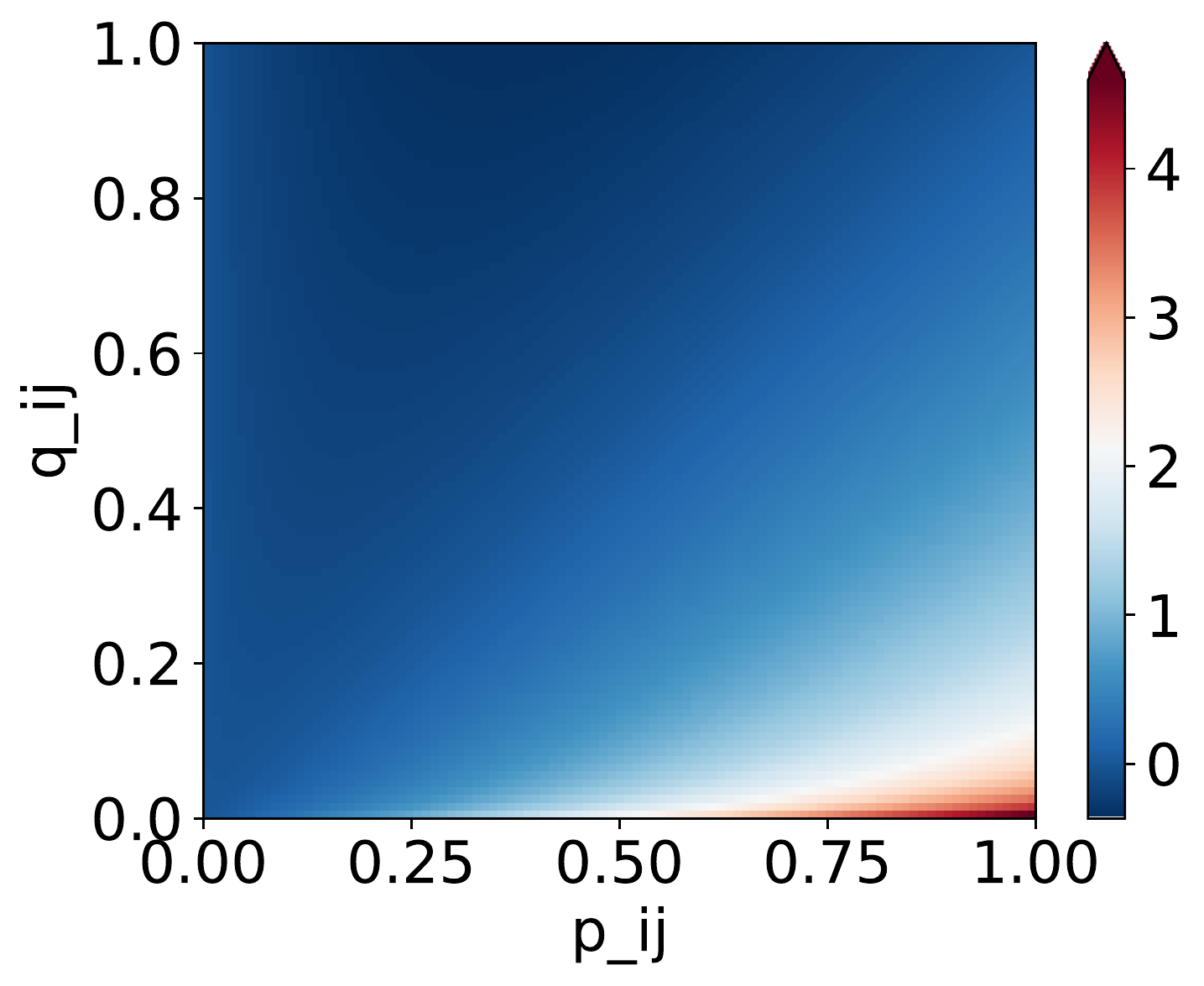}
       \end{minipage}         
       \begin{minipage}{0.195\textwidth}
       \includegraphics[width=\linewidth]{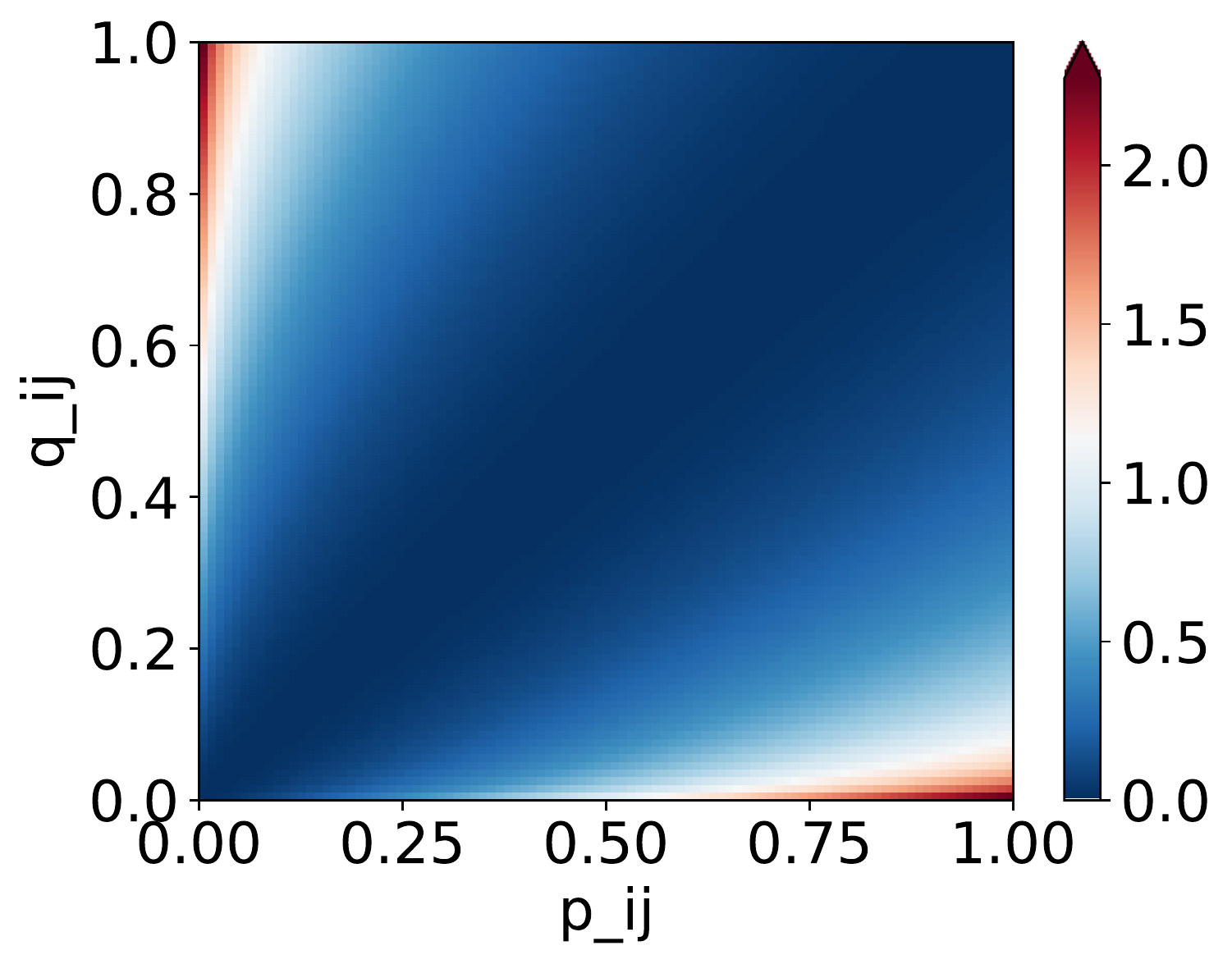}
       \end{minipage}  
       \begin{minipage}{0.195\textwidth}
       \includegraphics[width=\linewidth]{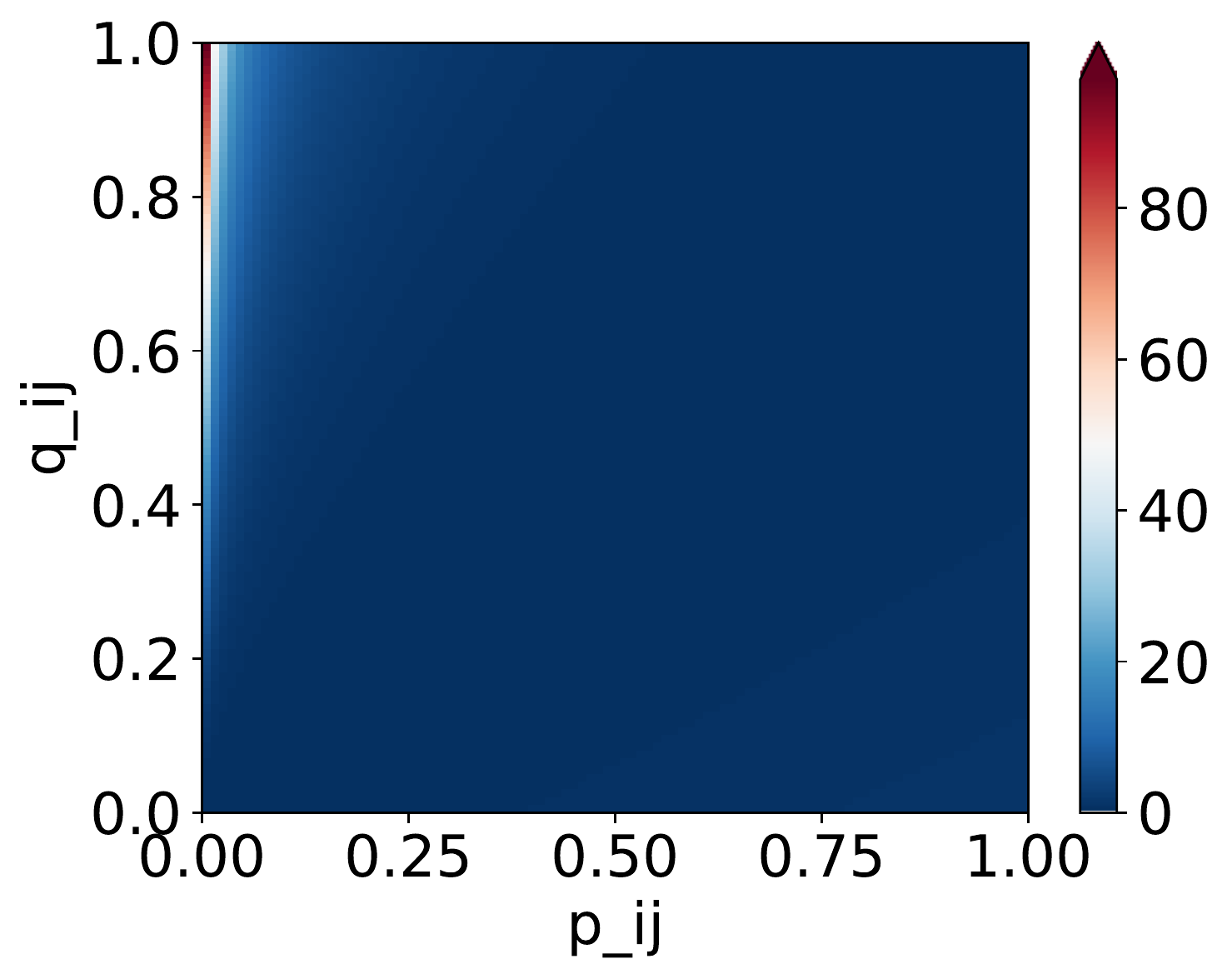}
       \end{minipage}  
       \begin{minipage}{0.195\textwidth}
       \includegraphics[width=\linewidth]{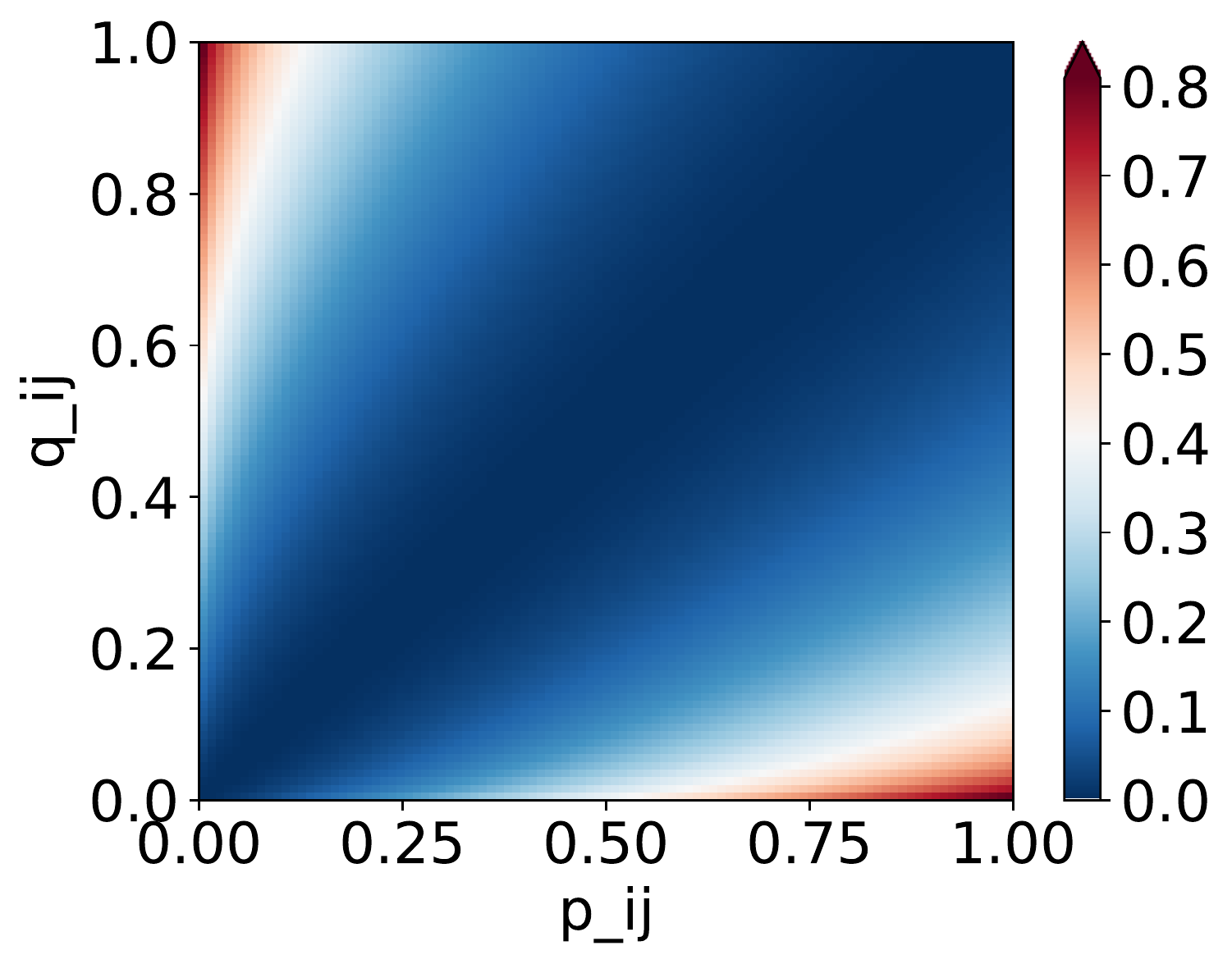}
       \end{minipage}  
   \end{minipage}             
   \begin{minipage}{\textwidth}
       \begin{minipage}{0.195\textwidth}
       \includegraphics[width=\linewidth]{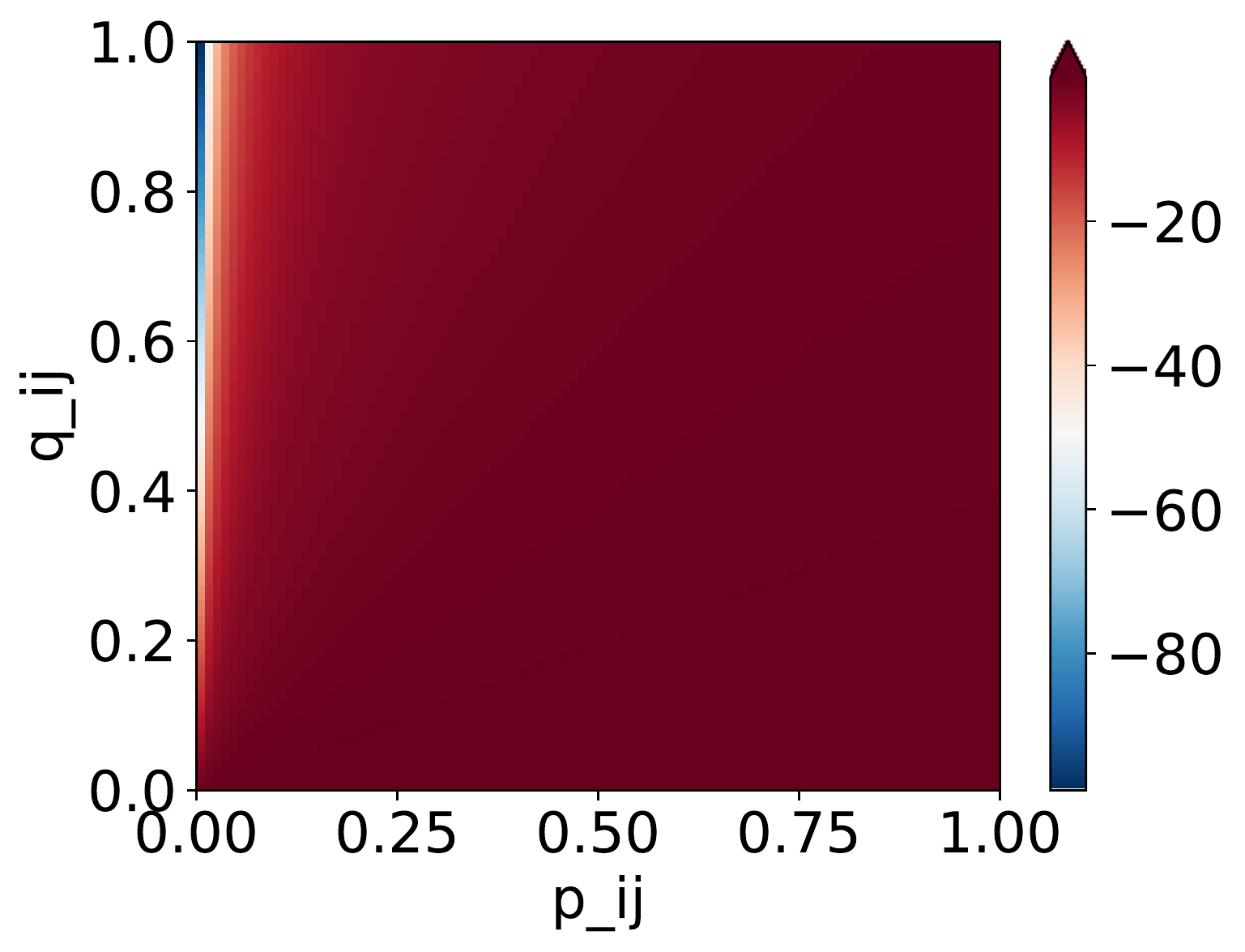}
       \subcaption*{KL}
       \end{minipage}   
       \begin{minipage}{0.195\textwidth}
       \includegraphics[width=\linewidth]{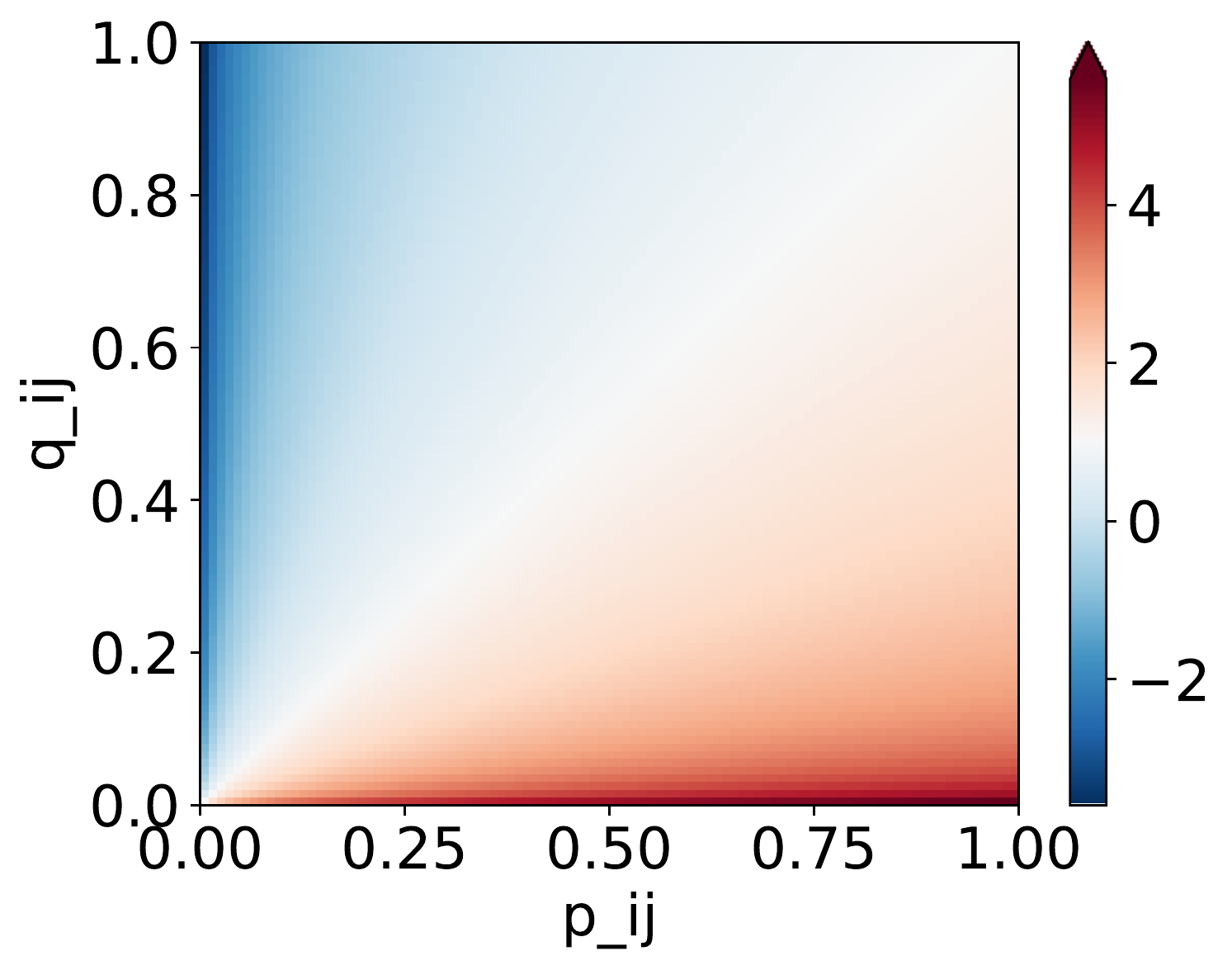}
       \subcaption*{RKL}
       \end{minipage}         
       \begin{minipage}{0.195\textwidth}
       \includegraphics[width=\linewidth]{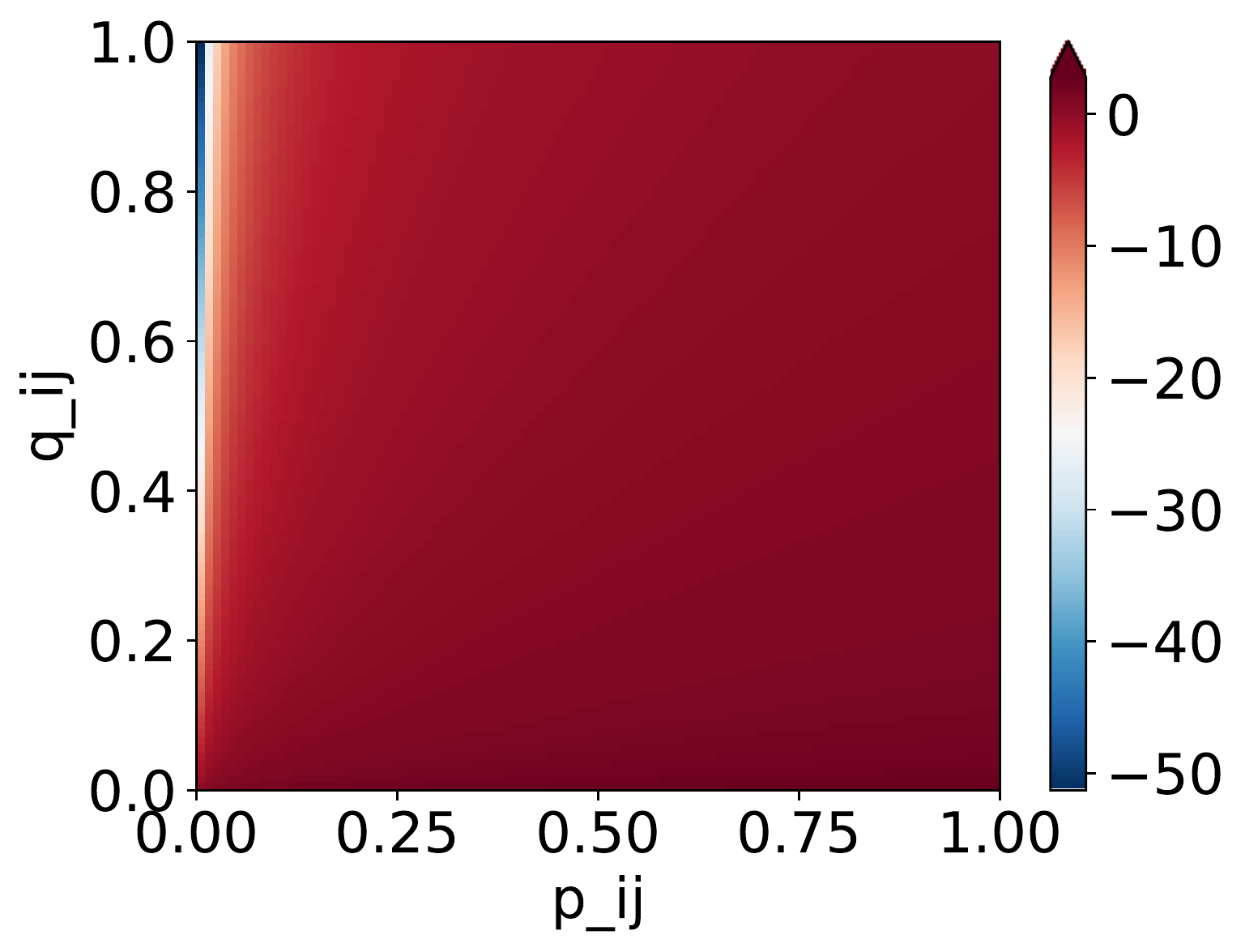}
       \subcaption*{JS}
       \end{minipage}  
       \begin{minipage}{0.195\textwidth}
       \includegraphics[width=\linewidth]{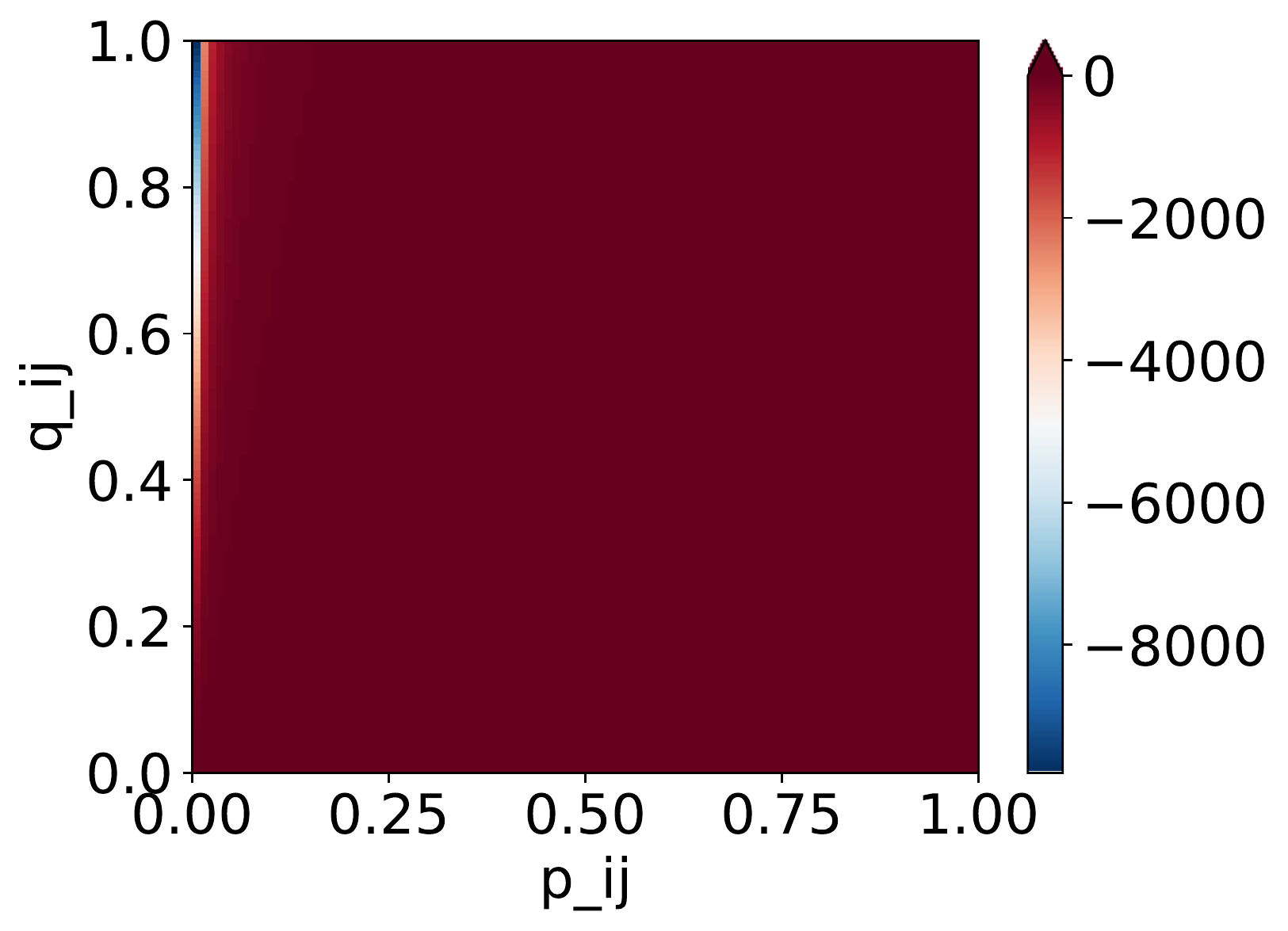}
       \subcaption*{CH}
       \end{minipage} 
       \begin{minipage}{0.195\textwidth}
       \includegraphics[width=\linewidth]{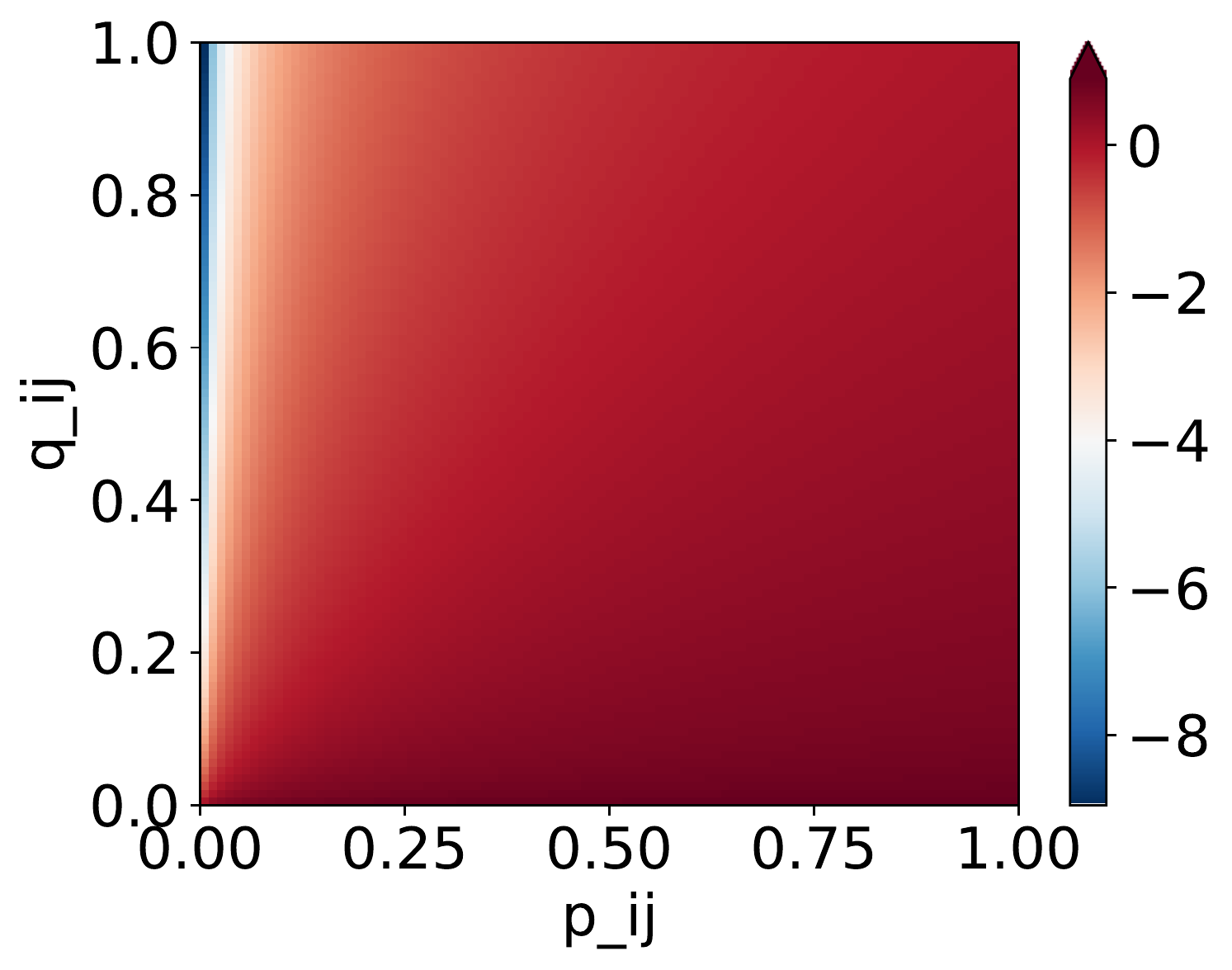}
       \subcaption*{HL}
       \end{minipage} 
   \end{minipage}     
   \caption{Top: $f$-divergence loss. Bottom: gradient of $f$-divergence.
    The color limit represents the magnitude of $f$-divergence (resp.\ gradient of $f$-divergence)
    of $p_{ij}$ and $q_{ij}$.} 
   \label{fig:div_p_vs_q}
   \vspace{-0.2cm}
\end{figure*}

\section{$f$-Divergence-based Stochastic Neighbor Embedding \label{sec:fSNE}}

KL-divergence is a special case of a broader class of divergences called $f$-divergences. A few popular special cases of $f$-divergences include
the reverse KL divergence, Jenson-Shannon divergence, Hellinger distance (HL), total variation distance and $\chi^2$-divergence. Of course, each instantiation compares discrepancy between the distributions differently \cite{nguyen2008} 
and 
it would be instructive to study what effects, if any, do these other divergences have on low-D visualizations of a given input. 
Formally $f$-divergence between two distributions $P$ and $Q$ (over the same measurable space $\Omega$) is defined as 
$$ 
     D_f(P||Q) := \int_\Omega f\left(\frac{P(x)}{Q(x)}\right)dQ(x),
$$ 
where $f$ is a convex function such that $f(1)=0$. Intuitively, $f$-divergence tells us the average odds-ratio between $P$ and $Q$ weighted by the function $f$. For the $t$-SNE objective, the 
generic form of $f$-divergence simplifies to 
\begin{align}
    J_{f}(y_1,\ldots,y_m) := D_f(P||Q) = \sum_{i\neq j} q_{ij} f\left(\frac{p_{ij}}{q_{ij}}\right).
    \label{eqn:fSNE_primal}
\end{align}
Table~\ref{tab:fSNE} shows a list of common instantiations of $f$-divergences and their corresponding $t$-SNE objectives, which we shall call $ft$-SNE.
  
Obviously, one expects different optimization objectives (i.e.\ different choices
of $f$) to produce different results. A more significant question is whether these differences have any significant qualitative effects on types of structure discovery.

An indication towards why the choice of $f$ might affect the type of structure revealed is to notice that $f$-divergences are typically asymmetric, and penalize 
the ratio $p_{ij}/q_{ij}$ (cf.\ Eq.\ \ref{eqn:fSNE_primal}) differently. KL-SNE (i.e.\ $f$ taken as KL-divergence, cf.\ Table \ref{tab:fSNE}) for instance 
penalizes pairs of nearby points in the original space getting mapped far away in the embedded space more heavily than faraway points being mapped nearby (since the corresponding $p_{ij} \gg q_{ij} \approx 0$). Thus KL-SNE optimization prefers visualizations that don't distort local neighborhoods. 
In contrast, SNE with the reverse-KL-divergence criterion, RKL-SNE, as the name suggests, emphasizes the opposite, and better captures global structure in the corresponding visualizations.

A nice balance between the two extremes is achieved by the JS- and HL-SNE (cf.\ Table \ref{tab:fSNE}), where JS is simply an arithmetic mean of the KL and RKL penalties, and HL is a sort of aggregated geometric mean. Meanwhile, CH-SNE can be viewed as relative version of the (squared) $L_2$ distance between the distributions, and is a popular choice for comparing bag-of-words models \cite{bow_chi_sq}. 

We can empirically observe how $p$ and $q$ similarities are penalized by divergence (see Figure~\ref{fig:div_p_vs_q}).
Our observation matches with our intuition: KL and CH are sensitive to high $p$ and low $q$, whereas RKL is sensitive to low $p$ and high $q$, and
JS and HL are symmetric.
The corresponding gradients w.r.t.\ $q$ show that all divergence are generally sensitive to when $p$ is high and $q$ is low.
However, RKL, JS, and HL provide much smoother gradient signals over $p > q$ space.
KL penalize strictly towards high $p$ and low $q$ and CH is much stricter towards $p \gg q$ space.
\\


\noindent \textbf{A Neighborhood-level Precision-Recall Analysis.} 
The optimization criterion of $ft$-SNE is a complex non-convex function that is
not conducive to a straightforward analysis without simplifying assumptions. 
To simplify the analysis, we consider pairs of points in a \emph{binary neighborhood}
setting, where, for each datapoint, other datapoints are either in its neighborhood, or not in its neighborhood.

Let $N_\epsilon(x_i)$ and $N_\epsilon(y_i)$ denote the neighbors of points
$x_i$ and $y_i$ by thesholding the pairwise similarities $p_{j|i}$ and $q_{j|i}$
at a fixed threshold $\epsilon$, respectively. Let $r_i := |N_\epsilon(x_i)|$ and $k_i := |N_\epsilon(y_i)|$ denote the number of true and retrieved neighbors. Our simplifying binary neighborhood assumption can be formalized as: 
\begin{flalign*}
\begin{aligned}
p_{ij} &:= \begin{cases}
a_i, & x_j \in N_\epsilon(x_i)  \\
b_i, & x_j \notin N_\epsilon(x_i)
\end{cases},
\end{aligned}
\begin{aligned}
q_{ij} &:= \begin{cases} 
c_i, & y_j \in N_\epsilon(y_i)  \\
d_i, & y_j \notin N_\epsilon(y_i)
\end{cases}.
\end{aligned}
\end{flalign*}
where $a_i$ and $c_i$ are large ($a_i \geq \frac{1-\delta}{r_i}$, $c_i \geq \frac{1-\delta}{k_i}$) and $b_i$ and $d_i$ are small ($b_i \leq \frac{\delta}{m-r_i-1}$, $d_i \leq \frac{\delta}{m-k_i-1}$), for small $\delta$. 

\begin{figure*}[htb]
\begin{center}
\begin{minipage}{.49\textwidth}
\includegraphics[width=\linewidth]{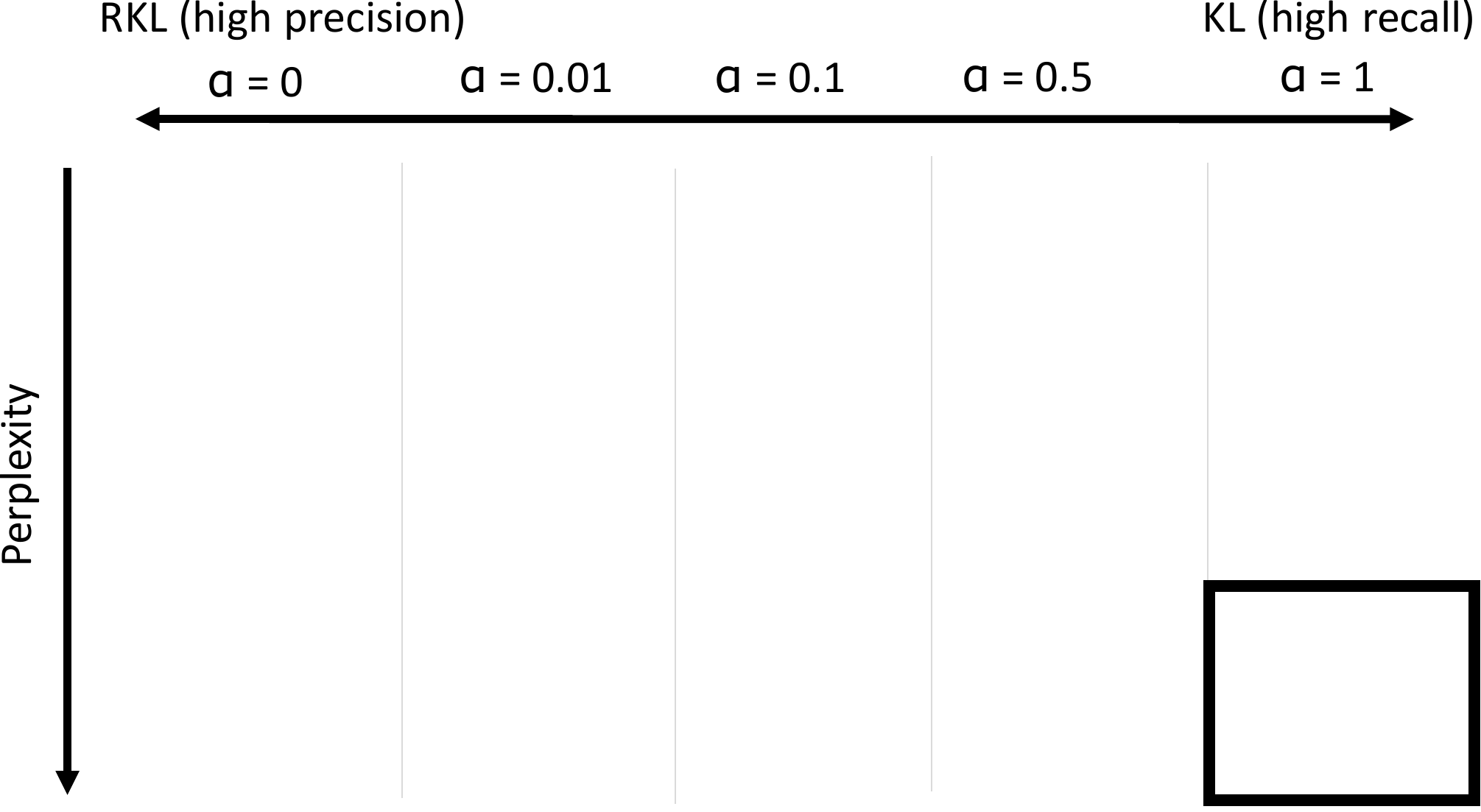}
\subcaption{3 well-separated Gaussian clusters}
\end{minipage}
\begin{minipage}{.49\textwidth}
\includegraphics[width=\linewidth]{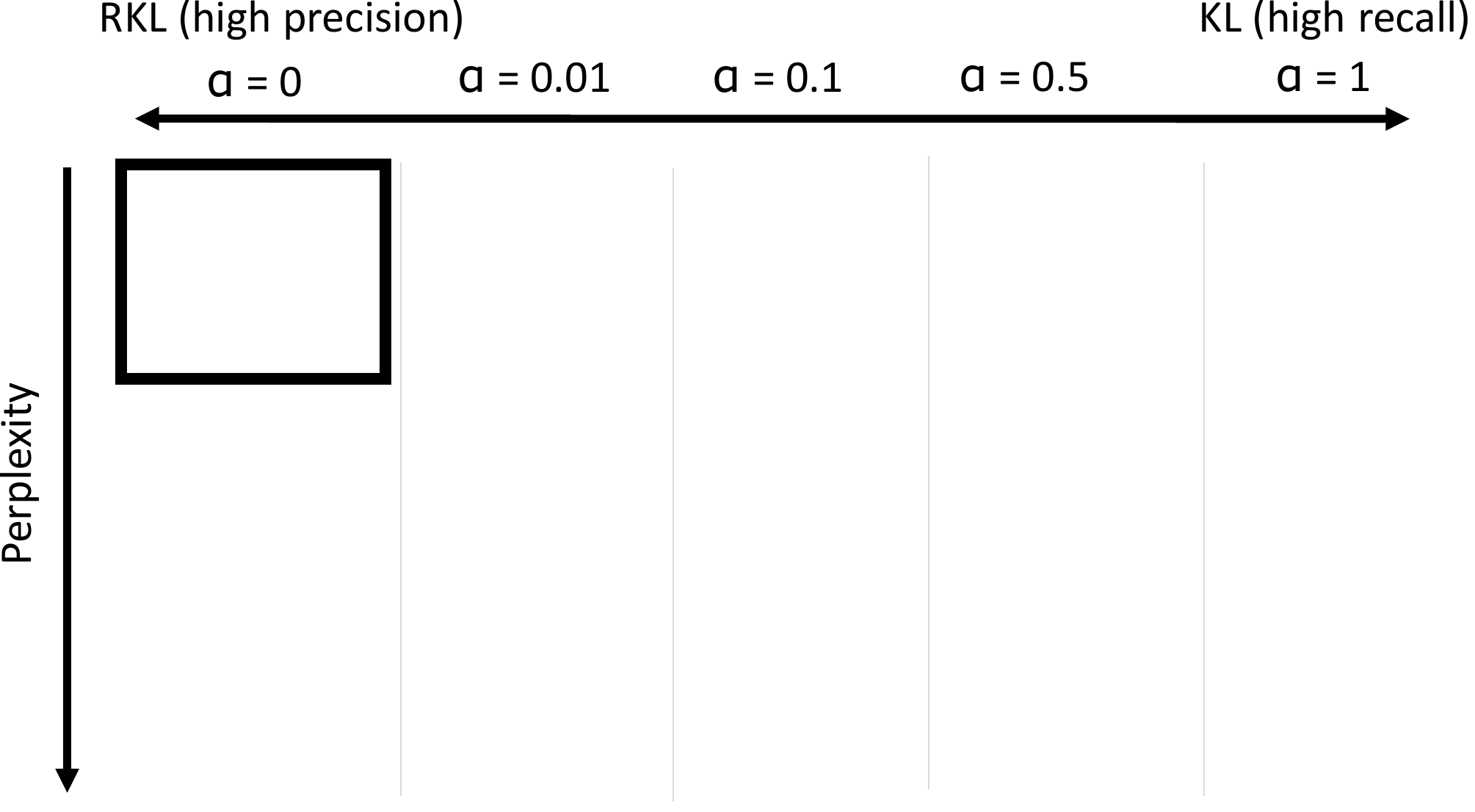}
\subcaption{Swiss roll manifold}
\end{minipage}
\end{center}
\caption{$ft$-SNE embeddings obtained with interpolated divergences between KL and RKL.
    The perplexity for each row corresponds to 10, 100, and 500 respectively.} \label{fig:syn_embeddings}
\vspace{-0.2cm}
\end{figure*}

In this binary formulation, we can rewrite each of the $f$-divergences in terms related to the embedding precision, the fraction of embedding-neighbors are true neighbors, and the recall, the fraction of true neighbors are also embedding-neighbors. Define $n_{\TP}^i := |N_\epsilon(x_i) \cap N_\epsilon(y_i)|$, $n_\textrm{FP}^i := |N_\epsilon(y_i) \setminus N_\epsilon(x_i)|$ and
$n_\textrm{FN}^i := |N_\epsilon(x_i) \setminus N_\epsilon(y_i)|$ to denote the
number of true-positive, false-positive and false-negative neighbors
respectively. In this notation, per-neighborhood precision is $n_{\TP}^i / k_i
= 1 - n_{\FP}^i/k_i$ and recall is $n_{\TP}^i/r_i = 1 - n_{\FN}^i/r_i$. This
information retrieval analysis has previously been performed for KL-SNE
\cite{Venna2010}. Novelly, we extend it to other $f$-divergences to understand
their assumptions. 

\begin{proposition}
    \label{prop1}
    Under the binary-neighborhood assumption, for $\delta$ sufficiently small,
\begin{enumerate}[(i)]
\item $J_{\KL} \propto  \Big( \sum_i \underbrace{{n^i_{\FN}}/{r_i}}_{1-\textrm{recall}} \Big)$, maximizes recall.
\item $J_{\RKL} \propto \Big( \sum_i \underbrace{{n^i_{\FP}}/{k_i}}_{1-\textrm{precision}} \Big)$ maximizes precision.
\item $J_{\JS} \propto J_{\KL}+J_{\RKL}$ balances precision and recall,
\item The first two terms of HL-SNE balance precision and recall (the coefficients are close to 1, since $\delta$ is small). The last term forces preservation of neighborhood sizes, and strongly penalizes small embedding neighborhoods when precision is high. 
\begin{align*}
J_{\HL} \propto \sum_i &\Big[ \underbrace{ \Big( \frac{n_{\FN}^i}{r_i} \Big)}_{1-\textrm{recall}}     \cdot ( 1 - O({(\delta r_i)}^{\frac{1}{2}})) \Big] 
                       +\Big[ \underbrace{ \Big( \frac{n_{\FP}^i}{k_i} \Big)}_{1-\textrm{precision}}  \cdot ( 1 - O({(\delta k_i)}^{\frac{1}{2}})) \Big] 
                       +\underbrace{\Big( \frac{n_{\TP}^i}{k_i} \Big) }_{\textrm{precision}} \cdot \underbrace{\Big(\sqrt{\frac{r_i}{k_i}} - 1 \Big)^2 }_{\substack{\textrm{neighborhood}\\\textrm{size ratio}} }.
\end{align*} 
\item 
CH-SNE is biased towards maximizing recall, since the multiplier of recall is much larger than that on precision. Like HL-SNE, the last term forces preservation of neighborhood sizes, and strongly penalizes small embedding neighborhoods when precision is high. 
\begin{align*}
J_{\CH} \propto \sum_i & \Big[ \underbrace{ \Big( \frac{n_{\FN}^i}{r_i} \Big)}_{1-\textrm{recall}}     \cdot \Big( \frac{m-k_i}{r_i \delta} \Big) \Big]  
+    \underbrace{ \Big( \frac{n_{\FP}^i}{k_i} \Big)}_{1-\textrm{precision}}  
 +    \Big[ \underbrace{\Big( \frac{n_{\TP}^i}{k_i} \Big) }_{\textrm{precision}} \cdot \underbrace{\Big({\frac{r_i}{k_i}} - 1 \Big)^2 }_{\substack{\textrm{neighborhood} \\ \textrm{size ratio}}} \Big].
\end{align*} 
\end{enumerate}
\end{proposition}
This proposition corroborates our intuition (see also Table \ref{tab:fSNE}), and provides a relationship between the proposed $ft$-SNE criteria and the types of neighborhood similarities that are preserved. KL-SNE maximizes neighborhood recall, while RKL-SNE maximizes neighborhood precision. All other criteria balance precision and recall in different ways. JS-SNE provides equal weight to precision and recall. HL-SNE gives approximately equal weight to precision and recall, with an extra term encouraging the original and embedding neighborhood sizes to match. This regularization term gives more severe penalties if the embedding neighborhood is much smaller than the original neighborhood than the reverse, and thus HL-SNE can be viewed as a regularized version of JS-SNE. CH-SNE gives more weight to maximizing recall, again with an extra term encouraging the original and embedding neighborhood sizes to match, and is and thus similar to a regularized version of KL-SNE.

Next, we connect these precision-recall interpretations of the various criteria to the types of intrinsic structure they preserve. Suppose the intrinsic structure within the data is clusters. A good embedding of this data would have points belonging to the same true cluster all grouped together in the visualization, but the specific locations of the embedded points within the cluster do not matter. Thus, cluster discovery requires good neighborhood recall, and one might expect KL-SNE to perform well. For neighborhood sizes similar to true cluster sizes, this argument is corroborated both theoretically by previous work \citep{Linderman2017,arora2018} and
empirically by our experiments (Experiments Section). Both theoretically and practically, the perplexity parameter---which is a proxy for neighborhood size---needs to be set so that the effective neighborhood size matches the cluster size for successful cluster discovery.

If the intrinsic structure within the data is a continuous manifold, a good embedding would preserve the smoothly varying structure, and not introduce artificial breaks in the data that lead to the appearance of clusters. Having a large neighborhood size (i.e.\ large perplexity) may not be conducive to this goal, again because
the SNE optimization criterion does not care about the specific mapped locations of
the datapoints within the neighborhood. Instead, it is more preferable to have
small enough neighborhood where the manifold sections are approximately
linear one require high precision in these small neighborhoods. Thus one might expect RKL-SNE to fare well manifold discovery tasks.
Indeed, this is also corroborated practically in our experiments. 
(To best of our knowledge, no theory work exists on this.)
\\

    \begin{table}[t]
      \centering
        {\small 
      \caption{Variational $ft$-SNE. }
      \label{tab:vfSNE}
      \begin{tabular}{lcccc}\hline
          $D_f(P\|Q)$                   & $f(t)$                            & $f^*(t)$                                                         & $h(x)$ \\\hline\hline
          Kullback-Leibler (KL)         & $t\log t$                             & $\exp(t-1)$                                                          & $x$\\
          Reverse-KL (RKL)              & $-\log t$                             & $-1-\log(-t)$                                                        & $-\exp(-x)$\\
          Jensen-Shannon (JS)           & $-(t+1)\log\frac{(1+t)}{2} + t\log t$ & $-\log(1-\exp(t))$                                                   & $\log(2) - \log\left(1+\exp(-x)\right)$\\
          Hellinger distance (HL)       & $(\sqrt{t} -1)^2$                     & $\frac{t}{1-t}$                                                      & $1-\exp(-x)$\\ 
          Chi-square ($\mathcal{X}^2$ or CS) & $(t-1)^2$		                        & $\frac{1}{4}t^2+t$                                                   & $x$\\          \hline
      \end{tabular}}
    \end{table}

\noindent \textbf{Variational $ft$-SNE for practical usage and improved optimization.}
The $ft$-SNE criteria can be optimized using gradient descent or one of its variants, e.g.~stochastic gradient descent, and KL-SNE is classically optimized in this way. The proposed $ft$-SNE criteria (including KL-SNE) are non-convex, and gradient descent may not converge to a good solution. We explored minimizing the $ft$-SNE criteria by expressing it in terms of its conjugate dual \citep{Nowozin2016,nguyen2008}:
$$
D_f(P||Q) = \sum_{i\neq j} \left[ q_{ij} \left(\sup_{h \in \mathcal{H}} h((x_i,x_j))\frac{p_{ij}}{q_{ij}} - f^*(h((x_i,x_j))) \right)\right]
$$
where $\mathcal{H}$ is the space of real-valued functions on the underlying measure space and $f^*$ is the Fenchel conjugate of $f$.
In this equation, the maximum operator acts per data point, making optimization infeasible. Instead, we optimize the variational lower bound
$$ 
        D_f(P||Q)\geq  \sup_{h \in \mathcal{H}} \sum_{i \neq j} \left[ h((x_i,x_j))p_{ij} - f^*(h((x_i, x_j)))q_{ij}\right],
$$
which is tight for sufficiently expressive $\mathcal{H}$. 
    In practice, one uses a parameteric hypothesis class $\bar{\mathcal{H}}$, and we use multilayer, fully-connected neural networks.
    Table~\ref{tab:vfSNE} shows a list of common instantiations of $f$-divergences and their corresponding $h(x)$ functions.
    Our variational form of $ft$-SNE objective (or $vft$-SNE) finally becomes the following minimax problem
    \begin{align}
\nonumber
J(&y_1,\ldots,y_m) = 
\min_{y_1, \ldots, y_m} \max_{\bar{h} \in \bar{\mathcal{H}}} \sum_{i \neq j} \left[ \bar{h} ((x_i, x_j))p_{ij} - f^*\left( \bar{h}((x_i,x_j))\right)q_{ij}\right].
        \label{eqn:vfsne}
    \end{align}     
We alternatively optimize $y_{1}, \ldots, y_{m}$ and $\bar{h}$  (see Algorithm~\ref{algo:vfsne_update_rule},
more details available in S.M.). 
    \begin{algorithm}[htp]
        \caption{Variational (Adversarial) SNE Optimization Algorithm}\label{euclid}
        \label{algo:vfsne_update_rule}
        \begin{algorithmic}[1]
            \Procedure{Optimization}{Dataset $\{X_{tr}, X_{vl}\}$, learning rate $\eta$, $f$-divergence $J$}
            \State Initialize the discriminant parameter $\phi$.
    
            \While {$\phi$ has not converged}
                \For {$j=1,\ldots, J$}
                    \State $\phi_{t+1} = \phi_t + \eta \nabla_\phi J$.
                \EndFor
                \For {$k=1,\ldots, K$}
                    \State $y^i_{t+1} = y^i_t - \eta_y \nabla_y J$.
                \EndFor
            \EndWhile
            \EndProcedure
        \end{algorithmic}
      \vspace{-0.1cm}
    \end{algorithm}

    \begin{table*}[t]
      \centering
      \caption{Best $ft$-SNE method for each dataset and criterion, according to maximum F-score in Figure~\ref{fig:results_diff_metric1} and \ref{fig:results_diff_metric2}.}
      \label{tab:results}
        {\footnotesize
      \begin{tabular}{l|c|cc|c|c}\hline
                                &               & \multicolumn{3}{|c|}{Data-Embeddings} & Class-Embedings     \\\hline
          Data                  & Type          & K-Nearest & K-Farthest                & F-Score on X-Y  & F-Score on Z-Y  \\\hline\hline
          MNIST (Digit 1)       & Manifold      & {\color{green} RKL}                 & {\color{green}RKL}      & {\color{green}RKL}     & - \\
          Face                  & Manifold      & {\color{gray}{HL},\color{green}RKL} &  {\color{green}RKL}     &  {\color{green}RKL}   & {\color{blue}JS} \\
          MNIST                 & Clustering    & {\color{red} KL}                    & {\color{red} KL}        &  {\color{brown}CS}         & {\color{red}KL}       \\        
          GENE                  & Clustering    & {\color{red} KL}                    & {\color{red} KL}        &  {\color{red}KL}           & {\color{red}KL}       \\        
          \multirow{2}{*}{20 News Groups}       & Sparse \& & \multirow{2}{*}{{\color{brown} CS}} & \multirow{2}{*}{{\color{brown}CS}}    & \multirow{2}{*}{{\color{brown}CS}} & \multirow{2}{*}{{\color{gray}HL}}   \\
                                & Hierachical  & & & \\
          \multirow{2}{*}{ImageNet (sbow)}      & Sparse \& & \multirow{2}{*}{{\color{brown} CS}} & \multirow{2}{*}{{\color{brown}CS}}   & \multirow{2}{*}{{\color{brown}CS}} & \multirow{2}{*}{{\color{red}KL}}   \\
                                & Hierachical  & & & \\\hline
      \end{tabular}}
      \vspace{-0.3cm}
    \end{table*}

\section{Experiments}
\label{sec:expts}
  
In this section, we compare the performance of the proposed $ft$-SNE methods in preserving different types of structure present in selected data sets. Next, we compare the efficacy of optimizing the primal versus the dual form of the $ft$-SNE. Details about the datasets, optimization parameters, and architectures are described in the Supplementary Material (S.M.).
\\

\noindent \textbf{Datasets. }
We compared the proposed $ft$-SNE methods on a variety of datasets with different latent structures. The MNIST dataset consists of images of handwritten digits from 0 to 9~\cite{lecun1998gradient}, thus the latent structure is clusters corresponding to each digit. We also tested on just MNIST images of the digit 1, which corresponds to a continuous manifold. The Face dataset, proposed in~\cite{Tenenbaum2000}, consists of rendered images of faces along a 3-dimensional manifold corresponding to up-down rotation, left-right rotation, and left-right position of the light source. The Gene dataset consists of RNA-Seq gene expression levels for patients with five different types of tumors~\cite{weinstein2013cancer}, and thus has a cluster latent structure. The 20-Newsgroups dataset consisted of text articles from a hierarchy of topics~\cite{joachims1996probabilistic}, and thus the latent structure corresponded to a hierarchical clustering. In addition, we used a bag-of-words representation of the articles, thus the feature representation is sparse (many of the features in the original representation will be 0). We also examined two synthetic datasets: the Swiss Roll dataset~\cite{Tenenbaum2000} which has a continuous manifold latent structure, and a simple dataset consisting of 3 Gaussian clusters in 2 dimensions, which has a cluster latent structure. Details of these datasets can be found in Appendix. 

\subsection{Comparison of $f$-divergences for SNE}

We developed several criteria for quantifying the performance of the different $ft$-SNE methods. Our criteria are based on the observation that, if the local structure is well-preserved, then the nearest neighbours in the original data space $\mathcal{X}$ should match the nearest neighbours in the embedded space $\mathcal{Y}$. In addition, many of our datasets include a known latent variable, e.g.~the discrete digit label for MNIST and the continuous head angle for Face. Thus, we also measure how well the embedded space captures the known structure of the latent space $\mathcal{Z}$. We define the neighbors $N_\epsilon(x_i)$, $N_\epsilon(y_i)$, and $N_\epsilon(z_i)$ of points $x_i$, $y_i$, and $z_i$ by thresholding the pairwise similarity $p_{j|i}$, $q_{j|i}$ and $r_{j|i}$, respectively, at a selected threshold $\epsilon$. Here, $r_{j|i} = r(z_j|z_i)$ is the pairwise similarity in the latent space $\mathcal{Z}$. For discrete labels, we define $r_{j|i} \propto \mathbb{I}(z_i = z_j)$. For continuous latent spaces, we use a t-distribution.

Using these definitions of neighbors, we can define precision and recall, considering the original $\mathcal{X}$ or latent $\mathcal{Z}$ spaces as true and the embedded space $\mathcal{Y}$ as the predicted:
\begin{align}
	\allowdisplaybreaks
\nonumber
\text{Precision}_X(\epsilon) &= \frac{1}{N}\sum^{N}_{i} \frac{| N_\epsilon(y_i) \cap N_\epsilon(x_i) |}{|N_\epsilon(y_i)|},  \quad 
 &\text{Precision}_Z(\epsilon)=\frac{1}{N}\sum^{N}_{i} \frac{| N_\epsilon(y_i) \cap N_\epsilon(z_i) |}{|N_\epsilon(y_i)|} \\
\nonumber
\text{Recall}_X(\epsilon) &= \frac{1}{N}\sum^{N}_{i} \frac{| N_\epsilon(y_i) \cap N_\epsilon(x_i) |}{|N_\epsilon(x_i)|}, \quad 
 &\text{Recall}_Z(\epsilon)=\frac{1}{N}\sum^{N}_{i} \frac{| N_\epsilon(y_i) \cap N_\epsilon(z_i) |}{|N_\epsilon(z_i)|}.
\end{align}

Alternatively, we can measure how well the embedded space preserves the nearest and farthest neighbor structure. Let $NN_K(x_i)$ and $NN_K(y_i)$ indicate the $K$ nearest neighbors and $FN_K(x_i)$ and $FN_K(y_i)$ indicate the $K$ farthest neighbors. We define
\begin{align}
\nonumber
\text{NN-Precision}(K) &= \frac{1}{NK}\sum^{N}_{i}|\text{NN}_K(y_i) \cap \text{NN}_K(x_i)|, \\ 
\nonumber
    \text{FN-Precision}(K) &= \frac{1}{NK}\sum^{N}_{i}|\text{FN}_K(y_i) \cap \text{FN}_K(X_i)|
\end{align}

\begin{figure*}[t]
        \begin{minipage}{\textwidth}
            \begin{minipage}{0.16\textwidth}
            \centering MNIST1
            \end{minipage}   
            \begin{minipage}{0.16\textwidth}
            \centering Face
            \end{minipage}   
            \begin{minipage}{0.16\textwidth}
            \centering MNIST
            \end{minipage}   
            \begin{minipage}{0.16\textwidth}
            \centering GENE
            \end{minipage}   
            \begin{minipage}{0.16\textwidth}
            \centering NEWS
            \end{minipage}   
            \begin{minipage}{0.16\textwidth}
            \centering SBOW
            \end{minipage}   
        \end{minipage}
        \begin{minipage}{\textwidth}
            \begin{minipage}{0.16\textwidth}
            \includegraphics[width=\linewidth]{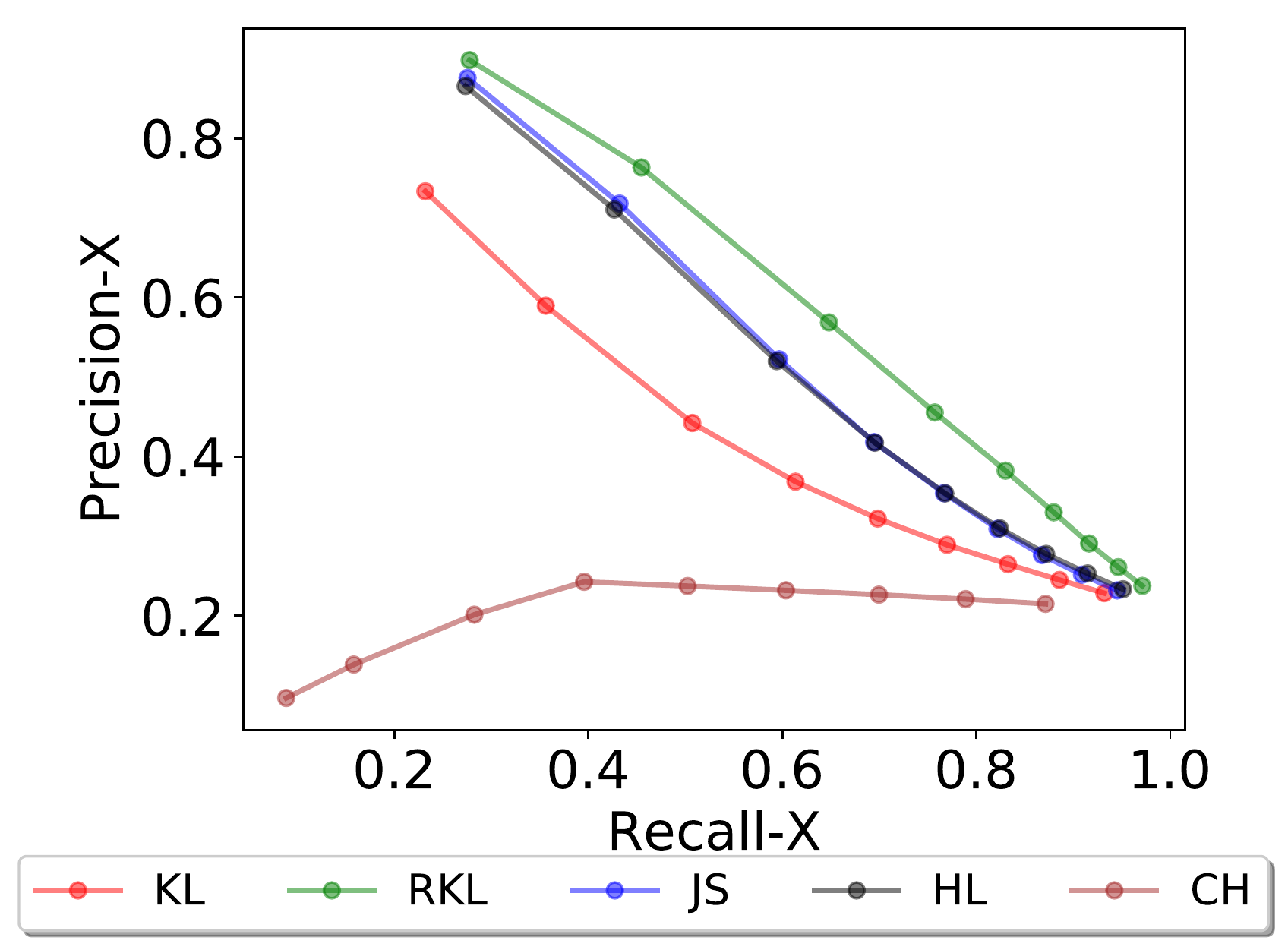}
            \vspace{-0.3cm}
            \end{minipage}
            \begin{minipage}{0.16\textwidth}
            \includegraphics[width=\linewidth]{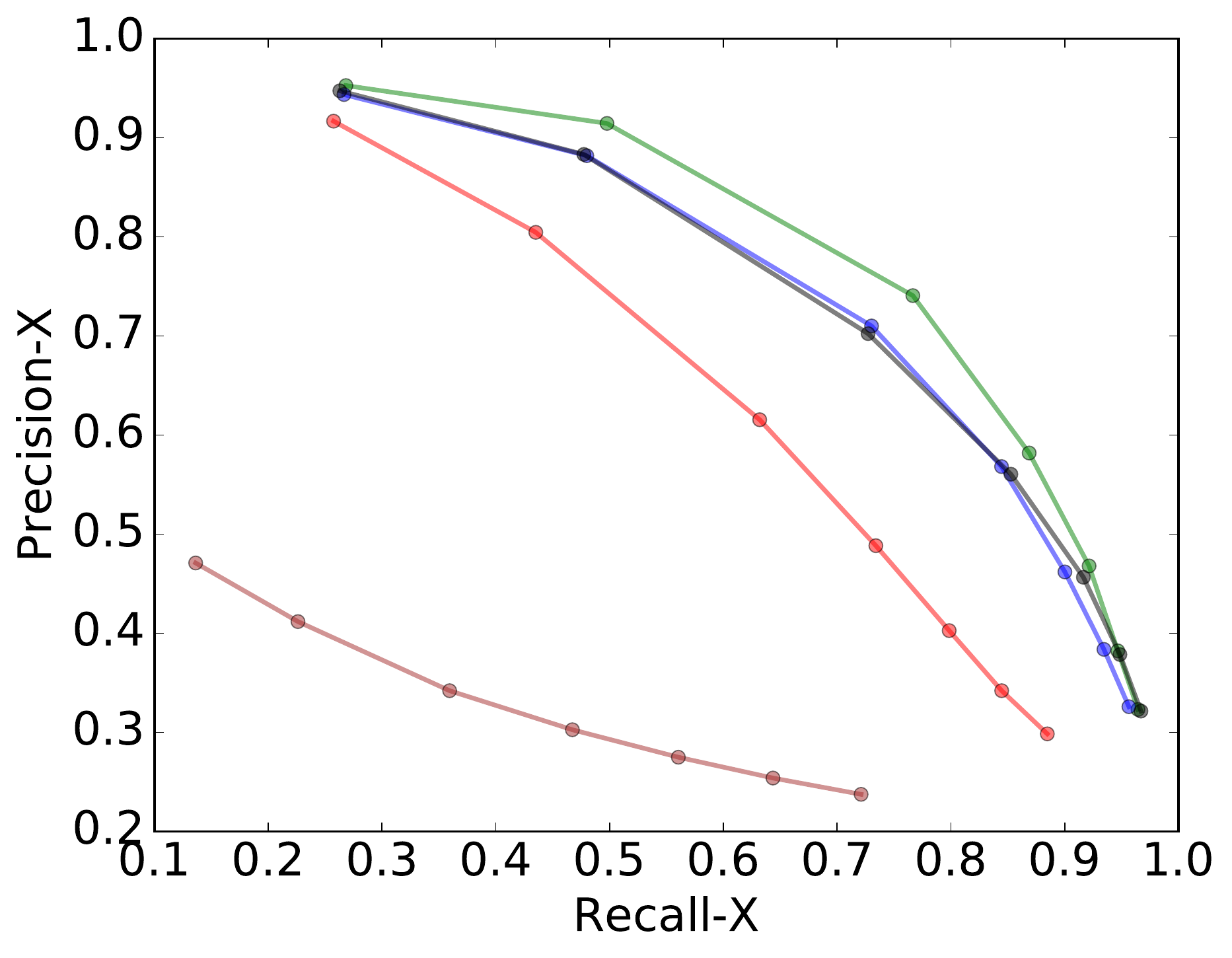}
            \vspace{-0.3cm}
            \end{minipage}         
            \begin{minipage}{0.16\textwidth}
            \includegraphics[width=\linewidth]{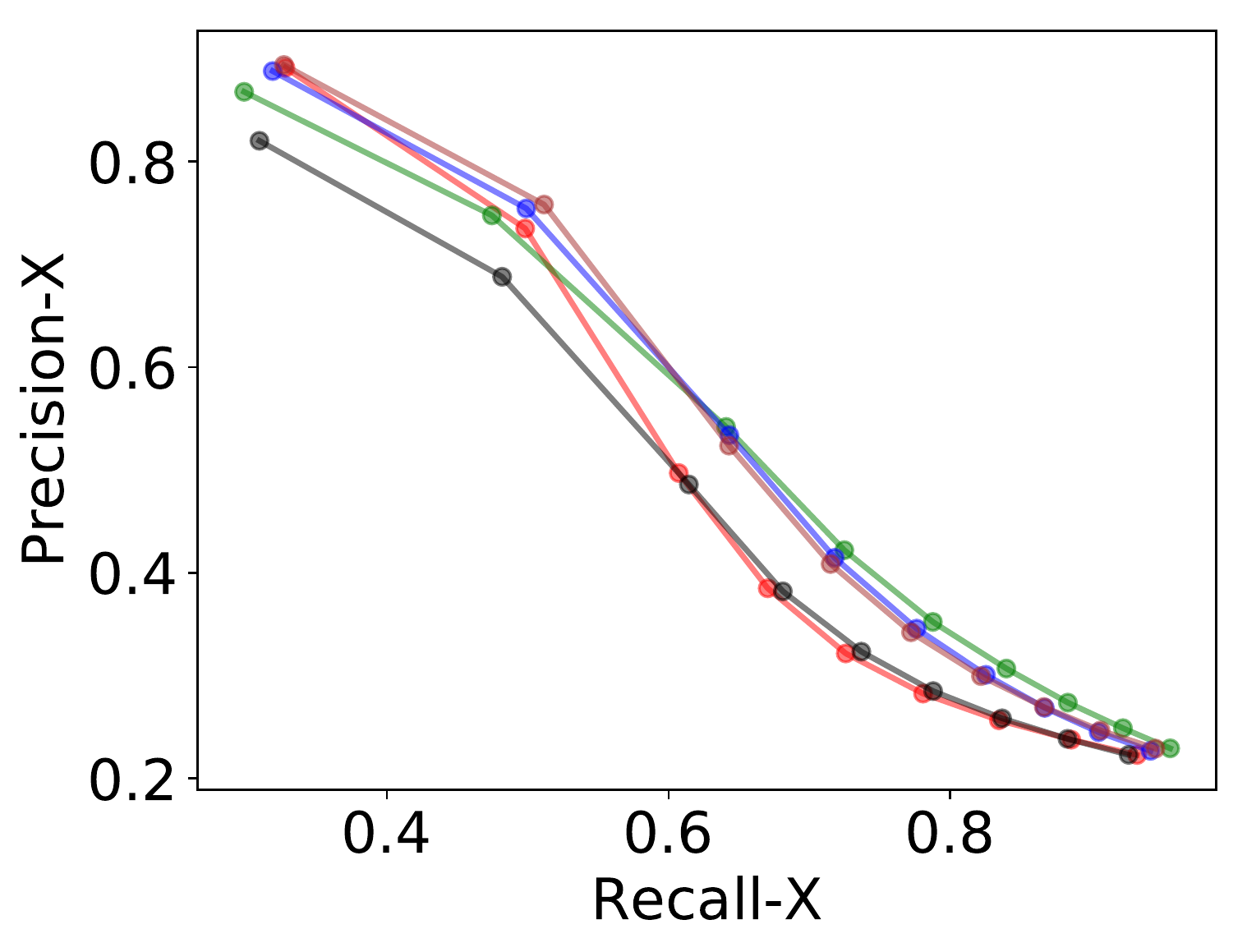}
            \vspace{-0.3cm}
            \end{minipage}         
            \begin{minipage}{0.16\textwidth}
            \includegraphics[width=\linewidth]{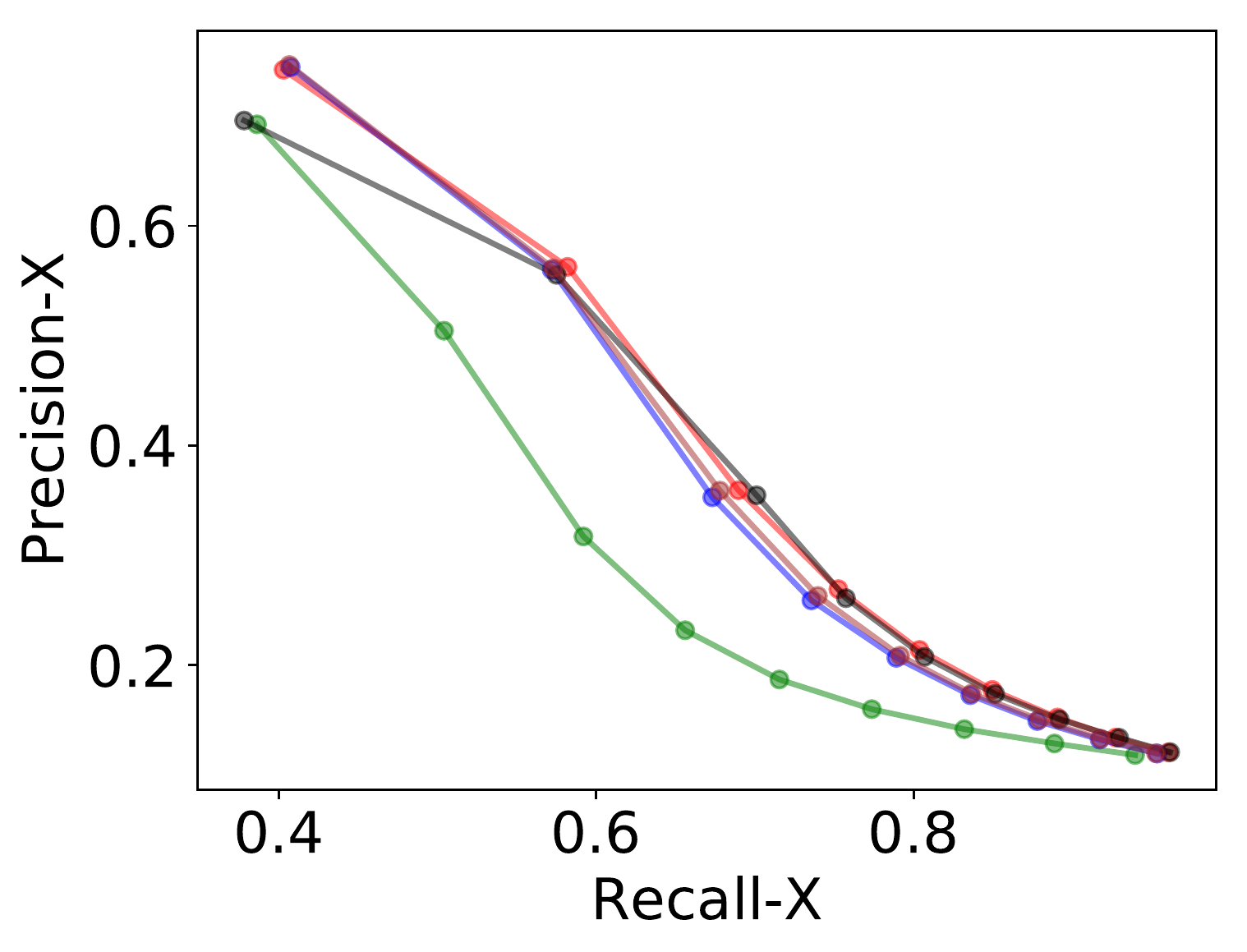}
            \vspace{-0.3cm}
            \end{minipage}   
            \begin{minipage}{0.16\textwidth}
            \includegraphics[width=\linewidth]{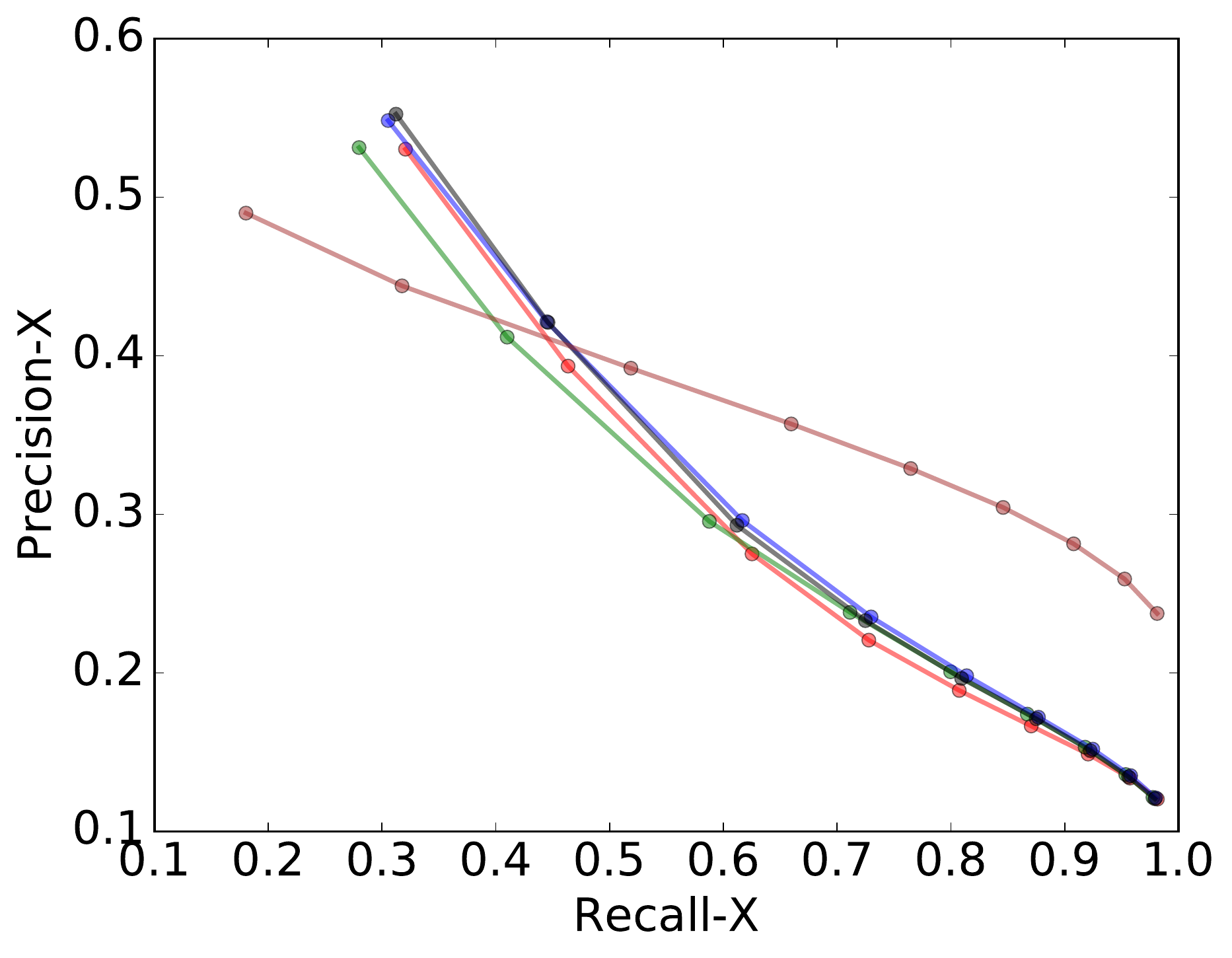}
            \vspace{-0.3cm}
            \end{minipage}  
            \begin{minipage}{0.16\textwidth}
            \includegraphics[width=\linewidth]{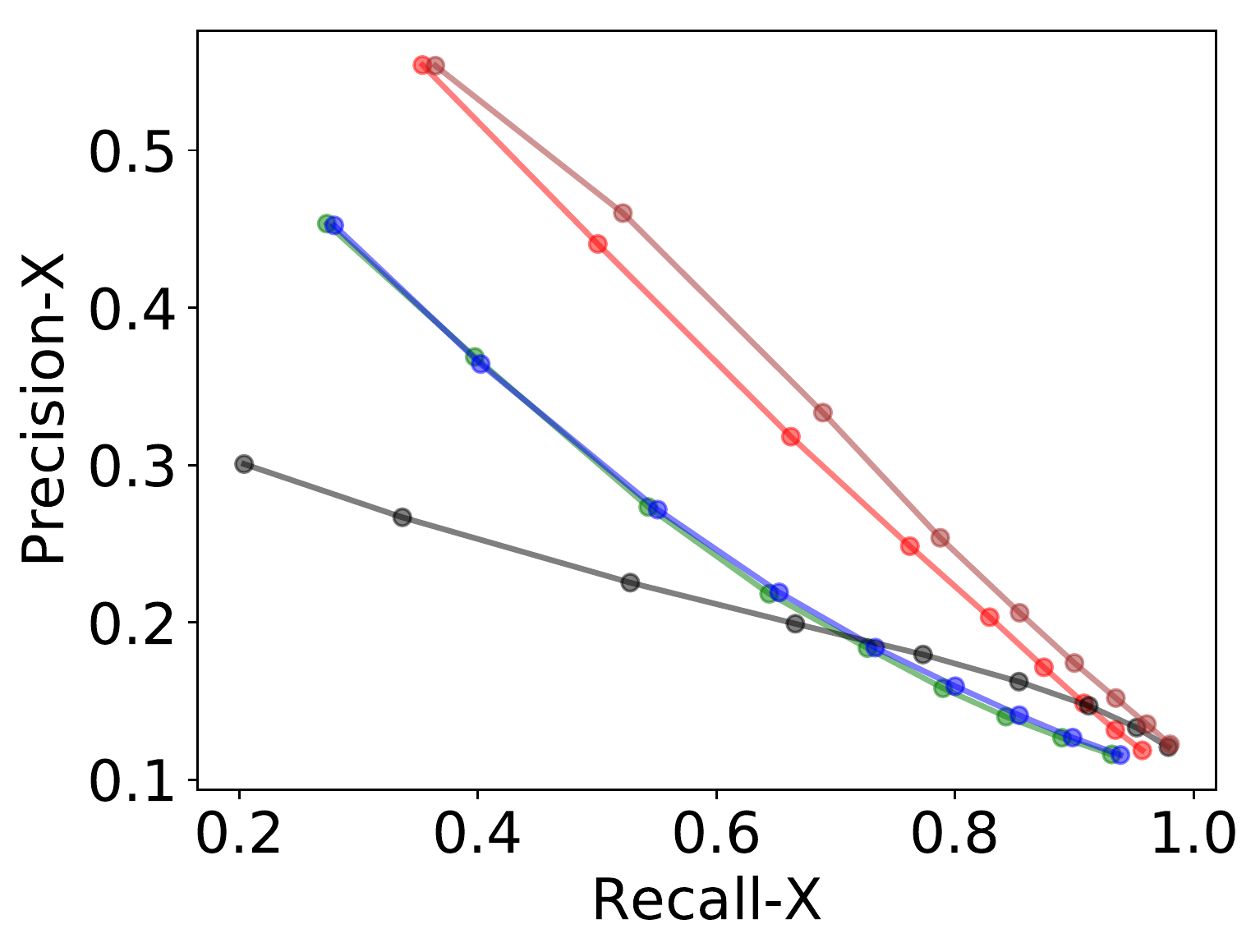}
            \vspace{-0.3cm}
            \end{minipage} 
            \vspace{-0.2cm}
            \subcaption{Precision-Recall curve for XY ($\text{Precision}_X(\epsilon)$ vs. $\text{Recall}_X(\epsilon)$)}
        \end{minipage}    
        \begin{minipage}{\textwidth}
            \begin{minipage}{0.16\textwidth}
            \hspace{0.19\textwidth}
            \end{minipage} 
            \begin{minipage}{0.16\textwidth}
            \includegraphics[width=\linewidth]{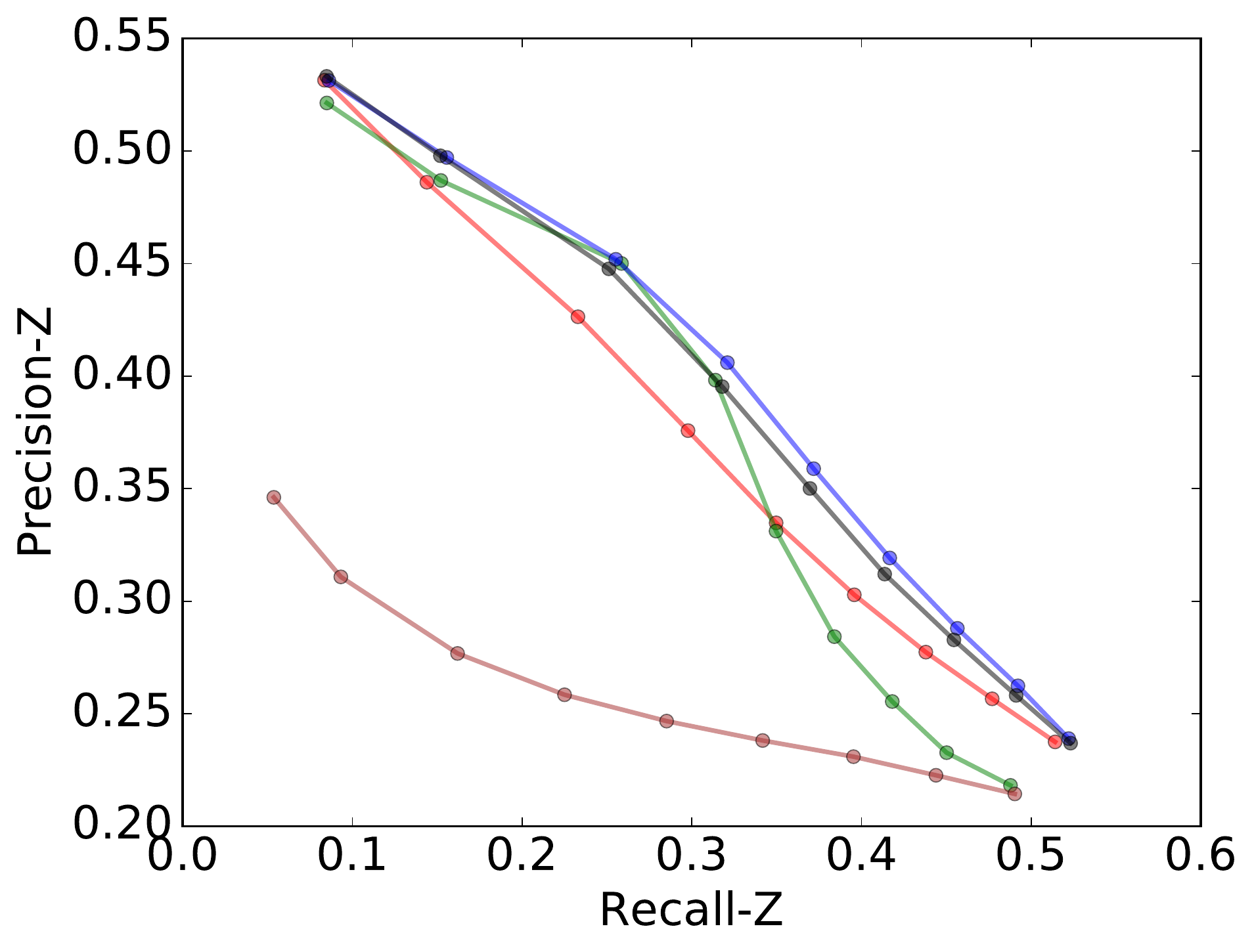}
            \vspace{-0.3cm}
            \end{minipage} 
            \begin{minipage}{0.16\textwidth}
            \includegraphics[width=\linewidth]{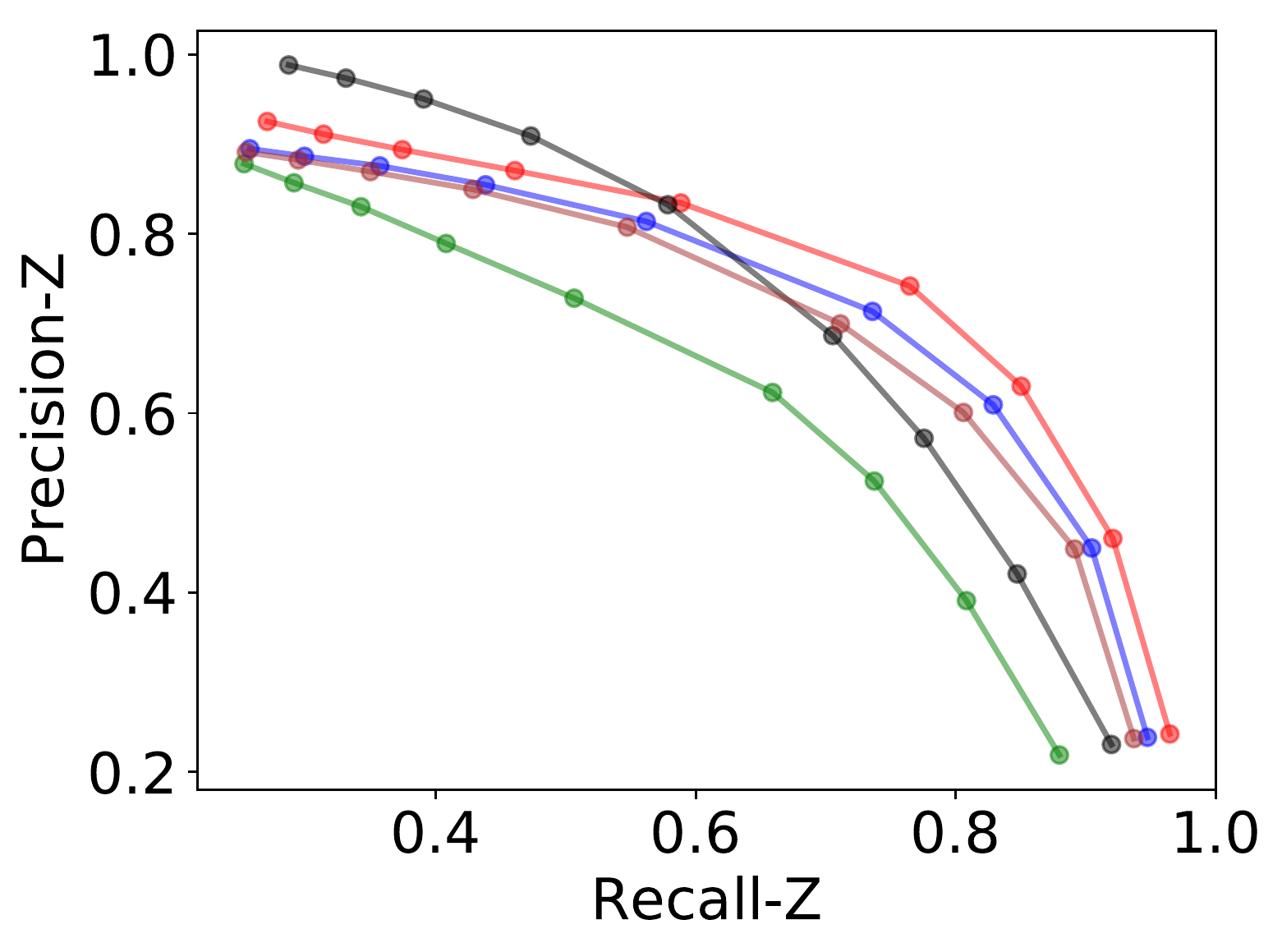}
            \vspace{-0.3cm}
            \end{minipage} 
            \begin{minipage}{0.16\textwidth}
            \includegraphics[width=\linewidth]{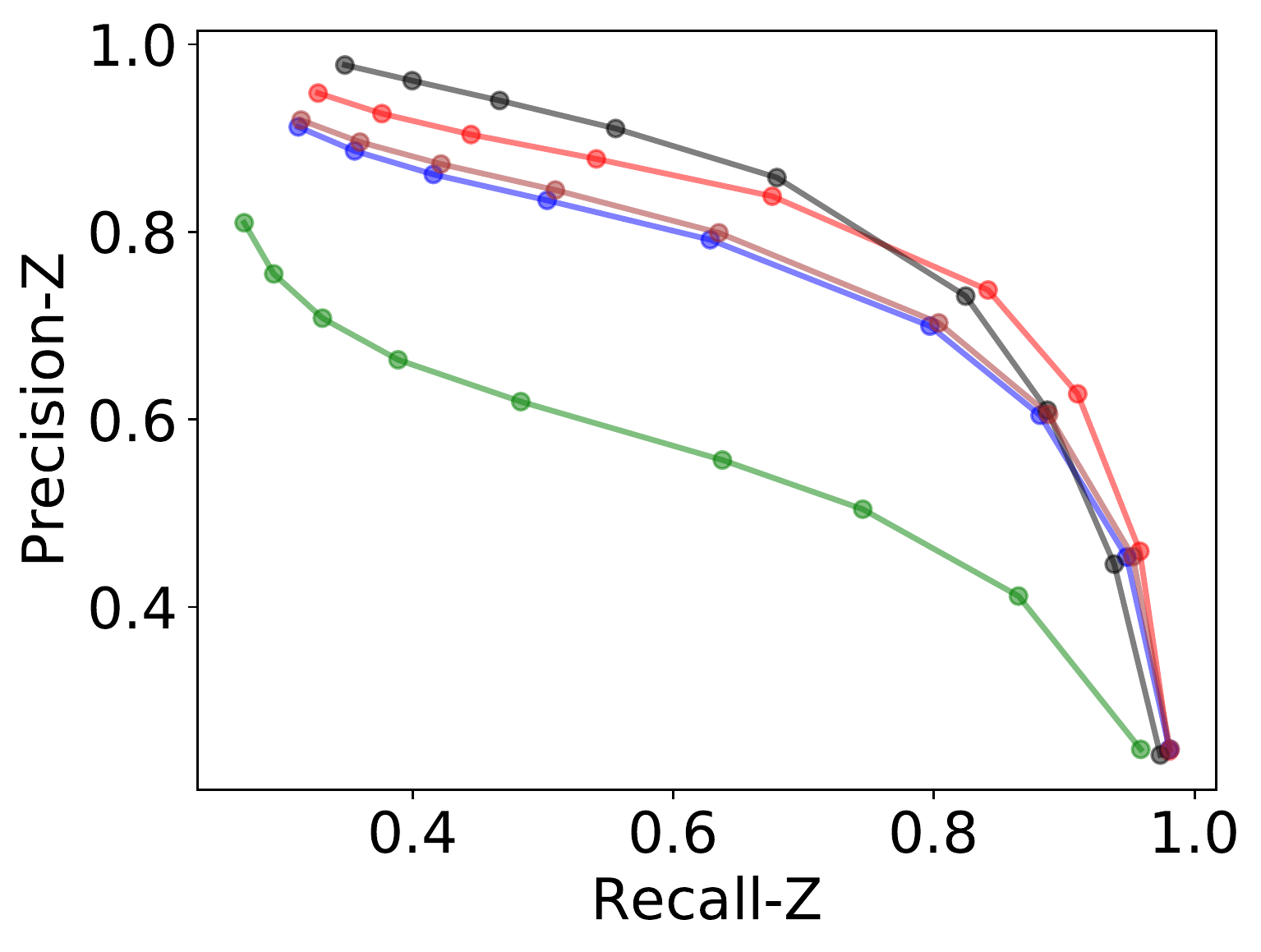}
            \vspace{-0.3cm}
            \end{minipage} 
            \begin{minipage}{0.16\textwidth}
            \includegraphics[width=\linewidth]{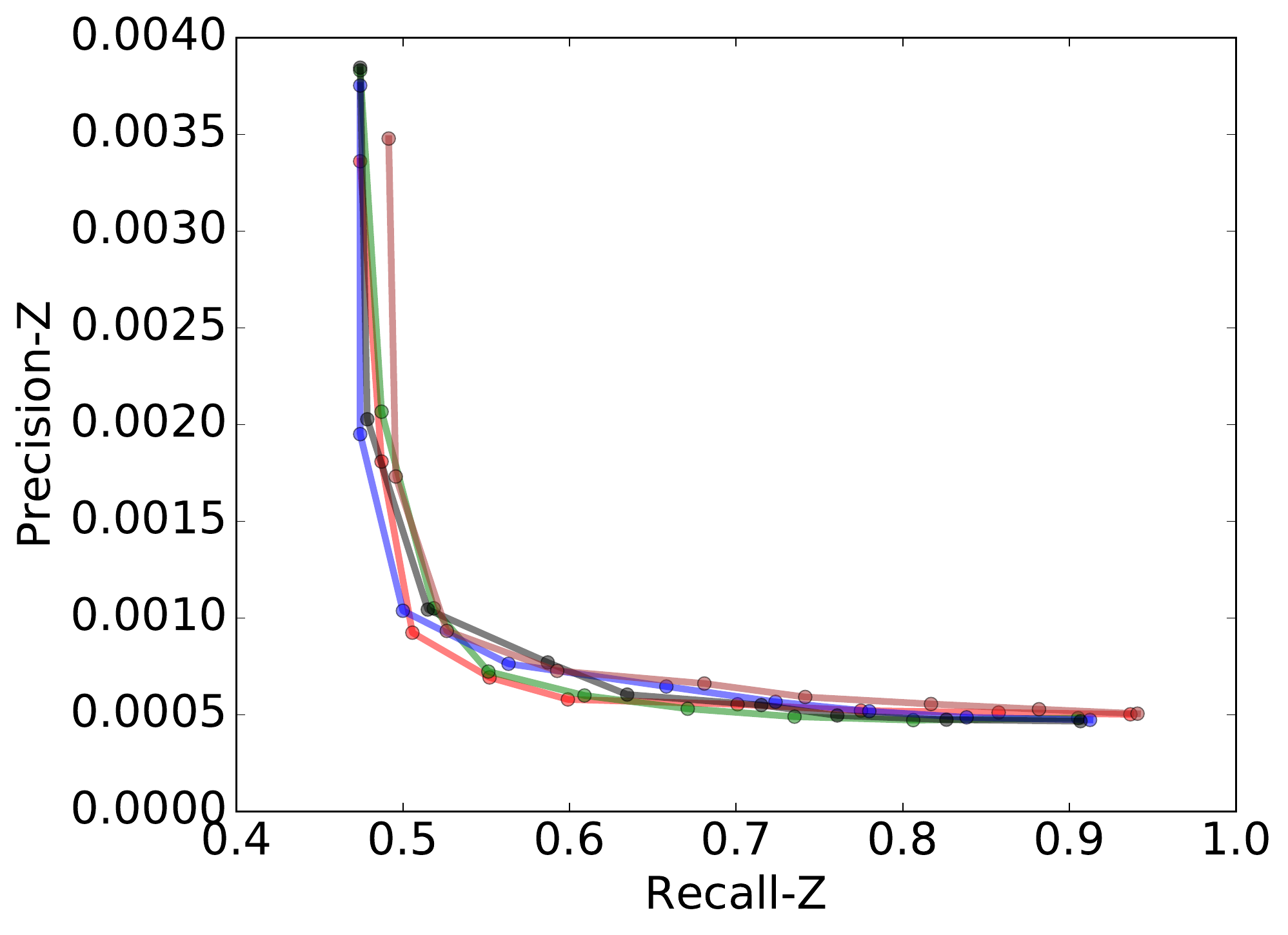}
            \vspace{-0.3cm}
            \end{minipage} 
            \begin{minipage}{0.16\textwidth}
            \includegraphics[width=\linewidth]{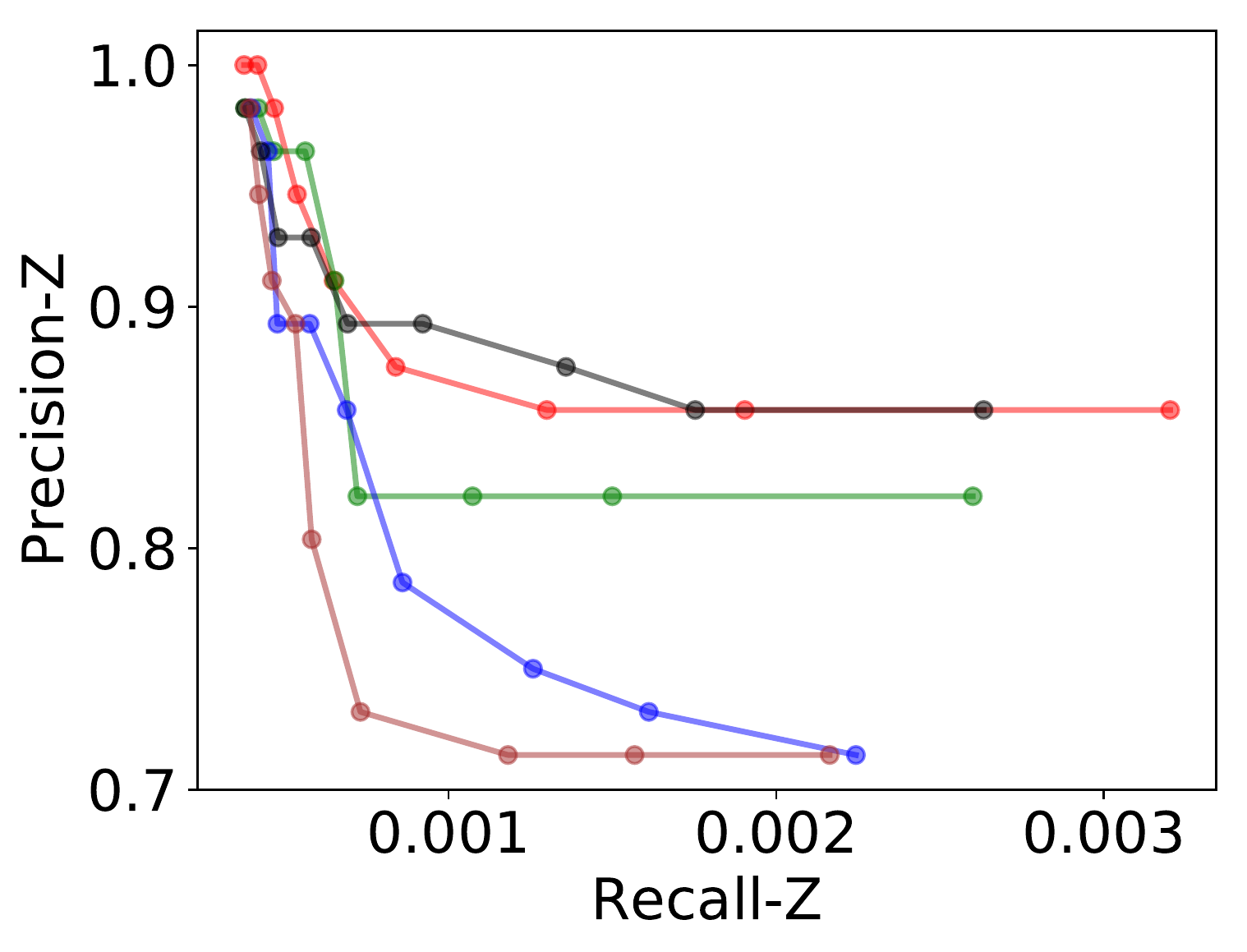}
            \vspace{-0.3cm}
            \end{minipage} 
            \vspace{-0.2cm}
            \subcaption{Precision-Recall curve for ZY ($\text{Precision}_Z(\epsilon)$ vs. $\text{Recall}_Z(\epsilon)$)}
        \end{minipage}     
        \caption{Precision-Recall curves for each of the proposed algorithms on all datasets. 
        Each row corresponds to different quantitative criteria, each column to a different dataset, and each line to a different algorithm.}
        \label{fig:results_diff_metric1}
        \vspace{-0.4cm}
    \end{figure*}

For each of the datasets, we produced Precision$_X(\epsilon)$-Recall$_X(\epsilon)$ and Precision$_Z(\epsilon)$-Recall$_Z(\epsilon)$ curves by varying $\epsilon$, and $\text{NN Precision}(K)$-$\text{FN Precision}(K)$ curves by varying $K$. Results are shown in Figure~\ref{fig:results_diff_metric1} and \ref{fig:results_diff_metric2}. Table~\ref{tab:results1}-\ref{tab:results12} summarizes these results by presenting the algorithm with the highest maximum f-score per criterion. 
For the two manifold datasets, MNIST-Digit-1 and Face, RKL and JS outperformed KL. This reflects the analysis 
(see Proposition 1) that RKL and JS emphasize global structure more than KL, and global structure preservation is more important for manifolds. Conversely, KL performs best on the two cluster datasets, MNIST and GENE. Finally, CH and HL performed best on the hierarchical dataset, News (cf.\ \citealp{bow_chi_sq}).

To better understand the relative strengths of KL and RKL, we qualitatively compared the embeddings resulting from interpolating between them: 
\begin{align}
\alpha KL\text{-SNE} + (1-\alpha) RKL\text{-SNE}
\end{align}
for $\alpha = 0, 0.05, 0.1, 0.5, 1.0$ ($\alpha=.5$ corresponds to JS).
Figure~\ref{fig:syn_embeddings} presents the embedding results for two
synthetic datasets: the Swiss Roll which is a continuous manifold, and three
Gaussian clusters, for a range of perplexity and $\alpha$ values.
%
We observe that RKL worked better for manifolds
with low perplexity while KL worked better clusters with larger perplexity (as
predicted in Section \ref{sec:fSNE}. In
addition, KL broke up the continuous Swiss Roll manifold into disjoint pieces,
which produces smoother embeddings compare to KL-SNE under low
perplexity. Finally, we did not see a continuous gradient in embedding results
as we changed $\alpha$. Instead, even for $\alpha = 0.1$, the Swiss Roll
embedding was more similar to the discontinuous KL embedding. For this
dataset, the embedding produced by JS was more similar to that produced by KL
than RKL. For the three Gaussian dataset, all algorithms separated the three
clusters, however KL and JS correctly formed circular clusters, while smaller
values of $\alpha$ resulted in differently shaped clusters.

    \begin{figure}[t]
            \begin{minipage}{0.325\textwidth}
            \includegraphics[width=\linewidth]{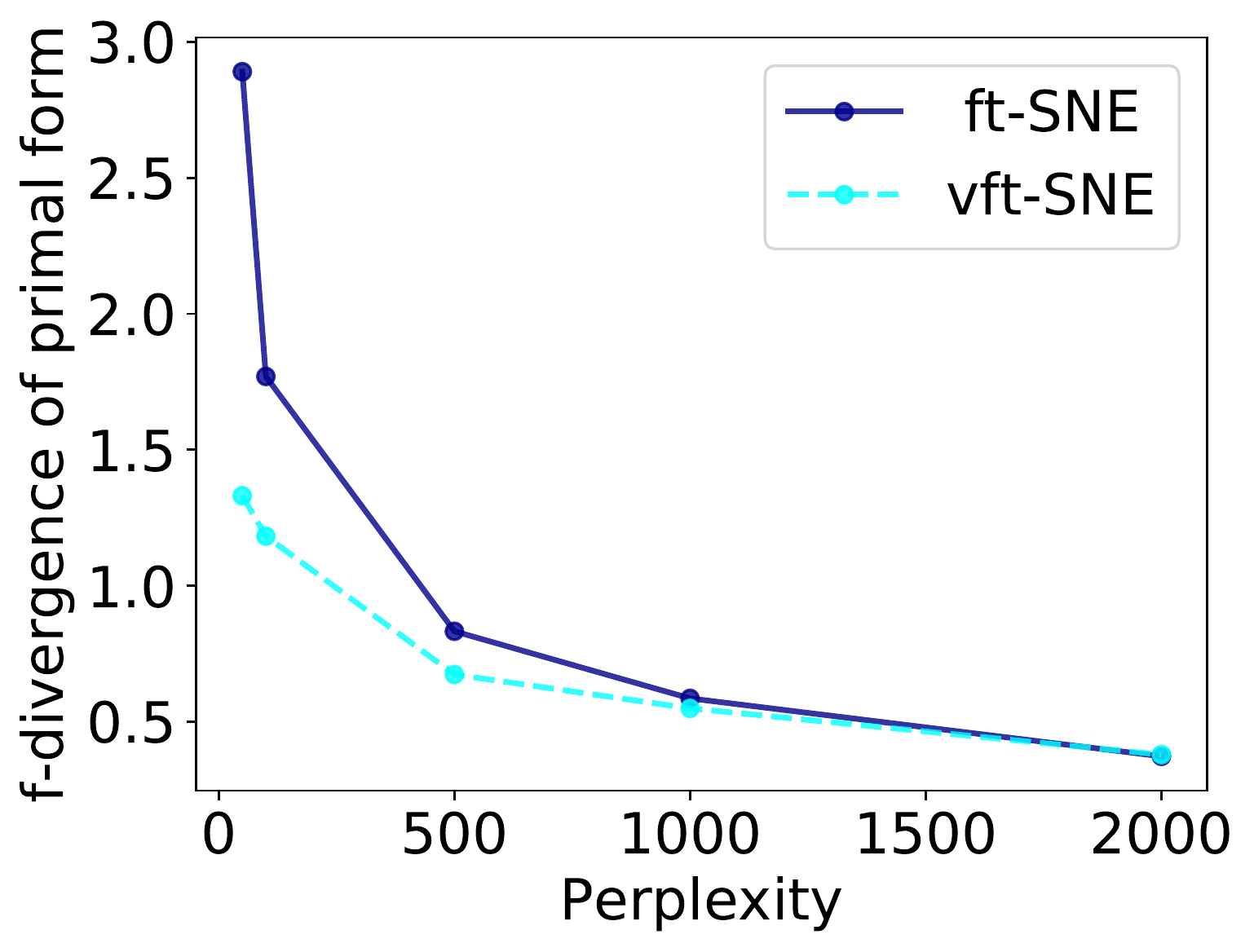}
            \vspace{-0.6cm}
            \subcaption*{KL}
            \end{minipage}
            \begin{minipage}{0.325\textwidth}
            \includegraphics[width=\linewidth]{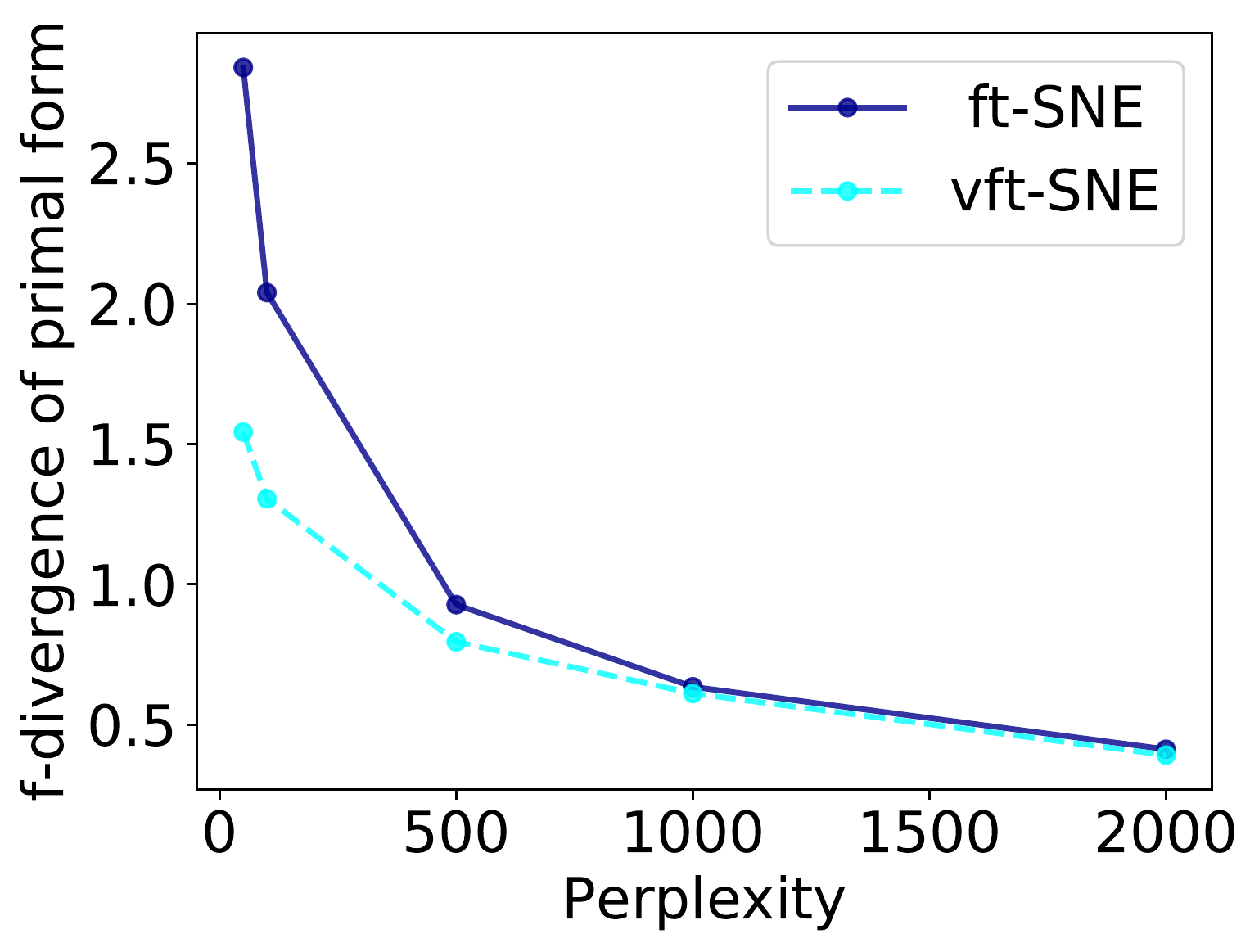}
            \vspace{-0.6cm}
            \subcaption*{JS}
            \end{minipage}            
            \begin{minipage}{0.325\textwidth}
            \includegraphics[width=\linewidth]{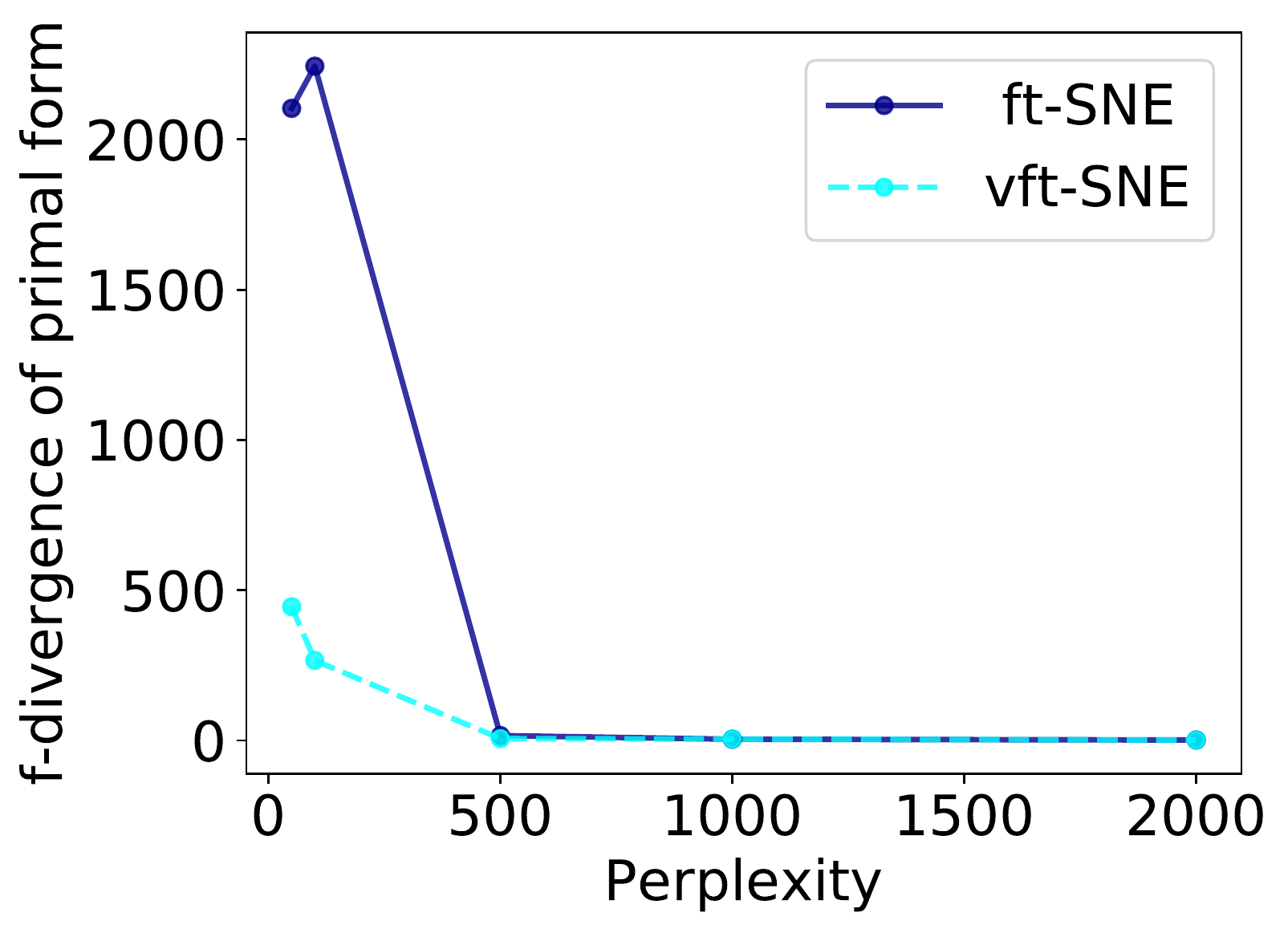}
            \vspace{-0.6cm}
            \subcaption*{CH}
            \end{minipage} 
        \vspace{-0.2cm}
        \caption{Log of the primal $ft$-SNE loss for the $ft$-SNE and $vft$-SNE algorithms for different perplexities on MNIST. 
                 The number of updates were set to J:K=10:10 and two hidden layer (10-20) deep ReLU neural network.}
        \label{fig:opt_perp}
        \vspace{-0.2cm}
    \end{figure}
    \begin{figure*}[t]
        \centering
            \begin{minipage}{0.28\textwidth}
            \includegraphics[width=\linewidth]{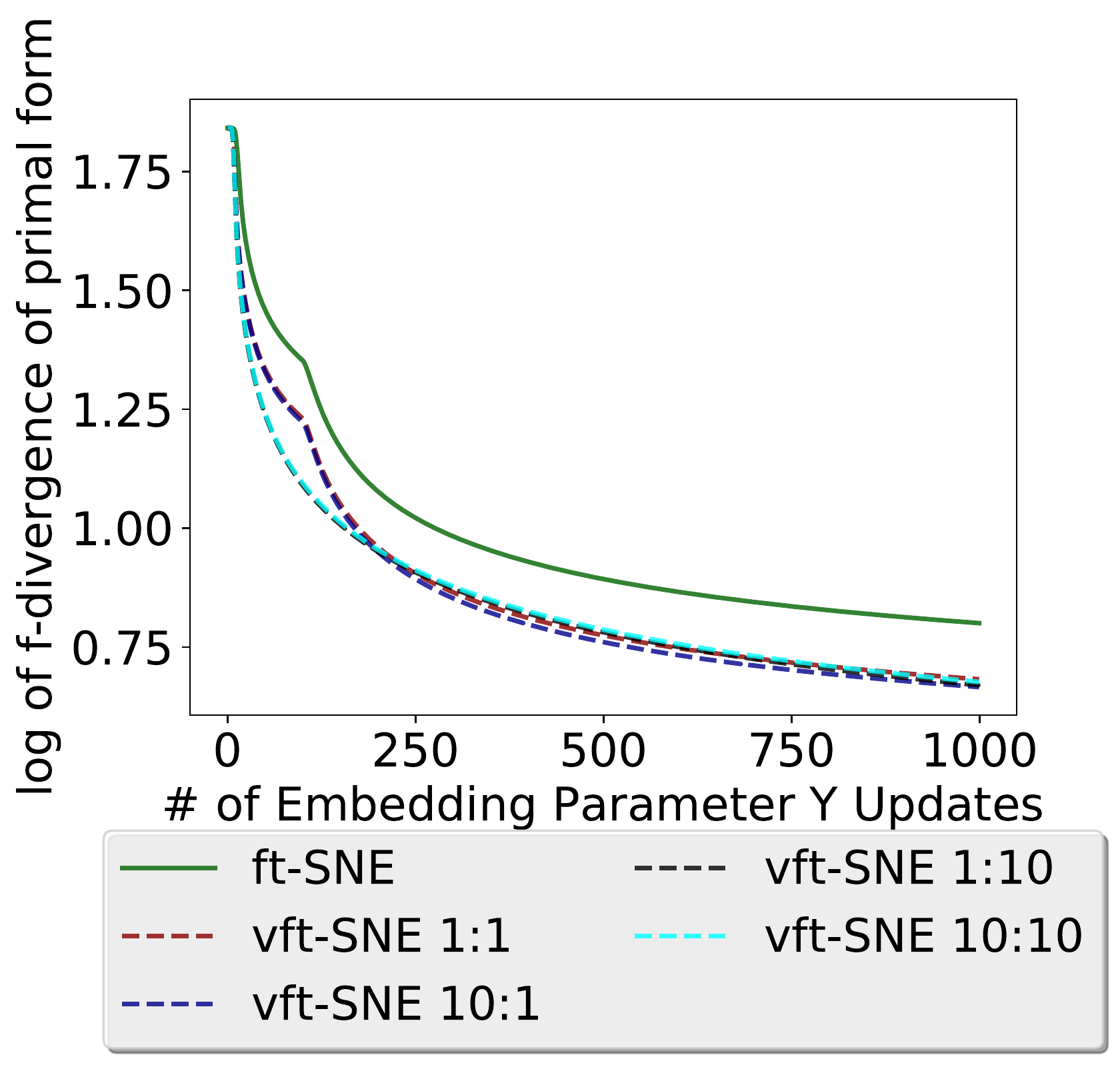}
            \vspace{-0.5cm}
            \subcaption{Number of updates ($J$:$K$)}
            \end{minipage}            
            \begin{minipage}{0.33\textwidth}
            \includegraphics[width=\linewidth]{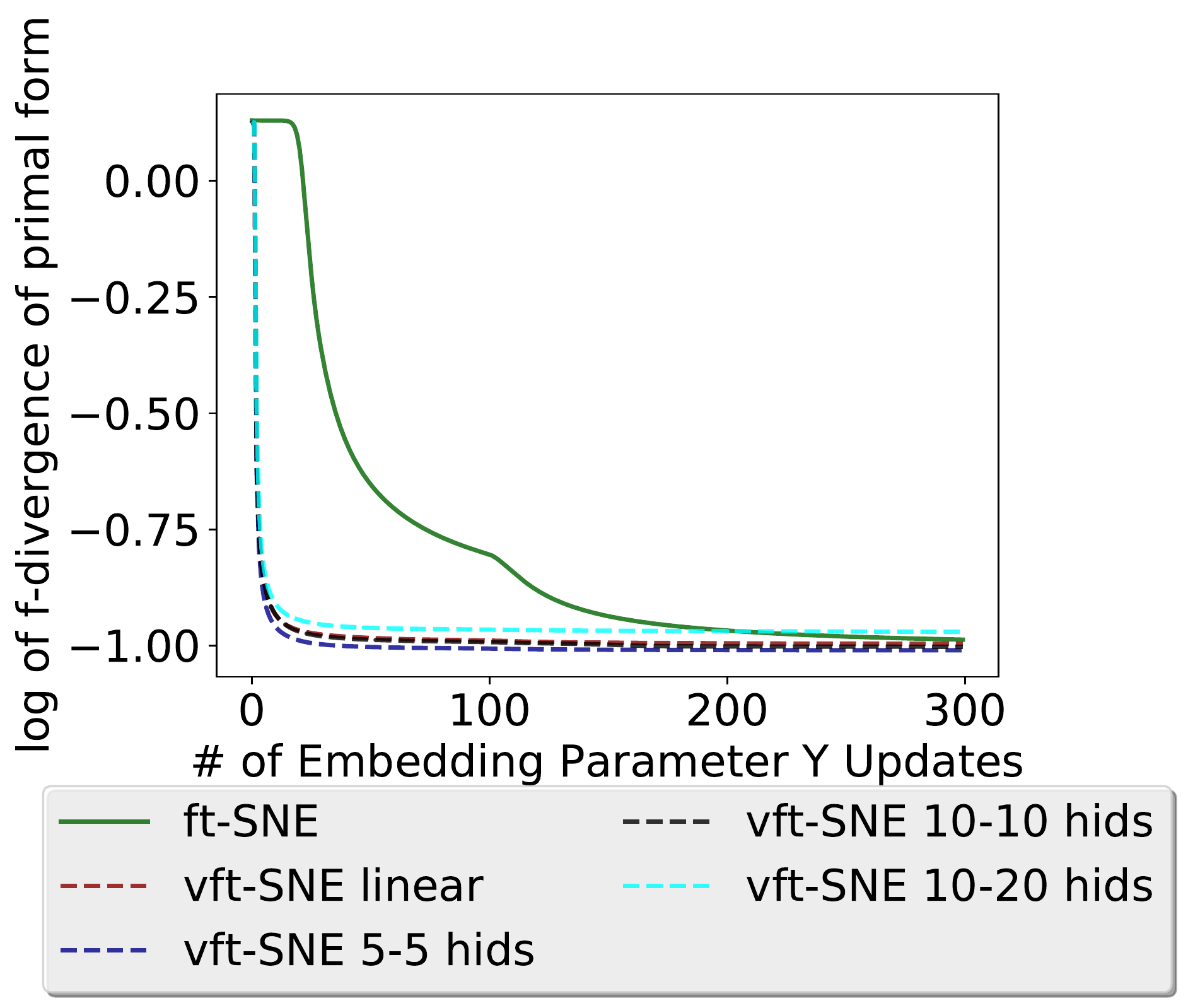}
            \vspace{-0.5cm}
            \subcaption{Network depth}
            \end{minipage}
            \begin{minipage}{0.35\textwidth}
            \includegraphics[width=\linewidth]{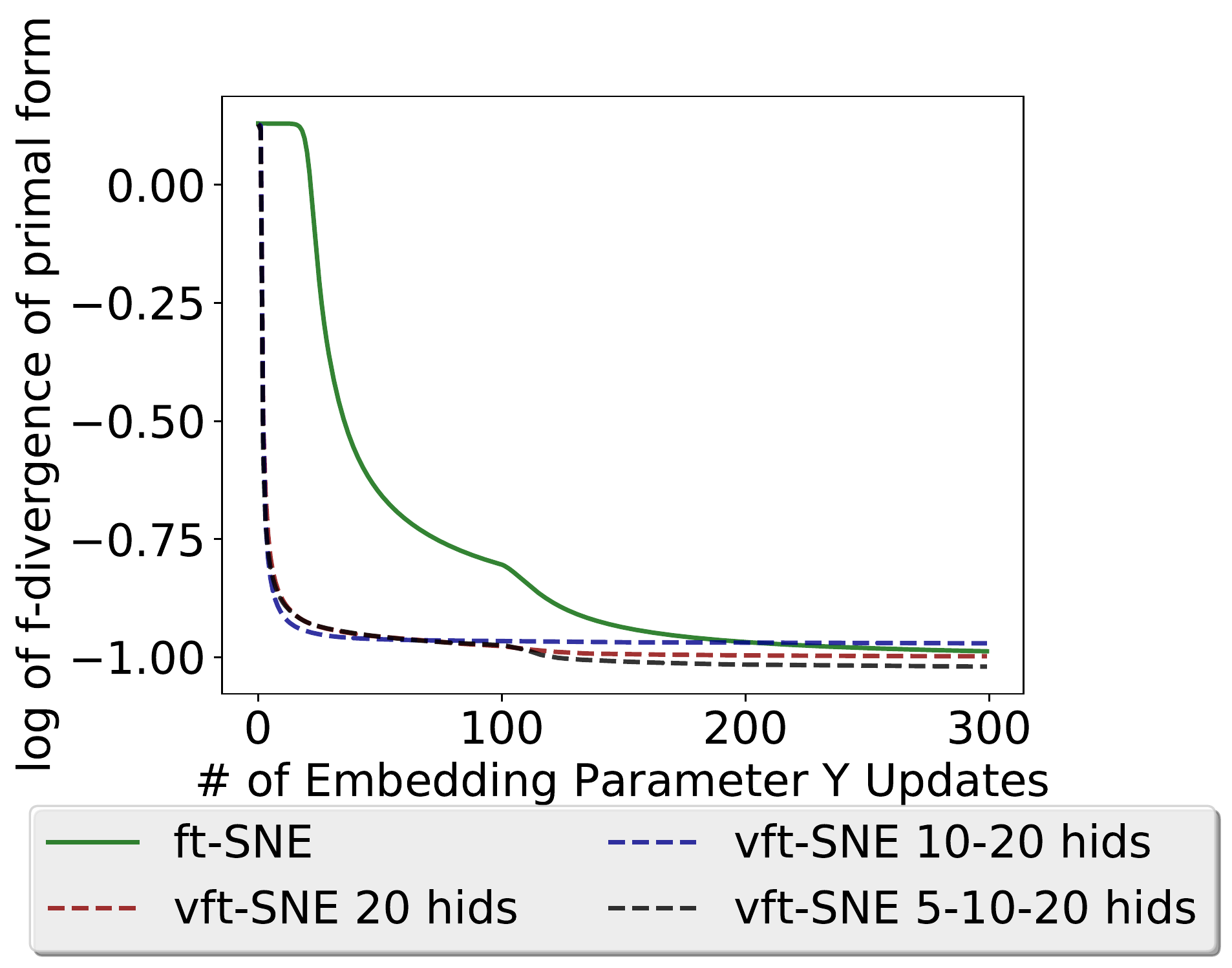}
            \vspace{-0.5cm}
            \subcaption{Network width}
            \end{minipage}
        \caption{Comparison of log $t$-SNE criterion for different parameter choices. All plots show results for the KL divergence on the MNIST dataset (perplexity 2,000), results for other divergences and datasets are in S.M. (a) Different numbers of updates to the discriminator and embedding weights with fixed network architecture (2 layers of 10 and 20 hidden units). (b) Different network widths, with fixed $J$:$K$ = 10:10. (c) Different network depths (widths specified in legend), $J$:$K$ = 10:10.}
        \label{fig:opt_fig}
        \vspace{-0.4cm}
    \end{figure*}
 
\subsection{Optimization of the primal vs.\ variational forms}

In this section, we quantitatively and qualitatively compared the efficacy of optimizing the primal, $ft$-SNE, versus the variational, $vft$-SNE, forms of the criteria. Quantitatively, we compared the primal $ft$-SNE criteria at solutions found using both methods during and after optimization.

Figure~\ref{fig:opt_perp} shows the log primal $ft$-SNE criteria of the final solutions using both optimization methods for different $f$-divergences and different perplexities for MNIST (Supp. Fig.~\ref{fig:SM_perp} shows results for other datasets). We found that for small perplexities $vft$-SNE outperforms $ft$-SNE, while this difference decreases as perplexity increases $ft$-SNE and $vft$-SNE converges to same loss values as the perplexity increases. However, even at perplexity $2000$, $vft$-SNE achieves a slightly lower loss than $ft$-SNE. This is surprising since $vft$-SNE minimizes a lower bound of $ft$-SNE, the criterion we are using for comparison, and suggests that optimizing the primal form using gradient descent can result in bad local minima. 

We next evaluated the performance of the $vft$-SNE algorithm as we vary some of the parameters of the method. Figure~\ref{fig:opt_fig}a compares the results as we vary the number of updates $J$ and $K$ to perform to the discriminator and embedding weights (Algorithm~\ref{algo:vfsne_update_rule}). For the KL divergence, we found that optimizing the variational form performed better for all choices of $J$ and $K$, both in terms of the rate of convergence and the final solution found. For the CH and JS divergences, nearly all choices of $J$ and $K$ resulted in faster optimization (see Supp. Fig.~\ref{fig:JJ_KK}). This plot is in terms of the number of updates, wall clock time is shown in Table~\ref{tab:speed_results} (S.M.). 

Figure~\ref{fig:opt_fig}b and~\ref{fig:opt_fig}c compares the results as we change the architecture of the discriminator. We experimented with a linear classifier and neural networks with 1-3 hidden layers of varying numbers of hidden units (network width). Figure~\ref{fig:opt_fig}a compares results results as we vary network width (architecture shown in Supp. Fig.~\ref{fig:disc_arch_size}) and Figure~\ref{fig:opt_fig}b compares results as we change network depth (architecture shown in Supp. Fig.~\ref{fig:disc_arch_depth}). We observed that the performance was largely consistent as we changed network architecture. The results for JS and CH-SNE are shown in Supp. Fig.~\ref{fig:arch}a and~\ref{fig:arch}b.



\section{Discussion and Related Work}
\label{sec:discussion}
Other divergences for $t$-SNE optimization have been explored previously.
Perhaps the first detailed study was done by 
\citet{tsne_div_bunte} where they explored divergences from various families
(Gamma- Bregman- and $f$-divergences) and their corresponding visualizations on
some image processing datasets.
\citet{div_improves_tsne} and 
\citet{alpha_beta_div_improves_micro_macro_struct} recently discussed how
different divergences can be used to find micro and macro relationships in
data. 
An interesting line of work by \citet{multiscale_tsne} and
\citet{trustworthy_dim_redux} highlights the issues of trustworthy structure
discovery and multi-scale visualizations to find local and global structures.

The work by \citet{Amid2015} is closely related where they study $\alpha$-divergences
from an informational retrieval perspective. Our work extends it to the general class of $f$-divergences and 
explores the relationships between data structure and the type of divergence used. 

It is worth emphasizing that no previous study makes an explicit connection
between the choice of divergence and the type of structure discovery. Our work makes this explicit
and should help a practitioner gain better insights about their data in the data exploration phase.
 Our work
goes a step further and attempts to
ameliorate the issues non-convex objective function in the $ft$-SNE criterion.
By studying the variational dual form, we can achieve 
better quality (locally optimal) solutions, which would be extremely beneficial to the practitioner.

\bibliographystyle{plainnat}
\bibliography{main}

\newpage

\appendix

\begin{appendices}

    \onecolumn
    \section{Precision and Recall}
    \begin{proof}
    \noindent For KL: 
    \begin{align*}
        J_{KL}(x_i) &= \sum_j  p_{ji} \left(\log\frac{p_{ji}}{q_{ji}}\right) \nonumber\\
                    &= \sum_{\substack{j\neq i,\\p_{ji}=a_i,\\q_{ji}=c_i}} a_i\log\left(\frac{a_i}{c_i}\right)\
                          + \sum_{\substack{j\neq i,\\p_{ji}=a_i,\\q_{ji}=d_i}} a_i\log\left(\frac{a_i}{d_i}\right) \
                          + \sum_{\substack{j\neq i,\\p_{ji}=b_i,\\q_{ji}=c_i}} b_i\log\left(\frac{b_i}{c_i}\right) \
                          + \sum_{\substack{j\neq i,\\p_{ji}=b_i,\\q_{ji}=d_i}} b_i\log\left(\frac{b_i}{d_i}\right) \nonumber\\
                    &= n^i_{TP}a_i\log\left(\frac{a_i}{c_i}\right) + n^i_{FN} a_i\log\left(\frac{a_i}{d_i}\right)\
                       n^i_{FP}b_i\log\left(\frac{b_i}{c_i}\right) + n^i_{TN} b_i\log\left(\frac{b_i}{d_i}\right) \nonumber
    \end{align*}
    where $n^i_{TP}, n^i_{FN}, n^i_{FP}, n^i_{TN}$ are number of true positives, false negatives (missed points), false positives, and true negatives respectively for point $x_i$.
    Given that $\delta$ is close to $0$, then the coefficient of $n^i_{FN}$ and $n^i_{FP}$ dominates the other terms,
    \begin{align*}
        J_{KL} &= n^i_{FN} a_i\log\left(\frac{a_i}{d_i}\right) + n^i_{FP} b_i\log\left(\frac{b_i}{c_i}\right) + O(\delta)\\
                &= n^i_{FN} \frac{1-\delta}{r_i}\log\left(\frac{1-\delta}{\delta}\frac{m-k_i-1}{r_i}\right) 
                   + n^i_{FP} \frac{\delta}{m - r_i -1 }\log\left(\frac{\delta}{1-\delta}\frac{k_i}{m-r_i-1}\right) + O(\delta).
    \end{align*}
        Again, the $\log \frac{1-\delta}{\delta}$ dominates the other logarithmic terms $\bigg(\log \left(\frac{m-r_i-1}{k_i}\right)$ and $\log \left(\frac{m-k_i-1}{r_i}\right) \bigg)$, so we have
    \begin{align*}
        J_{KL} &= \left(\frac{n^i_{FN}}{r_i} (1-\delta) - \frac{n^i_{FP}}{m - r_i -1} \delta \right) \log \left(\frac{1-\delta}{\delta}\right) + O(\delta) \nonumber\\
                &= \frac{n^i_{FN}}{r_i} C_0 = (1-\text{Recall}(i))C_0 + O(\delta)
    \end{align*}
    where $C_0 = \log \left(\frac{1-\delta}{\delta}\right)$.\\\\
 
    \noindent For Reverse KL: 
    \begin{align*}
        J_{RKL}(x_i) &= - \sum_j  q_{ji} \left(\log\frac{p_{ji}}{q_{ji}}\right) \nonumber\\
                    &= - \sum_{\substack{j\neq i,\\p_{ji}=a_i,\\q_{ji}=c_i}} c_i\log\left(\frac{a_i}{c_i}\right)\
                            - \sum_{\substack{j\neq i,\\p_{ji}=a_i,\\q_{ji}=d_i}} d_i\log\left(\frac{a_i}{d_i}\right) \
                            - \sum_{\substack{j\neq i,\\p_{ji}=b_i,\\q_{ji}=c_i}} c_i\log\left(\frac{b_i}{c_i}\right) \
                            - \sum_{\substack{j\neq i,\\p_{ji}=b_i,\\q_{ji}=d_i}} d_i\log\left(\frac{b_i}{d_i}\right) \nonumber\\
                    &= - n^i_{TP}c_i\log\left(\frac{a_i}{c_i}\right) - n^i_{FN} d_i\log\left(\frac{a_i}{d_i}\right)\
                        - n^i_{FP} c_i\log\left(\frac{b_i}{c_i}\right) - n^i_{TN} d_i\log\left(\frac{b_i}{d_i}\right) \nonumber
    \end{align*}
    where $n^i_{TP}, n^i_{FN}, n^i_{FP}, n^i_{TN}$ are number of true positives, false negatives (missed points), false positives, and true negatives respectively for point $x_i$.
    Given that $\delta$ is close to $0$, then the coefficient of $n^i_{FN}$ and $n^i_{FP}$ dominates the other terms,
    \begin{align*}
        J_{RKL} &= - n^i_{FN} d_i\log\left(\frac{a_i}{d_i}\right) - n^i_{FP} c_i\log\left(\frac{b_i}{c_i}\right) + O(\delta)\\
                &= - n^i_{FN} \frac{\delta}{m - k_i -1 }\log\left(\frac{1-\delta}{\delta}\frac{m-k_i-1}{r_i}\right) 
                    - n^i_{FP} \frac{1-\delta}{k_i}\log\left(\frac{\delta}{1-\delta}\frac{k_i}{m-r_i-1}\right) + O(\delta).
    \end{align*}
        Again, the $\log \frac{1-\delta}{\delta}$ dominates the other logarithmic terms $\bigg(\log \left(\frac{m-r_i-1}{k_i}\right)$ and $\log \left(\frac{m-k_i-1}{r_i}\right) \bigg)$, so we have
    \begin{align*}
        J_{RKL} &= \left(\frac{n^i_{FP}}{k_i}(1-\delta) - \frac{n^i_{FN}}{m - k_i -1} \delta \right) \log \left(\frac{1-\delta}{\delta}\right) + O(\delta) \nonumber\\
                &= \frac{n^i_{FP}}{r_i} C_0 = (1-\text{Precision}(i))C_0 + O(\delta)
    \end{align*}
    where $C_0 = \log \left(\frac{1-\delta}{\delta}\right)$.\\\\
   
    \noindent For Jensen-Shanon :
    \begin{align*}   
        J_{JS}(x_i) &= \frac{1}{2}\left(\sum_j p_{ij} \log\frac{2p_{ij}}{p_{ij}+q_{ij}} + \sum_j q_{ij} \log\frac{2q_{ij}}{p_{ij}+q_{ij}}\right)\\
                    &= -\frac{1}{2}\left(\sum_j p_{ij} \log\frac{p_{ij}+q_{ij}}{p_{ij}} + \sum_j q_{ij} \log\frac{p_{ij}+q_{ij}}{q_{ij}} + \log 4 \right)\\
                    &= -\frac{1}{2}\left(\sum_j p_{ij} \log\frac{q_{ij}}{p_{ij}} + \sum_j q_{ij} \log\frac{p_{ij}}{q_{ij}} + \log 4\right)\\
                    &= \frac{1}{2}\left(J_{KL}(x_i) + J_{RKL}(x_i) + \log 4\right)
    \end{align*}          

    \noindent For Chi-Square distance:
    \begin{align*}
        J_{\text{CS}}(x_i) &= \sum_j  q_{ji} \left(\frac{p_{ji}}{q_{ji}}\right)^2 \nonumber\\
                    &= \sum_{\substack{j\neq i,\\p_{ji}=a_i,\\q_{ji}=c_i}} c_i\left(\frac{a_i}{c_i}-1\right)^2 \
                            + \sum_{\substack{j\neq i,\\p_{ji}=a_i,\\q_{ji}=d_i}} d_i\left(\frac{a_i}{d_i}-1\right)^2 \
                            + \sum_{\substack{j\neq i,\\p_{ji}=b_i,\\q_{ji}=c_i}} c_i\left(\frac{b_i}{c_i}-1\right)^2 \
                            + \sum_{\substack{j\neq i,\\p_{ji}=b_i,\\q_{ji}=d_i}} d_i\left(\frac{b_i}{d_i}-1\right)^2 \nonumber\\
                    &= n^i_{TP}c_i\left(\frac{a_i}{c_i}-1\right)^2 + n^i_{FN} d_i\left(\frac{a_i}{d_i}-1\right)^2\
                        + n^i_{FP} c_i\left(\frac{b_i}{c_i}-1\right)^2 + n^i_{TN} d_i\left(\frac{b_i}{d_i}-1\right)^2. \nonumber
    \end{align*}
    Given that $\delta$ is near $0$, then the last term $\left(n^i_{TN} d_i\left(\frac{b_i}{d_i}-1\right)^2\right)$ gets eliminated. So, we have
    \begin{align*}
        J_{\text{CS}}(x_i) &= n^i_{TP}c_i\left(\frac{a_i}{c_i}-1\right)^2 + n^i_{FN} d_i\left(\frac{a_i}{d_i}-1\right)^2 + n^i_{FP} c_i\left(\frac{b_i}{c_i}-1\right)^2 + O(\delta) \nonumber\\
                    &=  n^i_{TP} \frac{1-\delta}{k_i} \left(\frac{r_i}{k_i} -1\right)^2\nonumber\\
                       &\qquad \qquad
                       +n^i_{FN} \frac{\delta}{m-k_i-1} \left( \frac{1-\delta}{\delta}\frac{m-k_i-1}{r_i} -1 \right)^2
                       +n^i_{FP} \frac{1-\delta}{k_i} \left( \frac{\delta}{1-\delta} \frac{k_i}{m-r_i-1}-1\right)^2 + O(\delta)\nonumber\\
                    &= \frac{n^i_{TP}}{k_i} (1-\delta) C_0 
                        + \frac{n^i_{FN}}{r_i} (1-\delta)C_1
                        + \frac{n^i_{FP}}{k_i} (1-\delta)  + O(\delta)\\
                    &= \text{Precision}(i) C_0+(1-\text{Precision}(i))+(1-\text{Recall}(i))C_1. 
    \end{align*}
    where $C_0 = (\frac{r_i}{k_i} - 1)^2$ and $C_1=\left(\frac{1-\delta}{\delta} \frac{m-k_i-1}{r_i}- 2\right)$.

    The proof layout is similar for Hellinger distance, except that it emphasize recall and has less strict penalities,
    \begin{align*}
        J_{\text{HL}}(x_i) &= n^i_{TP}c_i\left(\sqrt{\frac{a_i}{c_i}}-1\right)^2 + n^i_{FN} d_i\left(\sqrt{\frac{a_i}{d_i}}-1\right)^2 + n^i_{FP} c_i\left(\sqrt{\frac{b_i}{c_i}}-1\right)^2\nonumber\\ 
                       &=n^i_{TP} \frac{1-\delta}{k_i} \left(\sqrt{\frac{r_i}{k_i}}-1\right)^2\nonumber\\
                       &\qquad
                       +n^i_{FN} \frac{\delta}{m-k_i-1} \left( \sqrt{\frac{1-\delta}{\delta}\frac{m-k_i-1}{r_i}} -1 \right)^2                                                            
                       +n^i_{FP} \frac{1-\delta}{k_i} \left( \sqrt{\frac{\delta}{1-\delta} \frac{k_i}{m-r_i-1}}-1\right)^2 + O(\delta)\nonumber\\
                    &= \frac{n^i_{TP}}{k_i} (1-\delta) C_0 
                        + \frac{n^i_{FN}}{r_i} (1-\delta)C_1
                        + \frac{n^i_{FP}}{k_i} (1-\delta)C_2+ O(\delta)\\
                    &= \text{Precision}(i) C_0+(1-\text{Precision}(i))C_2+(1-\text{Recall}(i))C_1 + O(\delta).
    \end{align*}      
    where $C_{0} = \left(\sqrt{\frac{r_i}{k_i}} - 1\right)^2$, 
    $C_{2} = \left( 1 - 2\sqrt{\frac{r_i\delta}{1-\delta}}\right)$, and
    $C_{1} = \left( 1 - 2\sqrt{\frac{k_i\delta}{1-\delta}}\right)$.

    \end{proof}

    \clearpage

    \section{Variational $ft$-SNE}

    %

    It is standard to relax the optimization of the variational $ft$-SNE objective function in Eq~\ref{eqn:vfsne} by alternatively optimizing the paramters $\phi$ and $y_1, \ldots, y_m$.
    Algorithm~\ref{algo:vfsne_update_rule} alternatively updates $y_1, \ldots, y_m$ and $\phi$.
    The parametric hypothesis class $\bar{\mathcal{H}}$ is parameterized by $\phi$ (for instance, $\phi$ are the weights of the deep neural network).
    Remark that this is not guaranteed to return the same solution as the original minimax objective in Eq~\ref{eqn:vfsne}. 
    Thus it is possible that Algorithm~\ref{algo:vfsne_update_rule} can find a different solution depending on the choice of $J$ and $K$ and under different measures.


    \newpage
    \section{Experimental Supplementary Materials}
    \label{app:exp}

    \begin{figure}[htp]
        \centering
        \begin{minipage}{0.4\textwidth}
            \includegraphics[width=\linewidth]{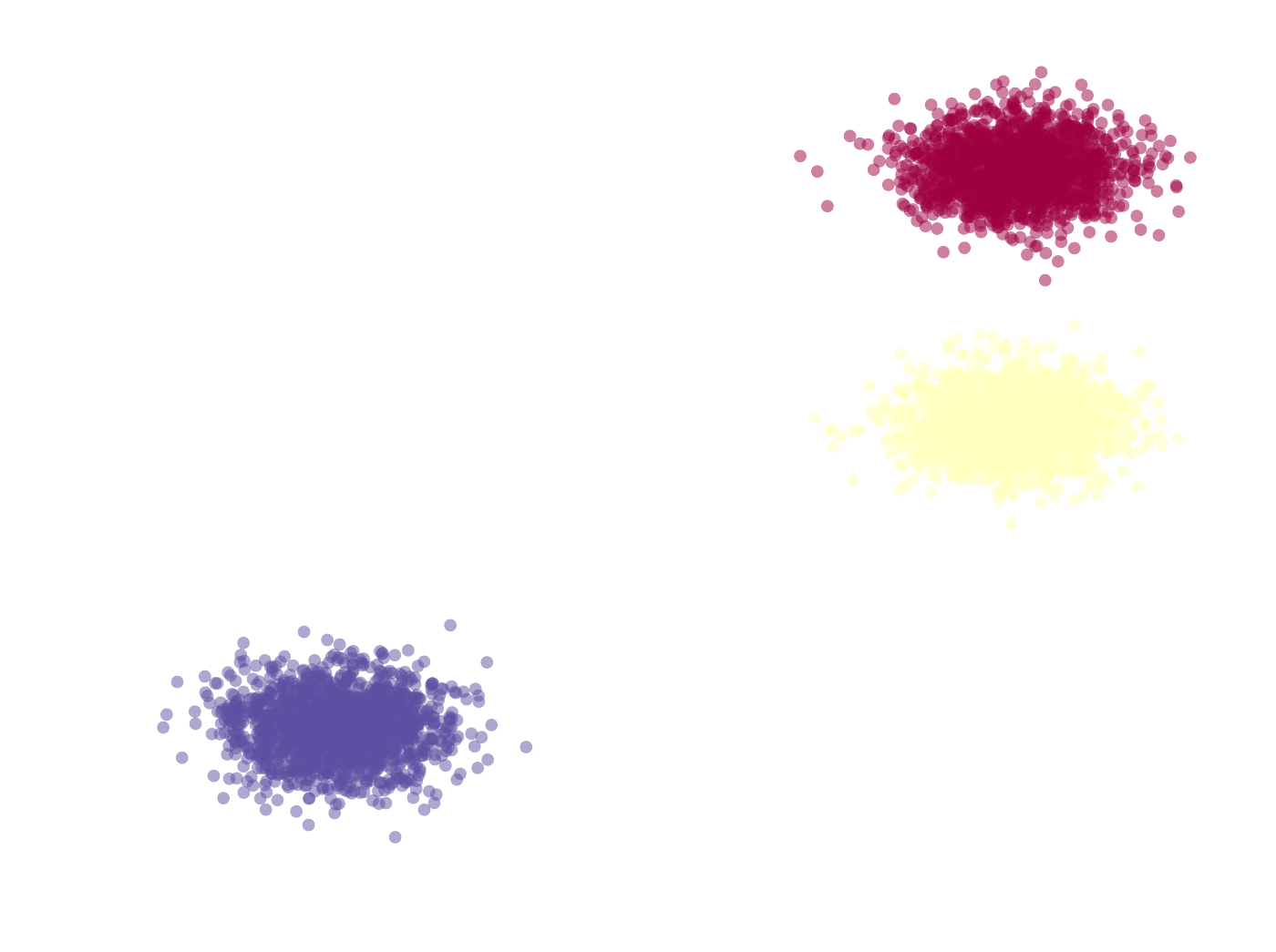}
            \vspace{-0.6cm}
            \label{fig:syn_blobs}
            \subcaption{Three Gaussian clusters}
        \end{minipage}   
        \begin{minipage}{0.4\textwidth}
            \includegraphics[width=\linewidth]{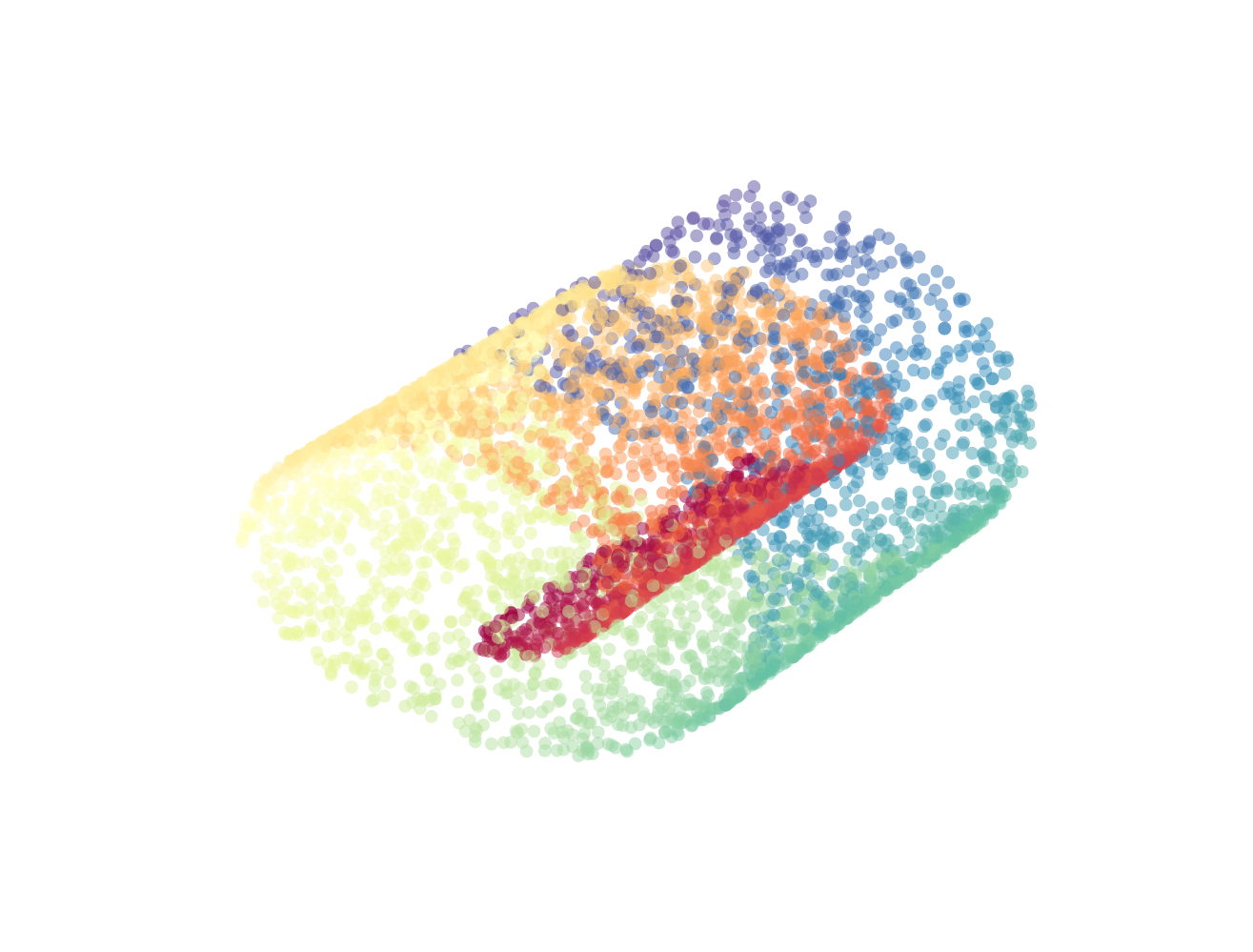}
            \vspace{-0.6cm}
            \label{fig:syn_swiss}
            \subcaption{Swiss Roll}
        \end{minipage}   
        \caption{Synthetic Datasets}
        \label{fig:toy_dataset}
    \end{figure}

    {\em Datasets}. Throughout the experiments, we diversified our datasets by selecting manifold, cluster, and hierical datasets.
    We first experimented with two synthetic datasets, swiss roll and three Gaussian cluter datasets (see S. M. Figure~\ref{fig:toy_dataset}).
    Thence, we conducted the set of experiments on FACE, MNIST, and 20 Newsgroups datasets.
    FACE and MNIST with single digits (MNIST1) fall under manifold datasets and MNIST and 20 Newsgroups fall under cluster and hierical cluster datasets. 
   
    \begin{itemize}[noitemsep,nolistsep]
        \item FACE contains 698 64 x 64 face images. 
            The face varies smoothly with respect to light intensities and poses.
            The face dataset used in isomap paper \cite{Tenenbaum2000}.
            We use face dataset as a manifold dataset.

        \item MNIST consists of 28 x 28 handwritten digits dataset with digits from 0 to 9. 
            MNIST data points were projected down to $30$ features using PCA. 
            We used MNIST as both clustering and manifold datasets. 
            For clustering dataset, we used 6,000 examples of first five digits (MNIST). 
            For manifold dataset, we used 6,000 examples of digits of ones (MNIST1).

        \item 20-NEWSGROUPS consists of 20 different news genres. Among 20 news genres, 
            some of the genres fall under the same abstract categories.
            The 20-newsgroup data are represented using bag of words.
            We used 6,000 new articles that fall under thirteen categories:
            {\em rec.autos, rec.motorcycles, rec.sport.baseball, rec.sport.hockey, 
            sci.crypt, sci.electronics, sci.med, sci.space, soc.religion.christian,
            talk.politics.guns, talk.politics.mideast, talk.politics.misc, and talk.religion.misc.}
            Hence, this dataset corresponds to sparse hierarchical clustering dataset.
    \end{itemize}

    {\em Optimization}. We use gradient decent method with momentum to optimize the $ft$-SNE. 
    We decreased the learning rate and momentum overtime as such
    $\epsilon_{t+1} = \frac{\epsilon_{t}}{(1+\frac{t}{\rho})}$ and 
    $\lambda_{t+1} = \frac{\lambda_{t}}{(1+\frac{t}{\eta})}$
    where $\epsilon_t$ and $\lambda_t$ are learning rate and momentum, and $\rho$ and $\eta$ are learning rate decay and momentum decay parameters.
    $t$-SNE has very tiny gradients in the beginning since all the parameters are intialize in the quite small domain (the initial embeddings are drawn from the Normal distribution with zero mean and $1*e^{-4}$ standard deviation). 
    However, once the embedding parameters spread, the gradients become relatively large compare to early stage. 
    Thus, the learning rate and momentum require to be adjusted appropriately over different stage of optimization.

    \newpage
    \subsection{More Experimental Results : Synthetic Data Experiments}

    \begin{figure}
        \begin{minipage}{\textwidth}
            \begin{minipage}{0.16\textwidth}
            \centering MNIST1
            \end{minipage}   
            \begin{minipage}{0.16\textwidth}
            \centering Face
            \end{minipage}   
            \begin{minipage}{0.16\textwidth}
            \centering MNIST
            \end{minipage}   
            \begin{minipage}{0.16\textwidth}
            \centering GENE
            \end{minipage}   
            \begin{minipage}{0.16\textwidth}
            \centering NEWS
            \end{minipage}   
            \begin{minipage}{0.16\textwidth}
            \centering SBOW
            \end{minipage}   
        \end{minipage}
        \begin{minipage}{\textwidth}
            \begin{minipage}{0.16\textwidth}
            \includegraphics[width=\linewidth]{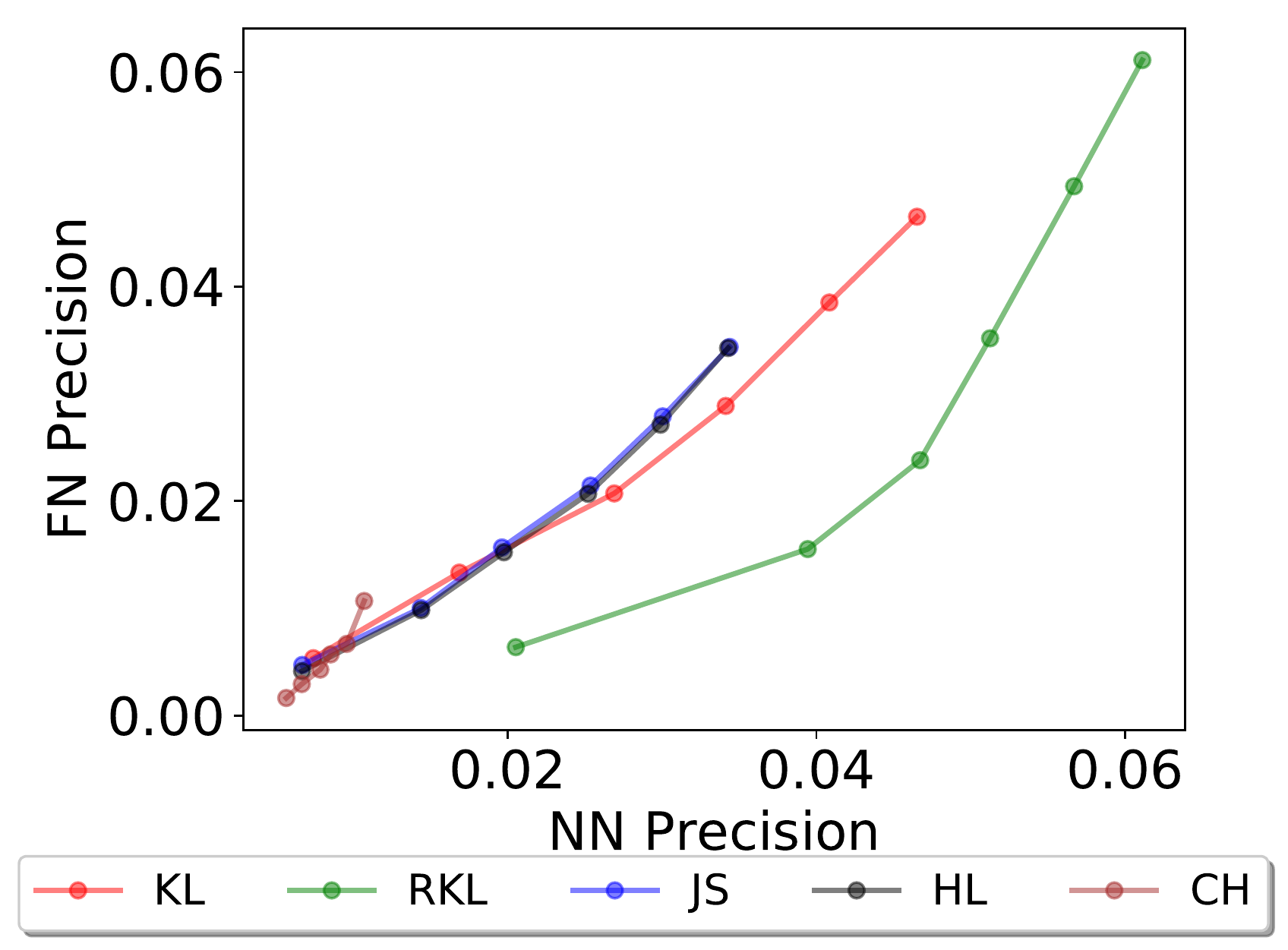}
            \vspace{-0.3cm}
            \end{minipage}   
            \begin{minipage}{0.16\textwidth}
                \includegraphics[width=\linewidth]{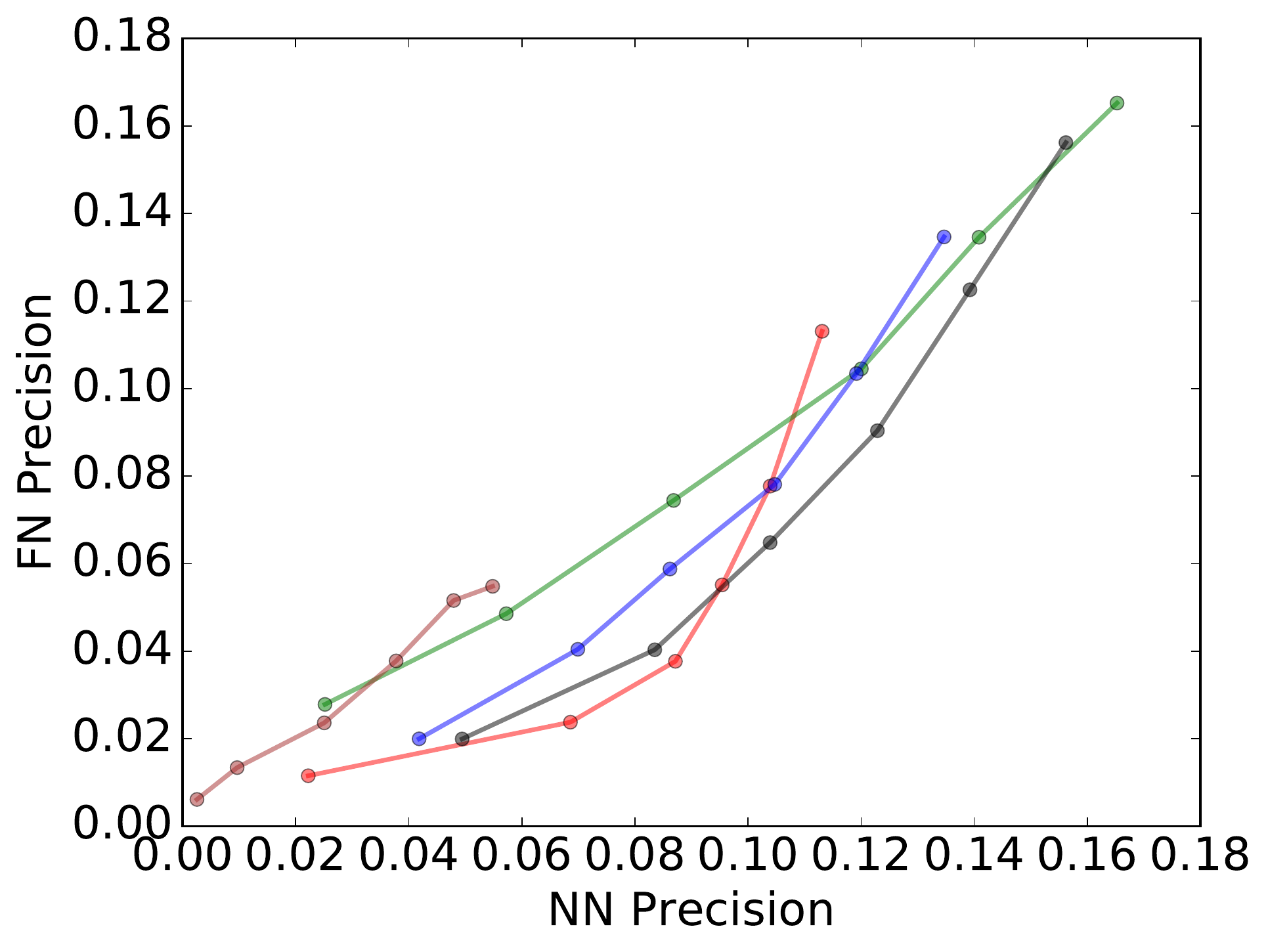}
            \vspace{-0.3cm}
            \end{minipage}   
            \begin{minipage}{0.16\textwidth}
            \includegraphics[width=\linewidth]{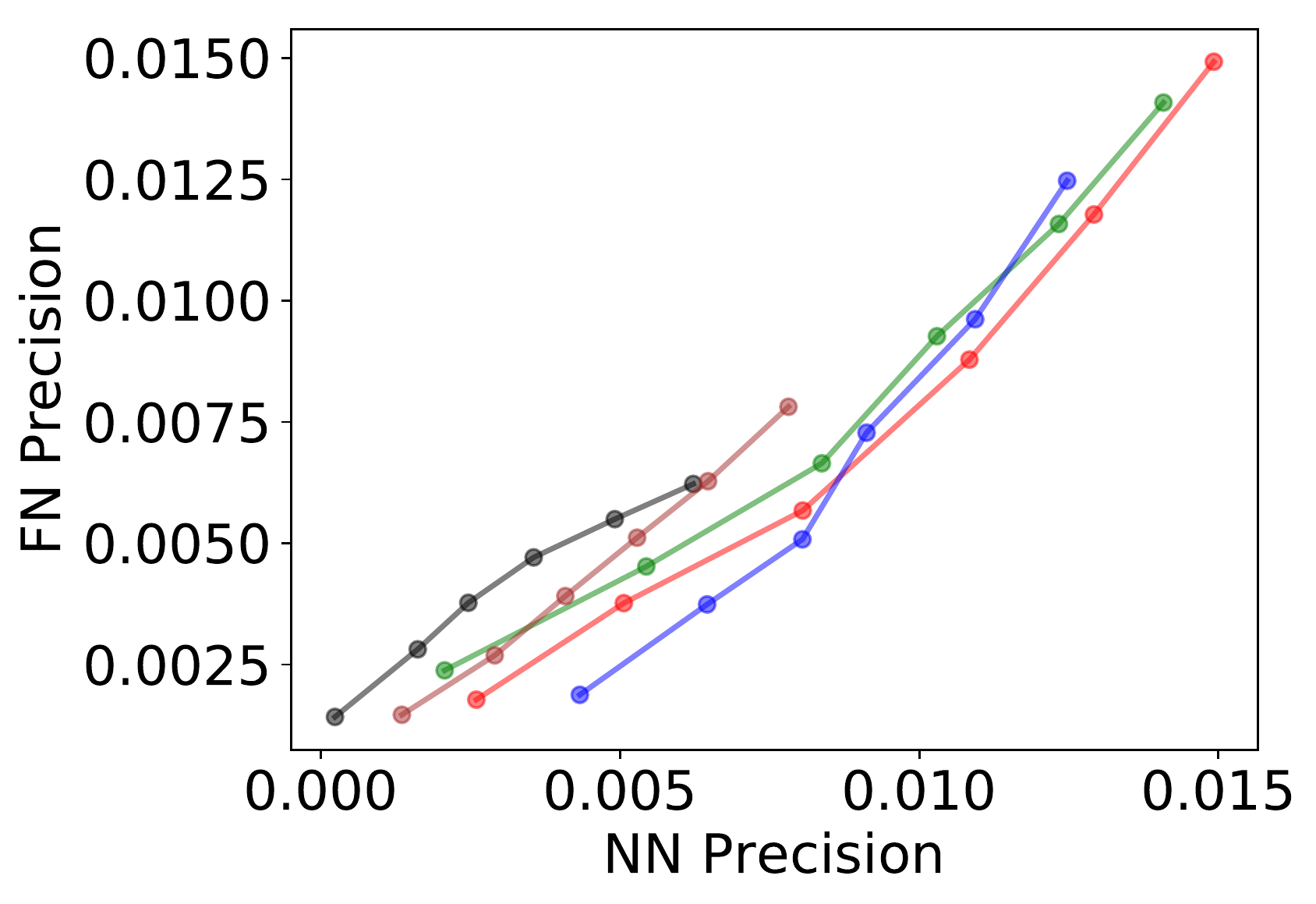}
            \vspace{-0.3cm}
            \end{minipage}   
            \begin{minipage}{0.16\textwidth}
            \includegraphics[width=\linewidth]{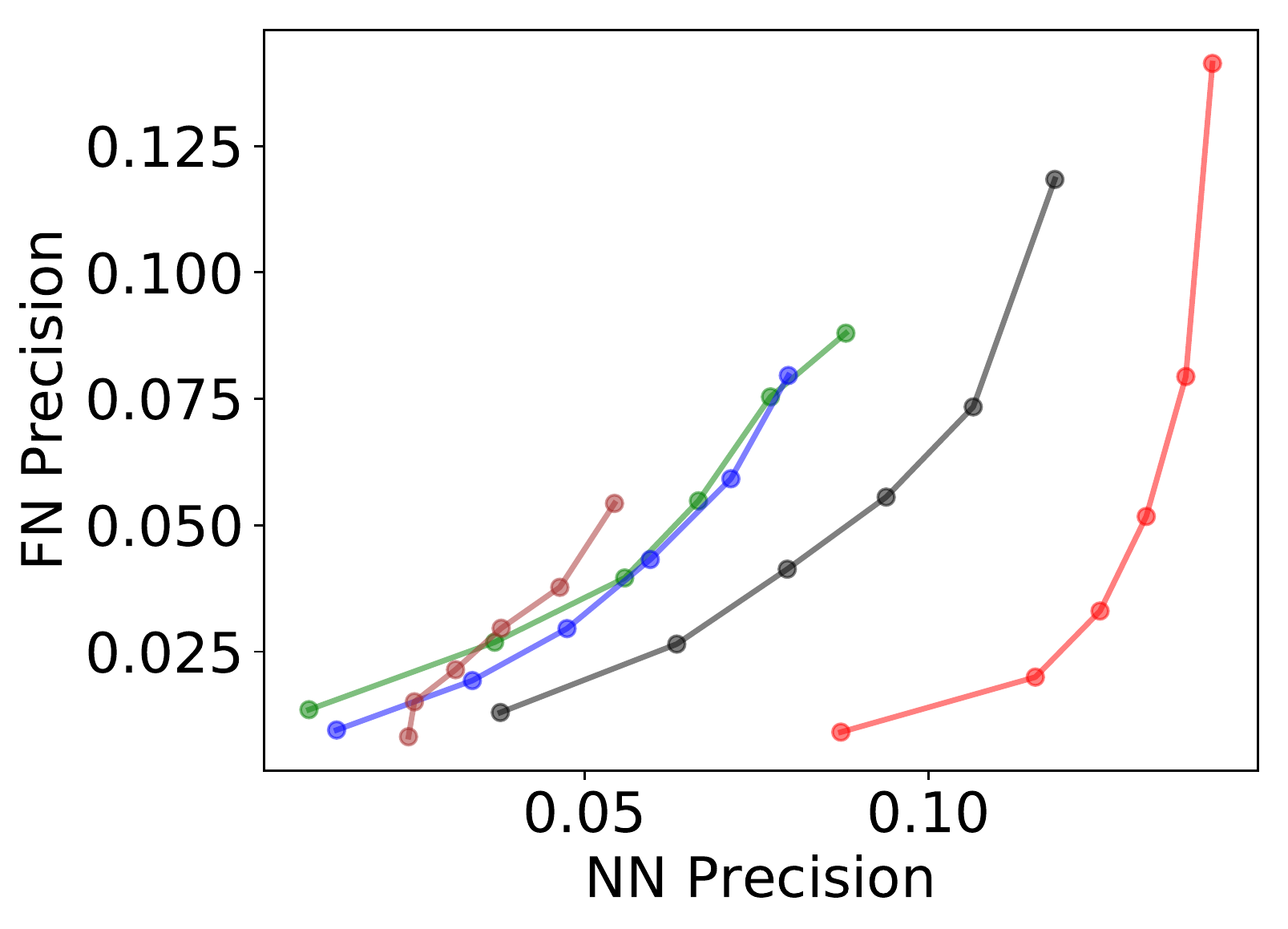}
            \vspace{-0.3cm}
            \end{minipage}   
            \begin{minipage}{0.16\textwidth}
            \includegraphics[width=\linewidth]{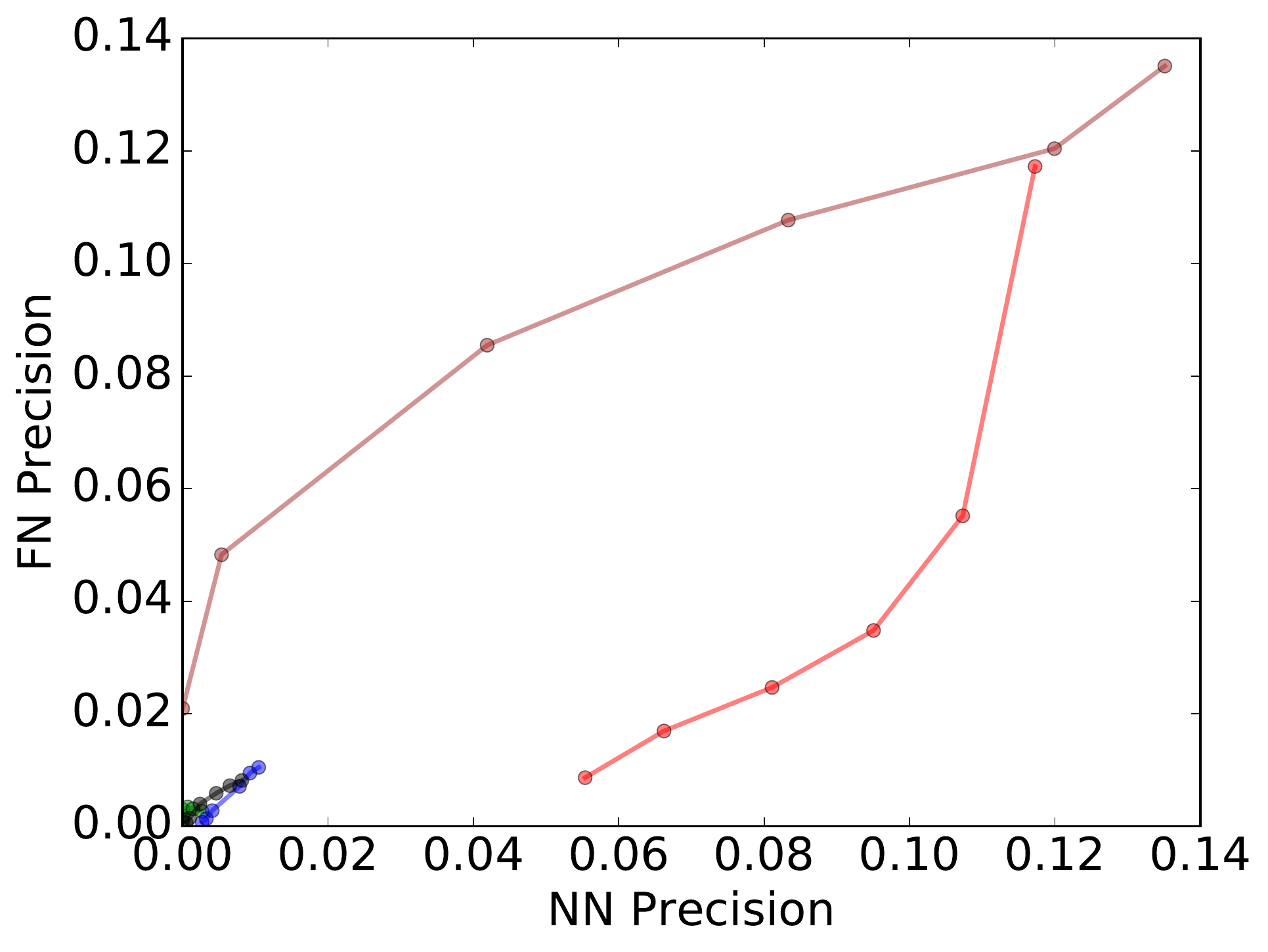}
            \vspace{-0.3cm}
            \end{minipage}   
            \begin{minipage}{0.16\textwidth}
            \includegraphics[width=\linewidth]{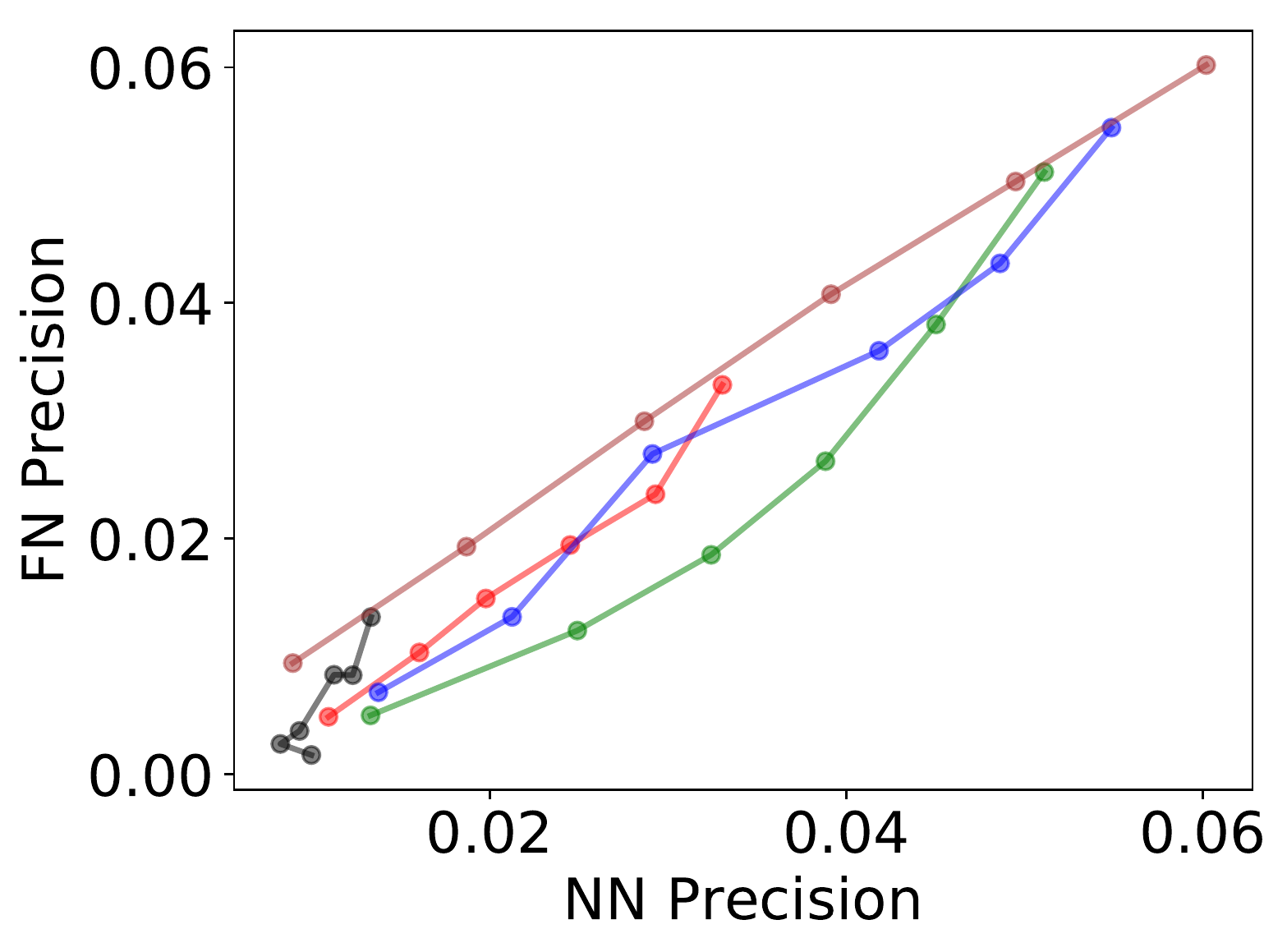}
            \vspace{-0.3cm}
            \end{minipage}   
            \subcaption{FN Precision vs. NN Precision}
        \end{minipage}
        \caption{Precision-Recall curves for each of the proposed algorithms on all datasets. Each row corresponds to a different dataset, each column to different quantitative criteria, and each line to a different algorithm.}
        \label{fig:results_diff_metric2}
    \end{figure}

    \begin{figure}[htp]
        \begin{minipage}{\textwidth}
            \begin{minipage}{0.195\textwidth}
                \subcaption*{RKL: $\alpha=0$}
            \end{minipage}   
            \begin{minipage}{0.195\textwidth}
            \subcaption*{$\alpha=0.01$}
            \end{minipage}         
            \begin{minipage}{0.195\textwidth}
            \subcaption*{$\alpha=0.1$}
            \end{minipage}  
            \begin{minipage}{0.195\textwidth}
            \subcaption*{JS: $\alpha=0.5$}
            \end{minipage}  
            \begin{minipage}{0.195\textwidth}
            \subcaption*{KL: $\alpha=1$}
            \end{minipage}  
        \end{minipage}     
        \begin{minipage}{\textwidth}
            \begin{minipage}{0.195\textwidth}
            \includegraphics[width=\linewidth]{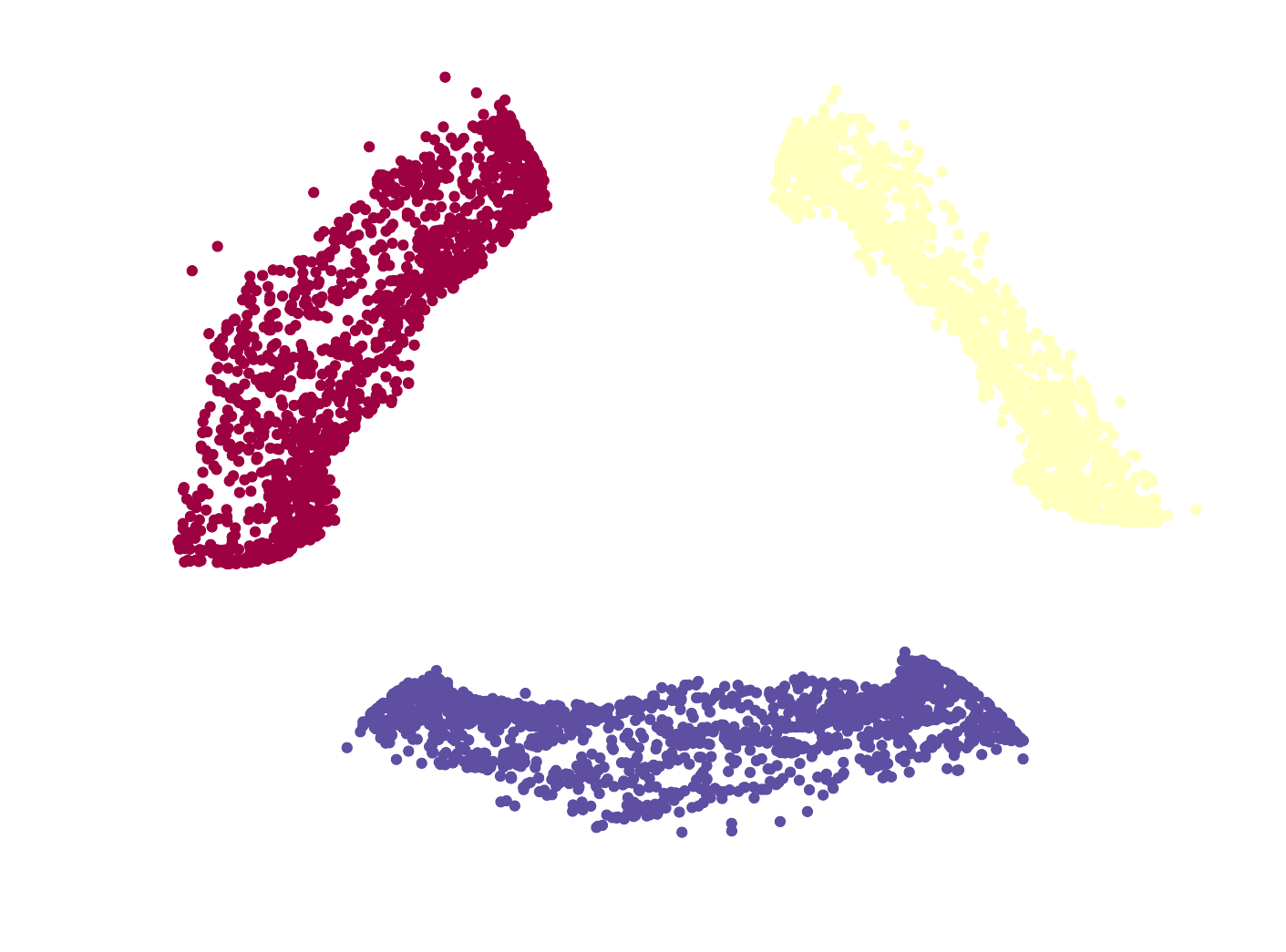}
            \end{minipage}   
            \begin{minipage}{0.195\textwidth}
            \includegraphics[width=\linewidth]{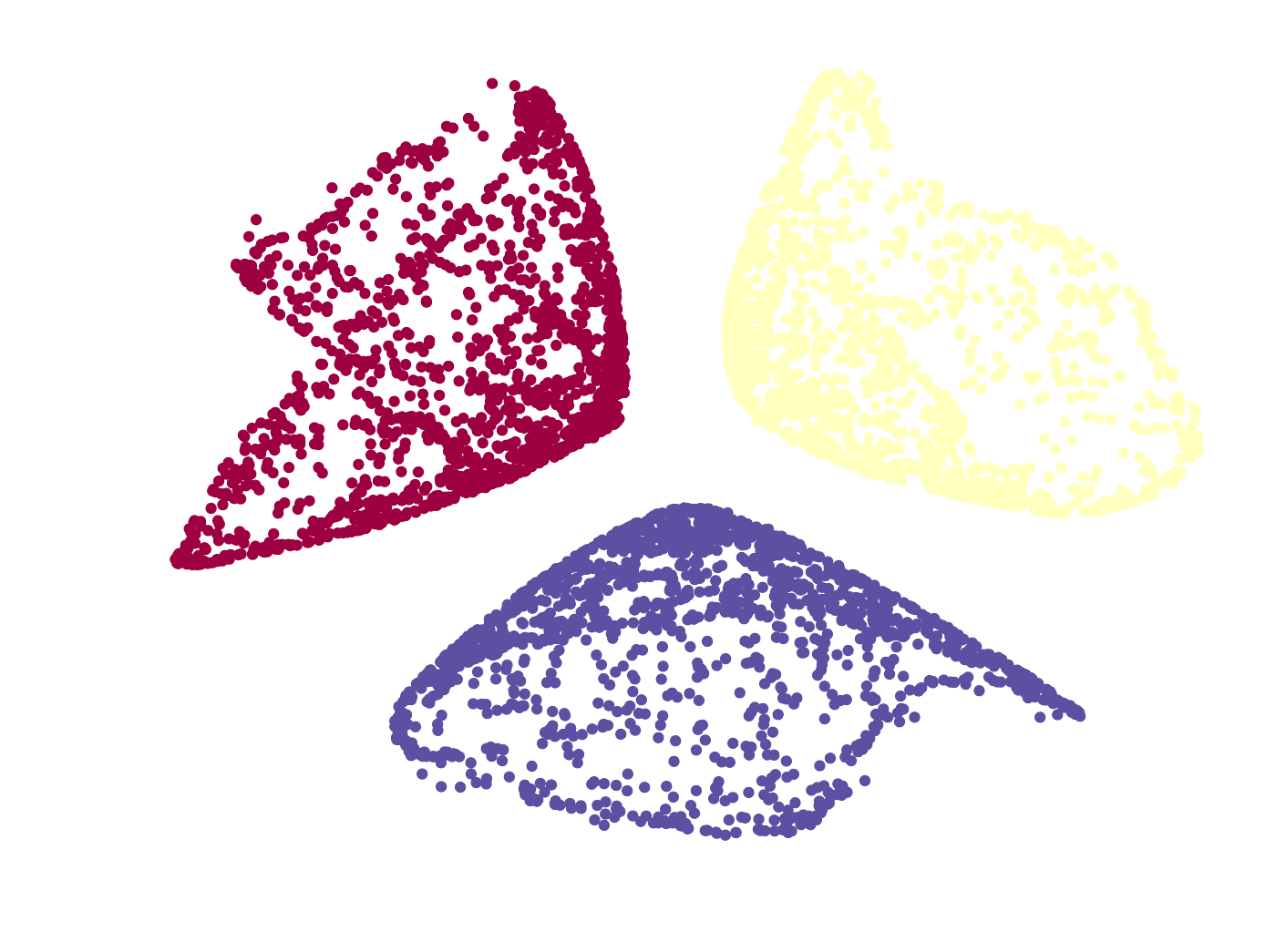}
            \end{minipage}         
            \begin{minipage}{0.195\textwidth}
            \includegraphics[width=\linewidth]{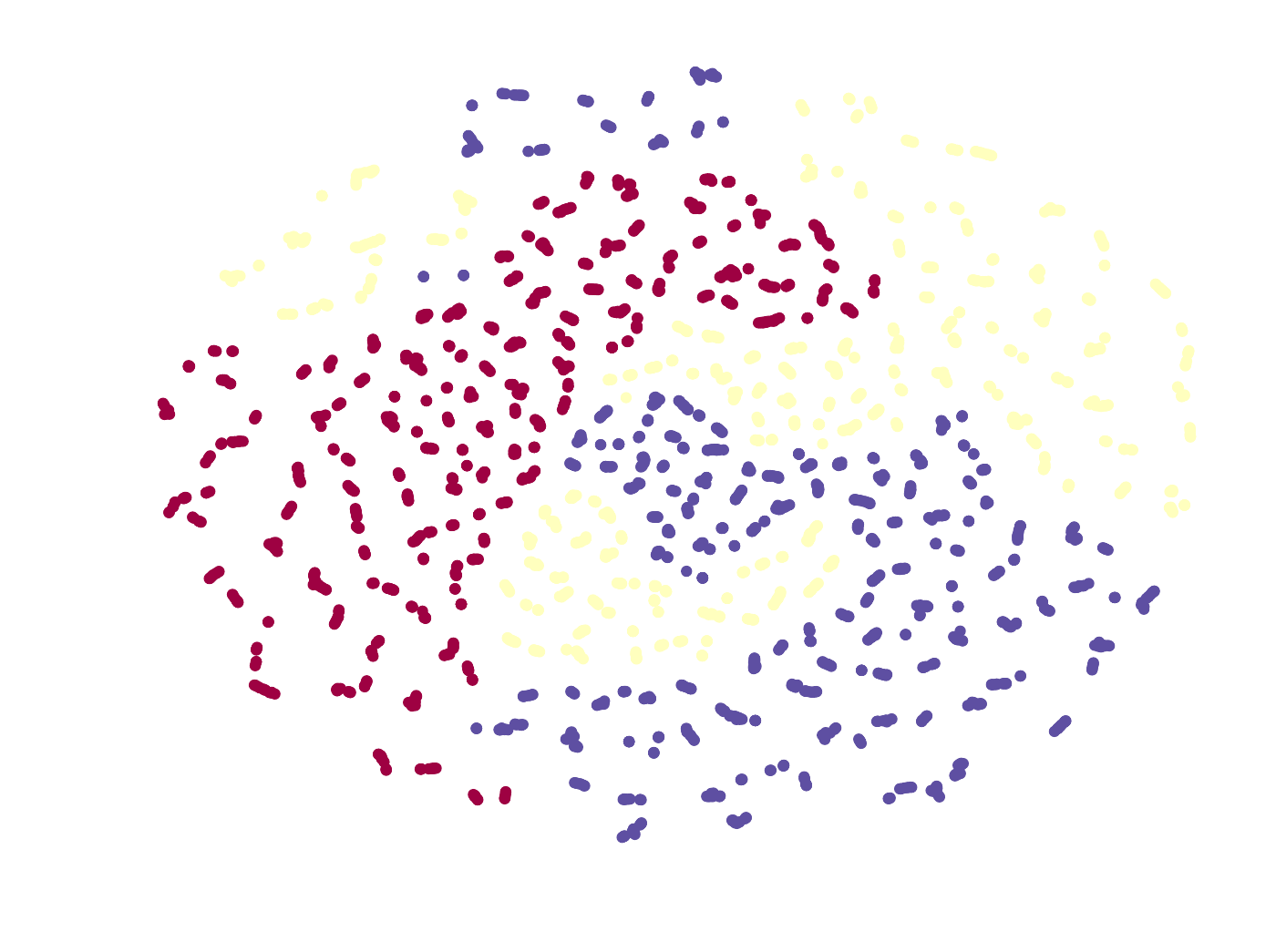}
            \end{minipage}  
            \begin{minipage}{0.195\textwidth}
            \includegraphics[width=\linewidth]{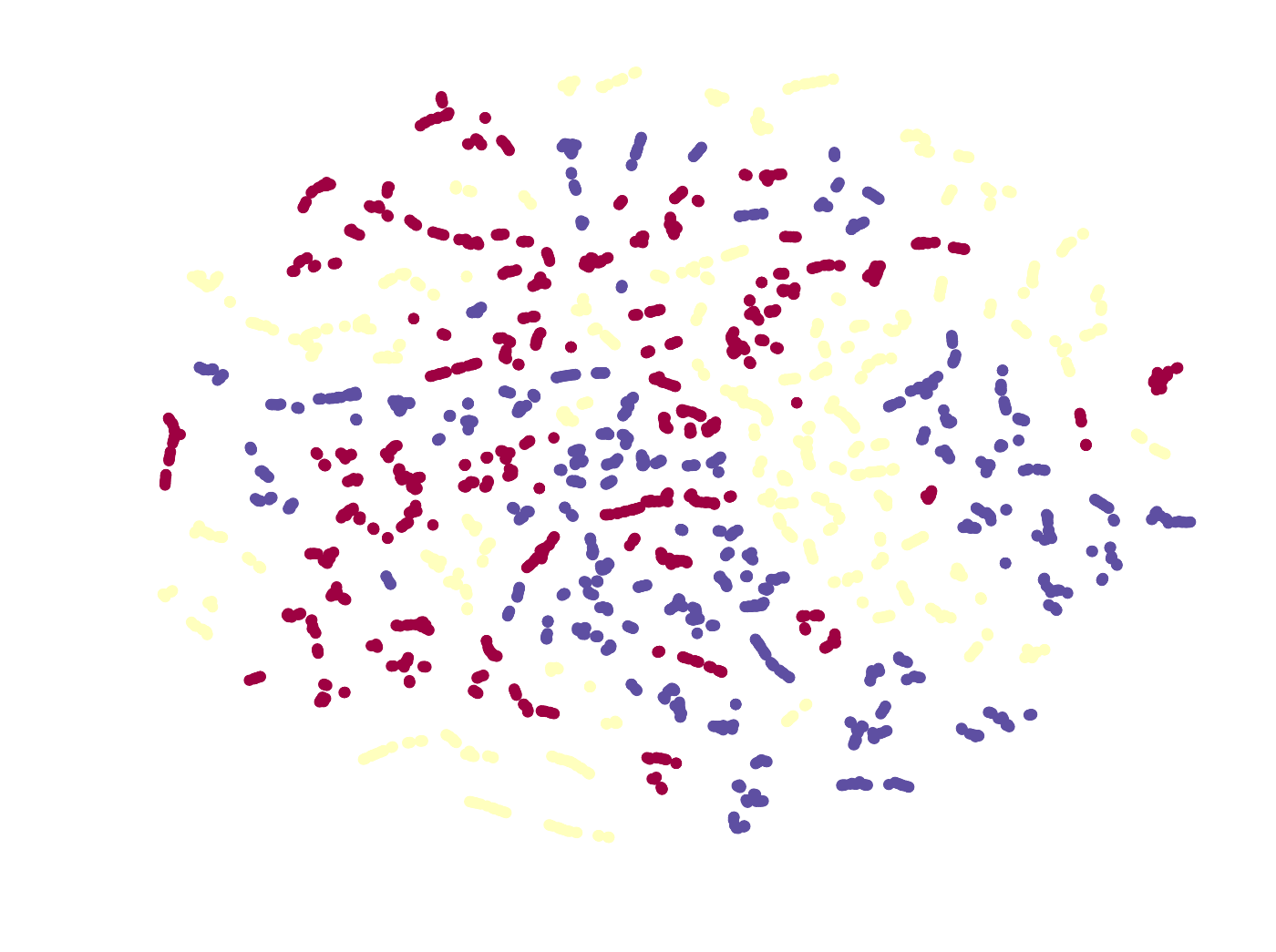}
            \end{minipage}  
            \begin{minipage}{0.195\textwidth}
            \includegraphics[width=\linewidth]{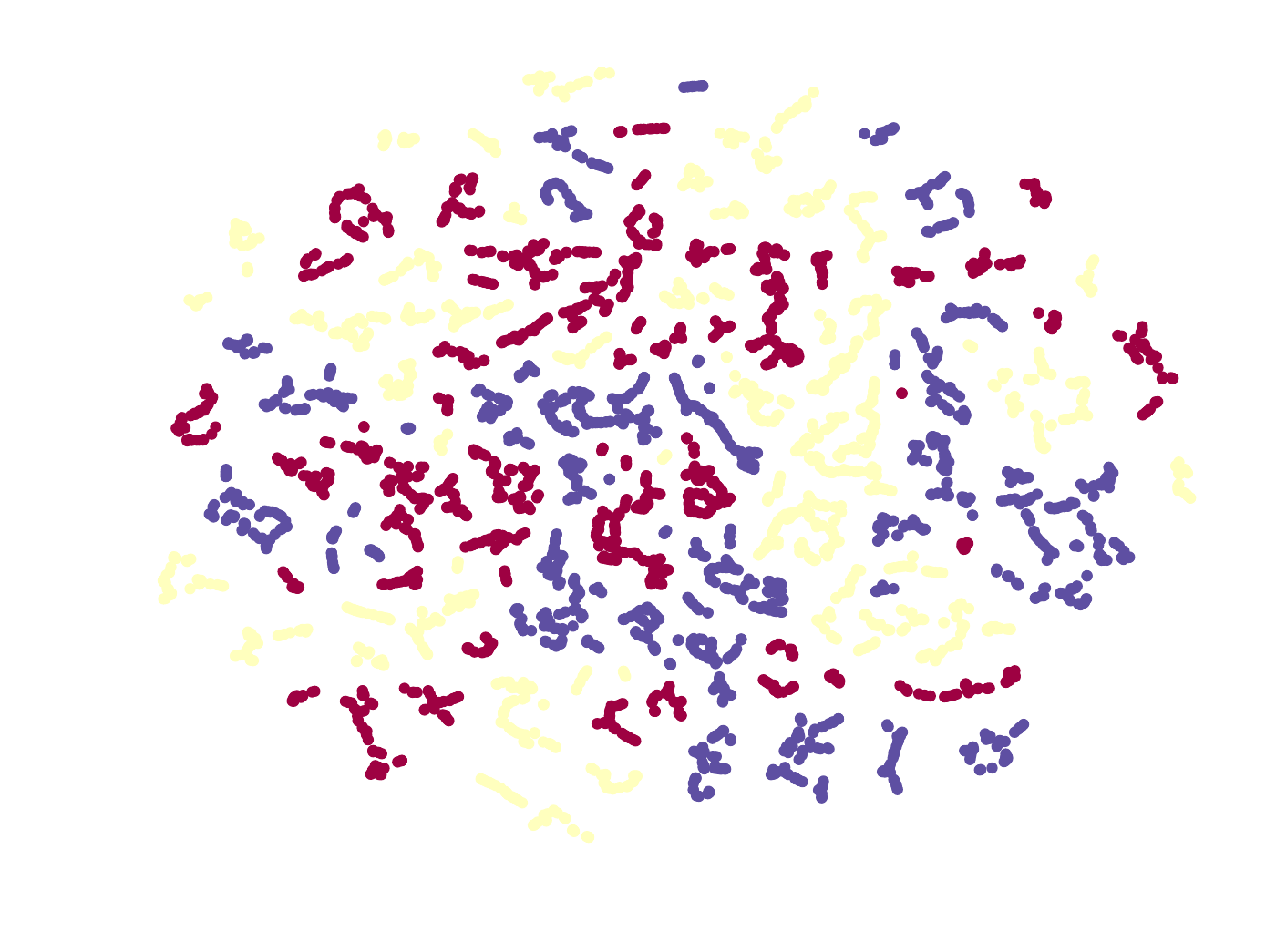}
            \end{minipage}  
            \subcaption*{Perplexity=10} 
        \end{minipage}       
        \begin{minipage}{\textwidth}
            \begin{minipage}{0.195\textwidth}
            \includegraphics[width=\linewidth]{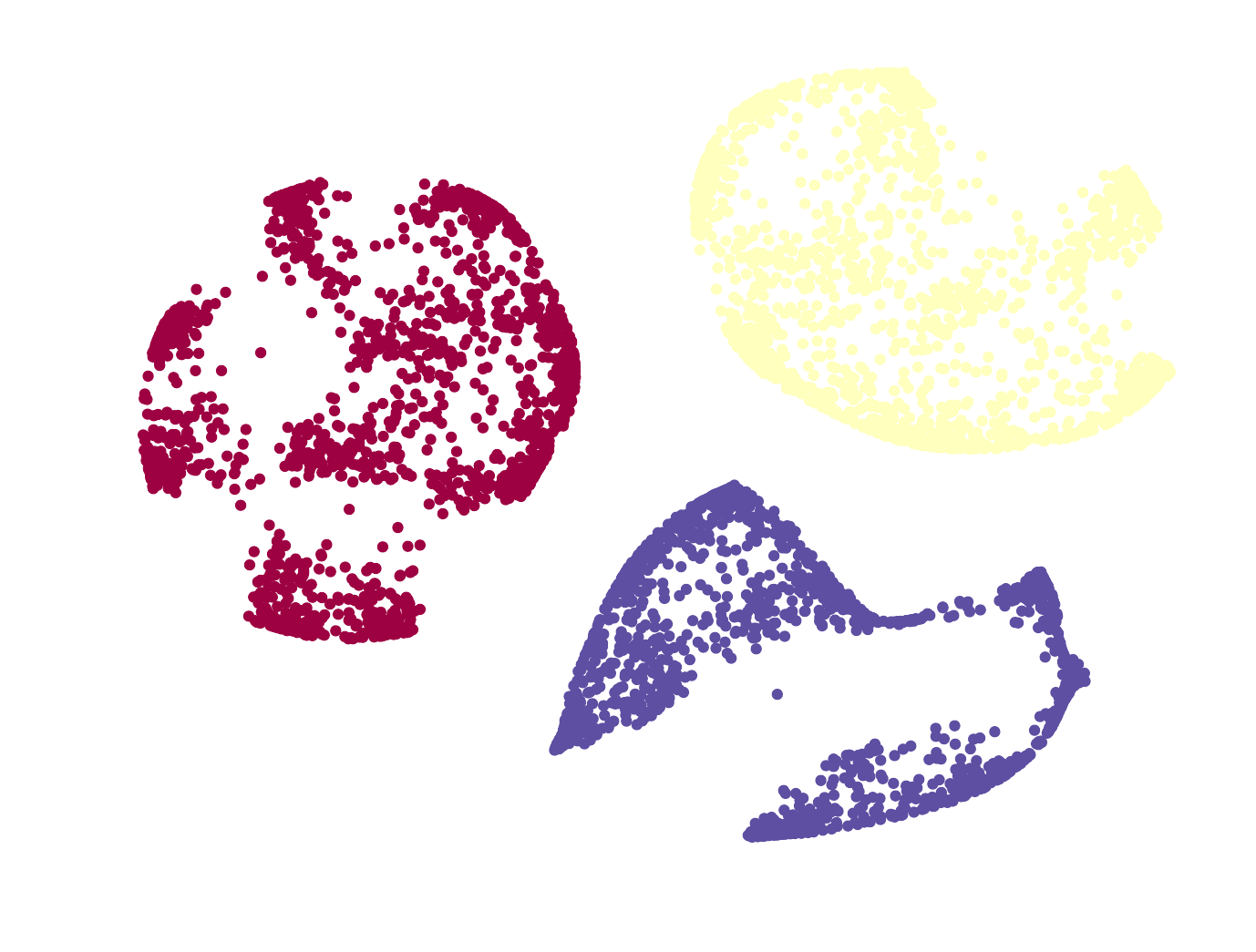}
            \end{minipage}   
            \begin{minipage}{0.195\textwidth}
            \includegraphics[width=\linewidth]{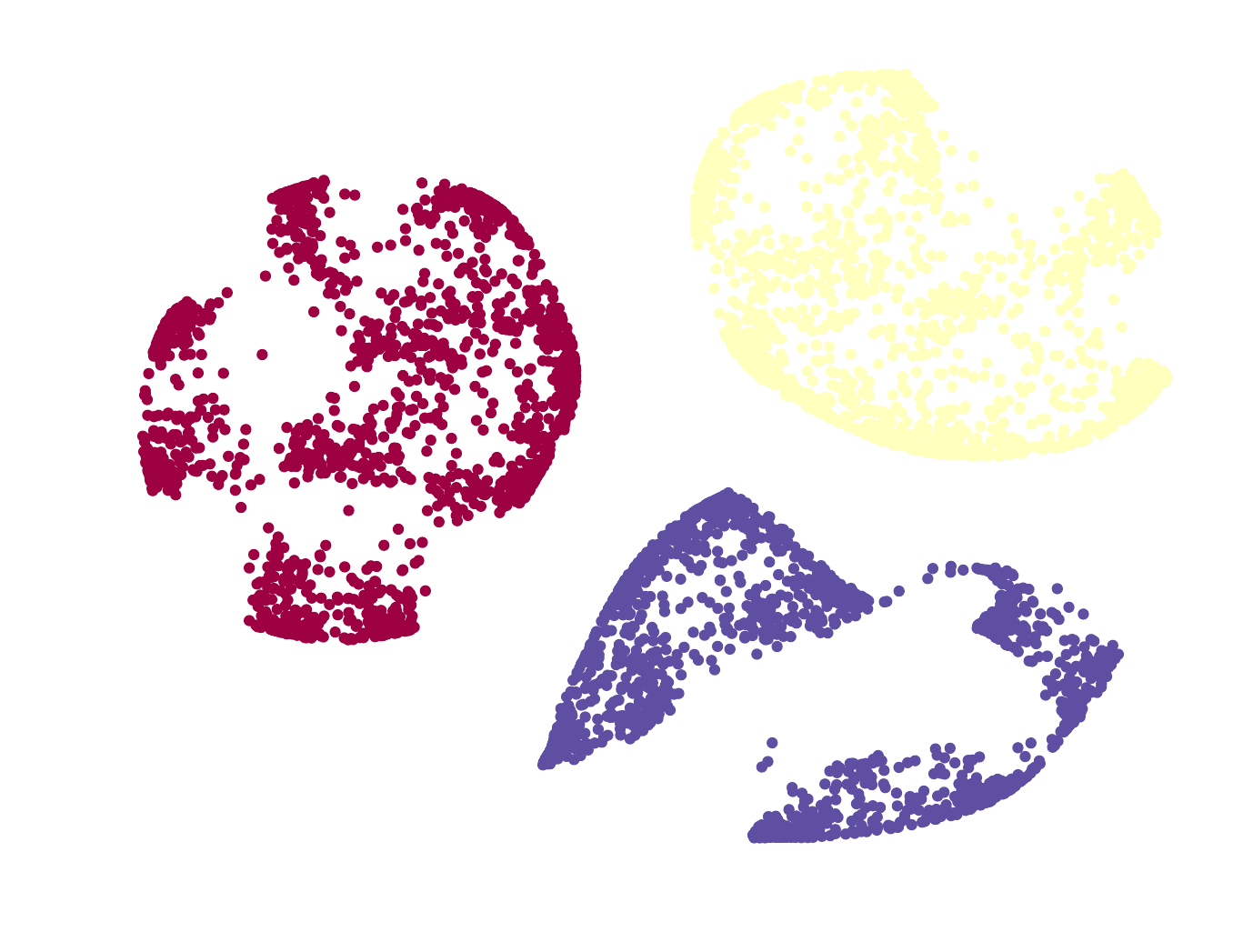}
            \end{minipage}         
            \begin{minipage}{0.195\textwidth}
            \includegraphics[width=\linewidth]{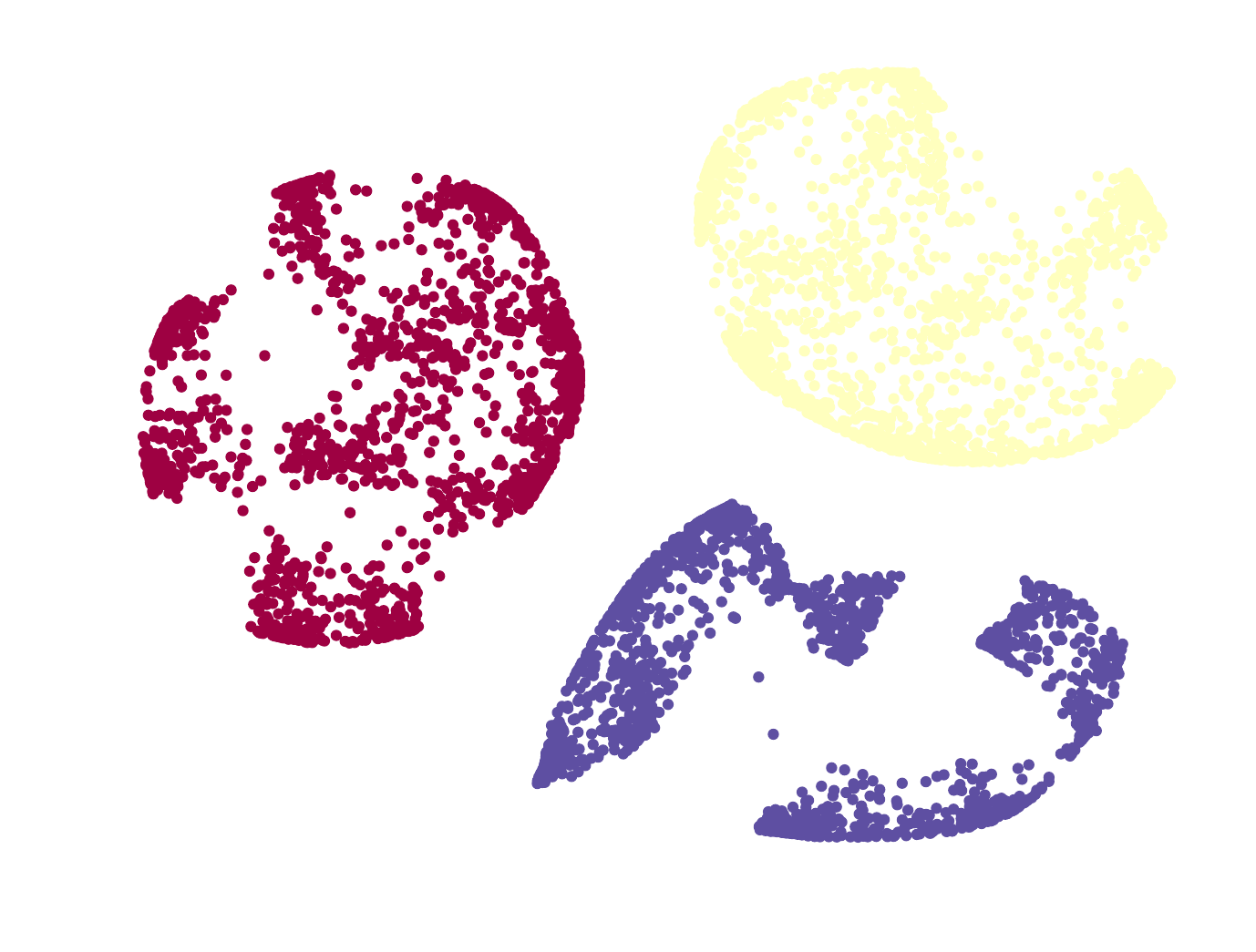}
            \end{minipage}  
            \begin{minipage}{0.195\textwidth}
            \includegraphics[width=\linewidth]{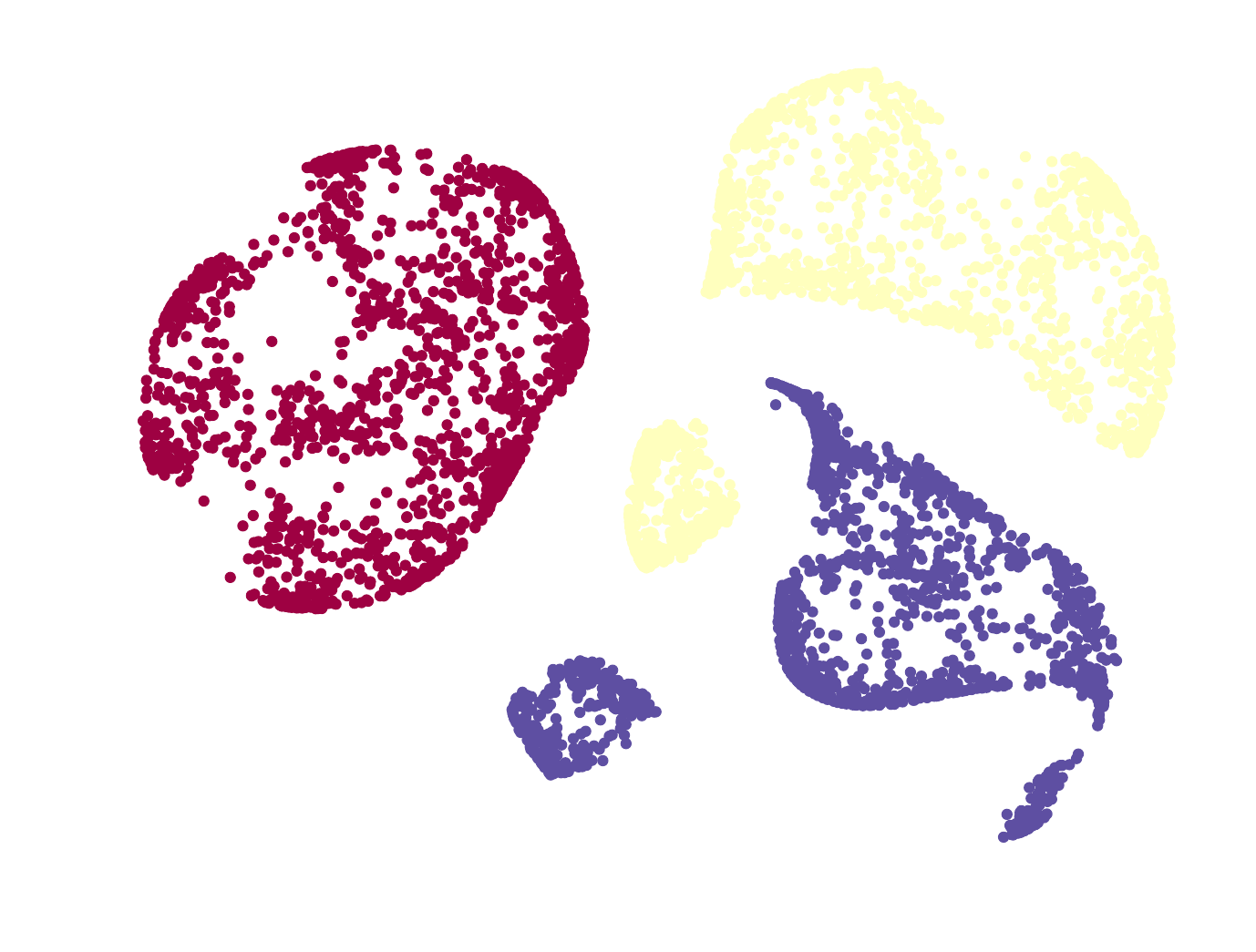}
            \end{minipage}  
            \begin{minipage}{0.195\textwidth}
            \includegraphics[width=\linewidth]{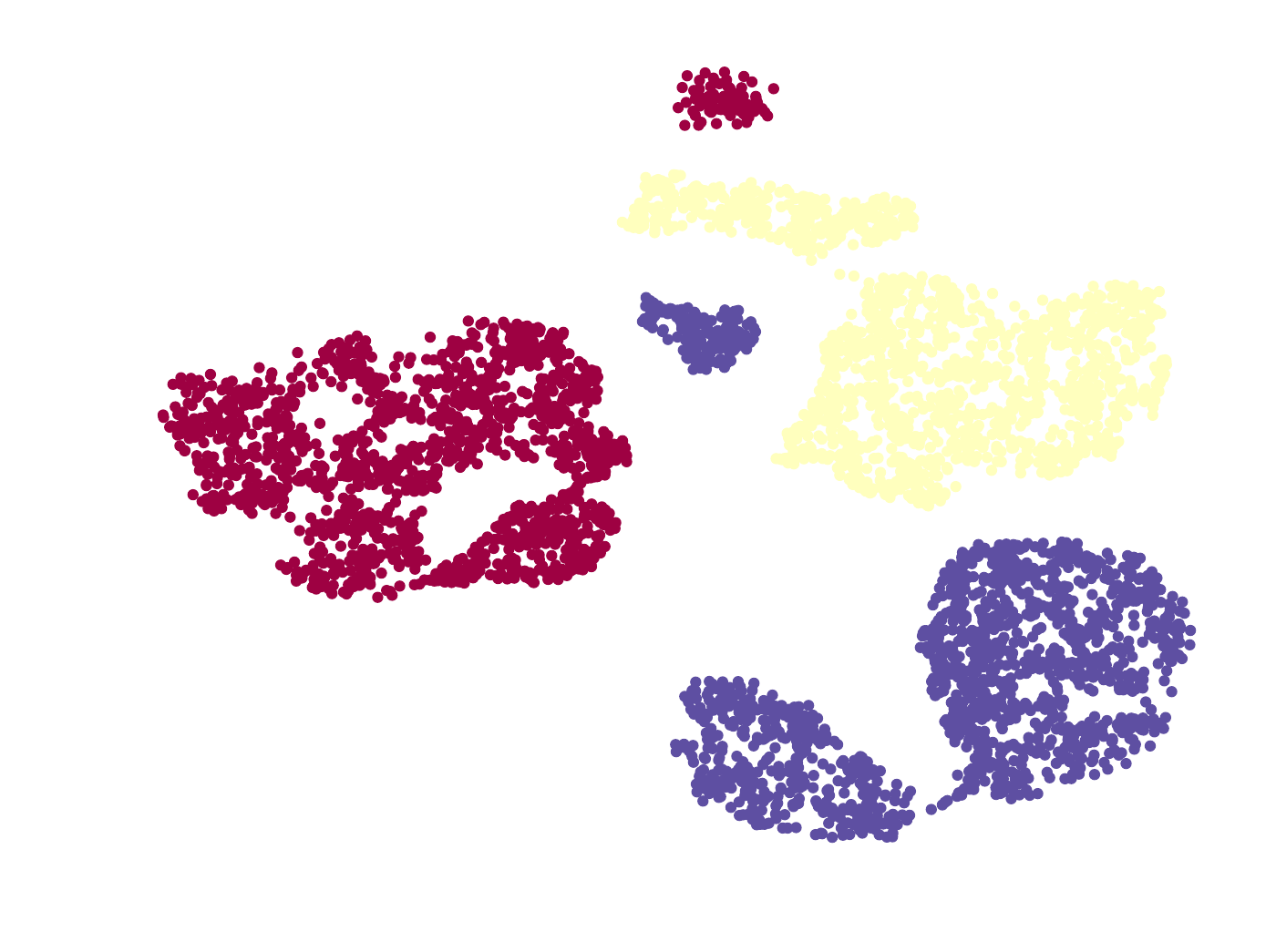}
            \end{minipage}  
            \subcaption*{Perplexity=100} 
        \end{minipage}       
        \begin{minipage}{\textwidth}
            \begin{minipage}{0.195\textwidth}
            \includegraphics[width=\linewidth]{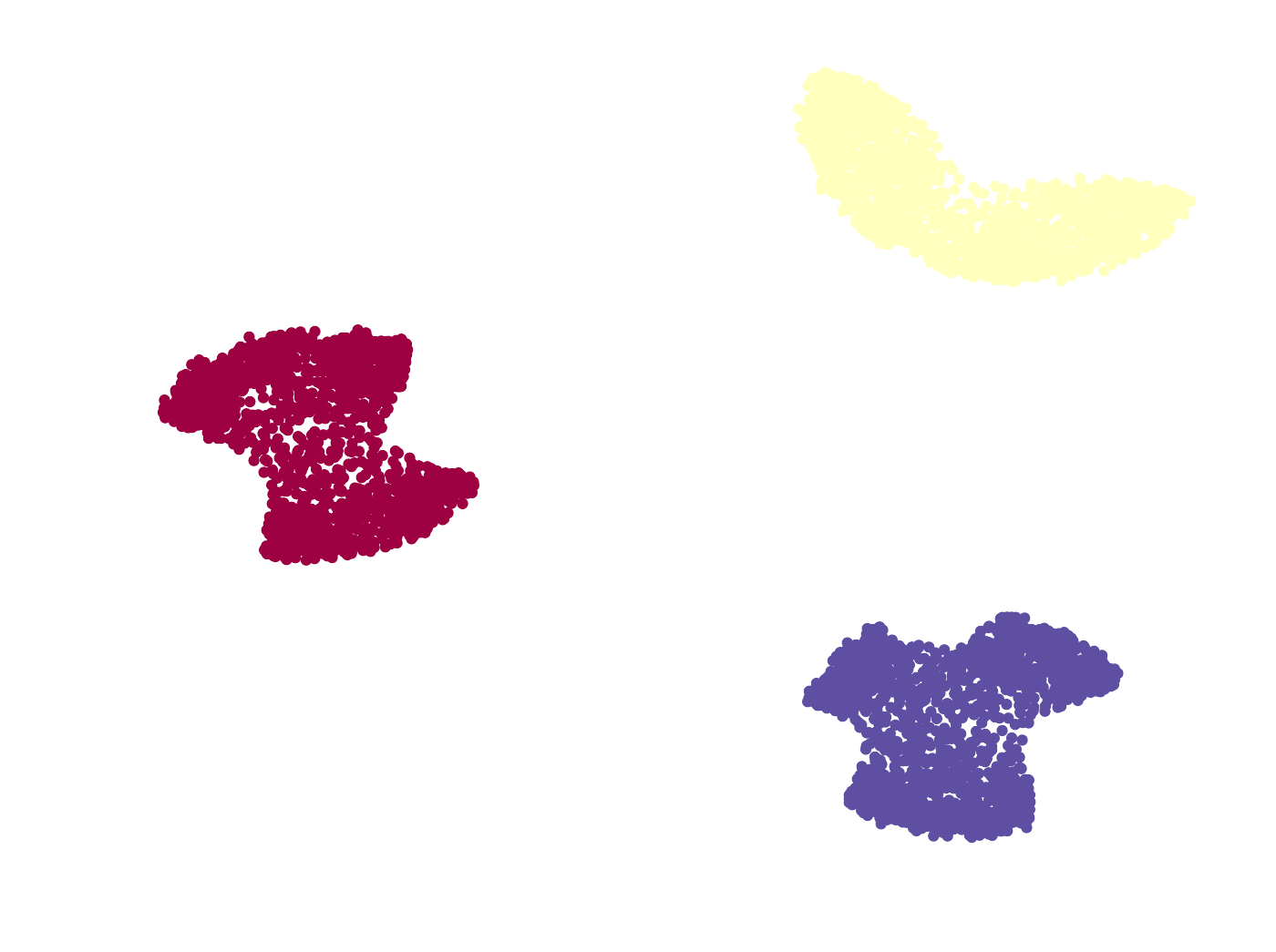}
            \end{minipage}   
            \begin{minipage}{0.195\textwidth}
            \includegraphics[width=\linewidth]{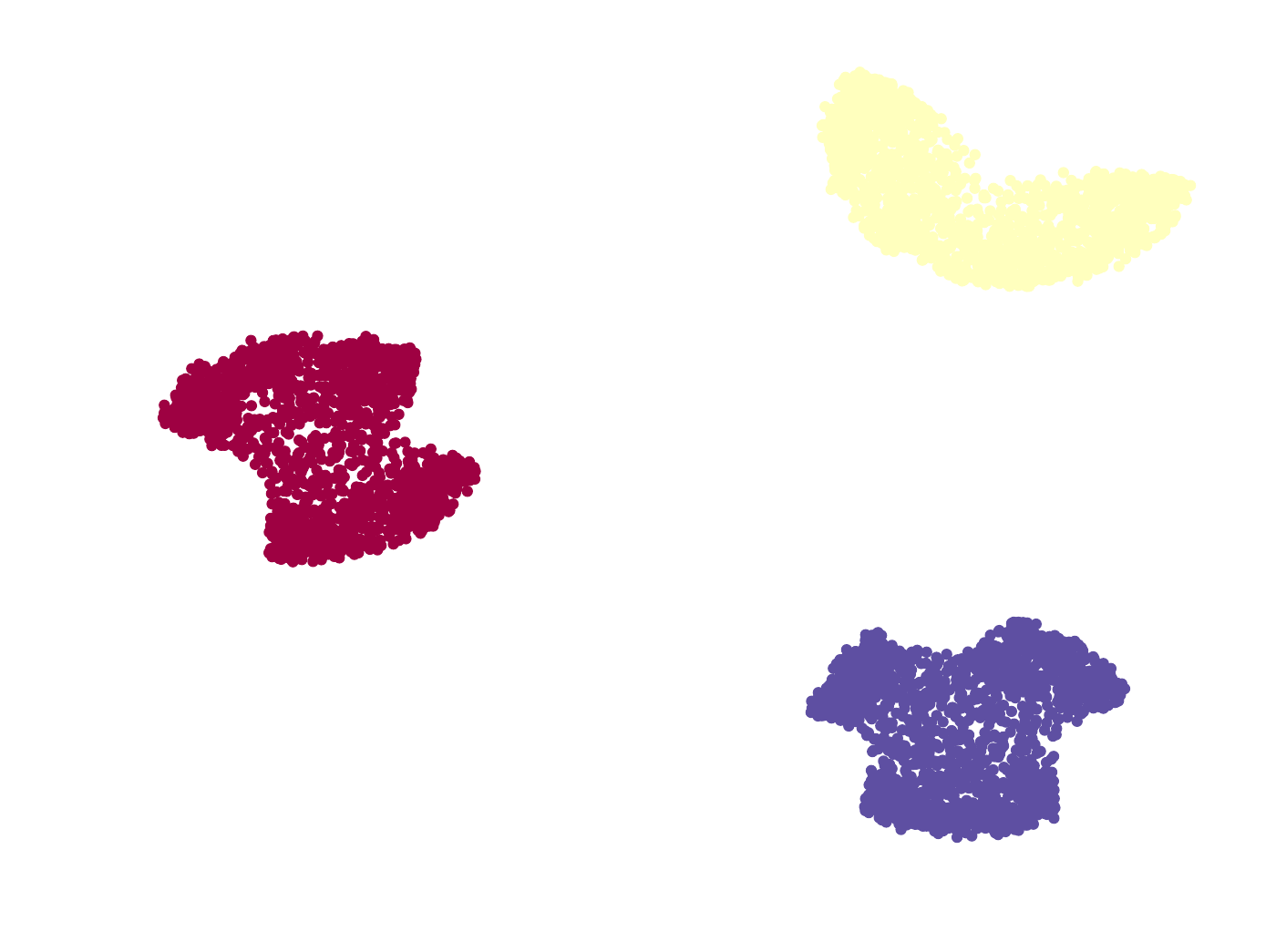}
            \end{minipage}         
            \begin{minipage}{0.195\textwidth}
            \includegraphics[width=\linewidth]{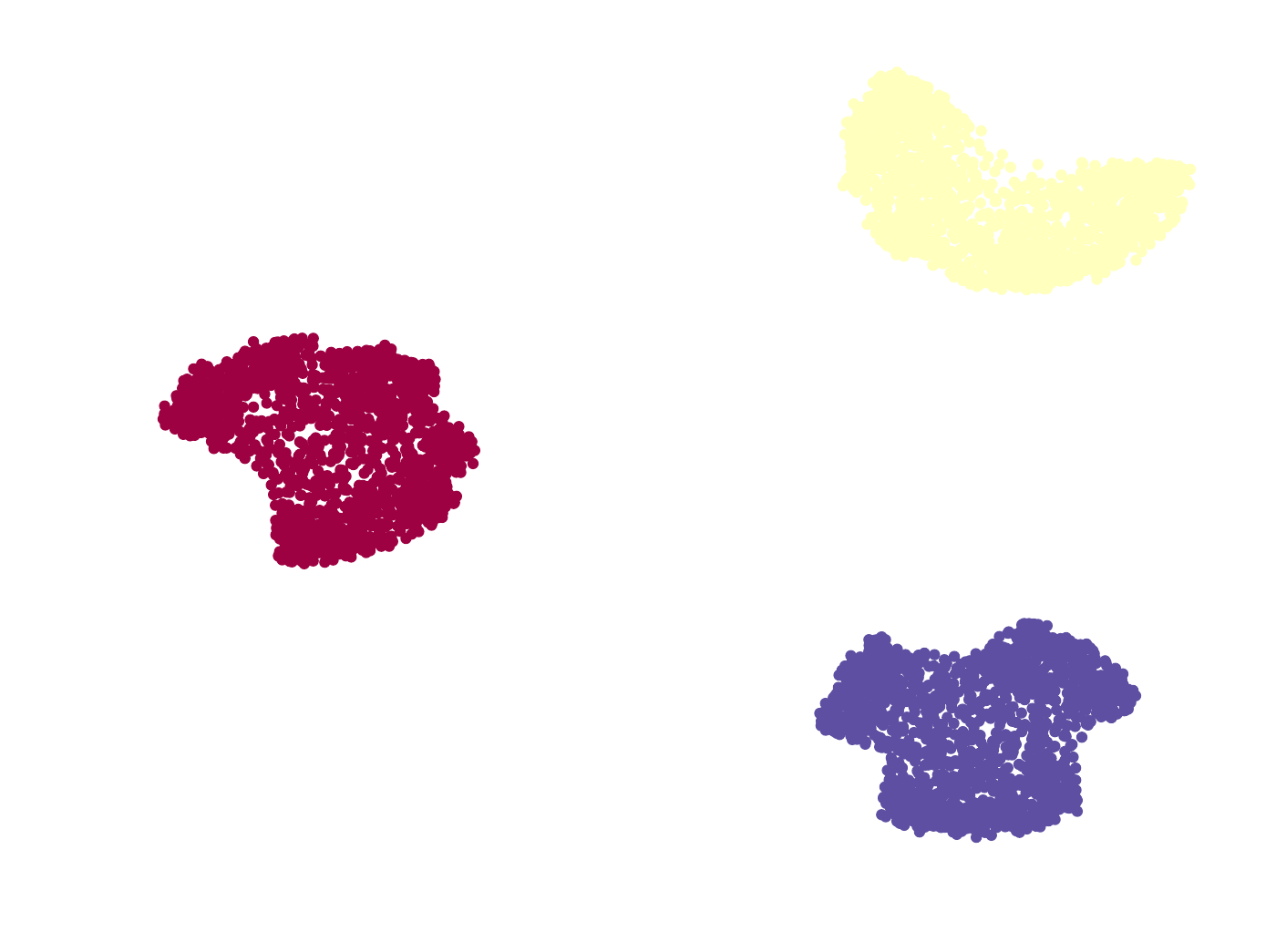}
            \end{minipage}  
            \begin{minipage}{0.195\textwidth}
            \includegraphics[width=\linewidth]{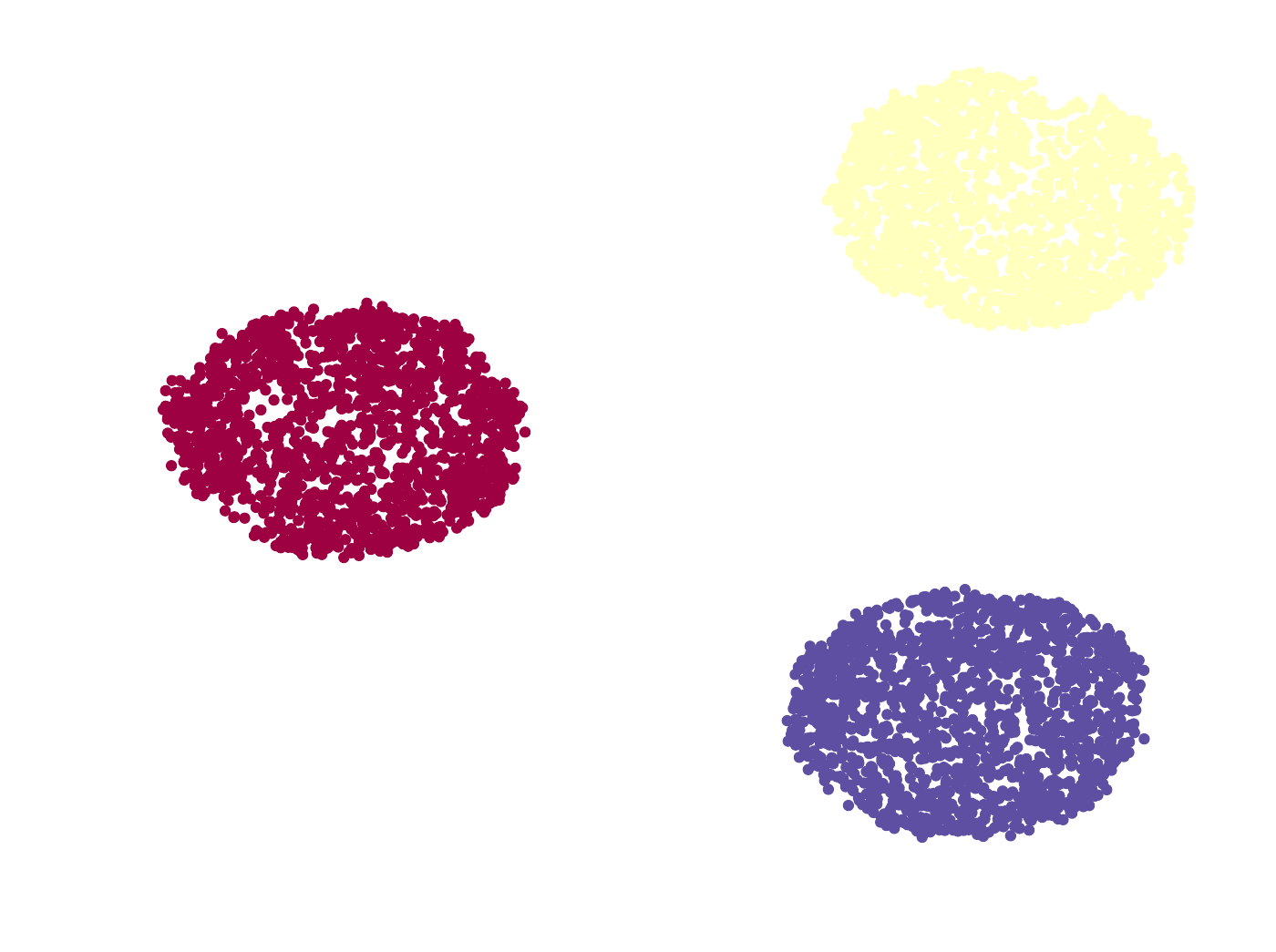}
            \end{minipage}  
            \begin{minipage}{0.195\textwidth}
            \includegraphics[width=\linewidth]{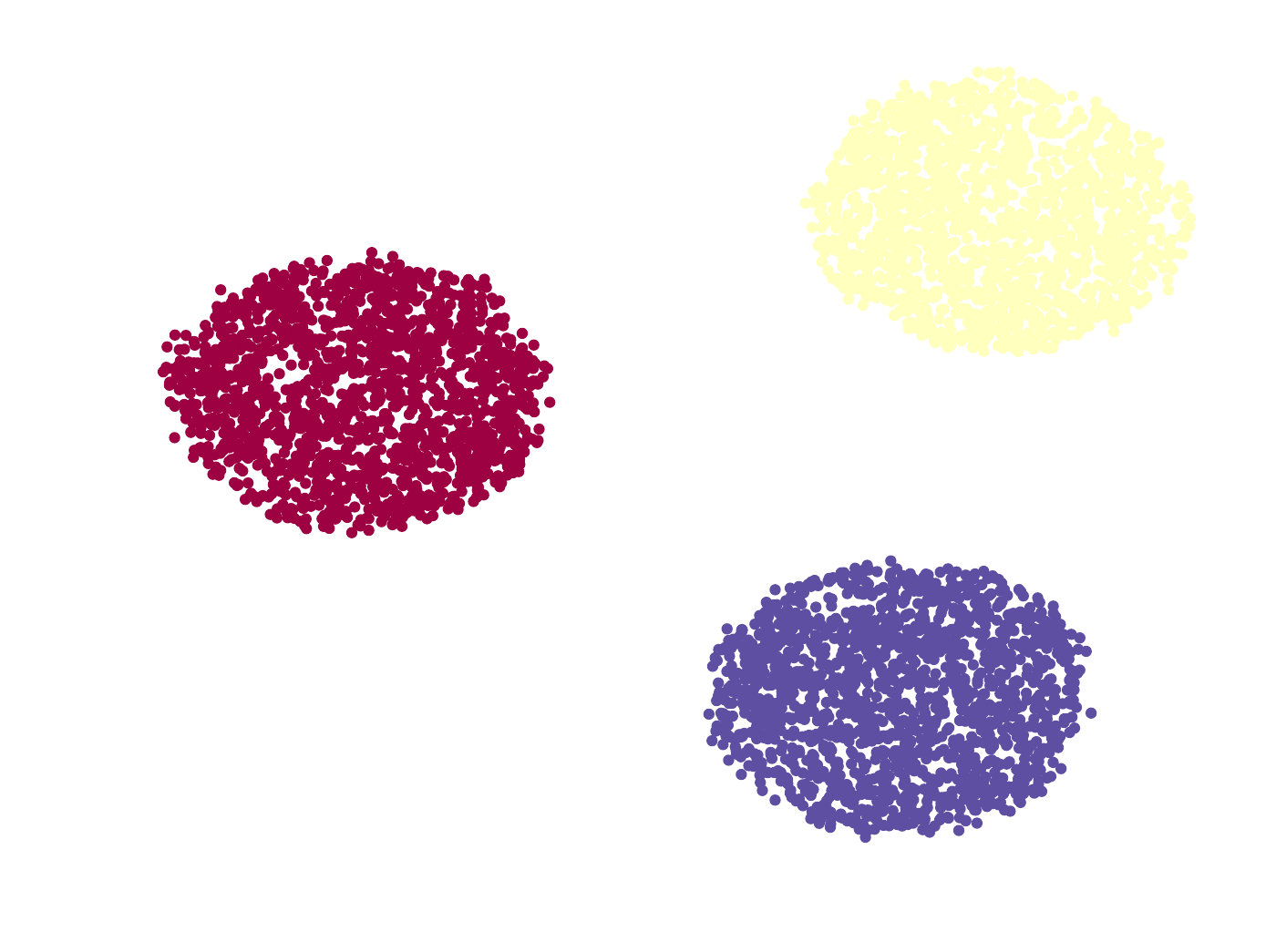}
            \end{minipage}  
            \subcaption*{Perplexity=500} 
        \end{minipage}       
        \caption{$t$-SNE embeddings on three Gaussian clusters}  
        \label{fig:syn_embeddings_blobs}
    \end{figure}
   \begin{figure}[htp]
        \begin{minipage}{\textwidth}
            \begin{minipage}{0.195\textwidth}
            \subcaption*{RKL: $\alpha=0$}
            \end{minipage}   
            \begin{minipage}{0.195\textwidth}
            \subcaption*{$\alpha=0.01$}
            \end{minipage}         
            \begin{minipage}{0.195\textwidth}
            \subcaption*{$\alpha=0.1$}
            \end{minipage}  
            \begin{minipage}{0.195\textwidth}
            \subcaption*{JS: $\alpha=0.5$}
            \end{minipage}  
            \begin{minipage}{0.195\textwidth}
            \subcaption*{KL: $\alpha=1$}
            \end{minipage}  
        \end{minipage}     
        \begin{minipage}{\textwidth}
            \begin{minipage}{0.195\textwidth}
            \includegraphics[width=\linewidth]{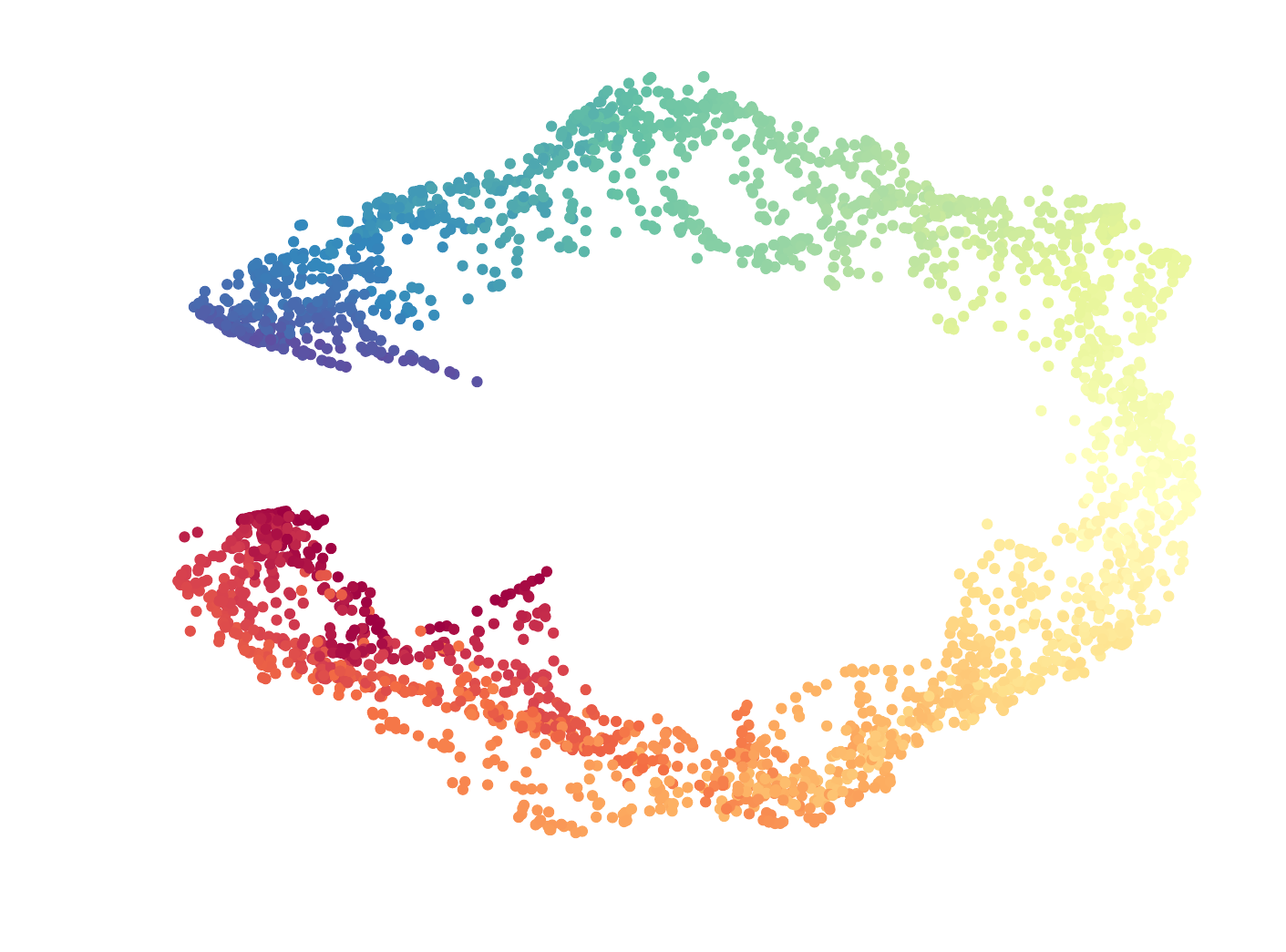}
            \end{minipage}   
            \begin{minipage}{0.195\textwidth}
            \includegraphics[width=\linewidth]{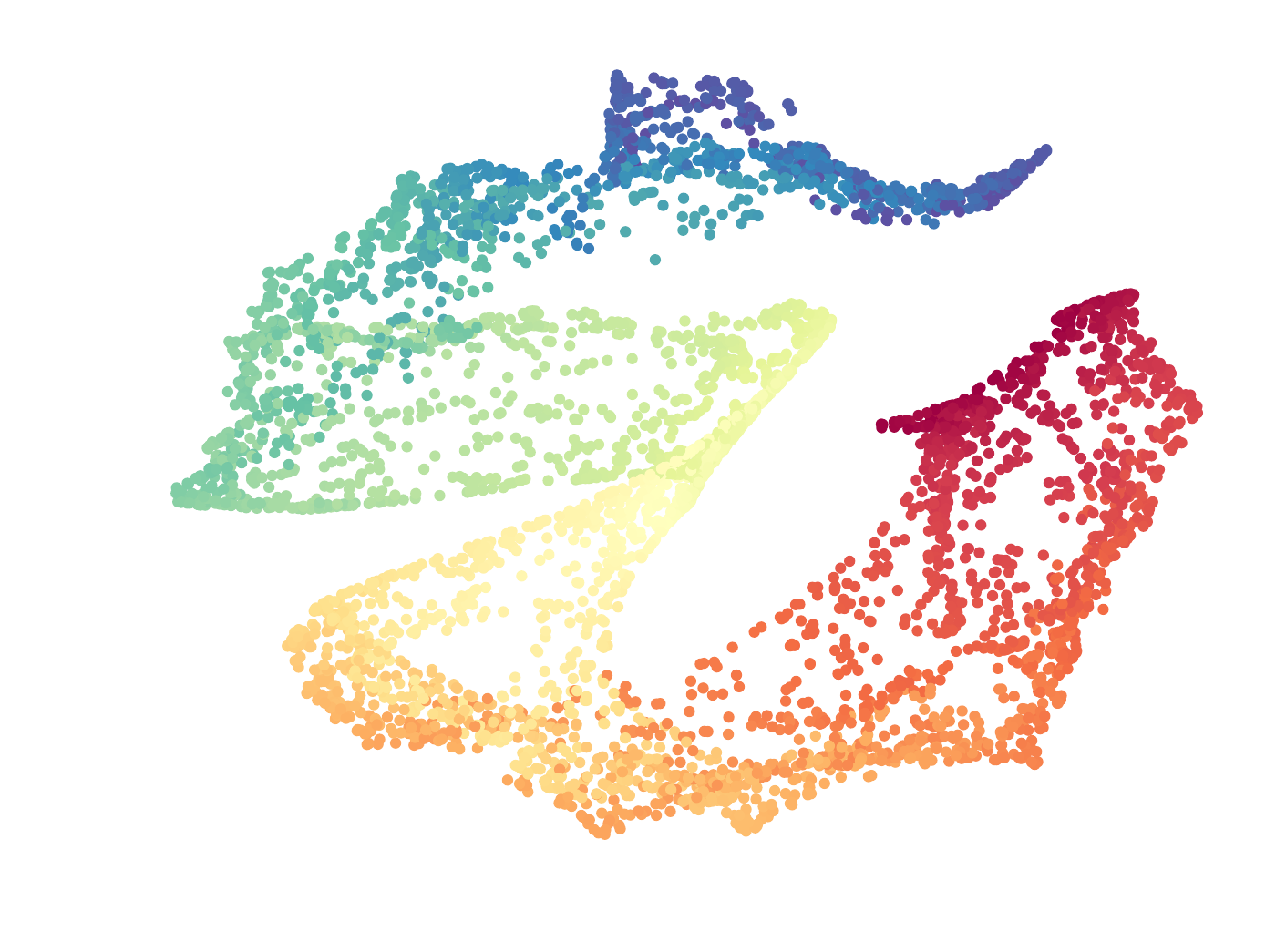}
            \end{minipage}         
            \begin{minipage}{0.195\textwidth}
            \includegraphics[width=\linewidth]{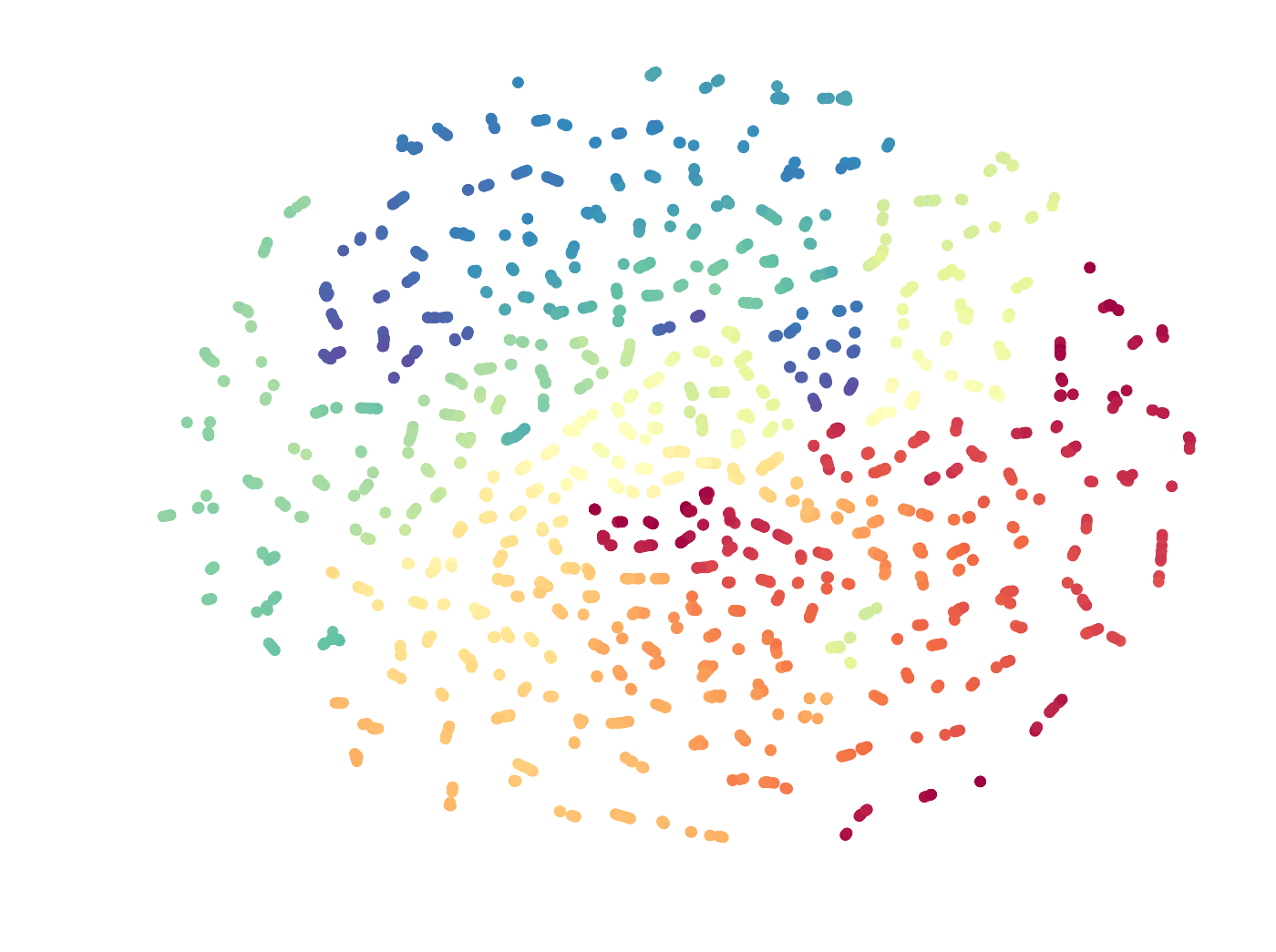}
            \end{minipage}  
            \begin{minipage}{0.195\textwidth}
            \includegraphics[width=\linewidth]{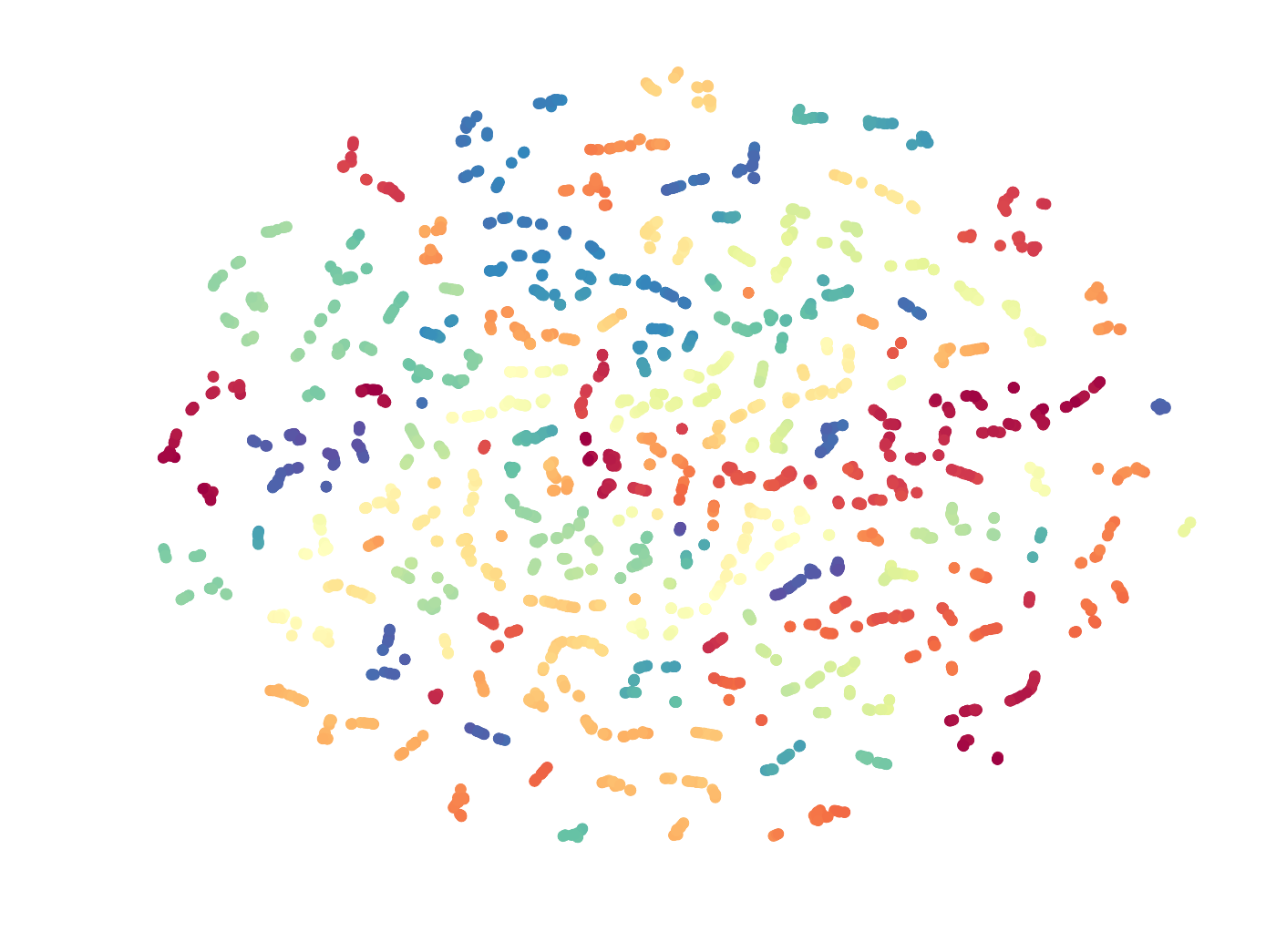}
            \end{minipage}  
            \begin{minipage}{0.195\textwidth}
            \includegraphics[width=\linewidth]{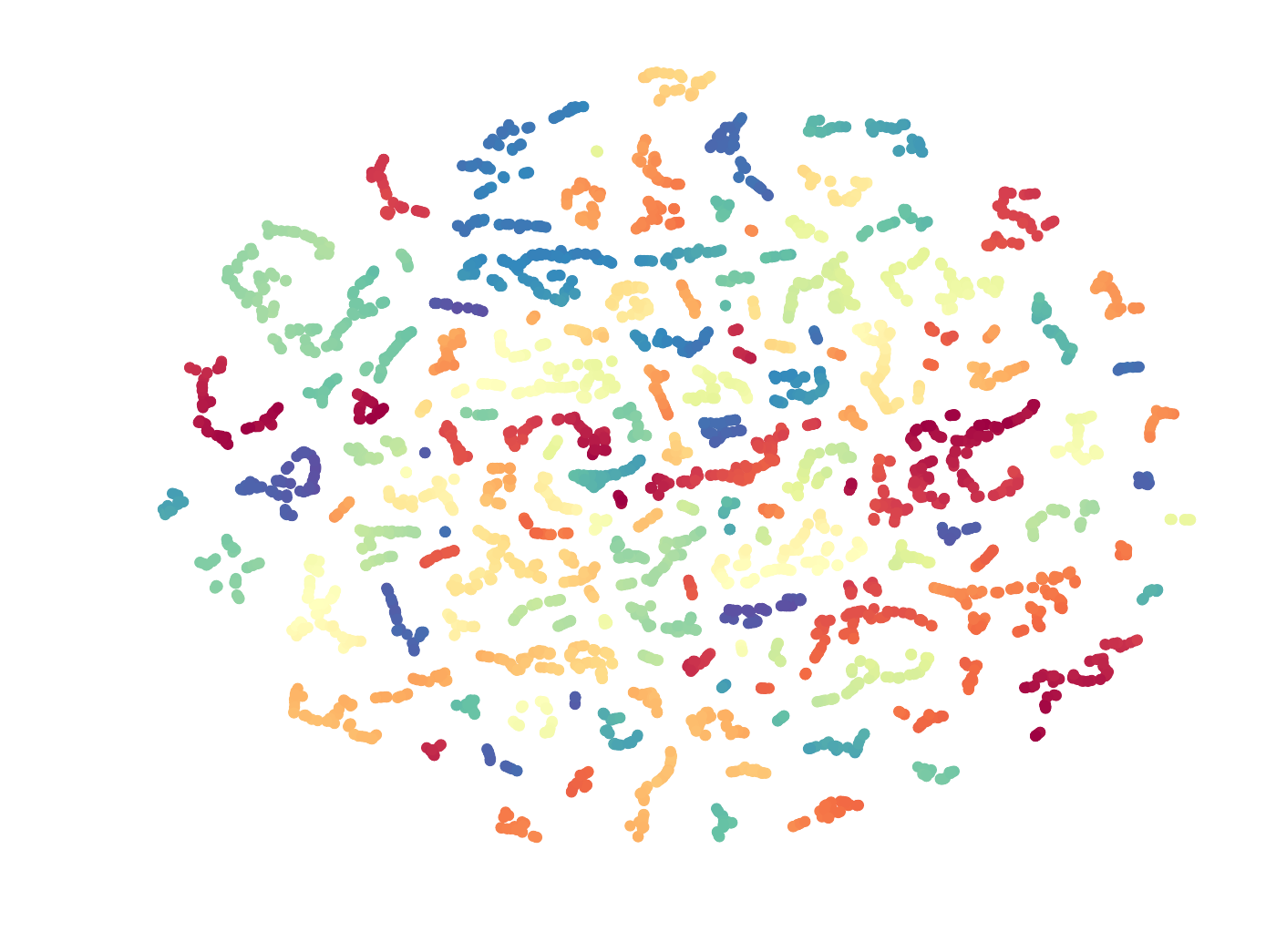}
            \end{minipage}  
            \subcaption{Perplexity=10} 
        \end{minipage}       
        \begin{minipage}{\textwidth}
            \begin{minipage}{0.195\textwidth}
            \includegraphics[width=\linewidth]{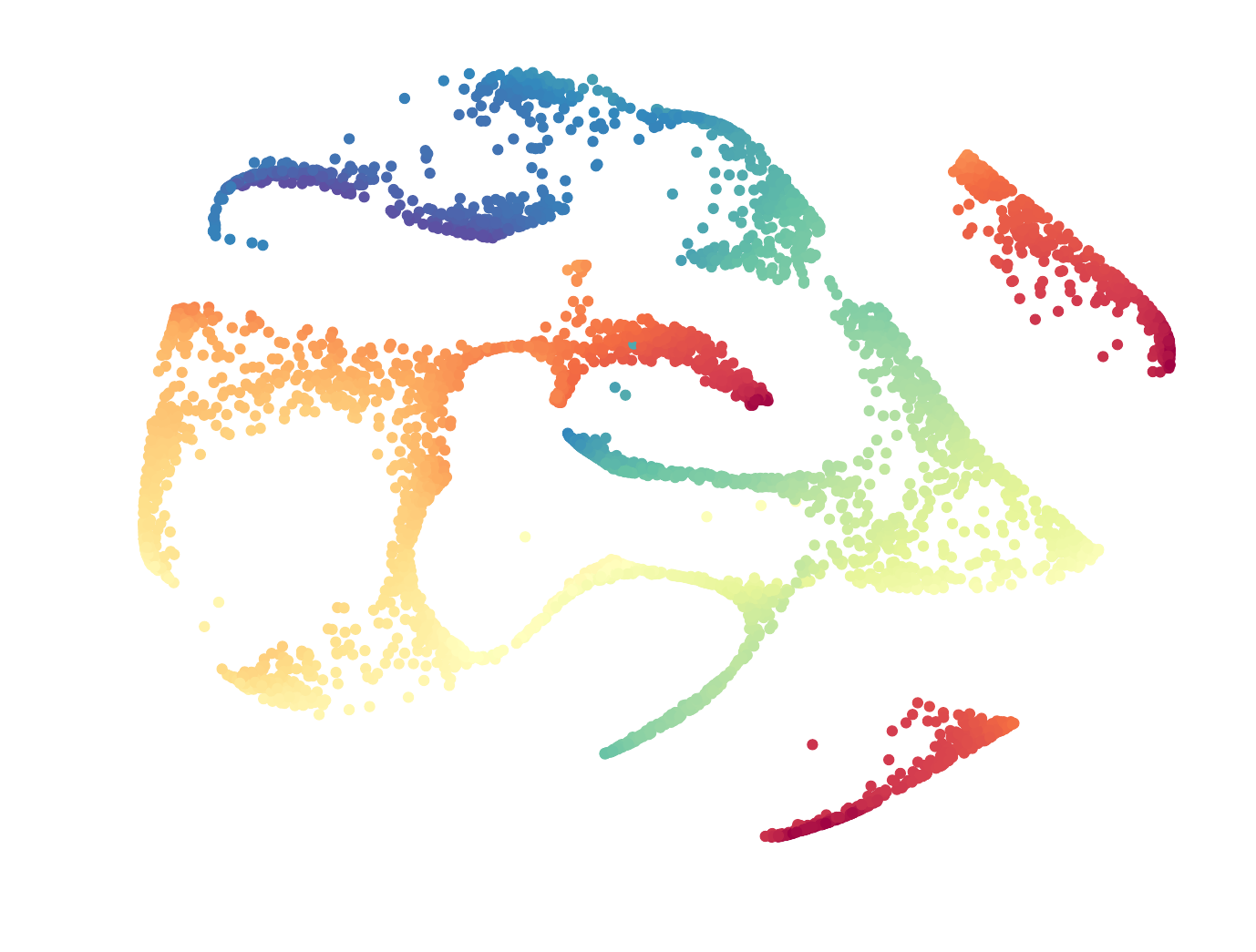}
            \end{minipage}   
            \begin{minipage}{0.195\textwidth}
            \includegraphics[width=\linewidth]{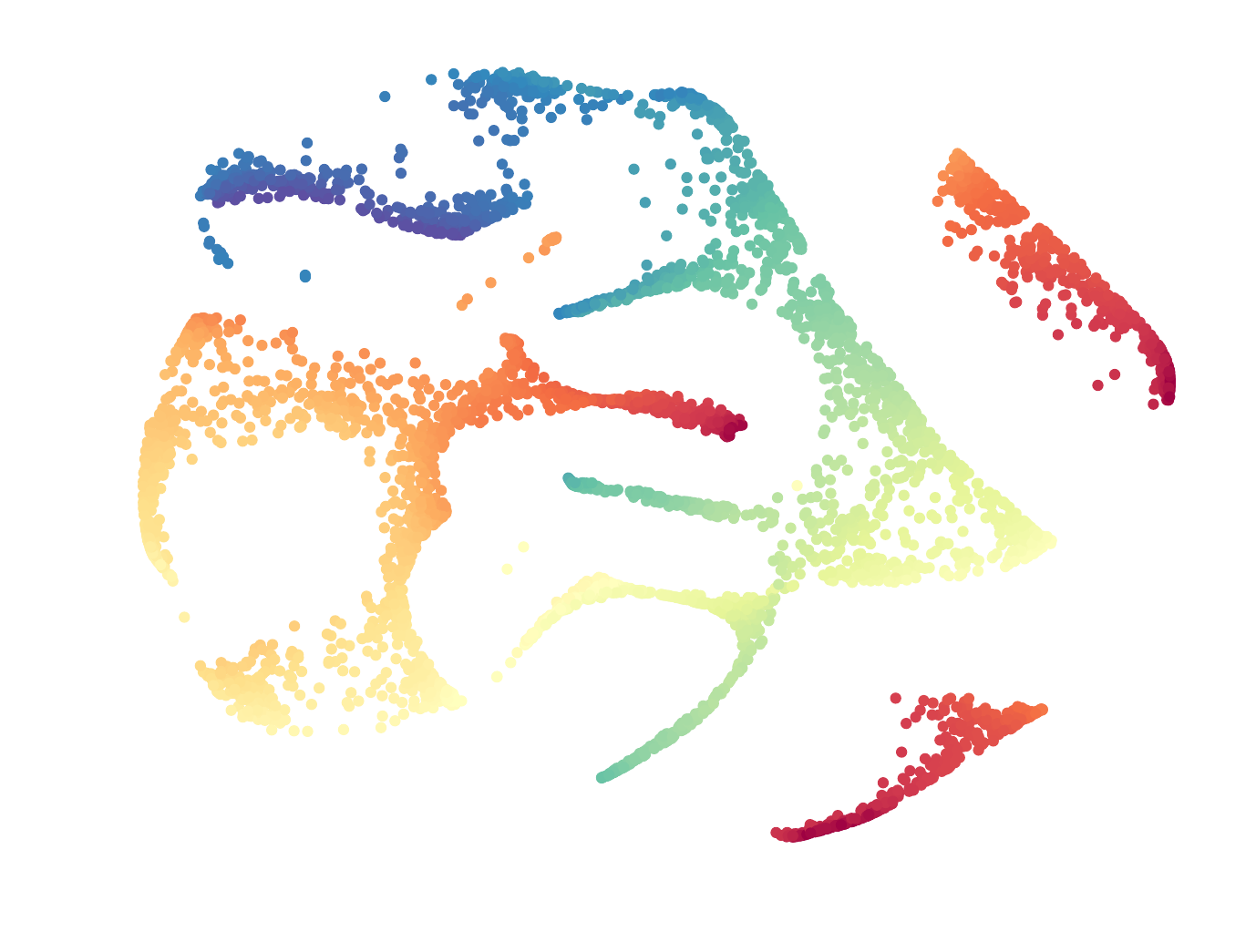}
            \end{minipage}         
            \begin{minipage}{0.195\textwidth}
            \includegraphics[width=\linewidth]{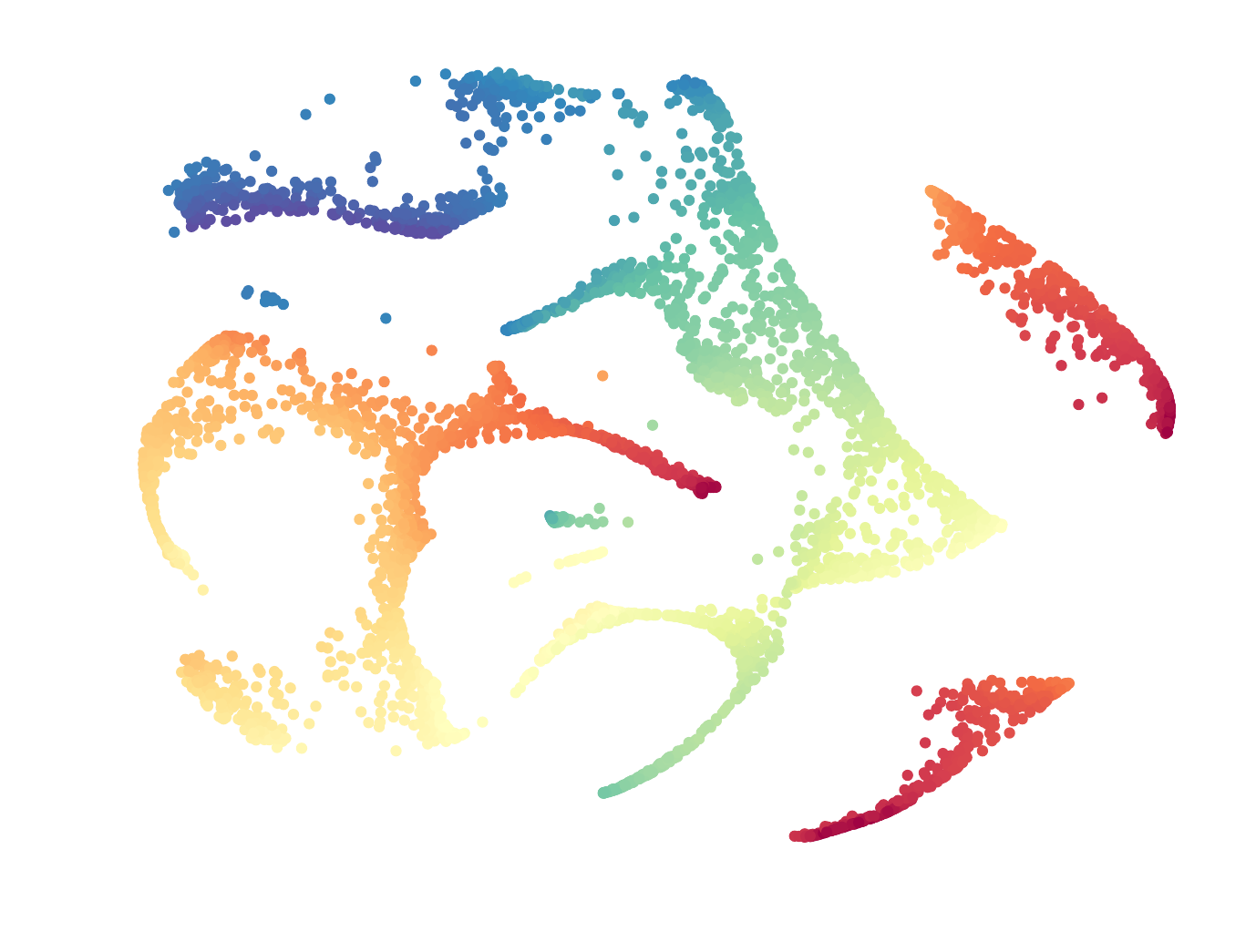}
            \end{minipage}  
            \begin{minipage}{0.195\textwidth}
            \includegraphics[width=\linewidth]{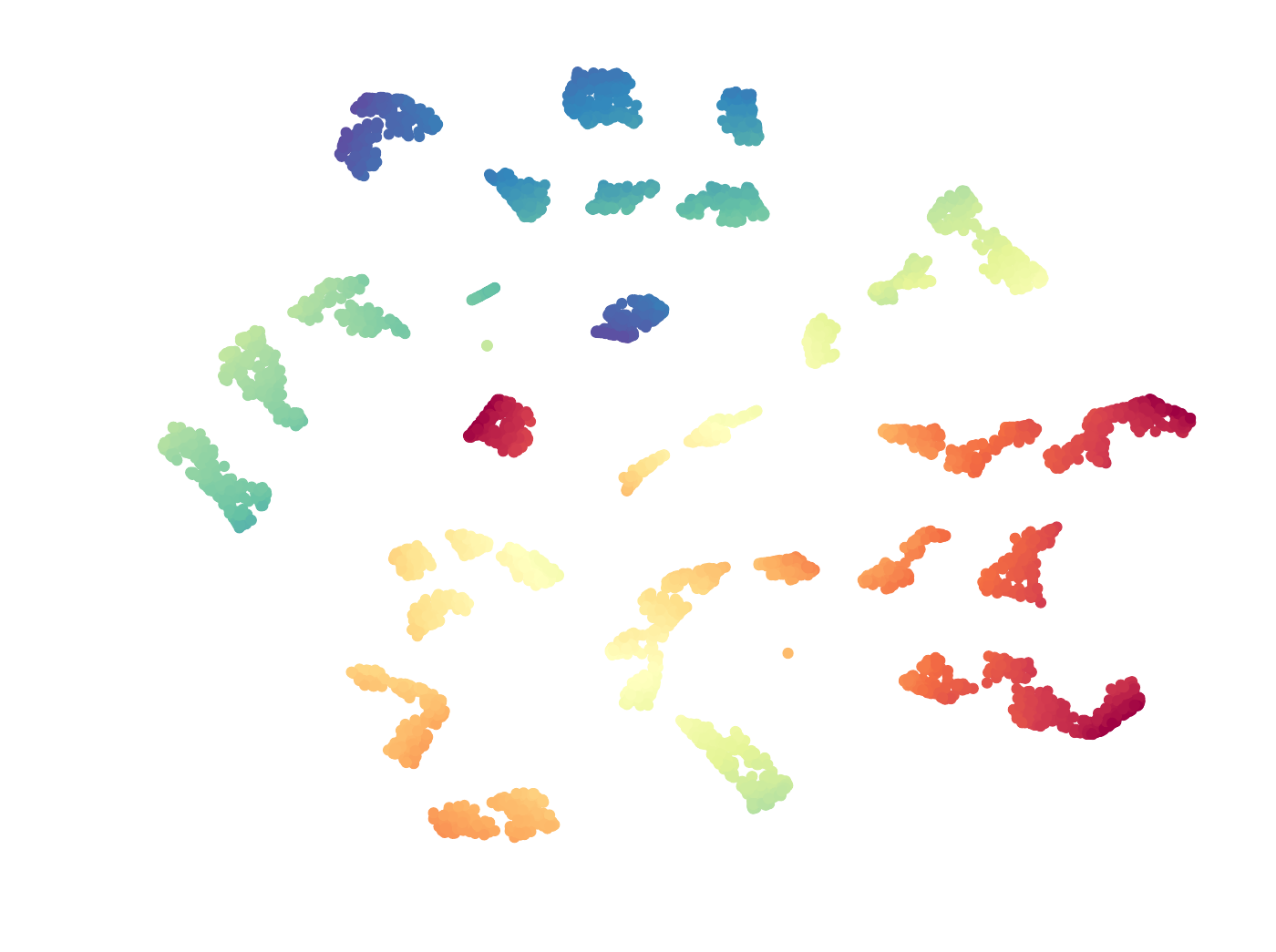}
            \end{minipage}  
            \begin{minipage}{0.195\textwidth}
            \includegraphics[width=\linewidth]{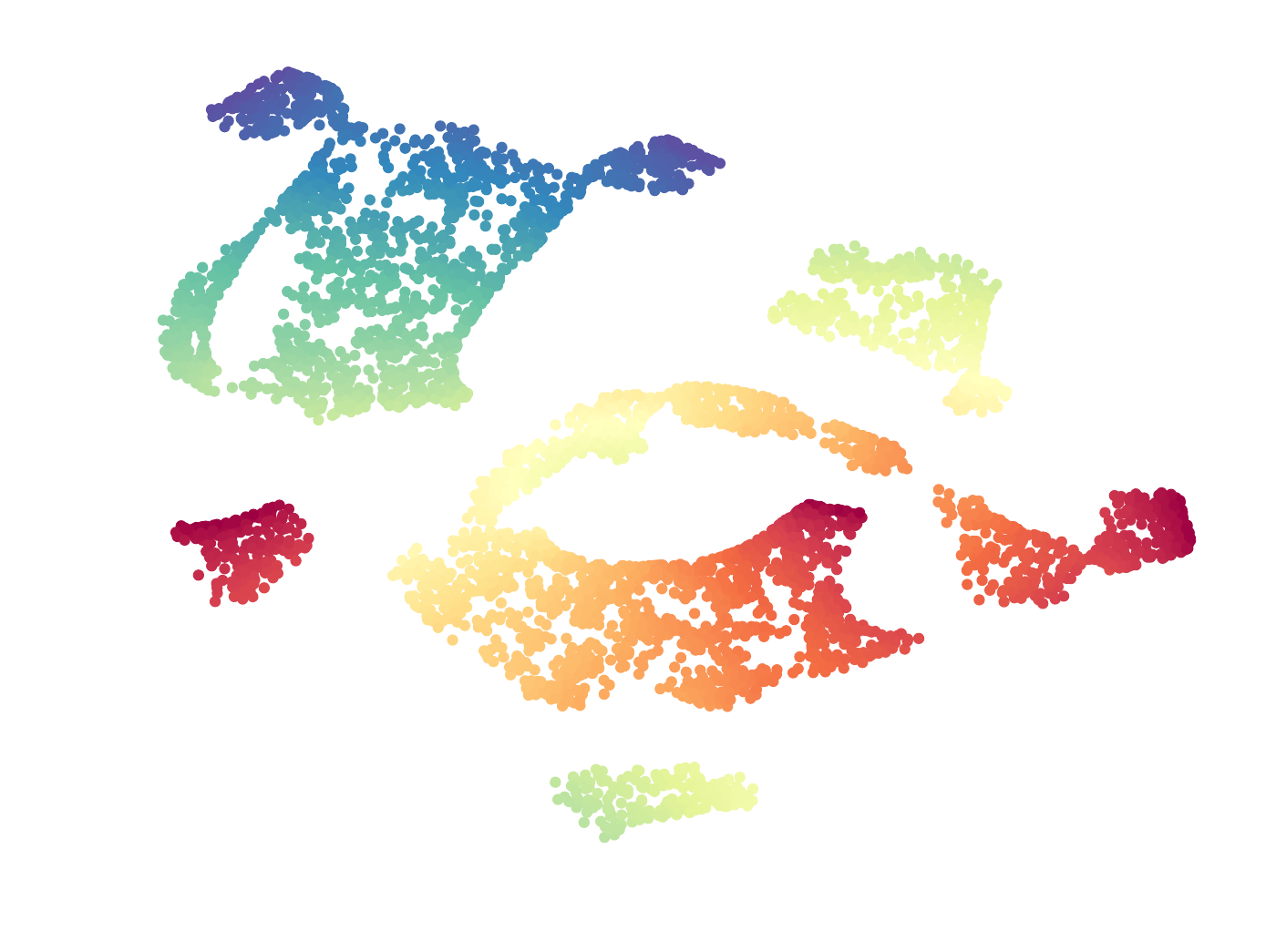}
            \end{minipage}  
            \subcaption{Perplexity=100} 
        \end{minipage}       
        \begin{minipage}{\textwidth}
            \begin{minipage}{0.195\textwidth}
            \includegraphics[width=\linewidth]{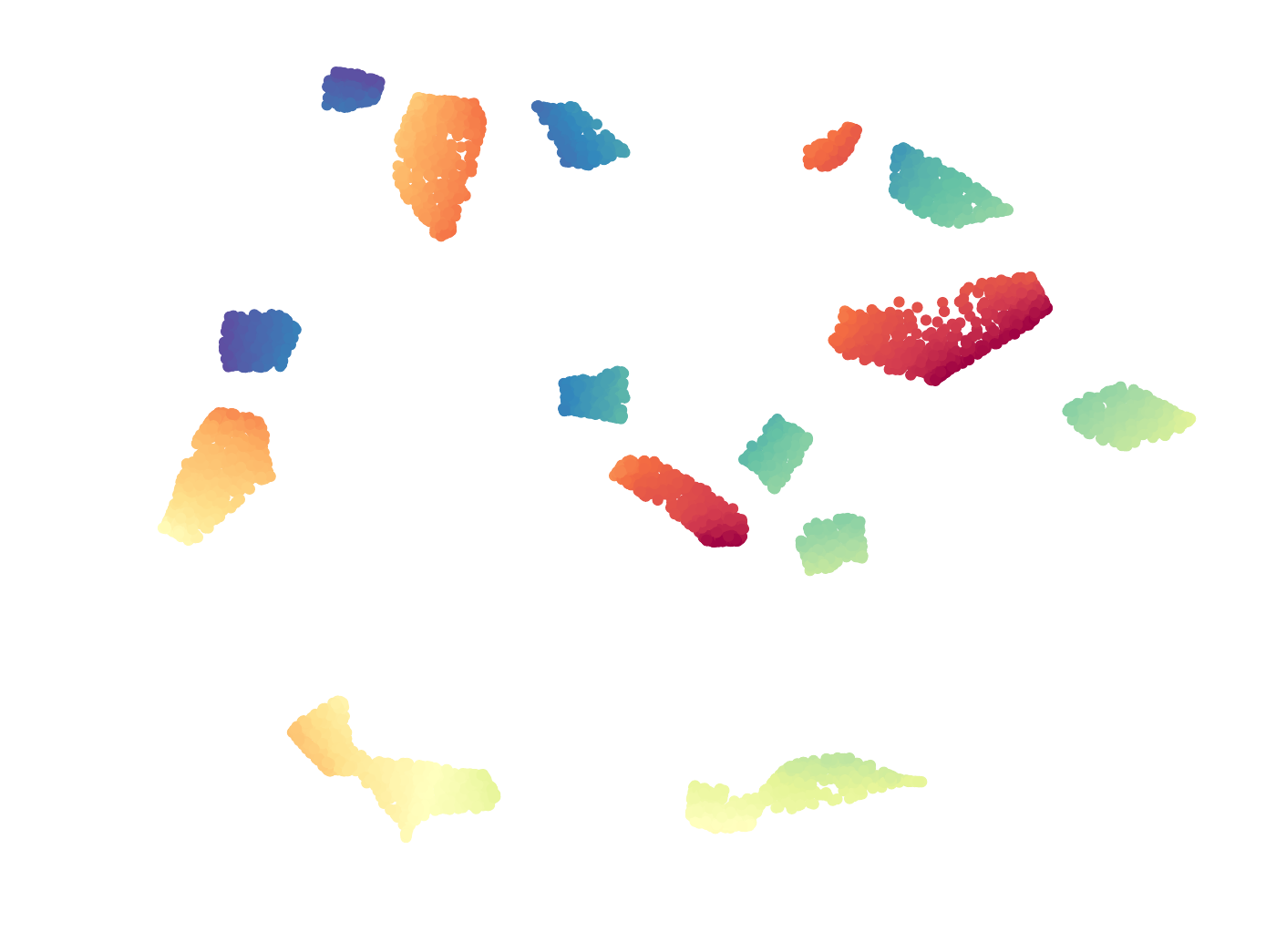}
            \end{minipage}   
            \begin{minipage}{0.195\textwidth}
            \includegraphics[width=\linewidth]{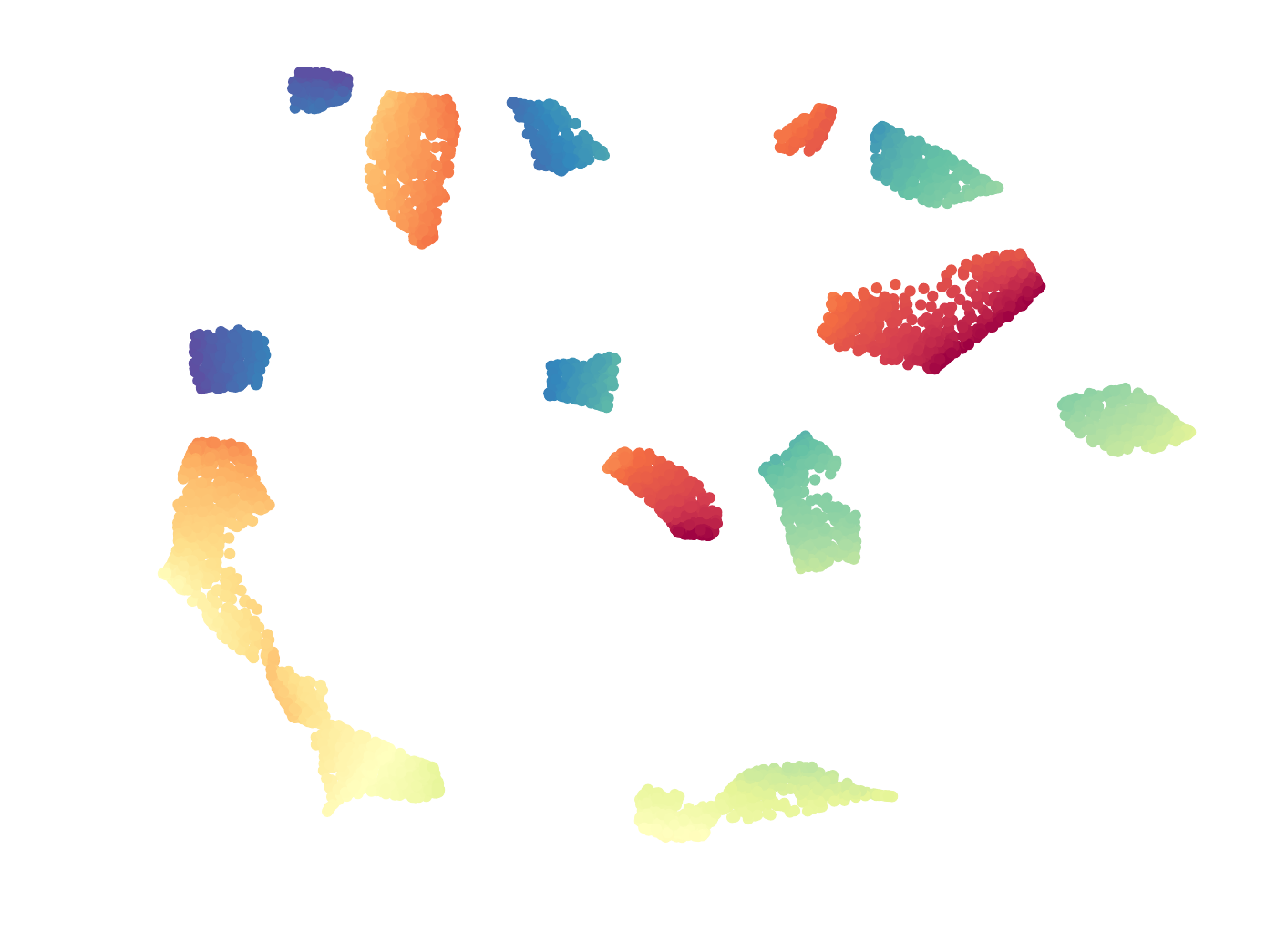}
            \end{minipage}         
            \begin{minipage}{0.195\textwidth}
            \includegraphics[width=\linewidth]{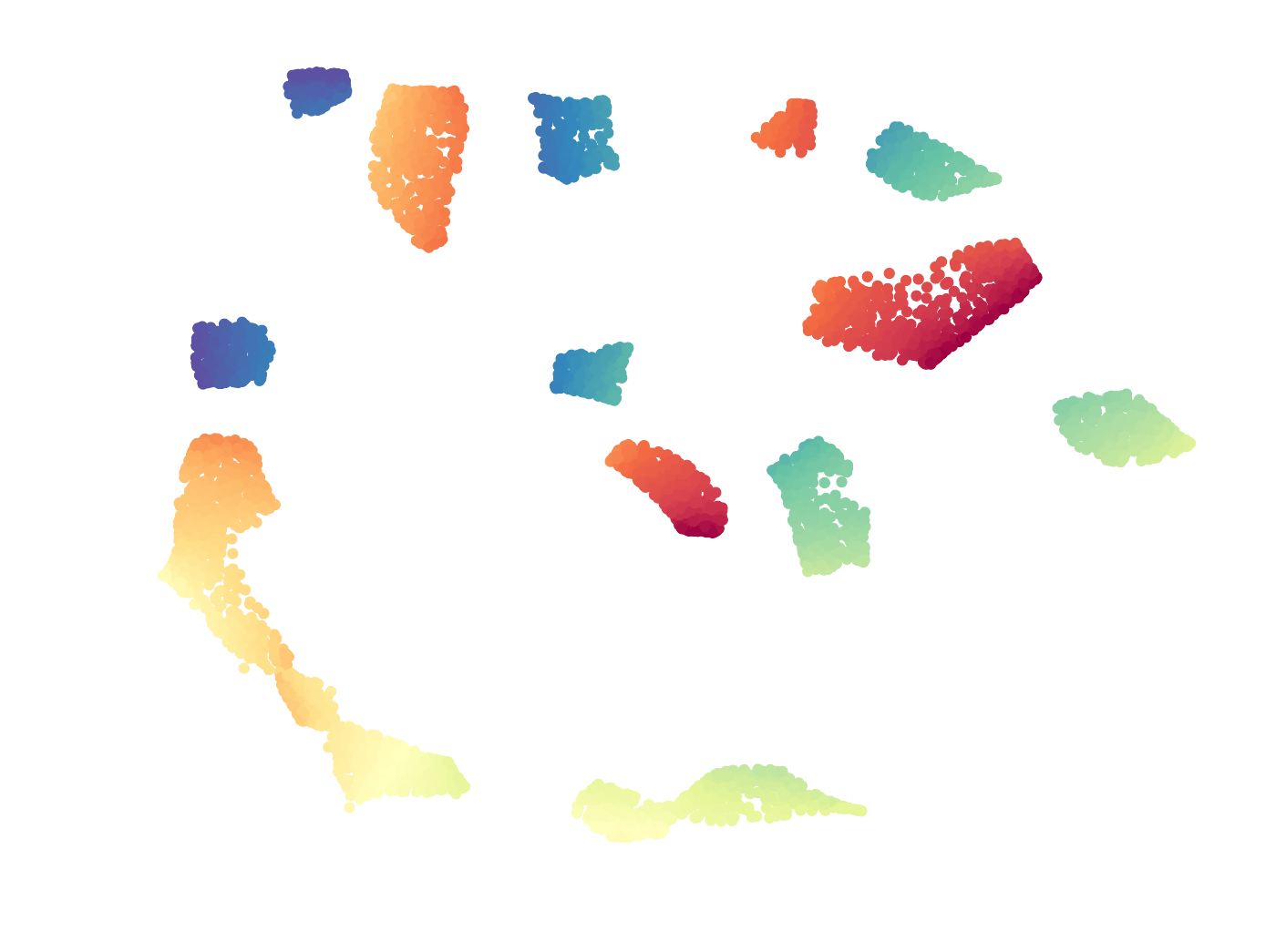}
            \end{minipage}  
            \begin{minipage}{0.195\textwidth}
            \includegraphics[width=\linewidth]{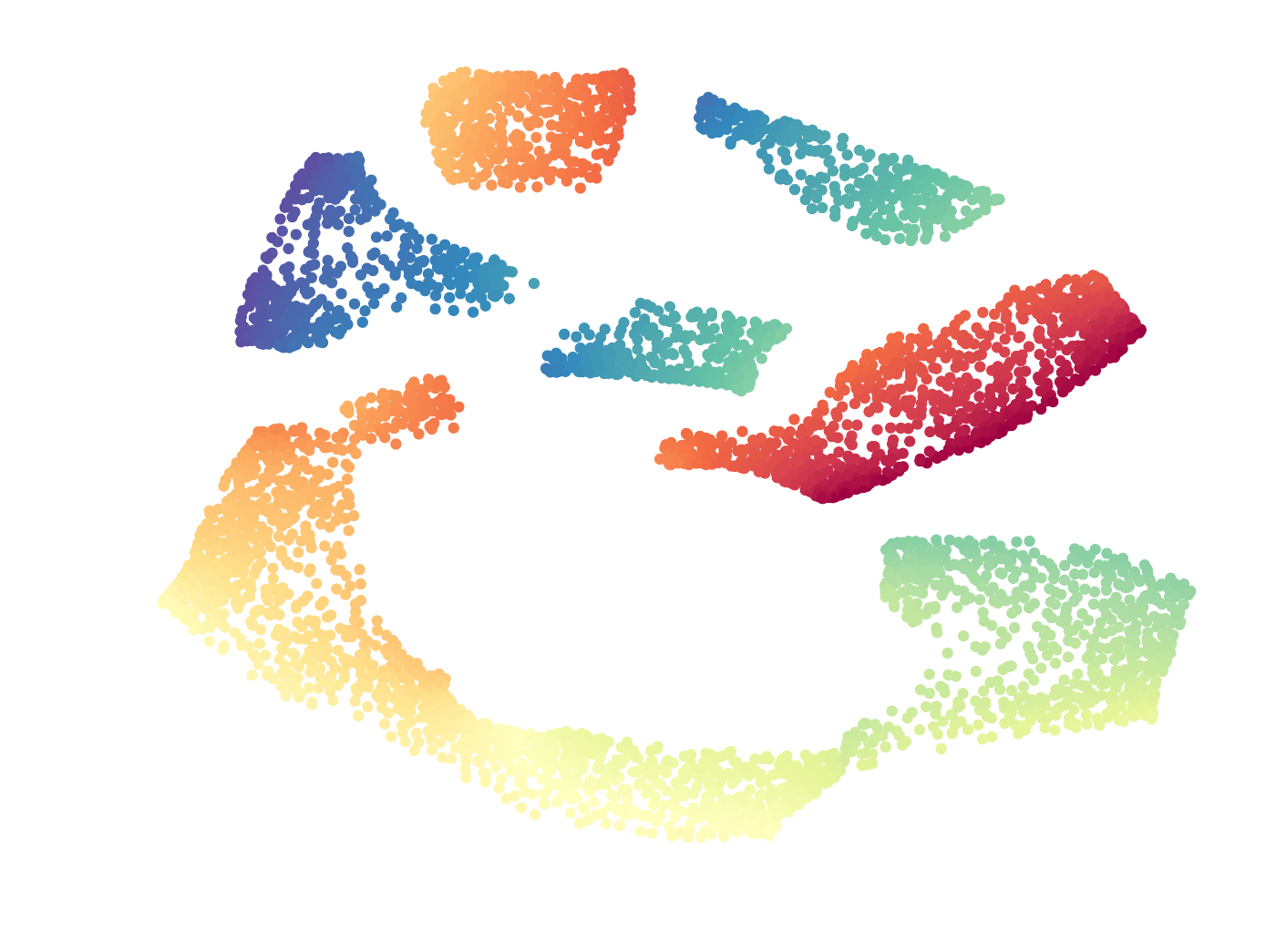}
            \end{minipage}  
            \begin{minipage}{0.195\textwidth}
            \includegraphics[width=\linewidth]{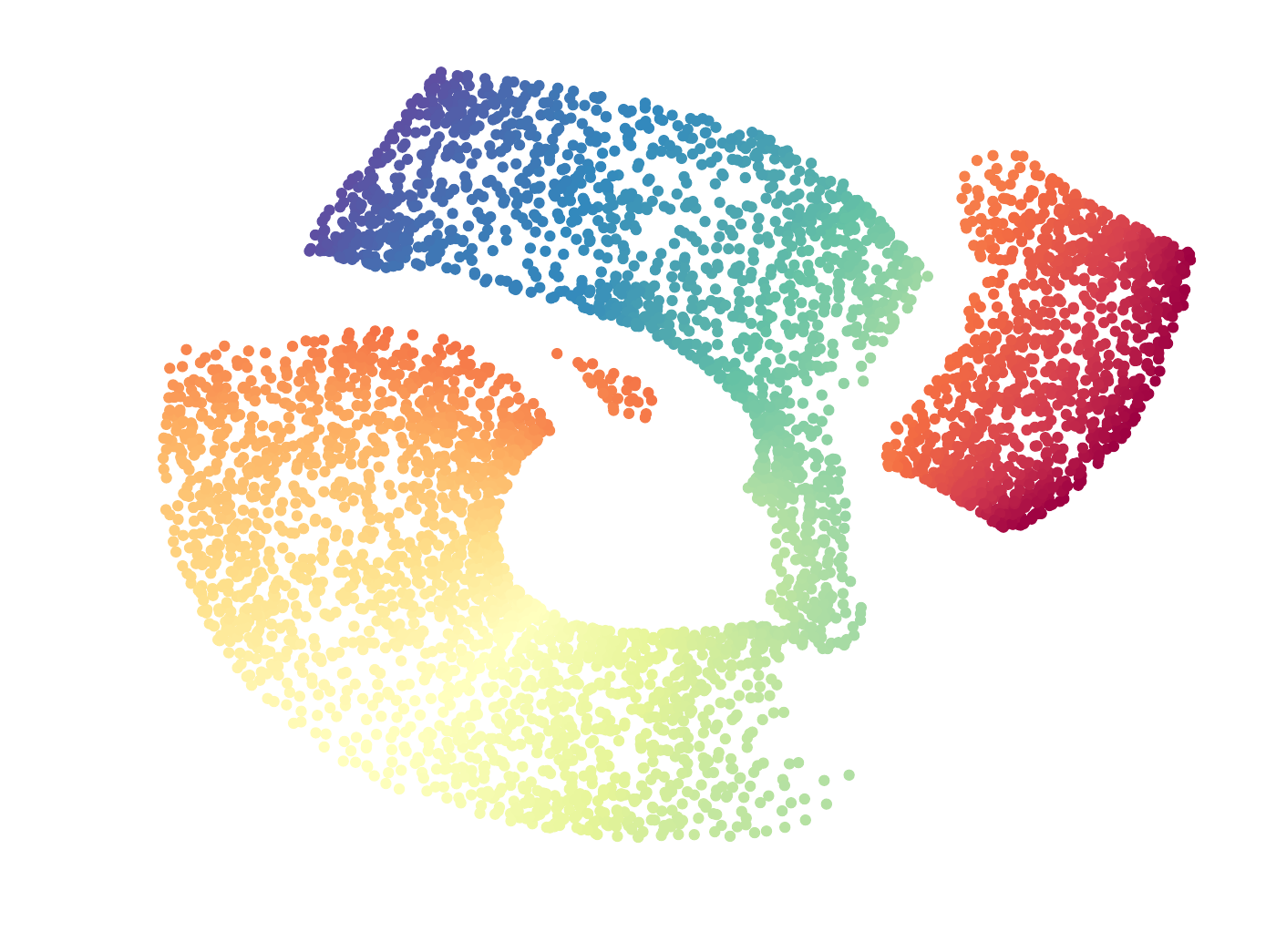}
            \end{minipage}  
            \subcaption{Perplexity=500} 
        \end{minipage}       
        \caption{$t$-SNE embeddings on Swiss Roll}  
        \label{fig:swiss_roll_embeddings}
    \end{figure}

    \pagebreak

    \begin{table}[htp]
    \begin{minipage}{0.49\textwidth}
      \centering
      \caption{F-Score on X-Y MNIST}
      \label{tab:results1}
        {\footnotesize
      \begin{tabular}{ccccc}\hline
         KL    &  RKL  &  JS    & HL   & CH \\ \hline\hline
          0.3524  & 0.4241 & 0.4190 &  0.4155 & 0.0922 \\
         0.4440  & 0.5699 & 0.5395 &  0.5332  & 0.1476 \\
         0.4724  & 0.6057 & 0.5570 &  0.5546  & 0.2349 \\
         0.4603  & 0.5687 & 0.5220 &  0.5217  & 0.3007 \\
         0.4407  & 0.5234 & 0.4843 &  0.4843  & 0.3222 \\
         0.4202  & 0.4798 & 0.4491 &  0.4503  & 0.3350 \\
         0.4016  & 0.4413 & 0.4189 &  0.4210  & 0.3420 \\
         0.3836  & 0.4090 & 0.3939 &  0.3964  & 0.3449 \\
         0.3667  & 0.3814 & 0.3722 &  0.3745  & 0.3446 \\\hline
      \end{tabular}}
      \vspace{-0.5cm}
    \end{minipage}
    \begin{minipage}{0.49\textwidth}
      \centering
      \caption{F-Score on X-Y FACE}
      \label{tab:results}
        {\footnotesize
      \begin{tabular}{ccccc}\hline
         KL    &  RKL  &  JS    & HL   & CH \\ \hline\hline
         0.4019 & 0.4186 & 0.4156 & 0.4115 & 0.2112 \\
         0.5648 & 0.6446 & 0.6216 & 0.6197 & 0.2919 \\
         0.6236 & 0.7534 & 0.7200 & 0.7146 & 0.3505 \\
         0.5865 & 0.6970 & 0.6793 & 0.6764 & 0.3674 \\
         0.5354 & 0.6207 & 0.6105 & 0.6095 & 0.3689 \\
         0.4870 & 0.5447 & 0.5441 & 0.5412 & 0.3642 \\
         0.4464 & 0.4841 & 0.4862 & 0.4826 & 0.3573 \\ \hline
      \end{tabular}}
      \vspace{-0.5cm}
    \end{minipage}
    \end{table}
    
    \begin{table}[htp]
    \begin{minipage}{0.49\textwidth}
      \centering
      \caption{F-Score on X-Y MNIST}
      \label{tab:results}
        {\footnotesize
      \begin{tabular}{ccccc}\hline
         KL    &  RKL  &  JS    & HL   & CH \\ \hline\hline
         0.4795 & 0.4444 & 0.4693 & 0.4494 & 0.4787 \\
         0.5938 & 0.5805 & 0.6006 & 0.5667 & 0.6109 \\
         0.5466 & 0.5872 & 0.5834 & 0.5426 & 0.5772 \\
         0.4891 & 0.5334 & 0.5256 & 0.4895 & 0.5200 \\
         0.4457 & 0.4868 & 0.4783 & 0.4494 & 0.4742 \\
         0.4149 & 0.4495 & 0.4409 & 0.4186 & 0.4391 \\
         0.3925 & 0.4180 & 0.4103 & 0.3950 & 0.4112 \\
         0.3749 & 0.3920 & 0.3856 & 0.3757 & 0.3879 \\
         0.3597 & 0.3699 & 0.3657 & 0.3595 & 0.3688 \\\hline
      \end{tabular}}
      \vspace{-0.5cm}
    \end{minipage}
    \begin{minipage}{0.49\textwidth}
      \centering
      \caption{F-Score on X-Y NEWS}
      \label{tab:results}
        {\footnotesize
      \begin{tabular}{ccccc}\hline
         KL    &  RKL  &  JS    & HL   & CH \\ \hline\hline
         0.3996 & 0.3665 & 0.3922 & 0.3989 & 0.2637 \\
         0.4256 & 0.4108 & 0.4328 & 0.4331 & 0.3703 \\
         0.3820 & 0.3933 & 0.4001 & 0.3964 & 0.4466 \\
         0.3387 & 0.3569 & 0.3559 & 0.3526 & 0.4633 \\
         0.3062 & 0.3209 & 0.3188 & 0.3163 & 0.4599 \\
         0.2796 & 0.2899 & 0.2877 & 0.2862 & 0.4476 \\
         0.2562 & 0.2626 & 0.2610 & 0.2594 & 0.4297 \\
         0.2346 & 0.2380 & 0.2370 & 0.2357 & 0.4076 \\
         0.2145 & 0.2162 & 0.2157 & 0.2148 & 0.3824 \\\hline 
      \end{tabular}}
      \vspace{-0.5cm}
    \end{minipage}
    \end{table}
    \begin{table}[htp]
    \begin{minipage}{0.49\textwidth}
      \centering
      \caption{F-Score on X-Y IMAGENET SBOW}
      \label{tab:results}
        {\footnotesize
      \begin{tabular}{ccccc}\hline
         KL    &  RKL  &  JS    & HL   & CH \\ \hline\hline
         0.4317 & 0.3411 & 0.3456 & 0.2431 & 0.4395 \\
         0.4686 & 0.3825 & 0.3823 & 0.2977 & 0.4889 \\
         0.4297 & 0.3635 & 0.3638 & 0.3157 & 0.4493 \\
         0.3747 & 0.3259 & 0.3280 & 0.3066 & 0.3838 \\
         0.3265 & 0.2932 & 0.2941 & 0.2914 & 0.3320 \\
         0.2867 & 0.2635 & 0.2659 & 0.2728 & 0.2919 \\
         0.2553 & 0.2403 & 0.2421 & 0.2532 & 0.2614 \\
         0.2306 & 0.2218 & 0.2222 & 0.2336 & 0.2373 \\
         0.2111 & 0.2065 & 0.2061 & 0.2149 & 0.2177 \\\hline 
      \end{tabular}}
      \vspace{-0.5cm}
    \end{minipage}
    \end{table}

    \begin{table}[htp]
    \centering
    \begin{minipage}{0.49\textwidth}
      \caption{F-Score on X-Z FACE}
      \label{tab:results}
        {\footnotesize
      \begin{tabular}{ccccc}\hline
         KL    &  RKL  &  JS    & HL   & CH \\ \hline\hline
         0.1444 & 0.1461 & 0.1482 & 0.1464 & 0.0930 \\
         0.2220 & 0.2318 & 0.2370 & 0.2327 & 0.1434 \\
         0.3012 & 0.3284 & 0.3262 & 0.3217 & 0.2043 \\
         0.3322 & 0.3511 & 0.3585 & 0.3524 & 0.2405 \\
         0.3421 & 0.3401 & 0.3652 & 0.3596 & 0.2645 \\
         0.3431 & 0.3267 & 0.3615 & 0.3557 & 0.2807 \\
         0.3396 & 0.3171 & 0.3531 & 0.3486 & 0.2915 \\
         0.3337 & 0.3068 & 0.3423 & 0.3384 & 0.2965 \\
         0.3249 & 0.3015 & 0.3278 & 0.3261 & 0.2983 \\\hline        
      \end{tabular}}
      \vspace{-0.5cm}
    \end{minipage}
    \begin{minipage}{0.49\textwidth}
      \centering
      \caption{F-Score on X-Z MNIST}
      \label{tab:results}
        {\footnotesize
      \begin{tabular}{ccccc}\hline
         KL    &  RKL  &  JS    & HL   & CH \\ \hline\hline
         0.3872 & 0.3503 & 0.3810 & 0.3686 & 0.3783 \\
         0.6137 & 0.5269 & 0.6007 & 0.5622 & 0.5967 \\
         0.7238 & 0.6128 & 0.7023 & 0.6584 & 0.6884 \\
         0.7531 & 0.6406 & 0.7246 & 0.6958 & 0.7054 \\
         0.6903 & 0.5974 & 0.6649 & 0.6826 & 0.6524 \\
         0.6027 & 0.5379 & 0.5795 & 0.6224 & 0.5698 \\
         0.5277 & 0.4850 & 0.5073 & 0.5537 & 0.4989 \\
         0.4667 & 0.4344 & 0.4472 & 0.4940 & 0.4414 \\
         0.4186 & 0.3923 & 0.3994 & 0.4448 & 0.3958 \\\hline
      \end{tabular}}
      \vspace{-0.5cm}
    \end{minipage}
    \end{table}
    \begin{table}[htp]
    \begin{minipage}{0.49\textwidth}
      \centering
      \caption{F-Score on X-Z NEWS}
      \label{tab:results}
        {\footnotesize
      \begin{tabular}{ccccc}\hline
         KL    &  RKL  &  JS    & HL   & CH \\ \hline\hline
         0.0066 & 0.0076 & 0.0074 & 0.0076 & 0.0069 \\
         0.0036 & 0.0041 & 0.0038 & 0.0040 & 0.0034 \\
         0.0018 & 0.0020 & 0.0020 & 0.0020 & 0.0018 \\
         0.0013 & 0.0014 & 0.0015 & 0.0015 & 0.0014 \\
         0.0011 & 0.0011 & 0.0012 & 0.0012 & 0.0013 \\
         0.0011 & 0.0010 & 0.0011 & 0.0011 & 0.0011 \\
         0.0010 & 0.0009 & 0.0010 & 0.0009 & 0.0011 \\
         0.0010 & 0.0009 & 0.0009 & 0.0009 & 0.0010 \\
         0.0010 & 0.0009 & 0.0009 & 0.0009 & 0.0010 \\\hline
      \end{tabular}}
      \vspace{-0.5cm}
    \end{minipage}
    \begin{minipage}{0.49\textwidth}
      \centering
      \caption{F-Score on X-Z IMAGENET SBOW}
      \label{tab:results12}
        {\footnotesize
      \begin{tabular}{ccccc}\hline
         KL    &  RKL  &  JS    & HL   & CH \\ \hline\hline
         0.0063 & 0.0051 & 0.0044 & 0.0052 & 0.0043 \\
         0.0025 & 0.0021 & 0.0025 & 0.0027 & 0.0023 \\
         0.0016 & 0.0014 & 0.0017 & 0.0018 & 0.0014 \\
         0.0012 & 0.0013 & 0.0013 & 0.0013 & 0.0011 \\
         0.0010 & 0.0011 & 0.0011 & 0.0011 & 0.0010 \\
         0.0009 & 0.0009 & 0.0009 & 0.0009 & 0.0009 \\
         0.0008 & 0.0008 & 0.0008 & 0.0008 & 0.0008 \\
         0.0007 & 0.0007 & 0.0007 & 0.0007 & 0.0007 \\
         0.0006 & 0.0006 & 0.0007 & 0.0006 & 0.0007 \\\hline
      \end{tabular}}
      \vspace{-0.5cm}
    \end{minipage}
    \end{table}

    \pagebreak
    \subsection{More Experimental Results : Optimization of the primal form versus variational form (Duality Gap) Analysis}

    \begin{table}[t]
      \centering
        \caption{Amount of time for $v$KL-SNE to achieve same level of loss as KL-SNE}
      \label{tab:speed_results}
        {\footnotesize
      \begin{tabular}{l|c|ccc}\hline
                                & KL-SNE & \multicolumn{3}{c}{$v$KL-SNE}     \\\hline
          Data                  &  -   & vSNE 20 hids & vSNE 10-20 hids & vSNE 5-10-20 hids \\\hline
          MNIST (Digit 1)       & 294s          & 230.1s       & {\bf 196.17s}   & 217.3s            \\\hline
          MNIST                 & 1280s         & 1239.84s   & {\bf 972.72s}   & 1171.05s          \\\hline
          News                  & {\bf 505.8s}  & 2003.48s   & 1910.08s        & 1676.73s          \\\hline
      \end{tabular}}
        \vspace{-0.5cm}
    \end{table}
    \begin{figure}[htp]
        \centering
        \includegraphics[width=0.35\linewidth]{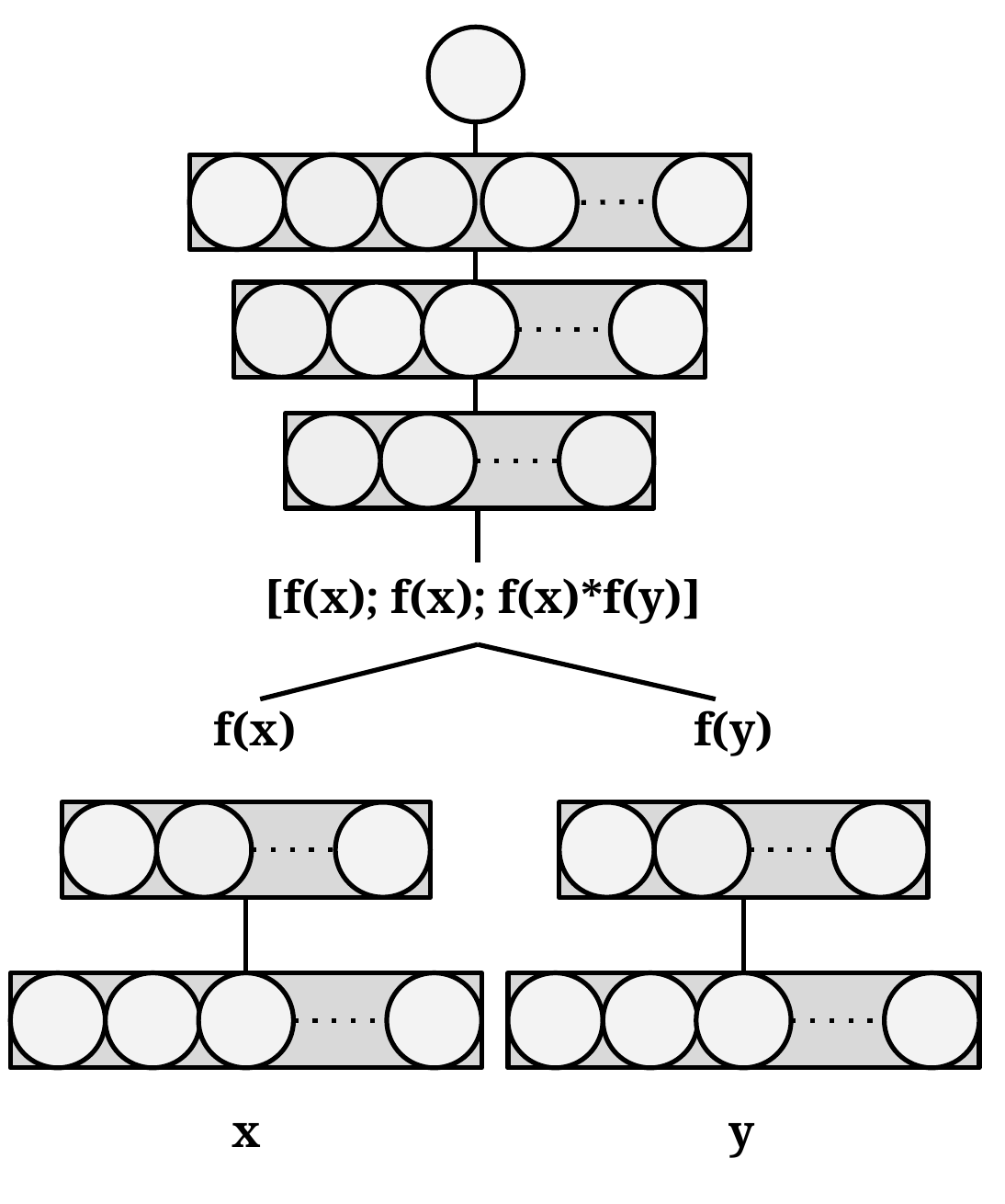}
        \vspace{-0.2cm}
        \caption{Discriminant Architecture \label{fig:disc_arch}}
    \end{figure}

    \begin{figure}[htp]
        \begin{minipage}{\textwidth}
            \begin{minipage}{0.325\textwidth}
            \includegraphics[width=\linewidth]{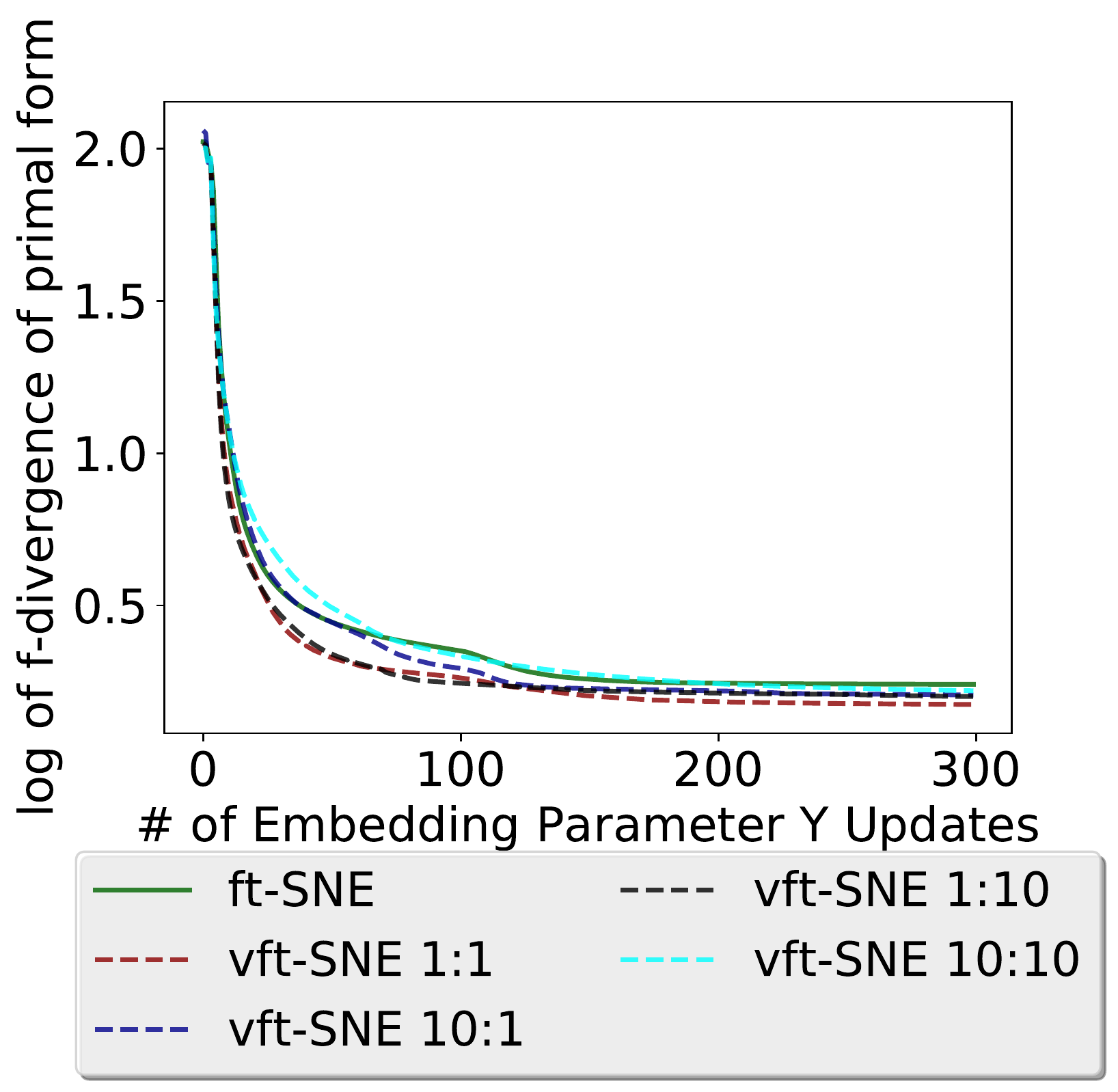}
            \vspace{-0.5cm}
            \subcaption*{CS-SNE}
            \end{minipage}            
            \begin{minipage}{0.325\textwidth}
            \includegraphics[width=\linewidth]{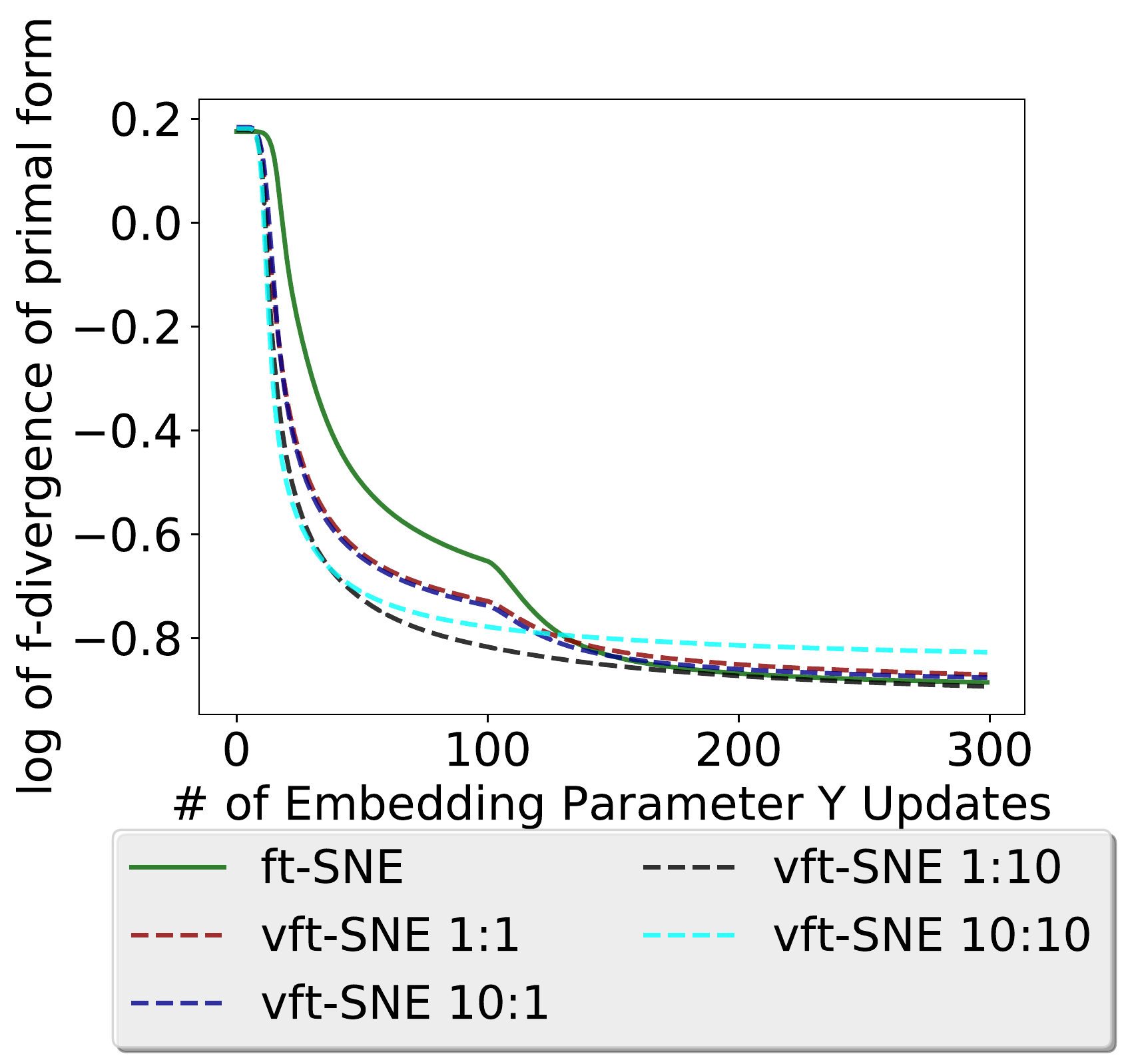}
            \vspace{-0.5cm}
            \subcaption*{JS-SNE}
            \end{minipage}            
            \begin{minipage}{0.325\textwidth}
            \includegraphics[width=\linewidth]{./figs/opt/mnist_cost_curves_JJKK_kl.pdf}
            \vspace{-0.5cm}
            \subcaption*{KL-SNE}
            \label{fig:kl_jjkk}
            \end{minipage}
            \subcaption{MNIST}
        \end{minipage}        
        \begin{minipage}{0.5\textwidth}
            \begin{minipage}{0.485\textwidth}
            \includegraphics[width=\linewidth]{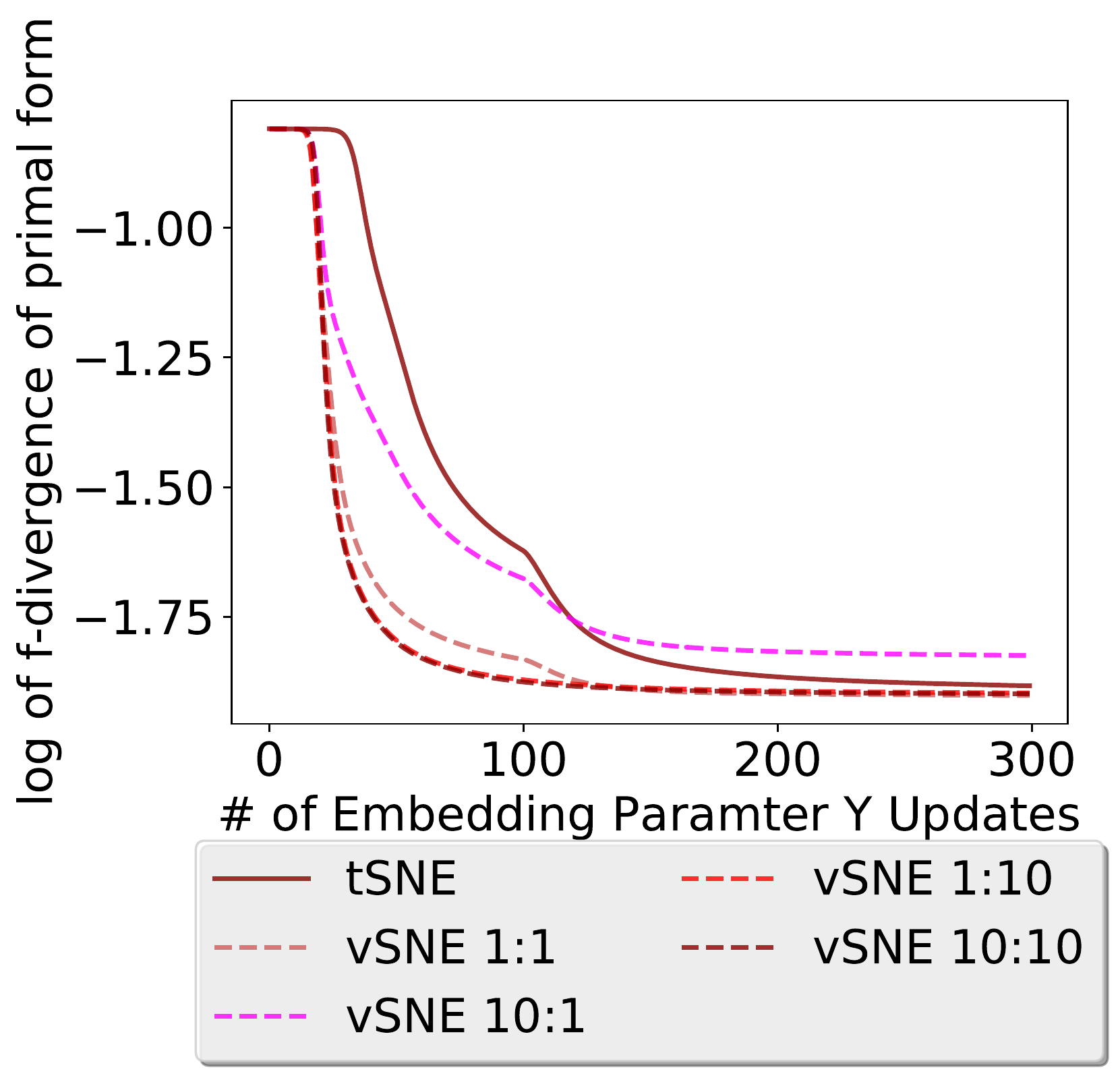}
            \vspace{-0.5cm}
            \subcaption*{MNIST1}
            \end{minipage}
            \begin{minipage}{0.485\textwidth}
            \includegraphics[width=\linewidth]{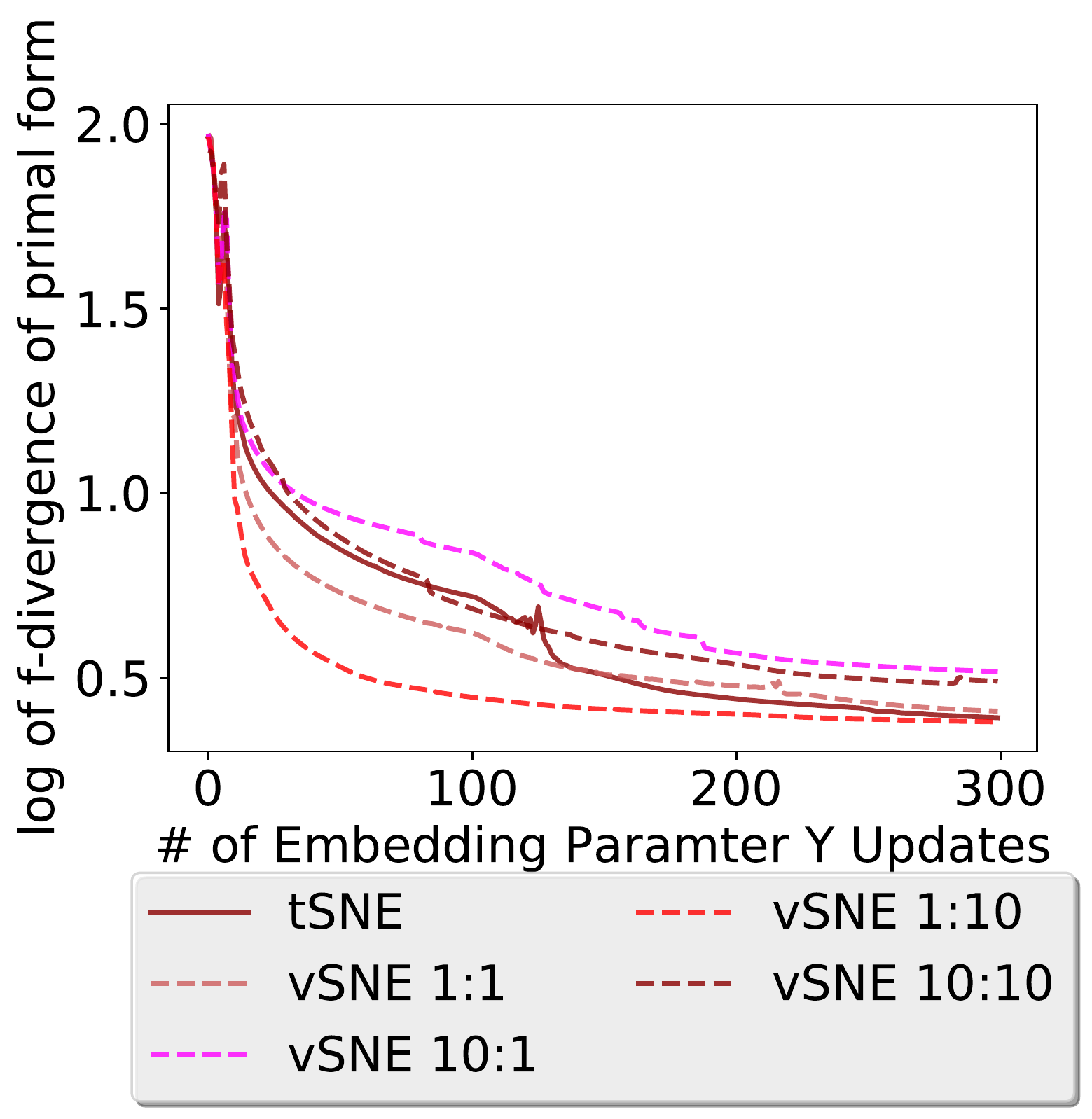}
            \vspace{-0.5cm}
            \subcaption*{NEWS}
            \end{minipage} 
            \subcaption{Chi-Square}
        \end{minipage}        
        \begin{minipage}{0.5\textwidth}
            \begin{minipage}{0.485\textwidth}
            \includegraphics[width=\linewidth]{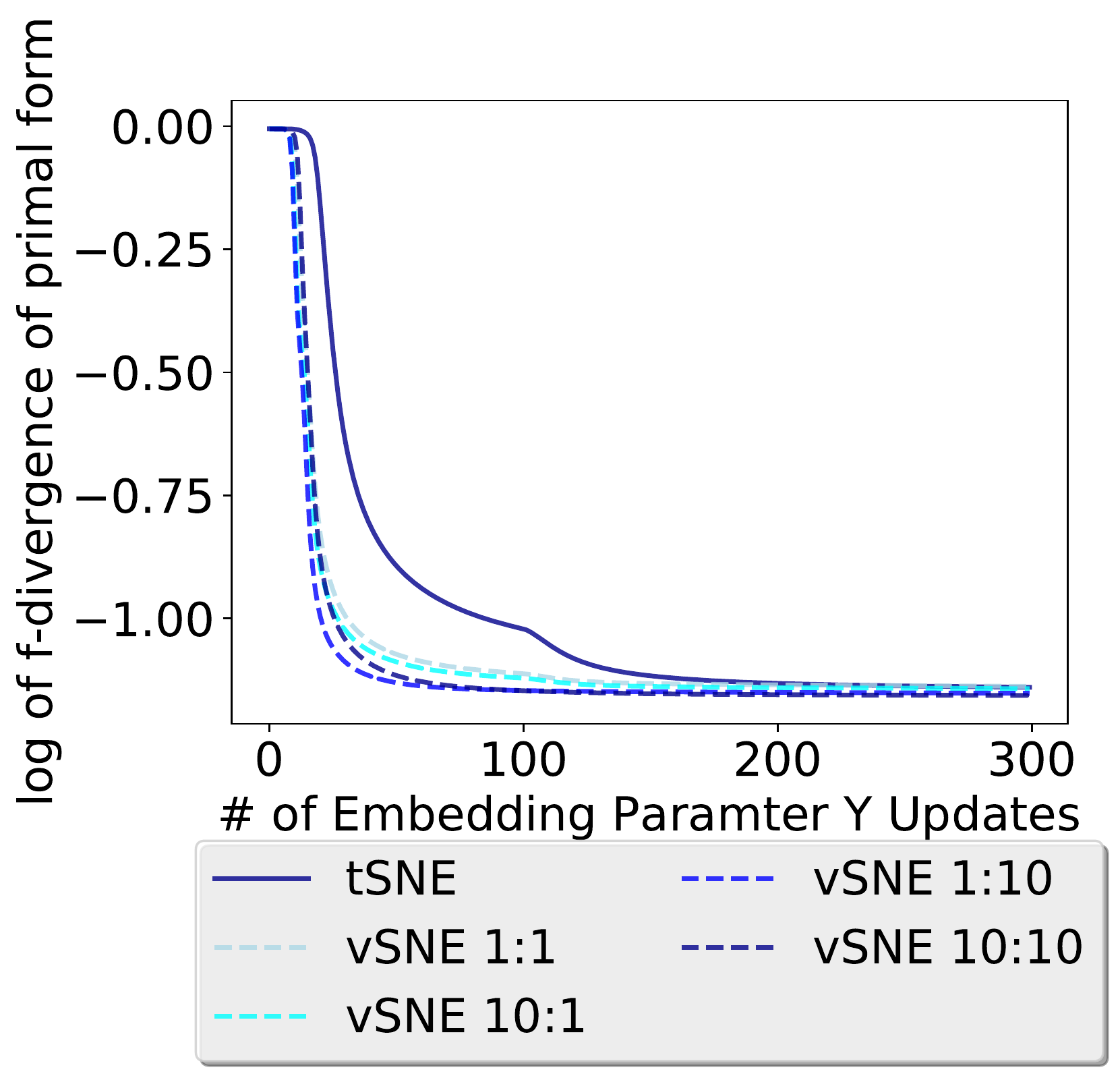}
            \vspace{-0.5cm}
            \subcaption*{MNIST1}
            \end{minipage}
            \begin{minipage}{0.485\textwidth}
            \includegraphics[width=\linewidth]{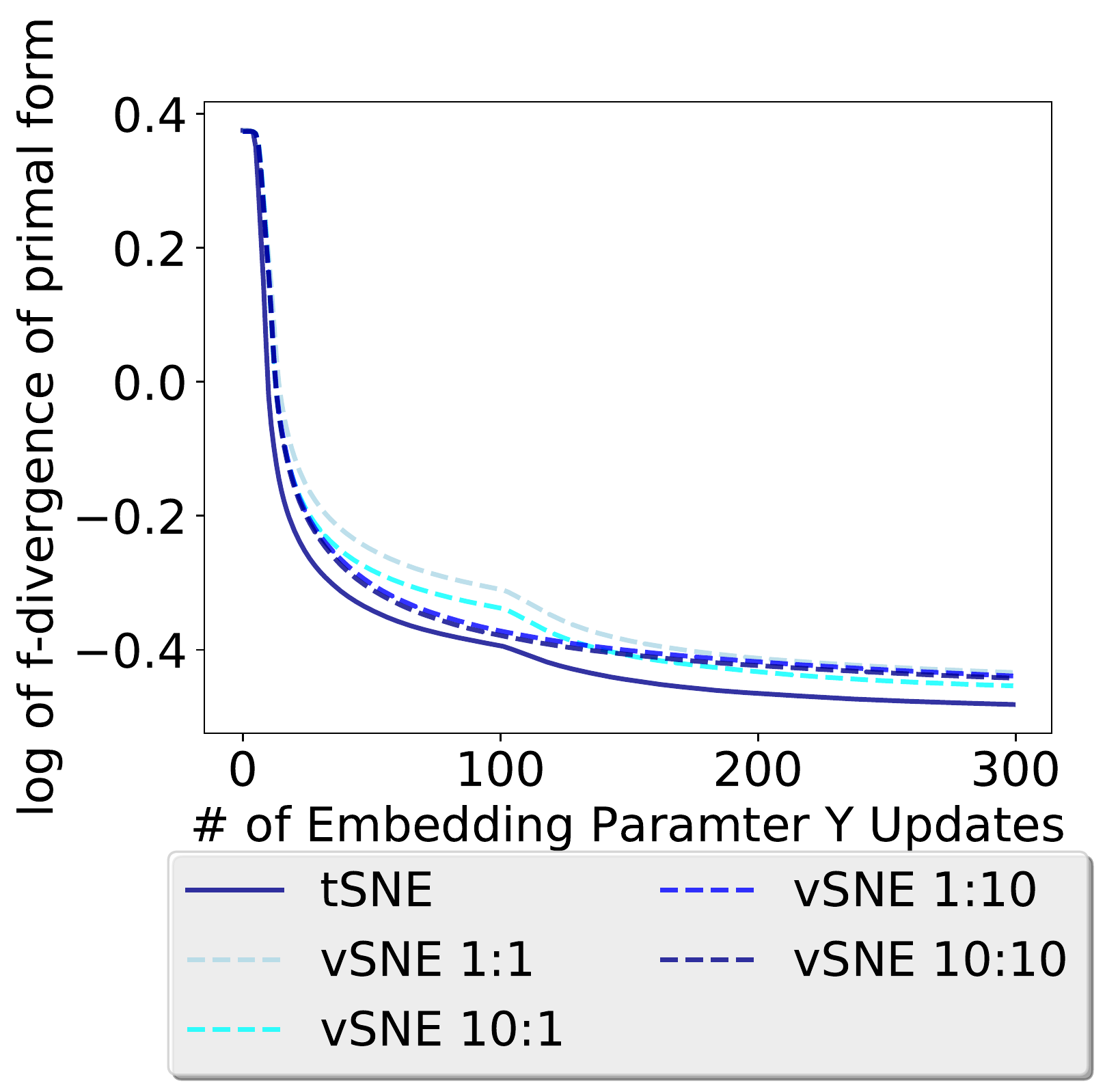}
            \vspace{-0.5cm}
            \subcaption*{NEWS}
            \end{minipage} 
        \subcaption{Jensen-Shannon}
        \end{minipage}        
        \caption{Log $ft$-SNE criterion during optimization for different choices of $J$ and $K$ on MNIST, MNIST1, and NEWS. 
                Two hidden layer (10-20) deep ReLU neural network was used as the discriminator and perplexity was set to 2,000.}
        \label{fig:JJ_KK}
    \end{figure}

    \begin{figure}[htp]
        \centering
        \begin{minipage}{0.325\textwidth}
            \includegraphics[width=\linewidth]{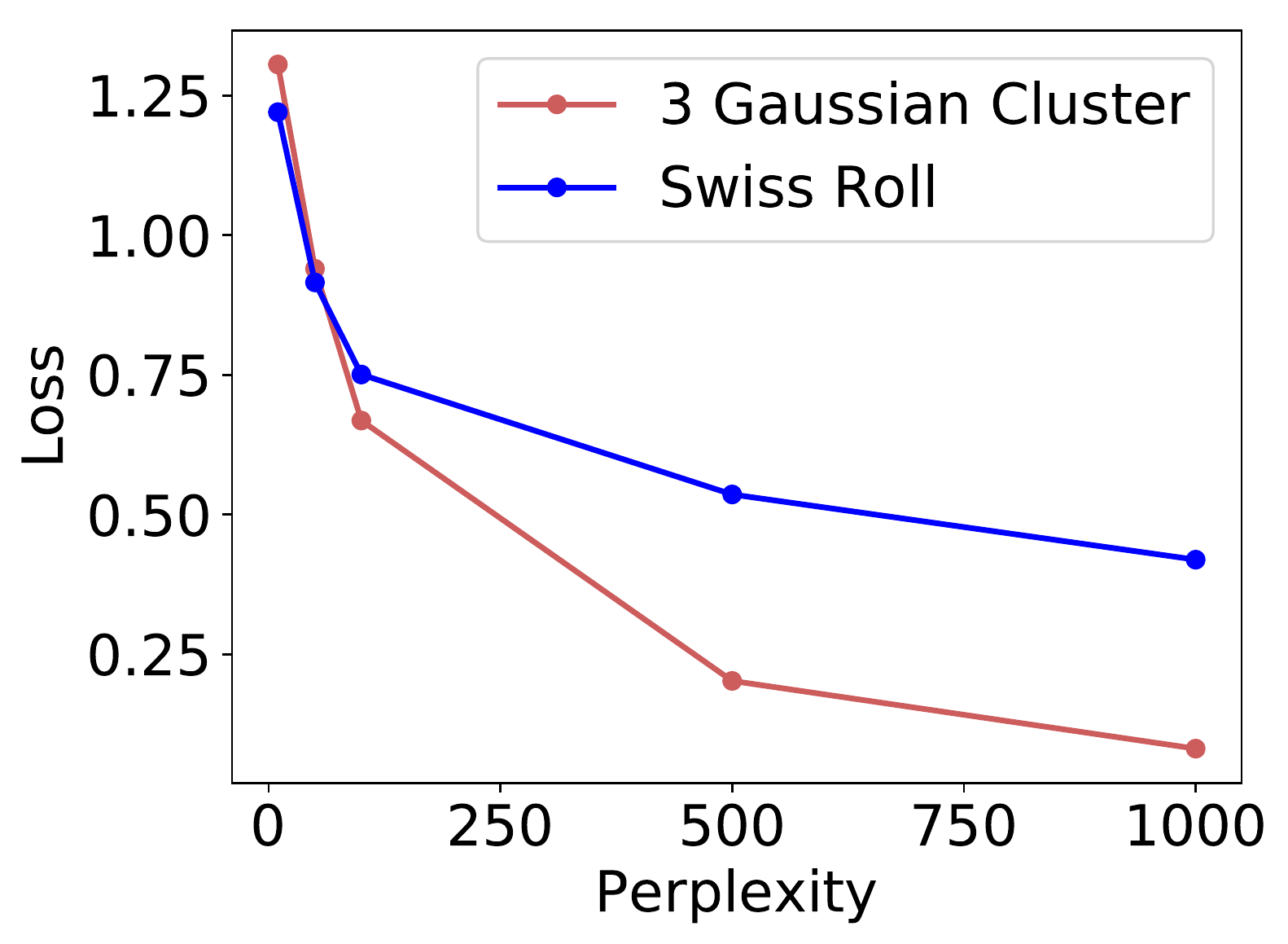}
            \vspace{-0.5cm}
            \subcaption*{KL-SNE}
         \end{minipage}  
        \begin{minipage}{0.325\textwidth}
            \includegraphics[width=\linewidth]{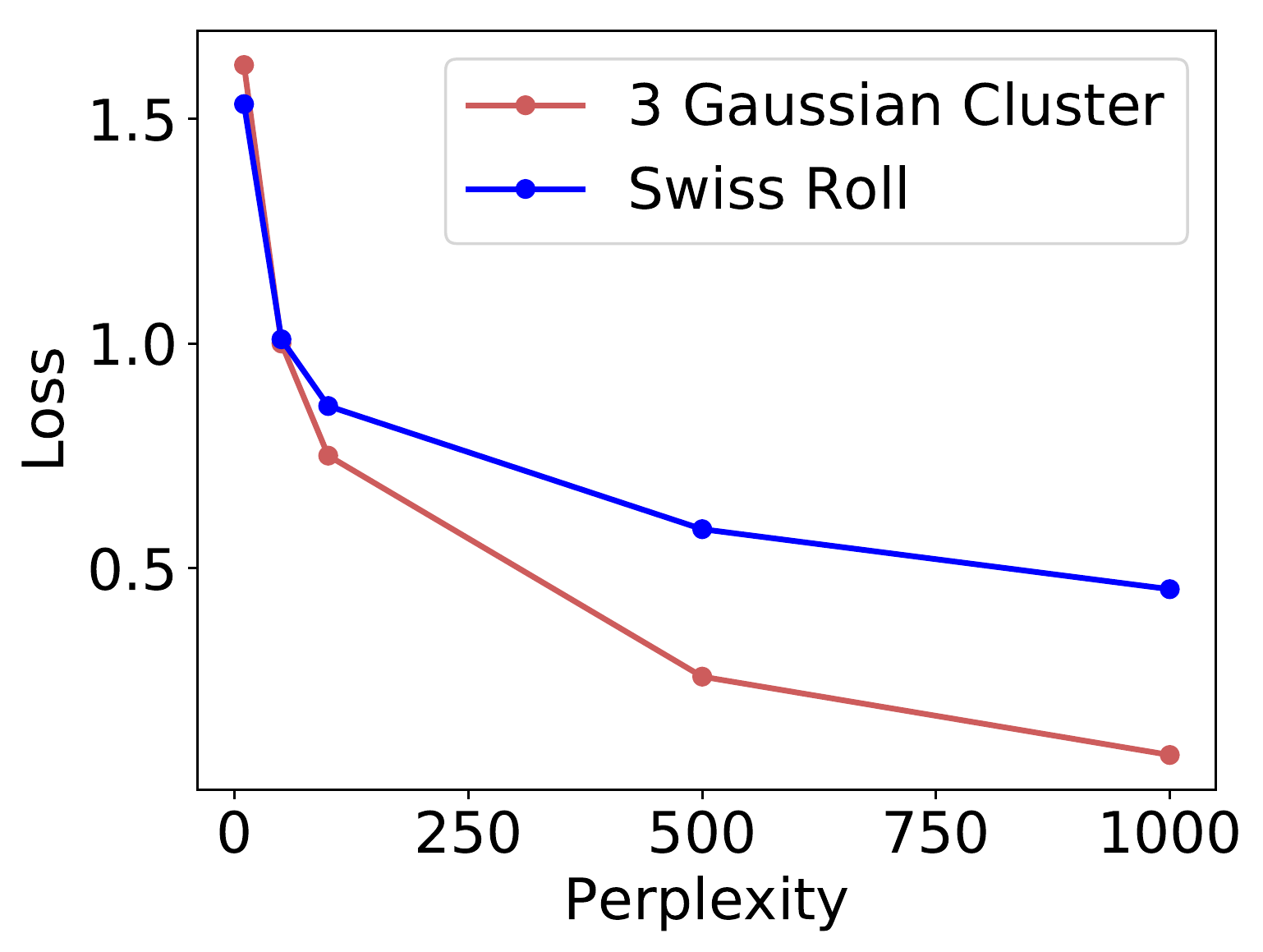}
            \vspace{-0.5cm}
            \subcaption*{JS-SNE}
        \end{minipage} 
        \begin{minipage}{0.325\textwidth}
            \includegraphics[width=\linewidth]{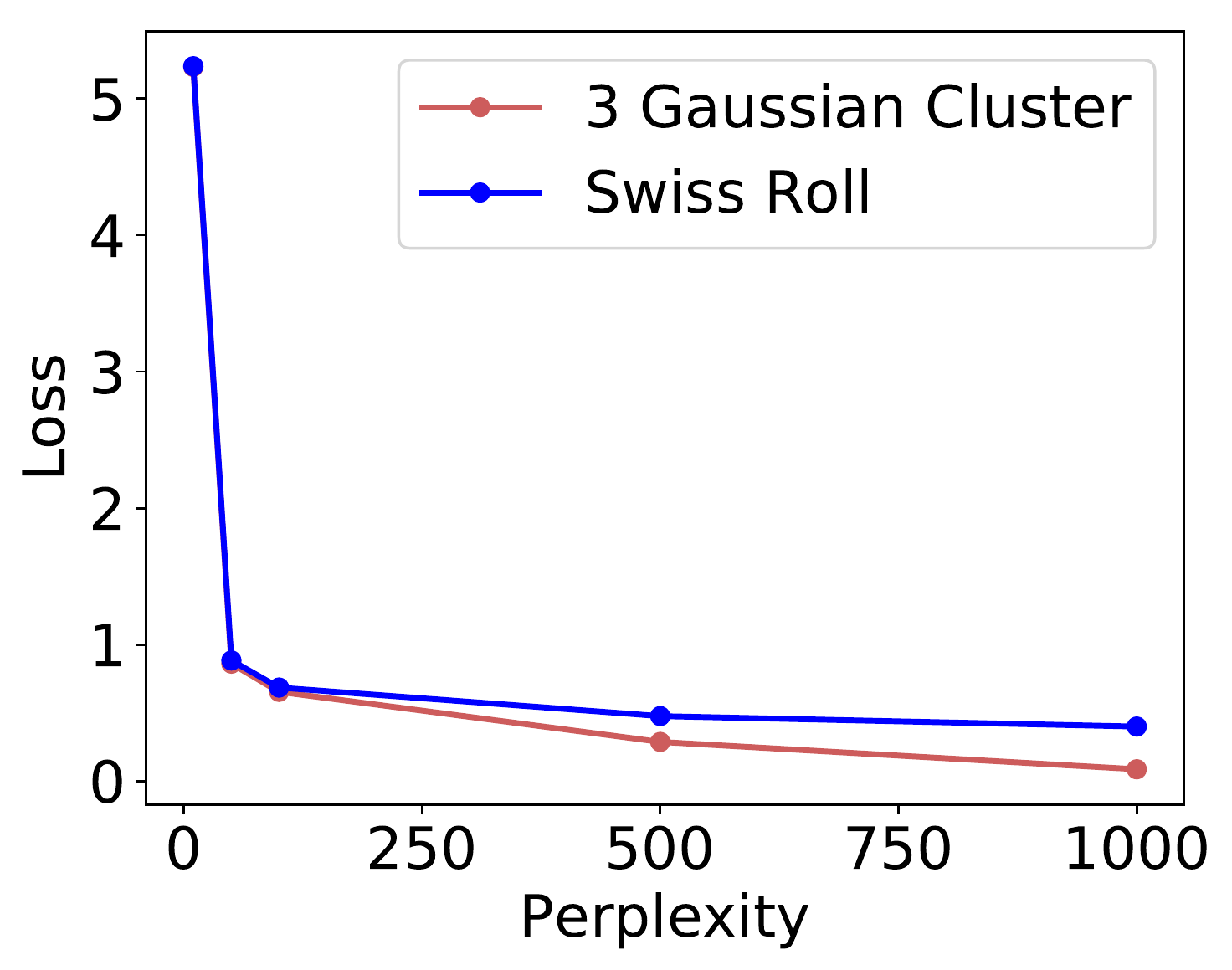}
            \vspace{-0.5cm}
            \subcaption*{RKL-SNE}
        \end{minipage} 
        \caption{$ft$-SNE Loss with respect to different perplexities on three Gaussian cluster and swiss roll datasets.}
        \label{fig:syn_perp}
    \end{figure}

    \begin{figure}[htp]
        \begin{minipage}{0.5\textwidth}
            \begin{minipage}{0.485\textwidth}
            \includegraphics[width=\linewidth]{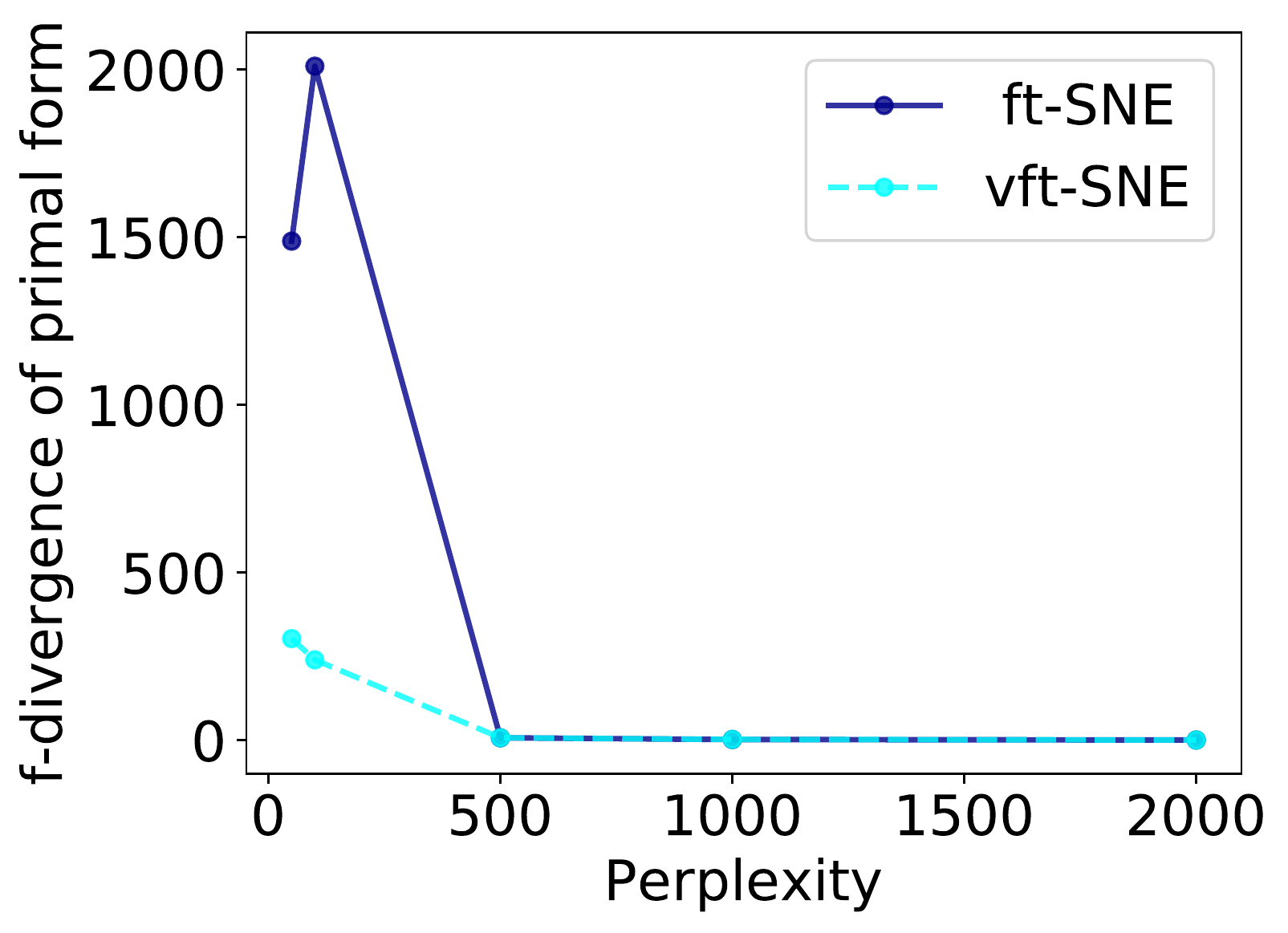}
            \vspace{-0.5cm}
            \subcaption*{MNIST1}
            \end{minipage}
            \begin{minipage}{0.485\textwidth}
            \includegraphics[width=\linewidth]{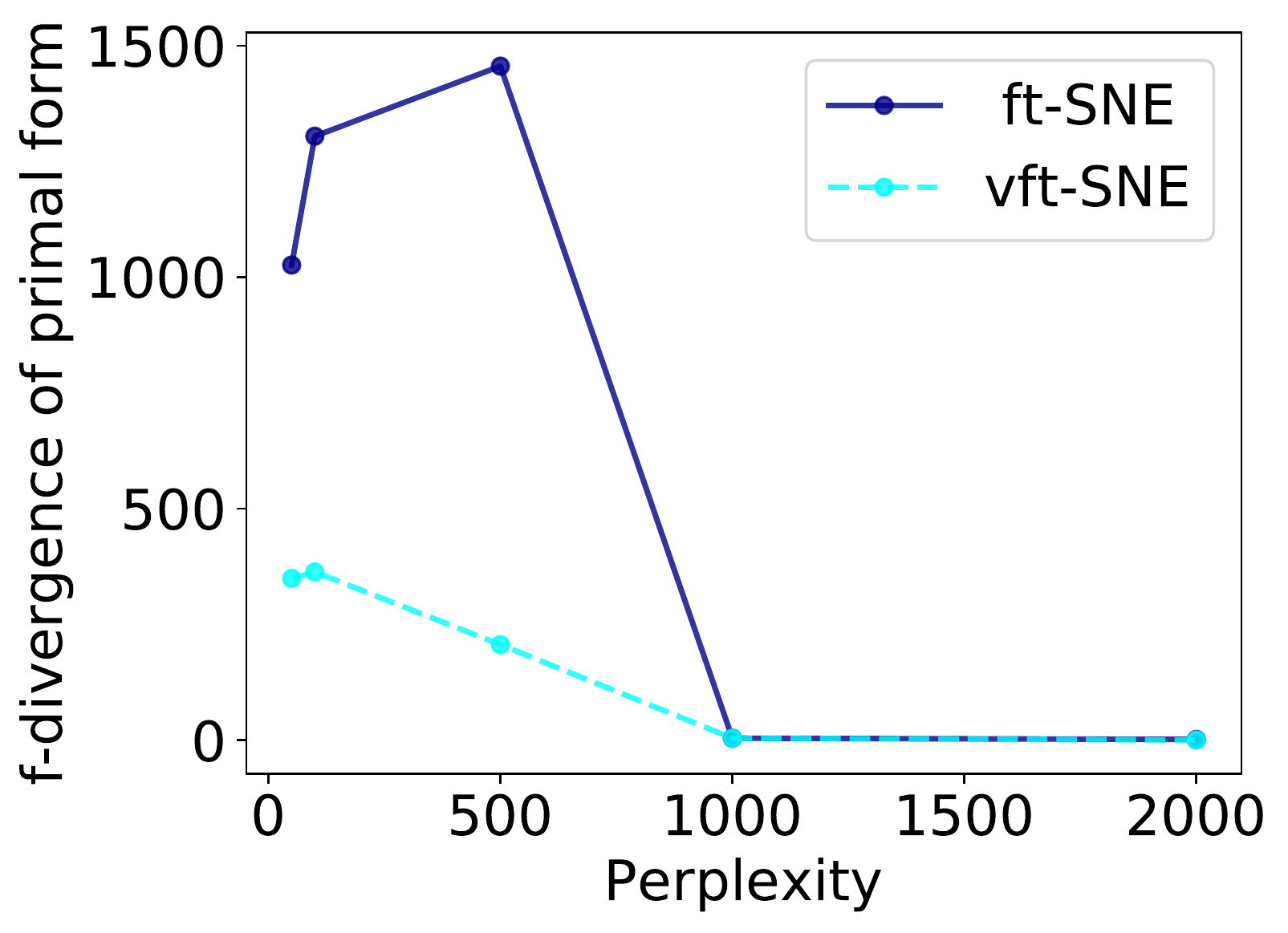}
            \vspace{-0.5cm}
            \subcaption*{NEWS}
            \end{minipage} 
            \subcaption{Chi-Square}
            \label{fig:perp_ch}
        \end{minipage}        
        \begin{minipage}{0.5\textwidth}
            \begin{minipage}{0.485\textwidth}
            \includegraphics[width=\linewidth]{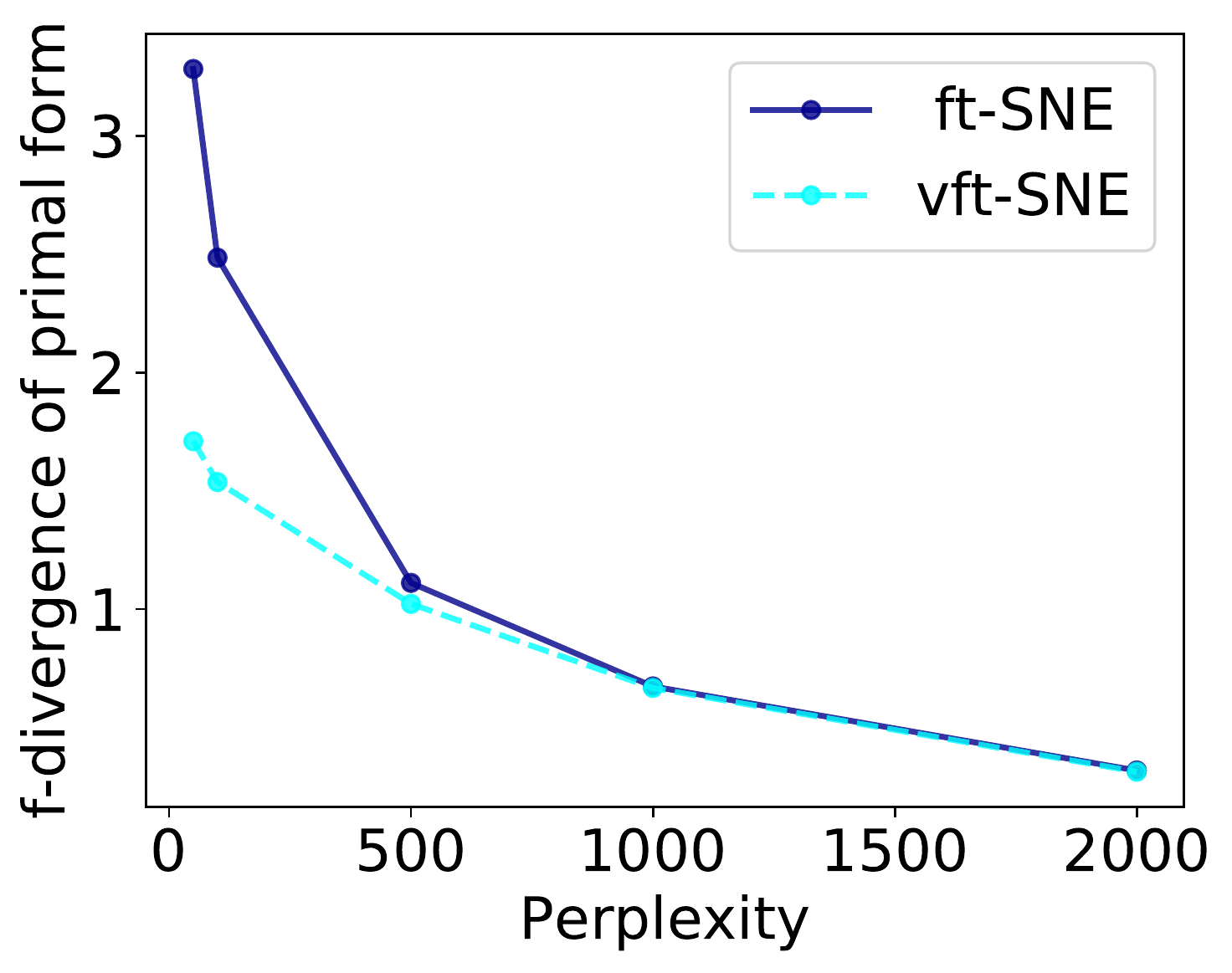}
            \vspace{-0.5cm}
            \subcaption*{MNIST1}
            \end{minipage}            
            \begin{minipage}{0.485\textwidth}
            \includegraphics[width=\linewidth]{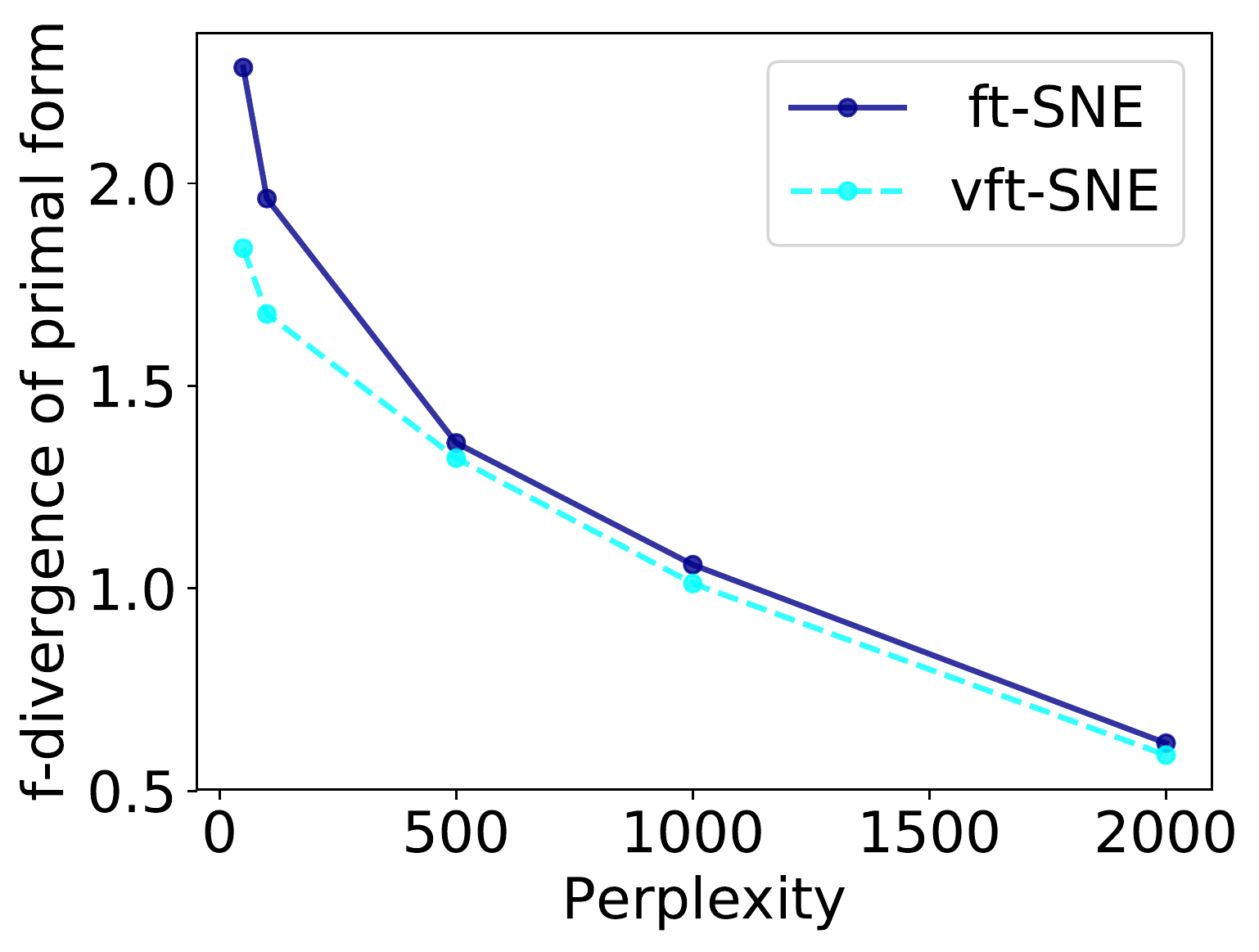}
            \vspace{-0.5cm}
            \subcaption*{NEWS}
            \end{minipage} 
        \subcaption{Jensen-Shannon}
        \label{fig:perp_js}
        \end{minipage}        
        \caption{$ft$-SNE and $vft$-SNE Loss with respect to different perplexities  MNIST1 and NEWS.}
        \label{fig:SM_perp}
    \end{figure}

    \pagebreak
    \subsubsection{More Experimental Results with different discriminant functions \label{app:sm_arch_exp}}
    {\em Architecture} Optimizing under $vf$-SNE require having a discriminant function.
    Throughout the experiments, we used deep neural network the discriminator,
    The architecture tology is defined as $D(x_i,x_j) = g([f(x_i) + f(x_j); f(x_i)\odot f(x_j)])$ (depicted in S. M. Figure~\ref{fig:disc_arch}).
    $f(\cdot)$ is the neural network that encodes the pair of data points, $f(x)$ and $f(y)$, and
    $g(\cdot)$ is the neural network that takes $[f(x)+f(y); f(x)\odot f(y)]$ and outputs the score value.
    Our architecture is invariant to the ordering of data points (i.e., $D(x_i,x_j) = D(x_j,x_i)$).
    We used 10 hidden layers and 20 hidden layers for $f(\cdot)$ and $g(\cdot)$ in the experiments except when we experiments with expressibility of discriminant function.

    \begin{figure}[htp]
        \centering
        \vspace{0.2cm}
        \includegraphics[width=\linewidth]{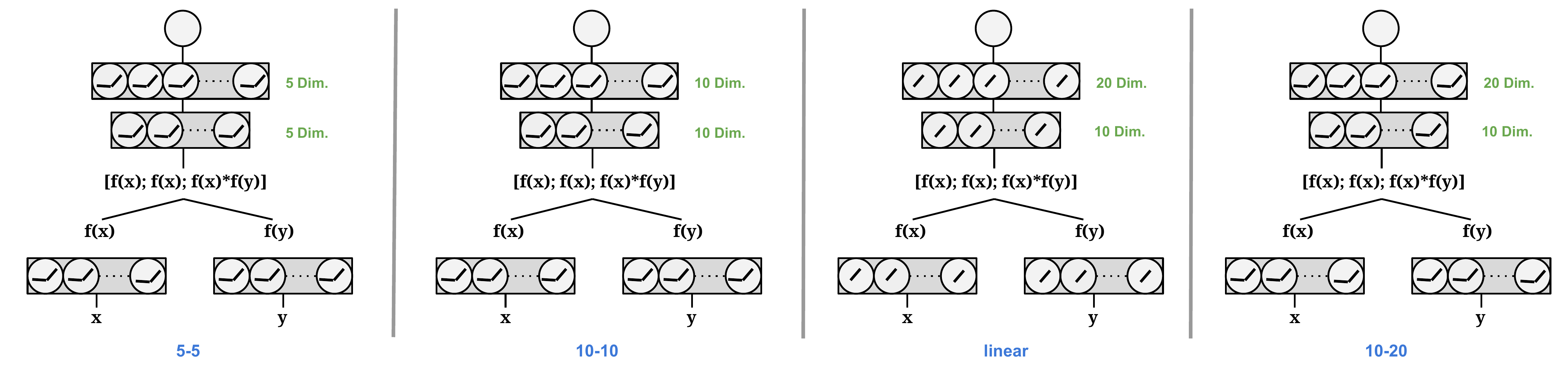}
        \vspace{-0.6cm}
        \caption{Discriminant Architecture : \# of Hidden Layer Size \label{fig:disc_arch_size}}

        \includegraphics[width=\linewidth]{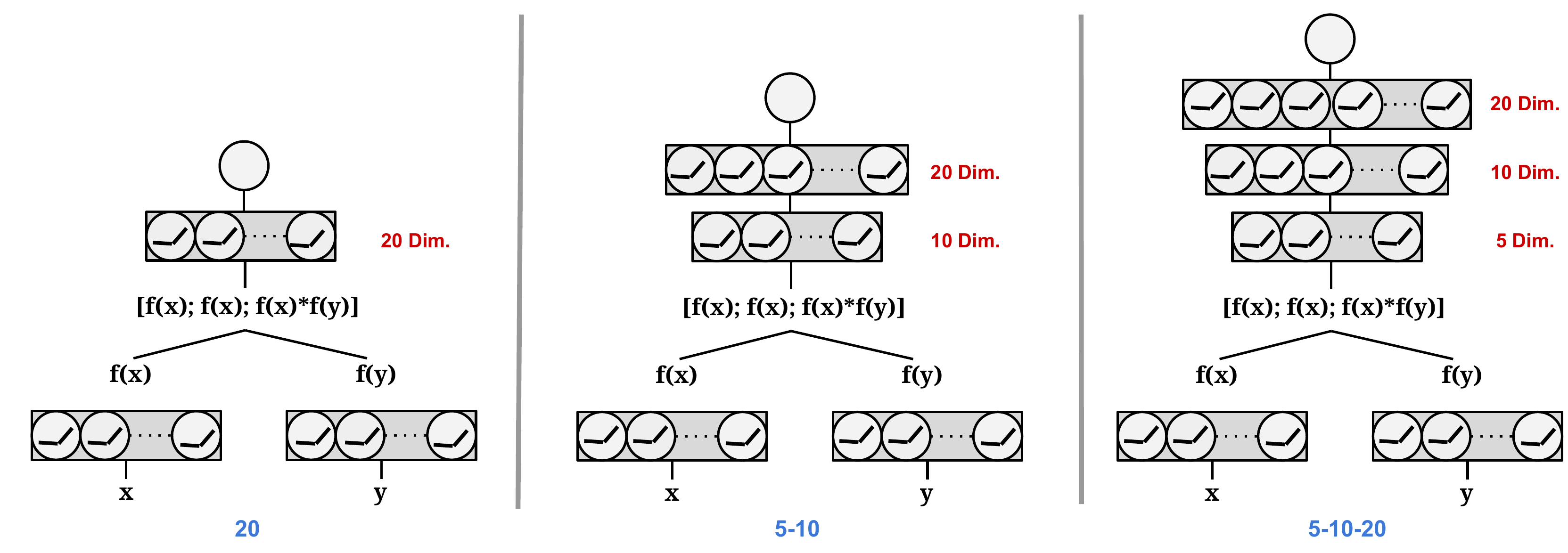}
        \vspace{-0.6cm}
        \caption{Discriminant Architecture : \# of Hidden Layer Depth \label{fig:disc_arch_depth}}
    \end{figure}

    \begin{figure}[htp]
        \centering
        \begin{minipage}{\textwidth}
            \begin{minipage}{0.32\textwidth}
            \includegraphics[width=\linewidth]{./figs/opt/mnist_cost_curves_numhids_kl.pdf}
            \vspace{-0.5cm}
            \subcaption*{KL-SNE}
            \end{minipage}
            \begin{minipage}{0.32\textwidth}
            \includegraphics[width=\linewidth]{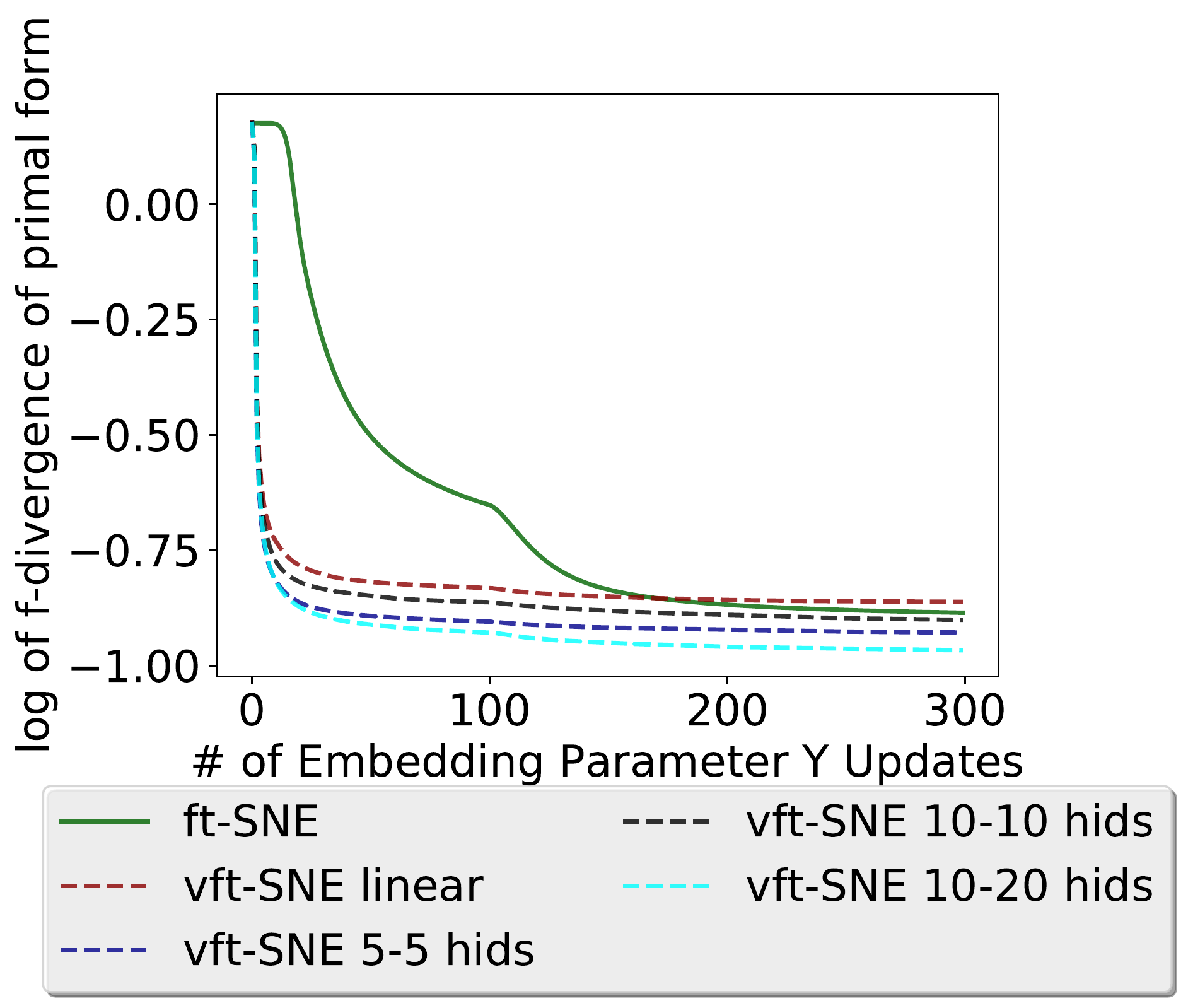}
            \vspace{-0.5cm}
            \subcaption*{JS-SNE}
            \end{minipage}            
            \begin{minipage}{0.32\textwidth}
            \includegraphics[width=\linewidth]{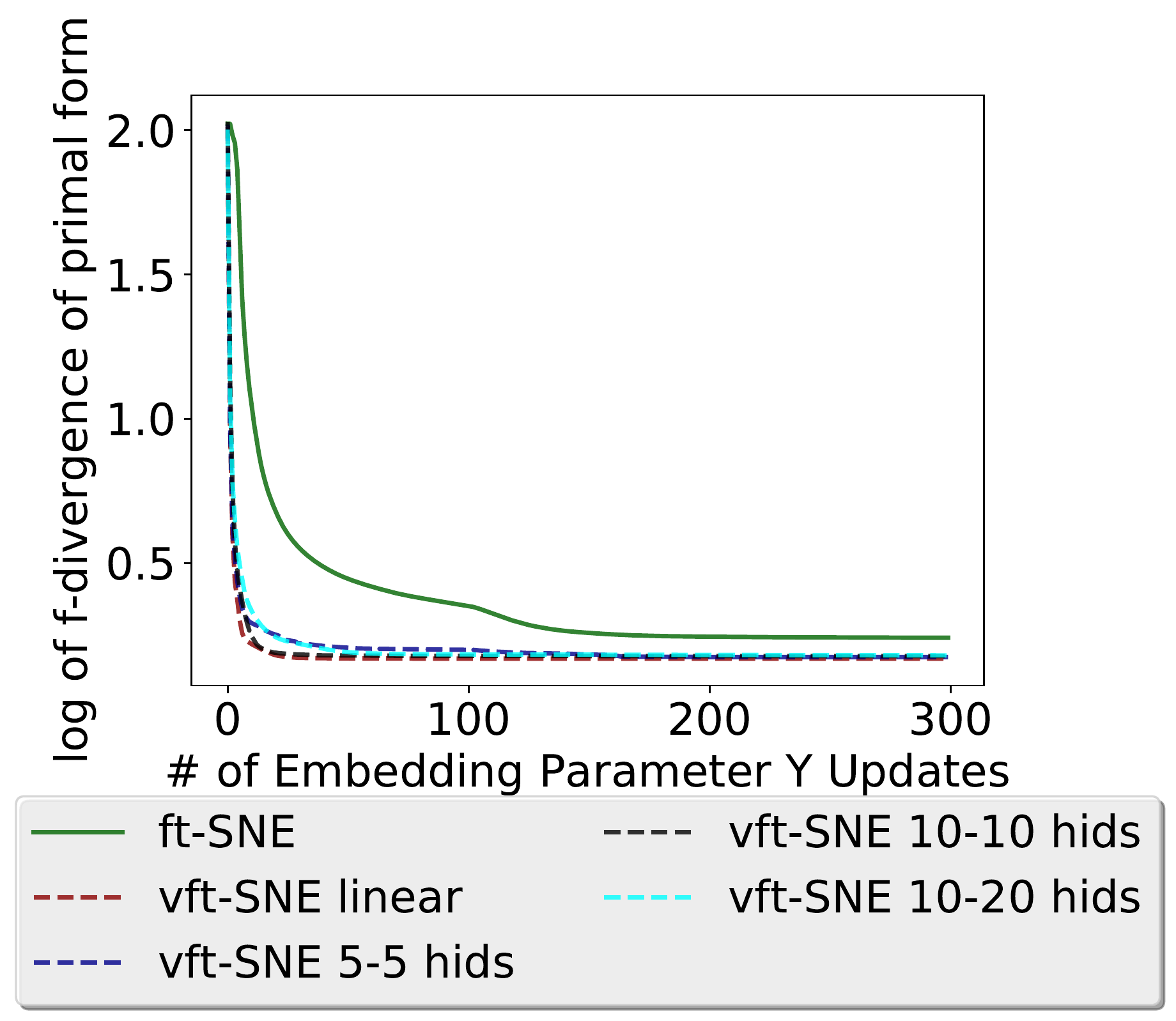}
            \vspace{-0.5cm}
            \subcaption*{CS-SNE}
            \end{minipage}
            \subcaption{Varying discriminator network width}
        \end{minipage}
        \begin{minipage}{\textwidth}
            \begin{minipage}{0.32\textwidth}
            \includegraphics[width=\linewidth]{./figs/opt/mnist_cost_curves_hids_depth_kl.pdf}
            \vspace{-0.5cm}
            \subcaption*{KL-SNE}
            \end{minipage}
            \begin{minipage}{0.32\textwidth}
            \includegraphics[width=\linewidth]{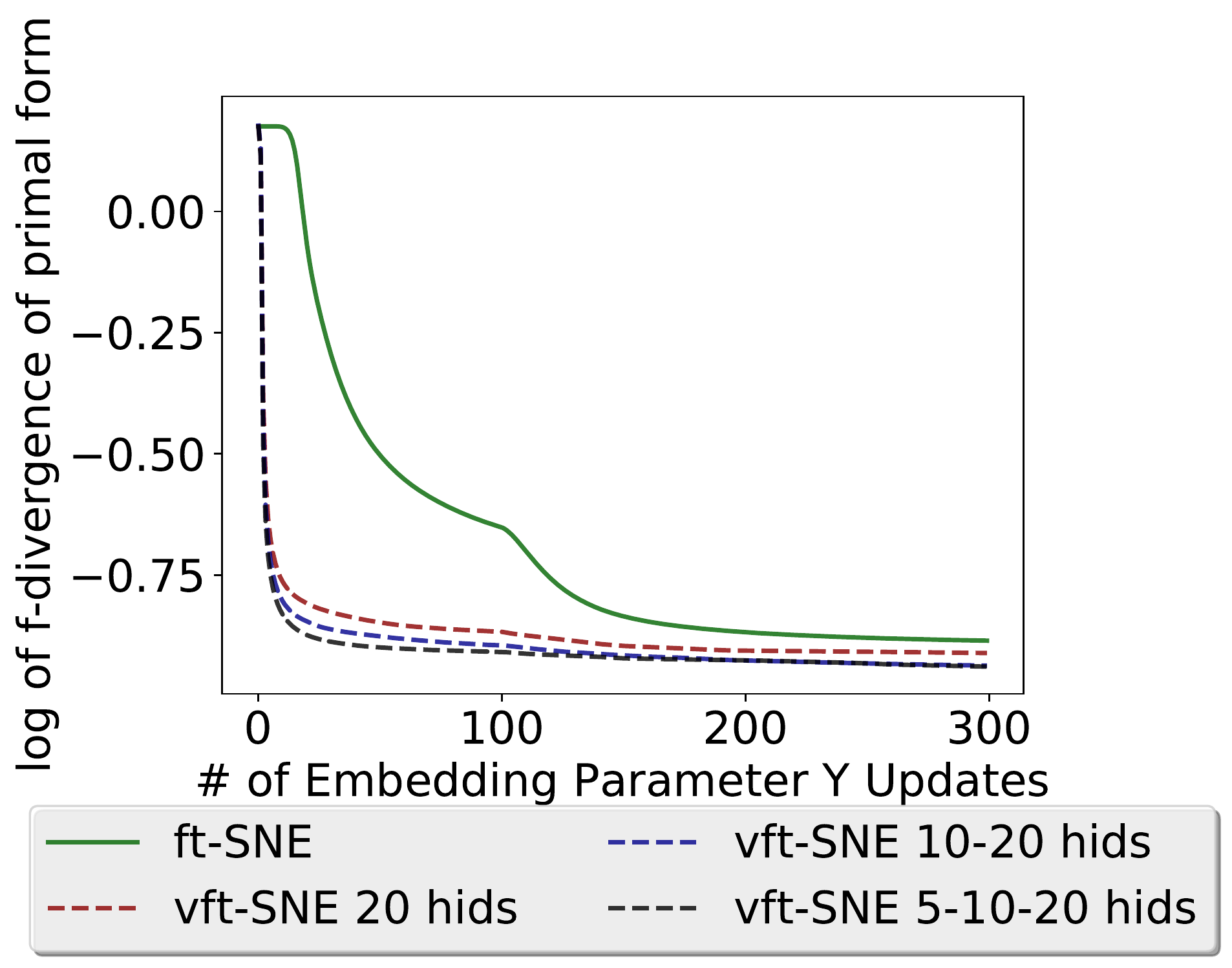}
            \vspace{-0.5cm}
            \subcaption*{JS-SNE}
            \end{minipage}            
            \begin{minipage}{0.32\textwidth}
            \includegraphics[width=\linewidth]{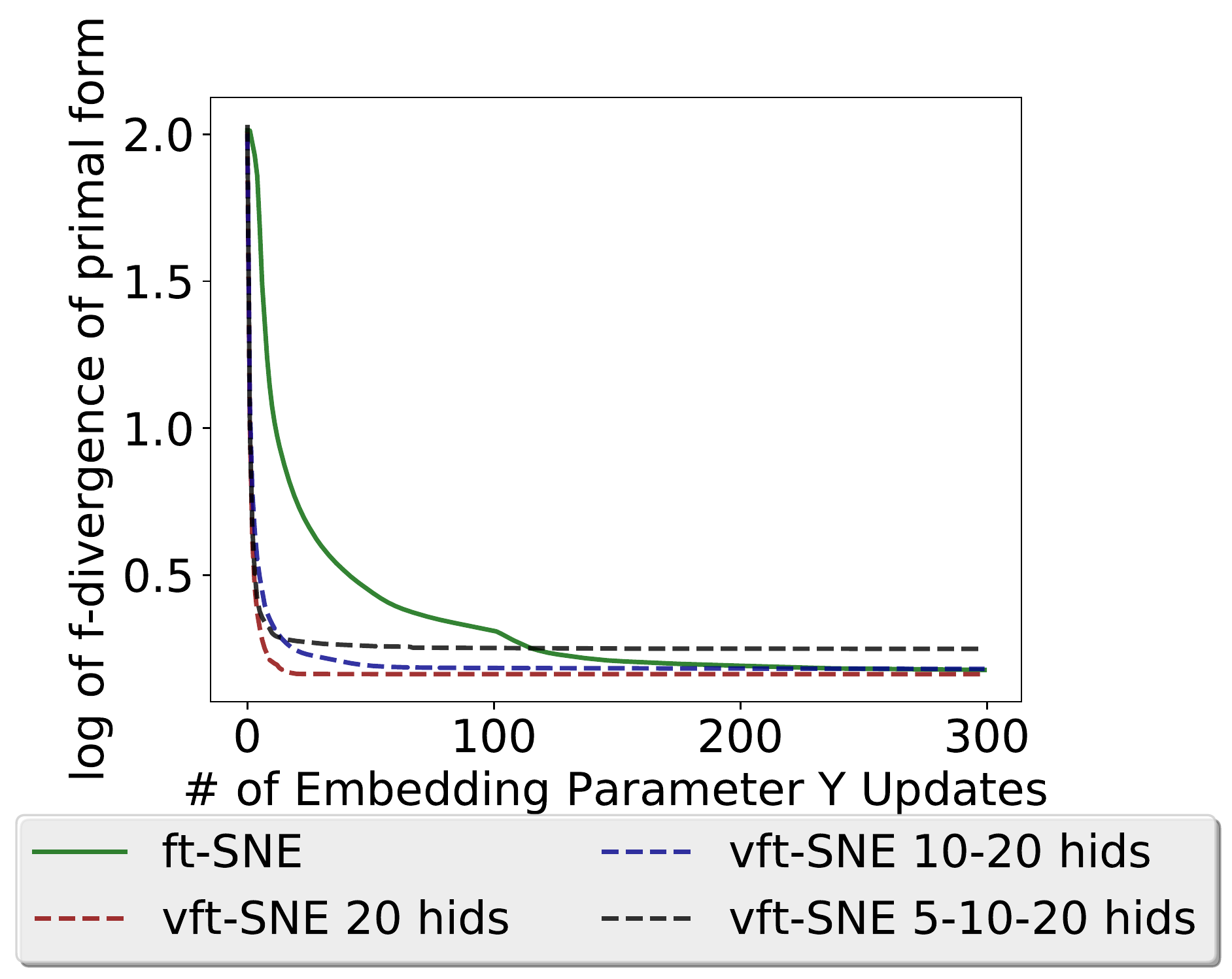}
            \vspace{-0.5cm}
            \subcaption*{CS-SNE}
            \end{minipage} 
            \subcaption{Varying discrimnator network depth}
        \end{minipage}
        \caption{Log $ft$-SNE criterion during optimization for different discriminator network architectures on MNIST. In (a), we compare two-layer networks of different widths (legend indicates number of units in first and second hidden layers). In (b), we compare networks with different depths (legend indicates number of units in each hidden layer). 
        The number of updates were set to J:K=10:10 and perplexity was set to 2,000.}
        \label{fig:arch}
        \begin{minipage}{0.495\textwidth}
            \begin{minipage}{0.485\textwidth}
            \includegraphics[width=\linewidth]{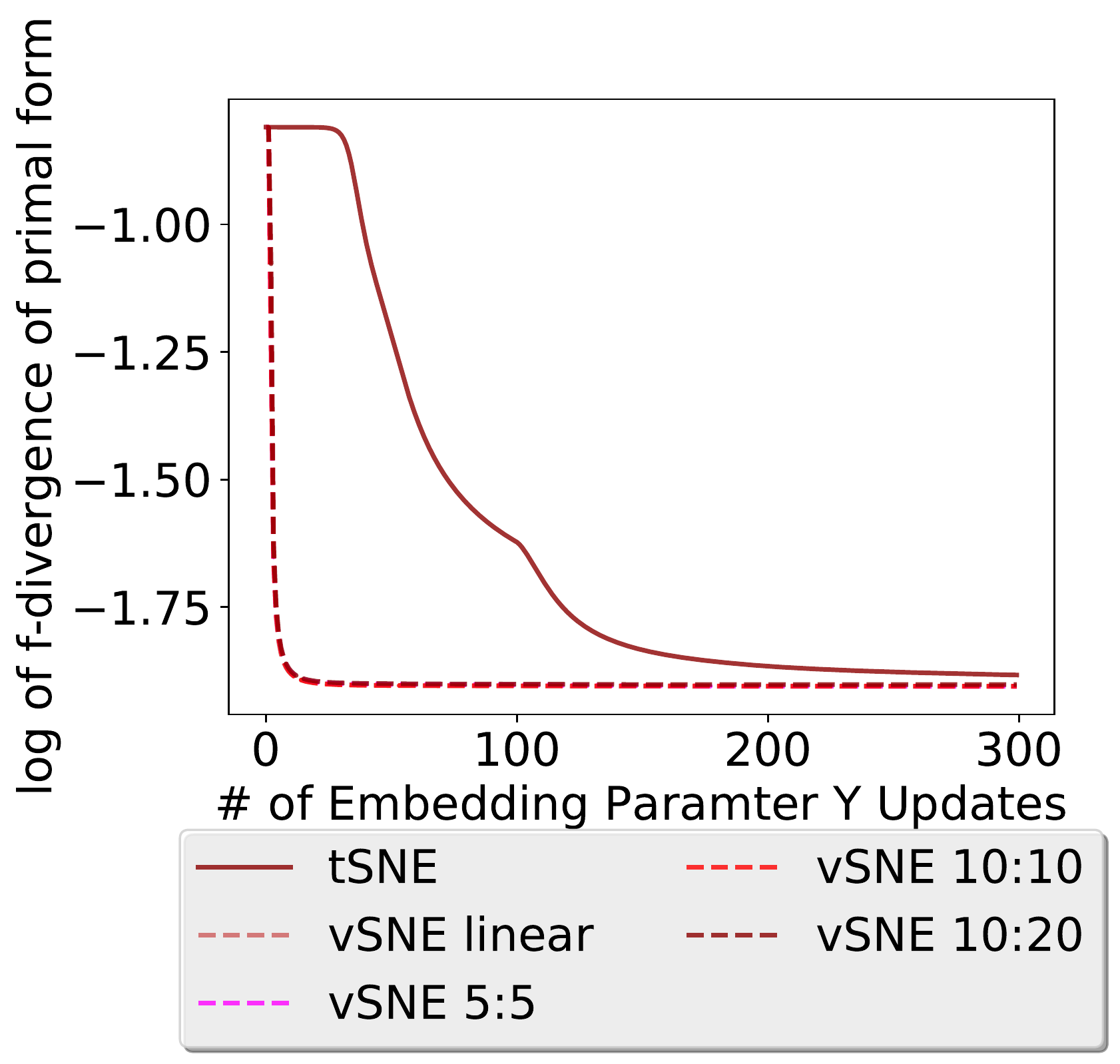}
            \vspace{-0.5cm}
            \subcaption*{MNIST1}
            \end{minipage}
            \begin{minipage}{0.485\textwidth}
            \includegraphics[width=\linewidth]{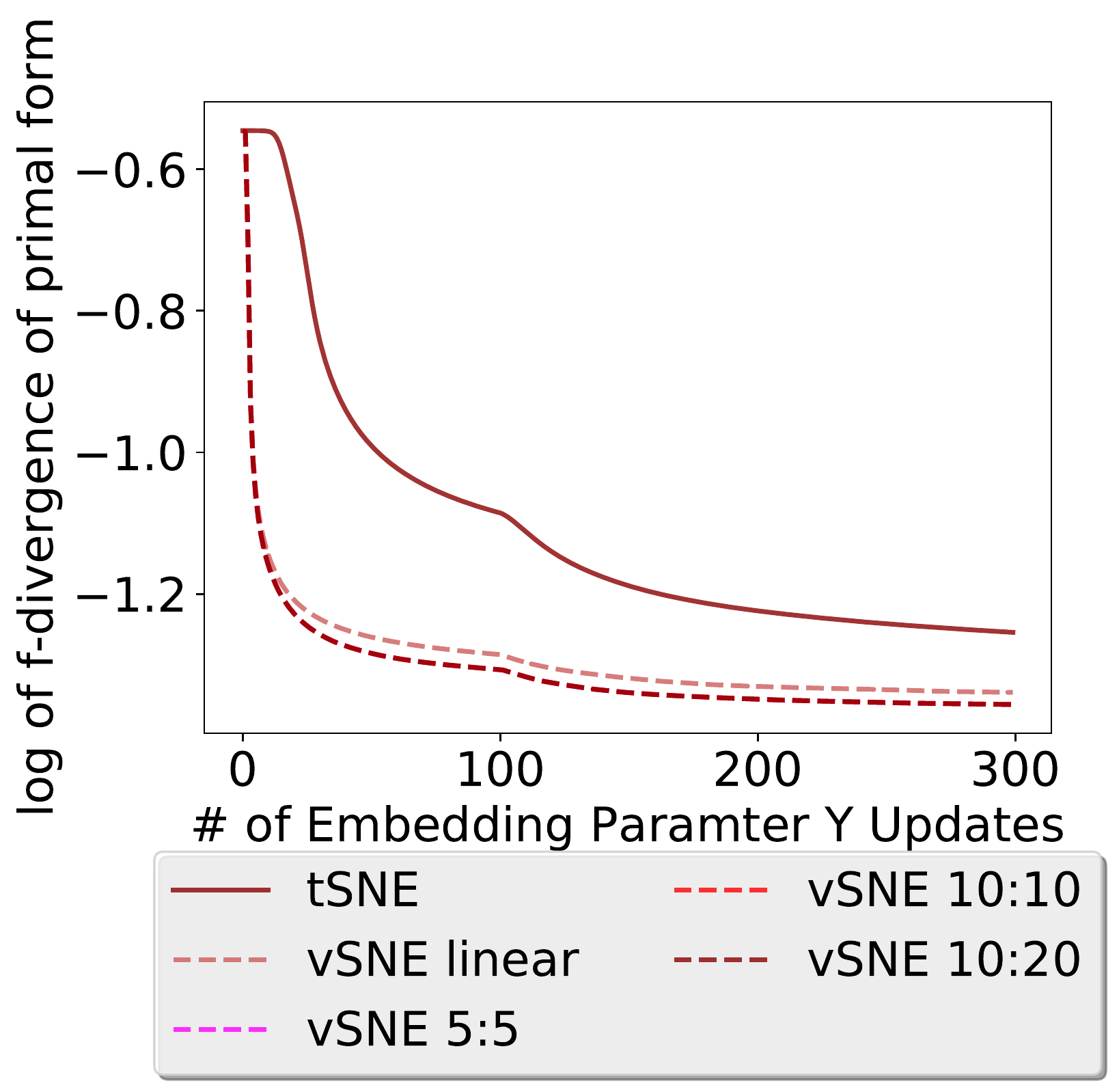}
            \vspace{-0.5cm}
            \subcaption*{NEWS}
            \end{minipage} 
            \subcaption*{CS-SNE}
        \end{minipage}
        \begin{minipage}{0.495\textwidth}
            \begin{minipage}{0.485\textwidth}
            \includegraphics[width=\linewidth]{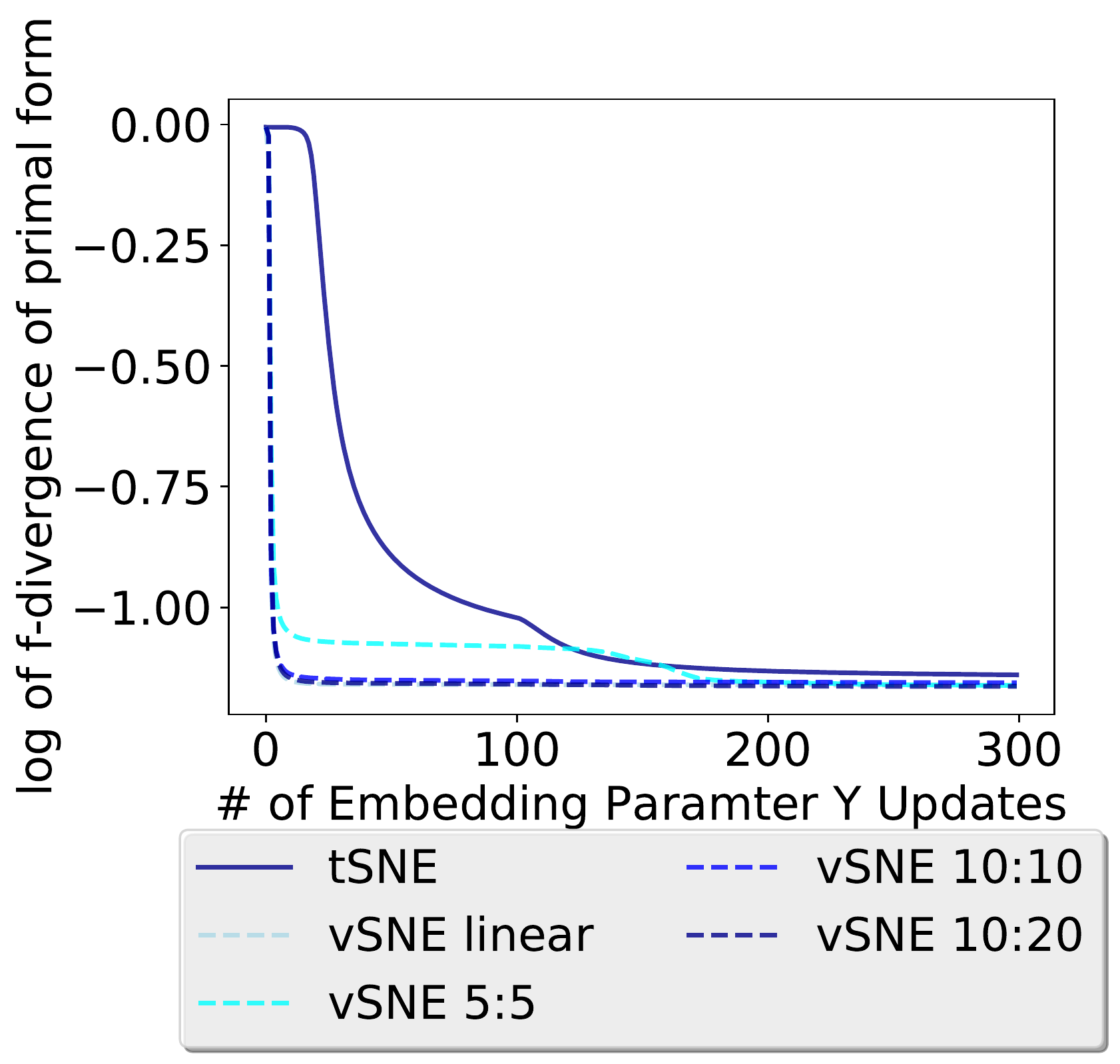}
            \vspace{-0.5cm}
            \subcaption*{MNIST1}
            \end{minipage}
            \begin{minipage}{0.485\textwidth}
            \includegraphics[width=\linewidth]{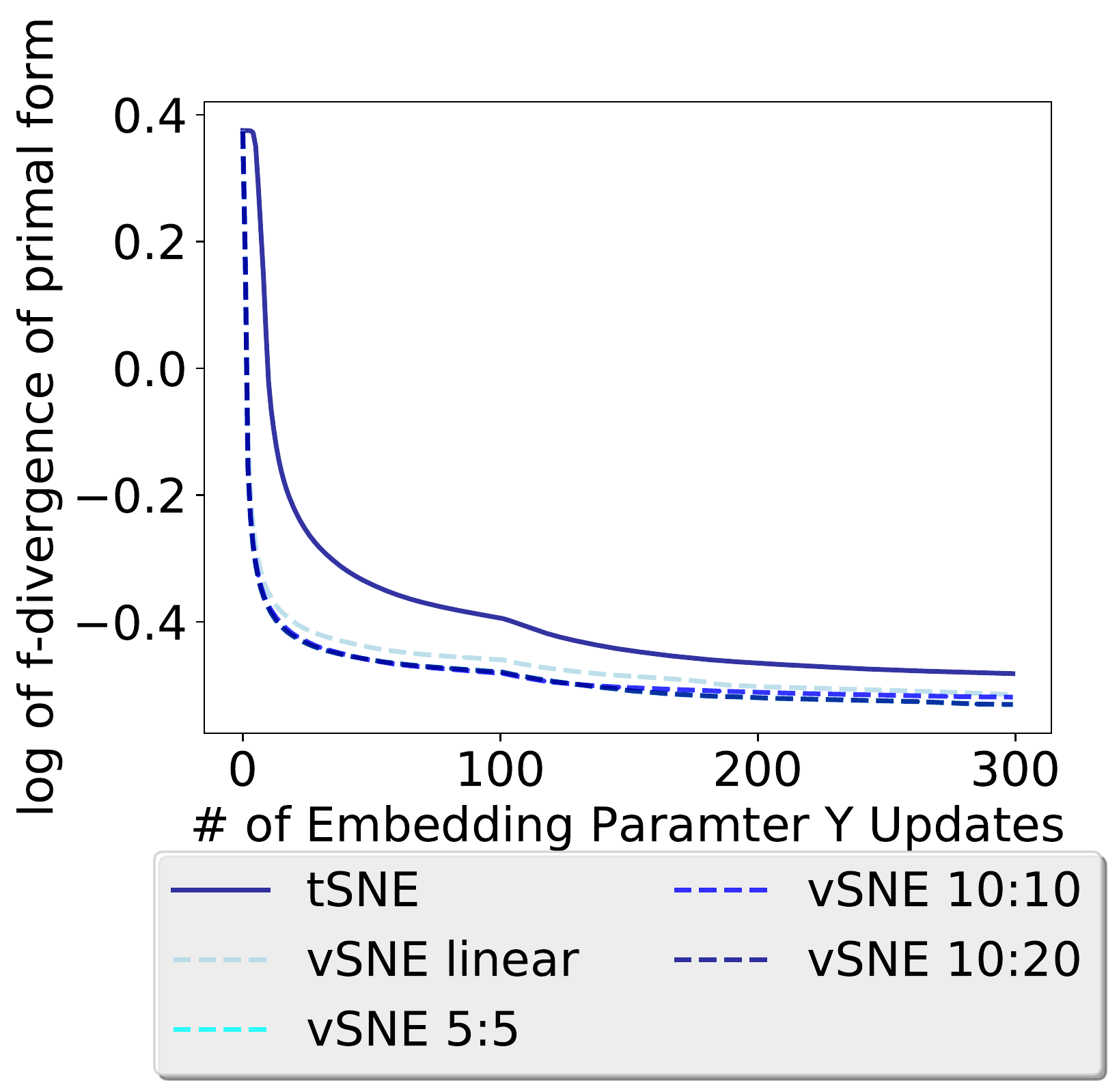}
            \vspace{-0.5cm}
            \subcaption*{NEWS}
            \end{minipage} 
            \subcaption*{JS-SNE}
        \end{minipage}        
        \caption{Log $ft$-SNE criterion during optimization for different discriminator network architectures of different widths on MNIST. }
        \label{fig:sm_arch}
    \end{figure}


\section{Embeddings}

\begin{figure}[htp]
        \centering
        \begin{minipage}{\textwidth}
            \begin{minipage}{0.49\textwidth}
                \includegraphics[width=\linewidth]{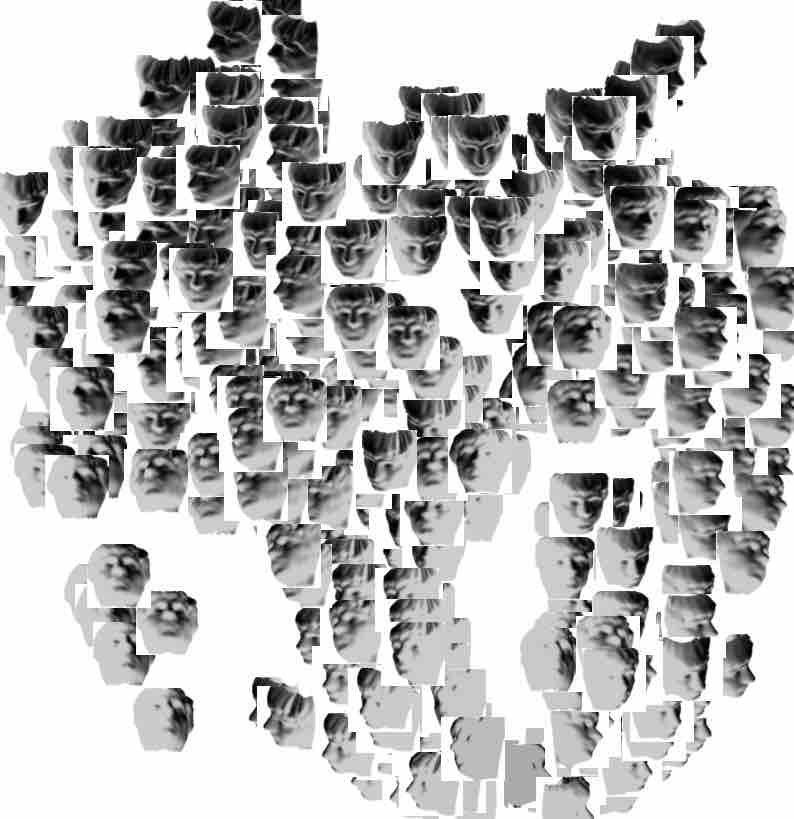}
            \end{minipage}
            \begin{minipage}{0.49\textwidth}
                \includegraphics[width=\linewidth]{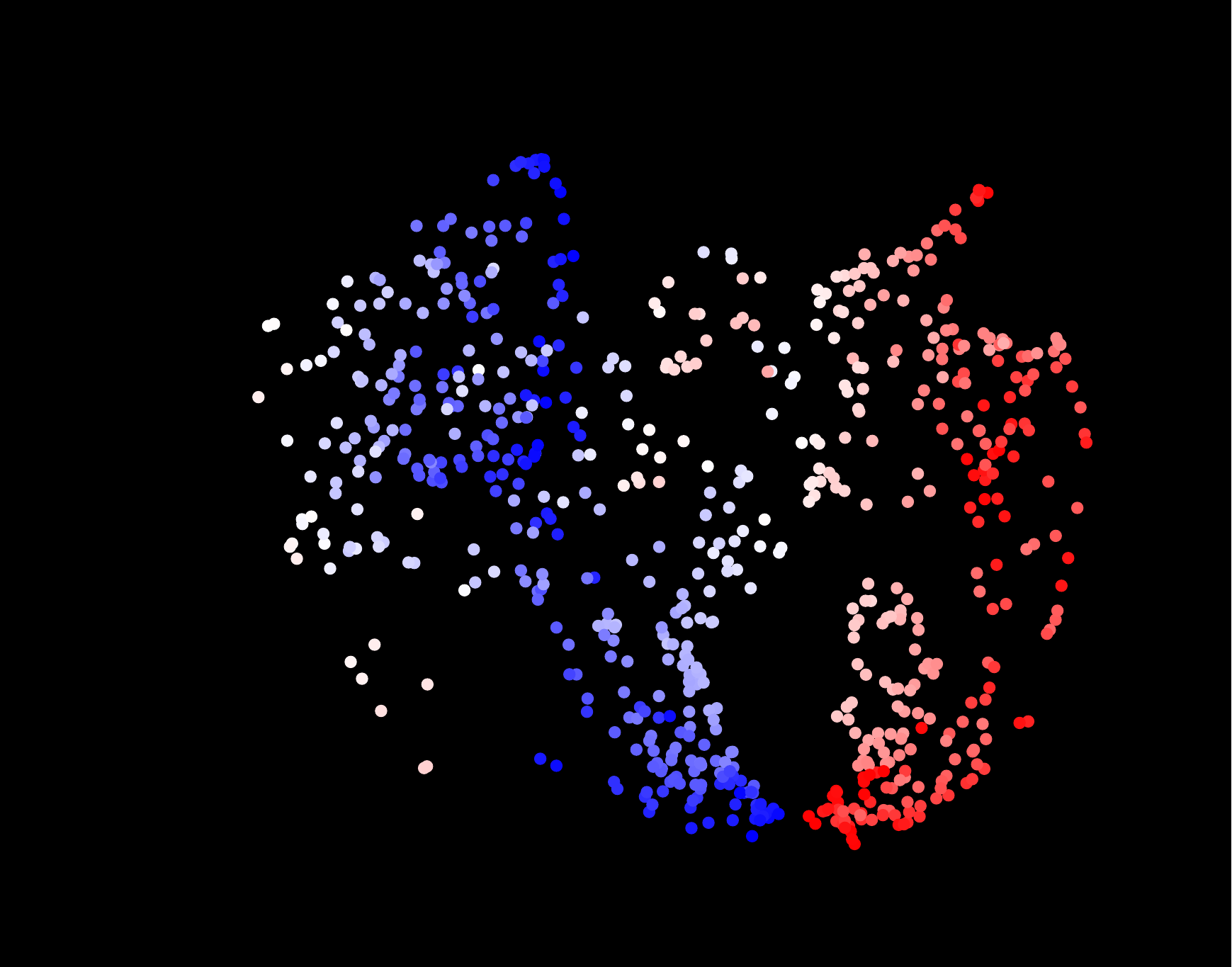}
            \end{minipage}  
            \vspace{-0.3cm}
            \subcaption{RKL-SNE \label{fig:kl_face}}
        \end{minipage}
        \begin{minipage}{\textwidth}
            \begin{minipage}{0.495\textwidth}
                \includegraphics[width=0.85\linewidth]{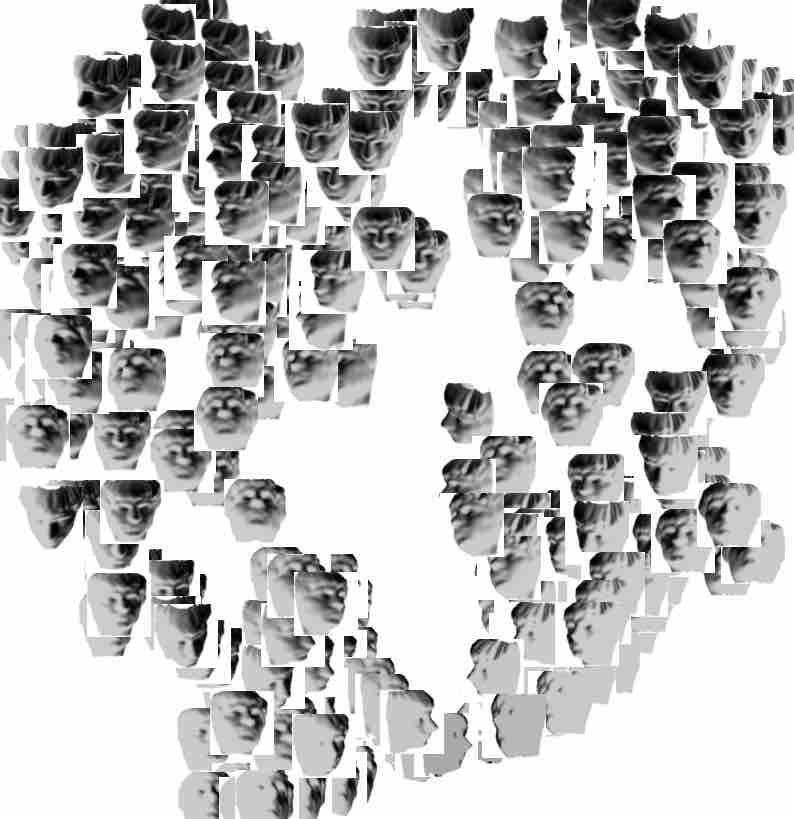}
            \end{minipage}
            \begin{minipage}{0.495\textwidth}
                \includegraphics[width=\linewidth]{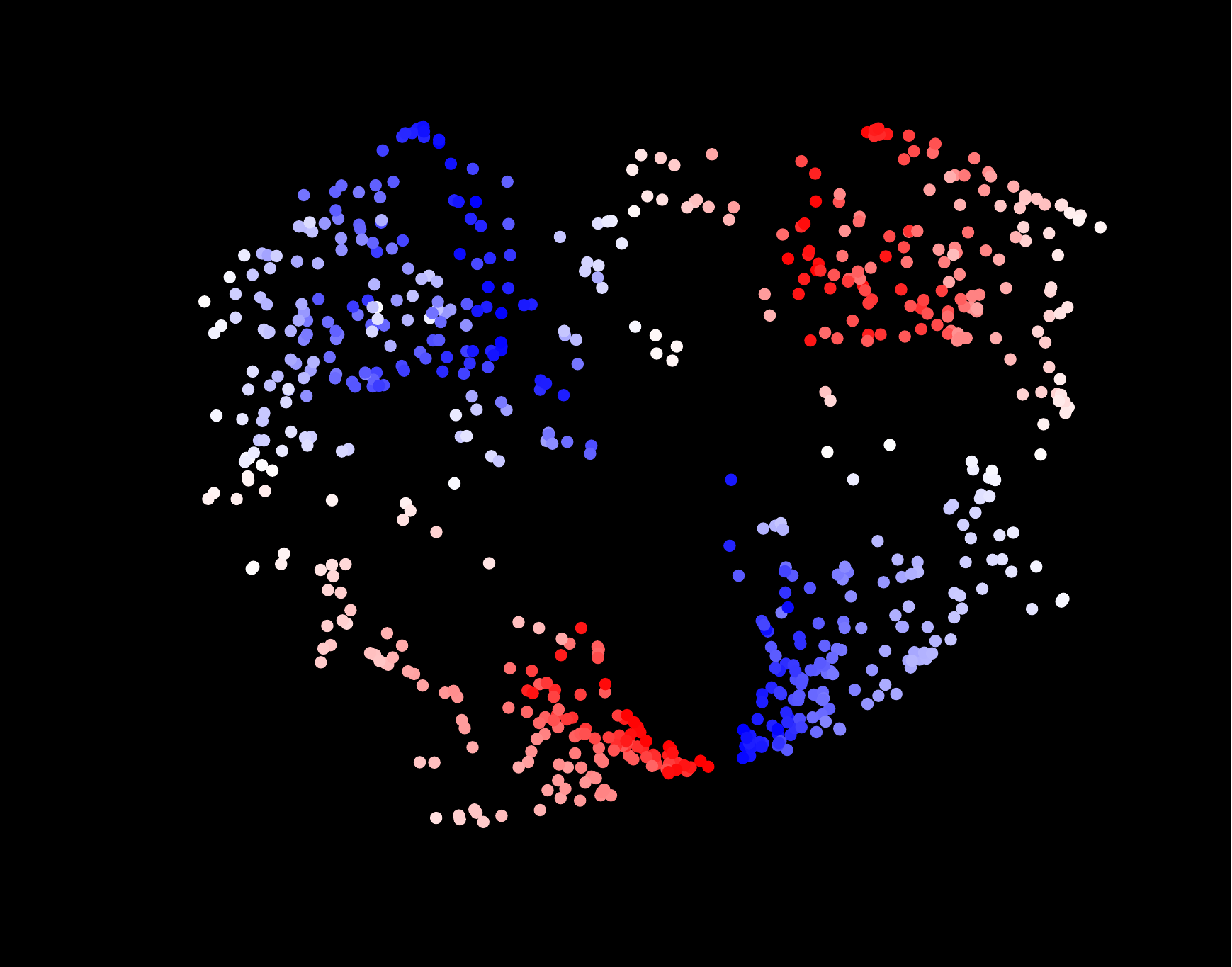}
            \end{minipage}  
            \vspace{-0.3cm}
            \subcaption{RKL-SNE \label{fig:rkl_face}}
        \end{minipage}  
        \caption{Face Embeddings with KL and RKL-SNE\label{fig:face_emb_kl_rkl}}
    \end{figure}

    Figure~\ref{fig:face_emb_kl_rkl} presnets the embeddings of KL-SNE and RKL-SNE.
    Note that KL-SNE generates spurious clusters on the bottom left of the embeddings,
    whereas RKL-SNE generated smooth embeddigns that captures the manifold structure.
    Note that in practice, we do not want to generate such a spurious cluster 
    because the practitioners can misinterpret the visualization of the dataset. 


\begin{figure}[htp]
    \captionsetup{justification=centering}
    \begin{minipage}{\textwidth} 
        \begin{minipage}{0.195\textwidth}
        \includegraphics[width=\linewidth]{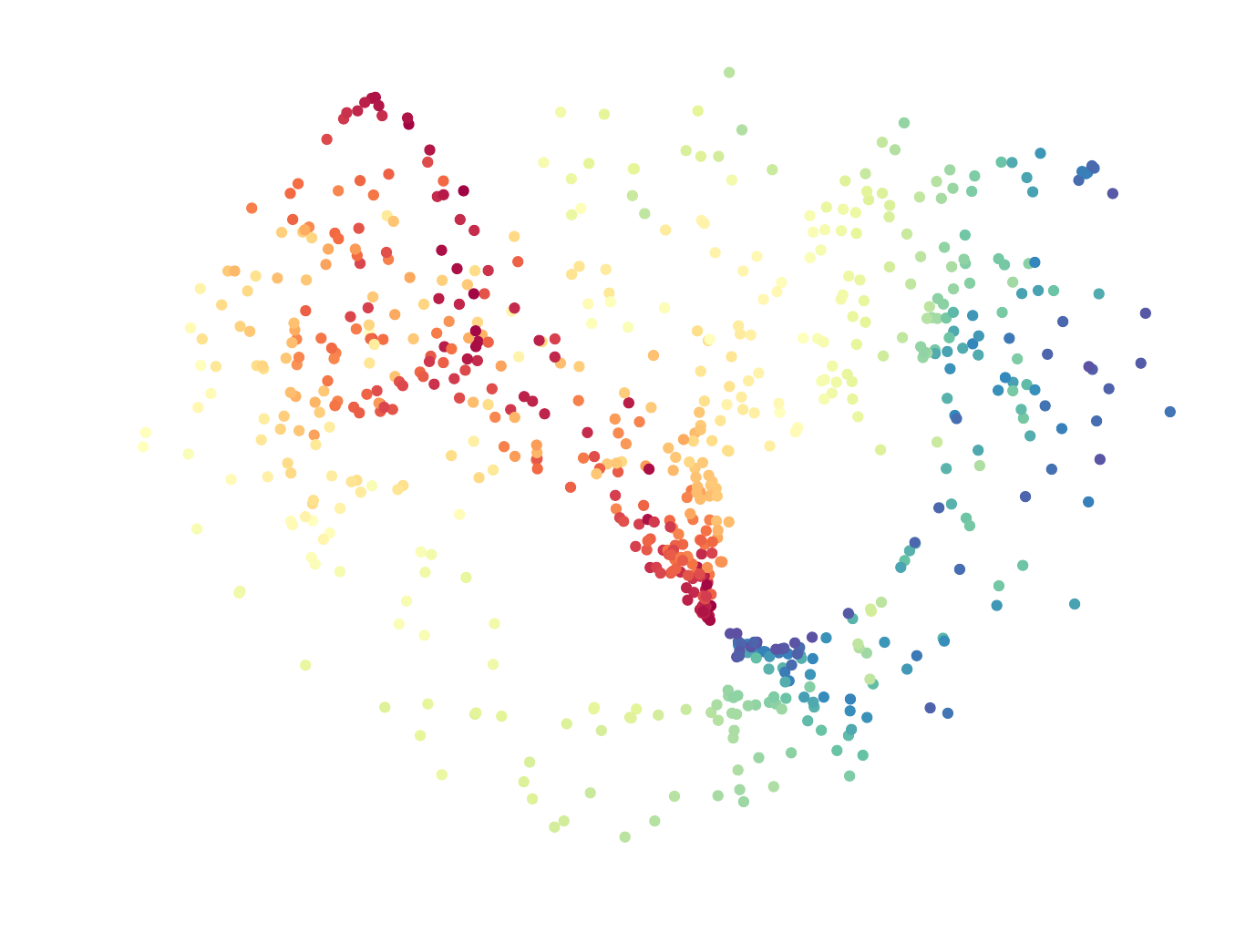}
        \vspace{-0.7cm}
        \subcaption*{KL-SNE}
        \end{minipage}
        \begin{minipage}{0.195\textwidth}
        \includegraphics[width=\linewidth]{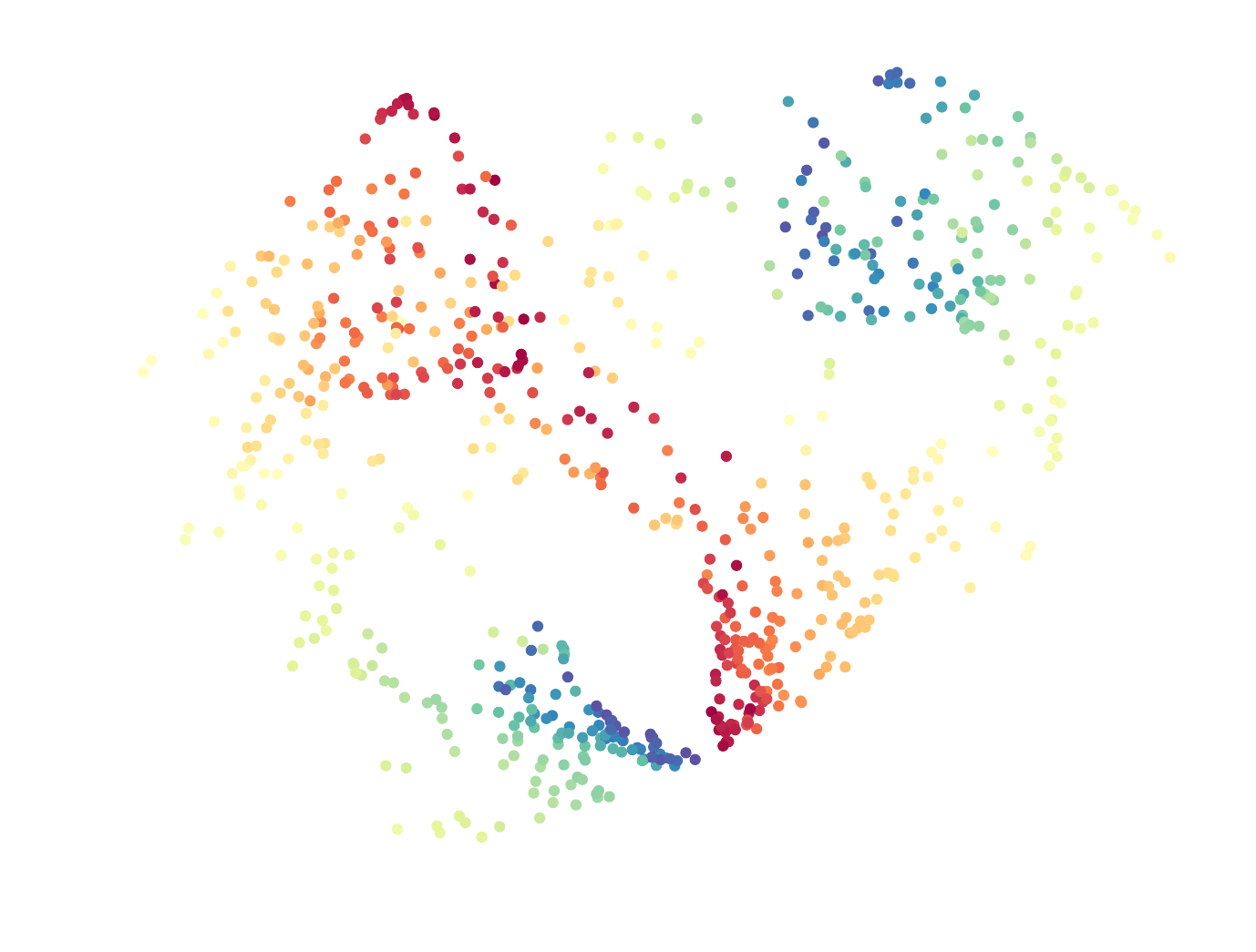}
        \vspace{-0.7cm}
        \subcaption*{RKL-SNE}
        \end{minipage}
        \begin{minipage}{0.195\textwidth}
        \includegraphics[width=\linewidth]{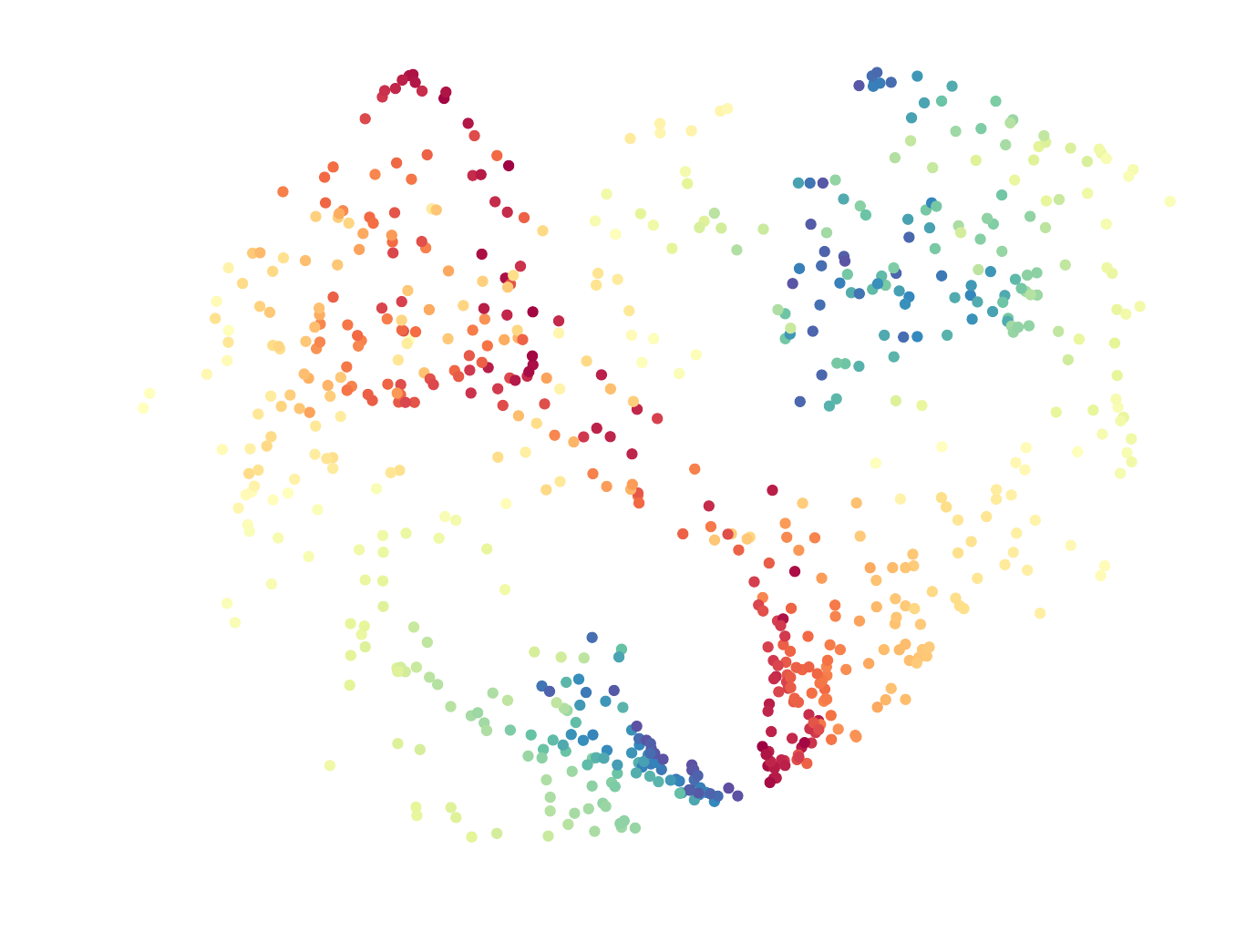}
        \vspace{-0.75cm}
        \subcaption*{JS-SNE}
        \end{minipage}
        \begin{minipage}{0.195\textwidth}
        \includegraphics[width=\linewidth]{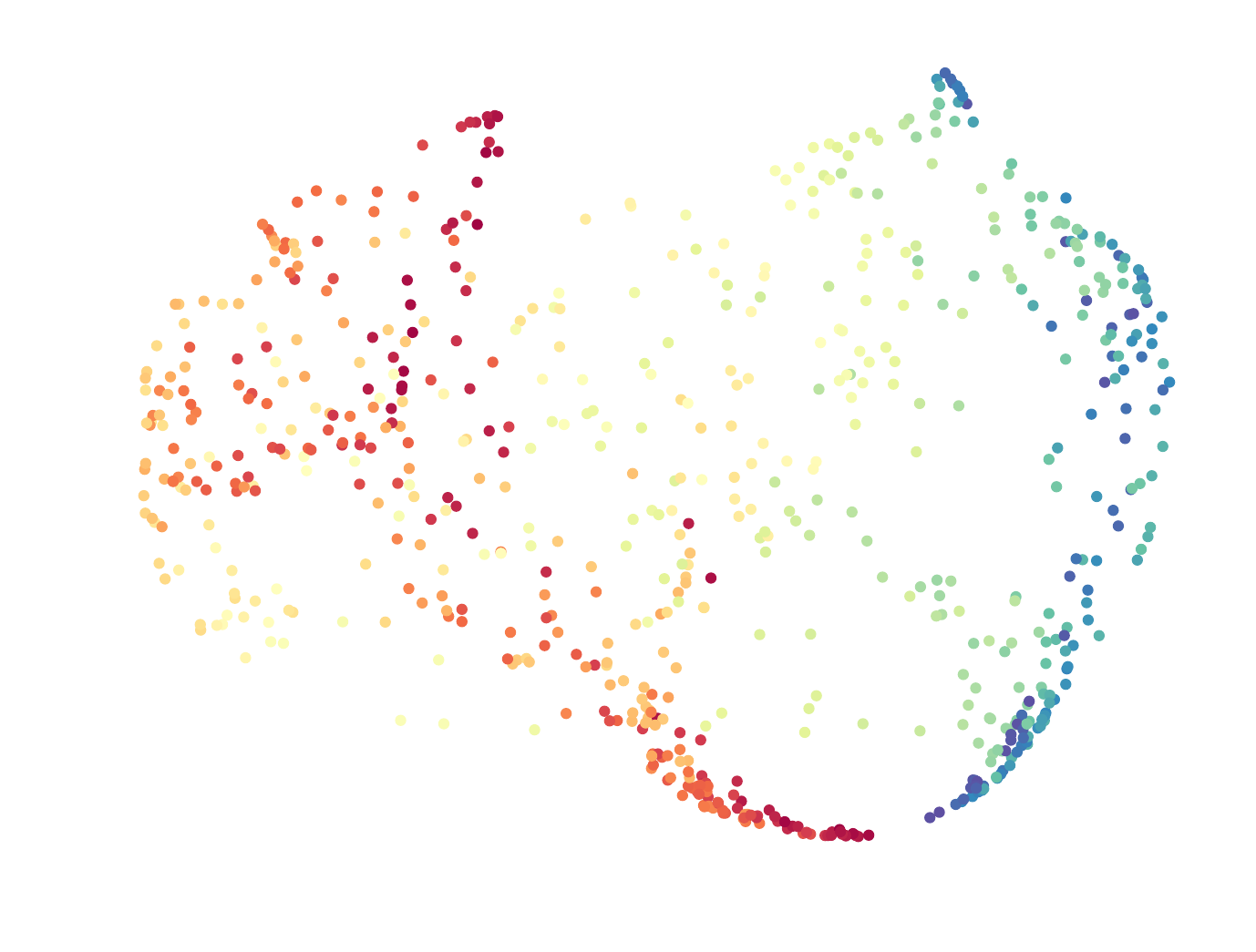}
        \vspace{-0.75cm}
        \subcaption*{HL-SNE}
        \end{minipage}
        \begin{minipage}{0.195\textwidth}
        \includegraphics[width=\linewidth]{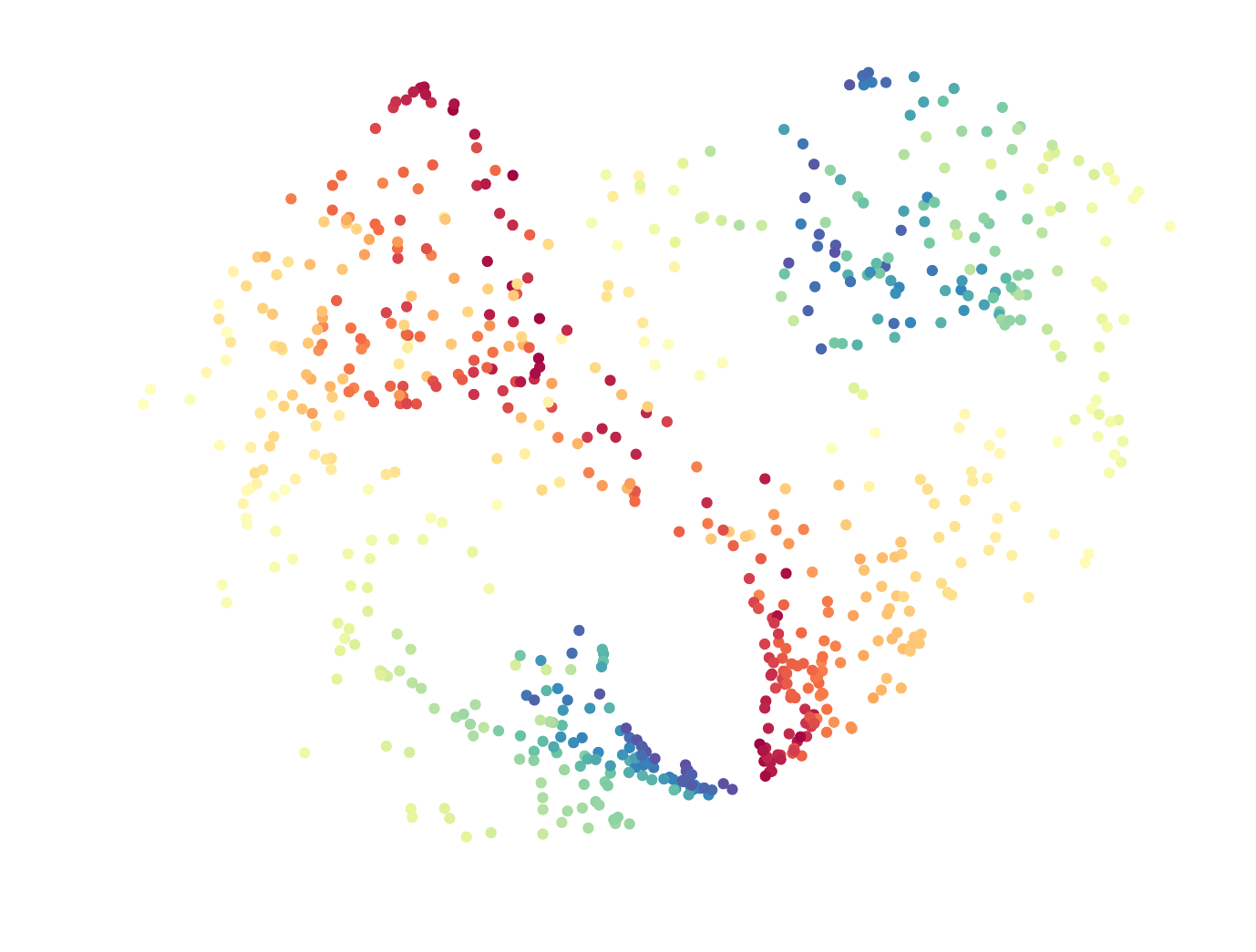}
        \vspace{-0.7cm}
        \subcaption*{CH-SNE}
        \end{minipage}
        \subcaption{Coloured based on pose 1}
    \end{minipage}
    \begin{minipage}{\textwidth} 
        \begin{minipage}{0.195\textwidth}
        \includegraphics[width=\linewidth]{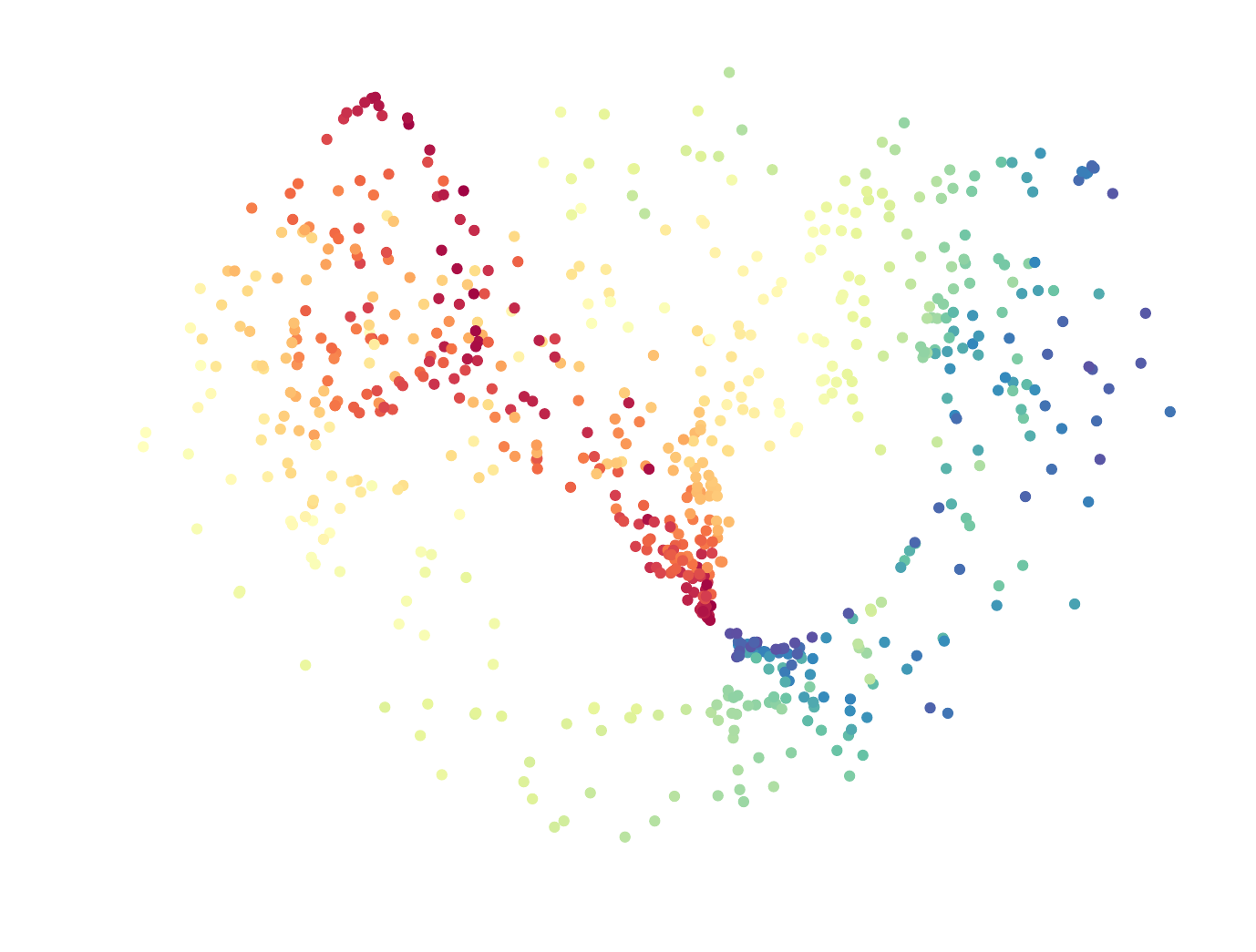}
        \vspace{-0.7cm}
        \subcaption*{KL-SNE}
        \end{minipage}
        \begin{minipage}{0.195\textwidth}
        \includegraphics[width=\linewidth]{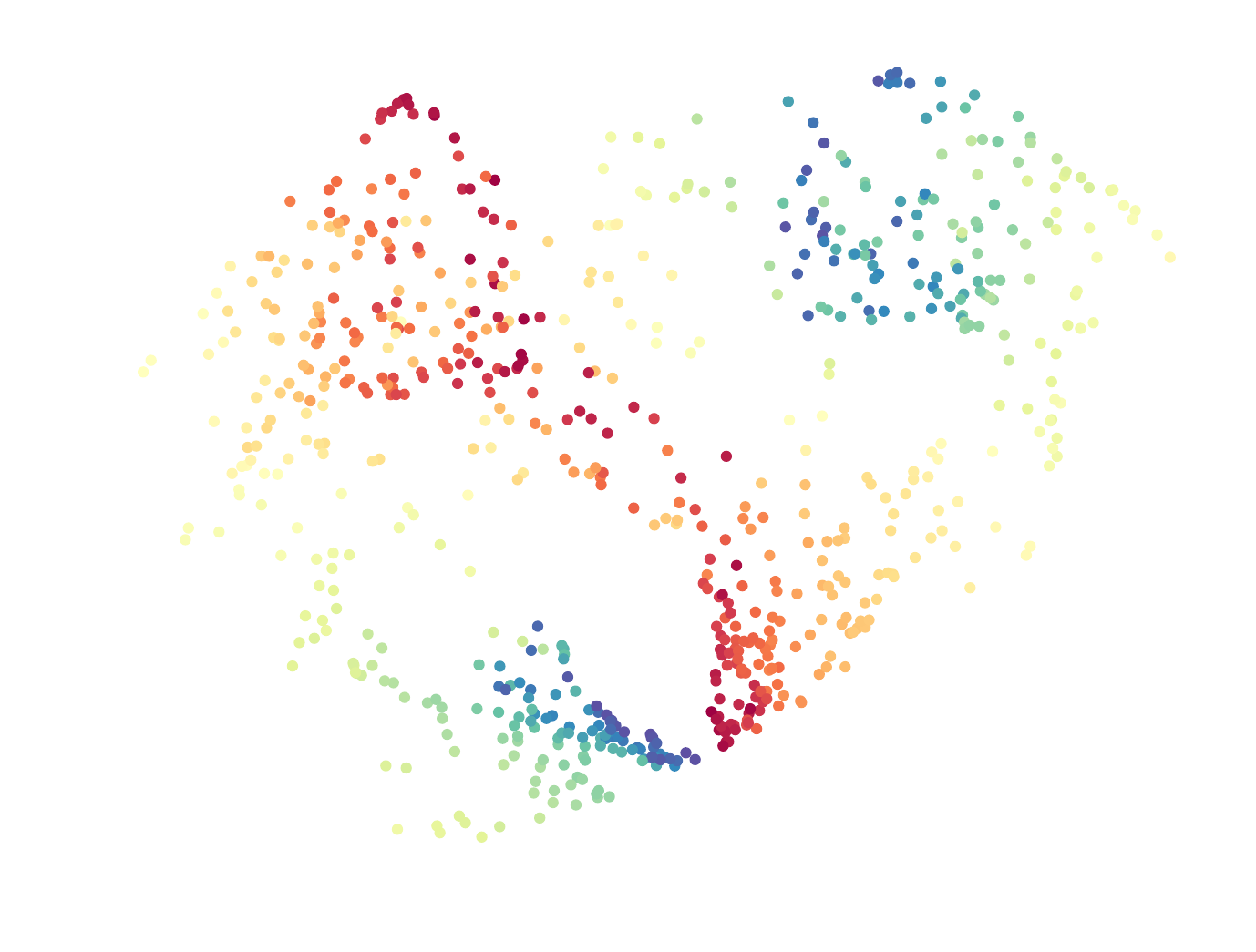}
        \vspace{-0.7cm}
        \subcaption*{RKL-SNE}
        \end{minipage}
        \begin{minipage}{0.195\textwidth}
        \includegraphics[width=\linewidth]{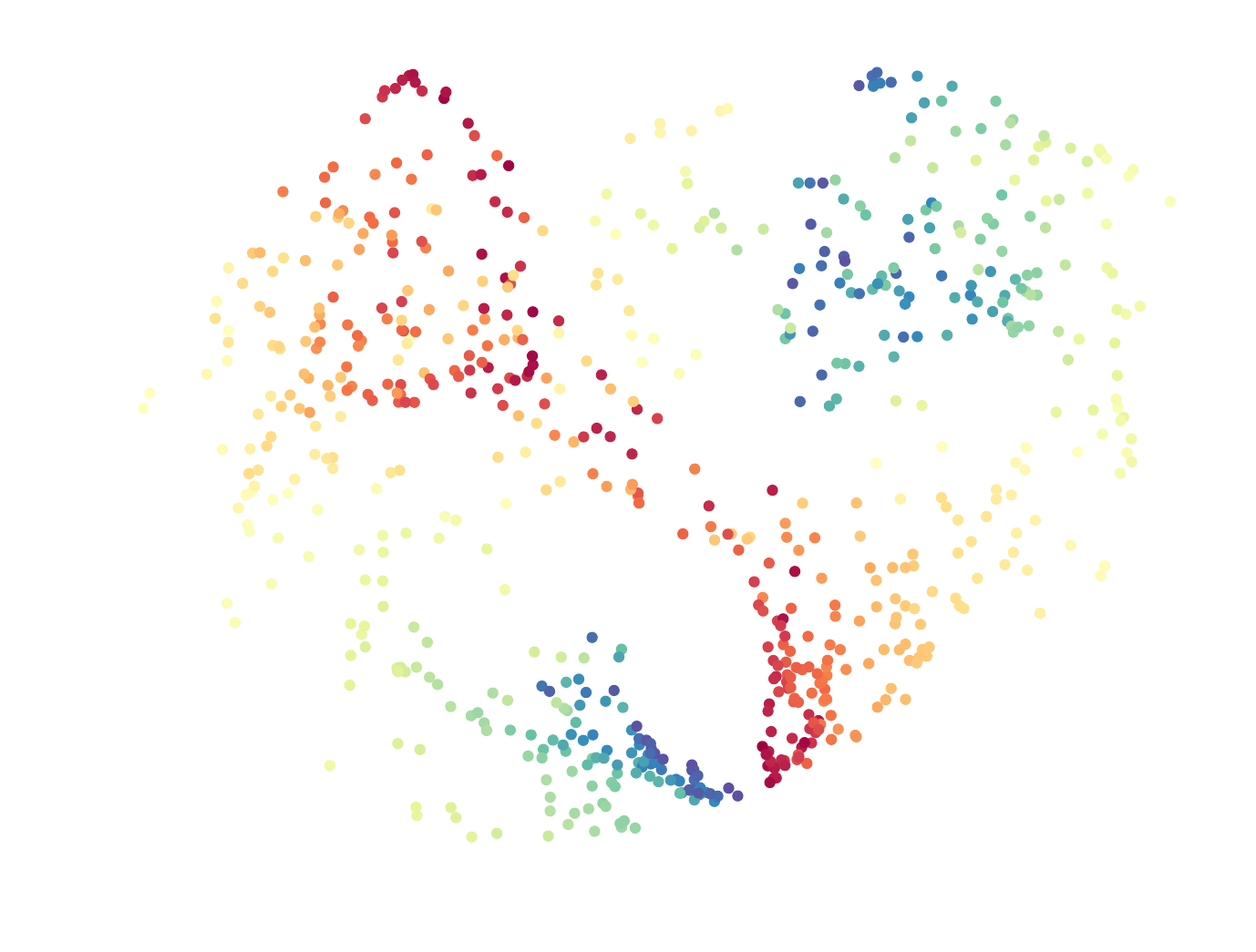}
        \vspace{-0.75cm}
        \subcaption*{JS-SNE}
        \end{minipage}
        \begin{minipage}{0.195\textwidth}
        \includegraphics[width=\linewidth]{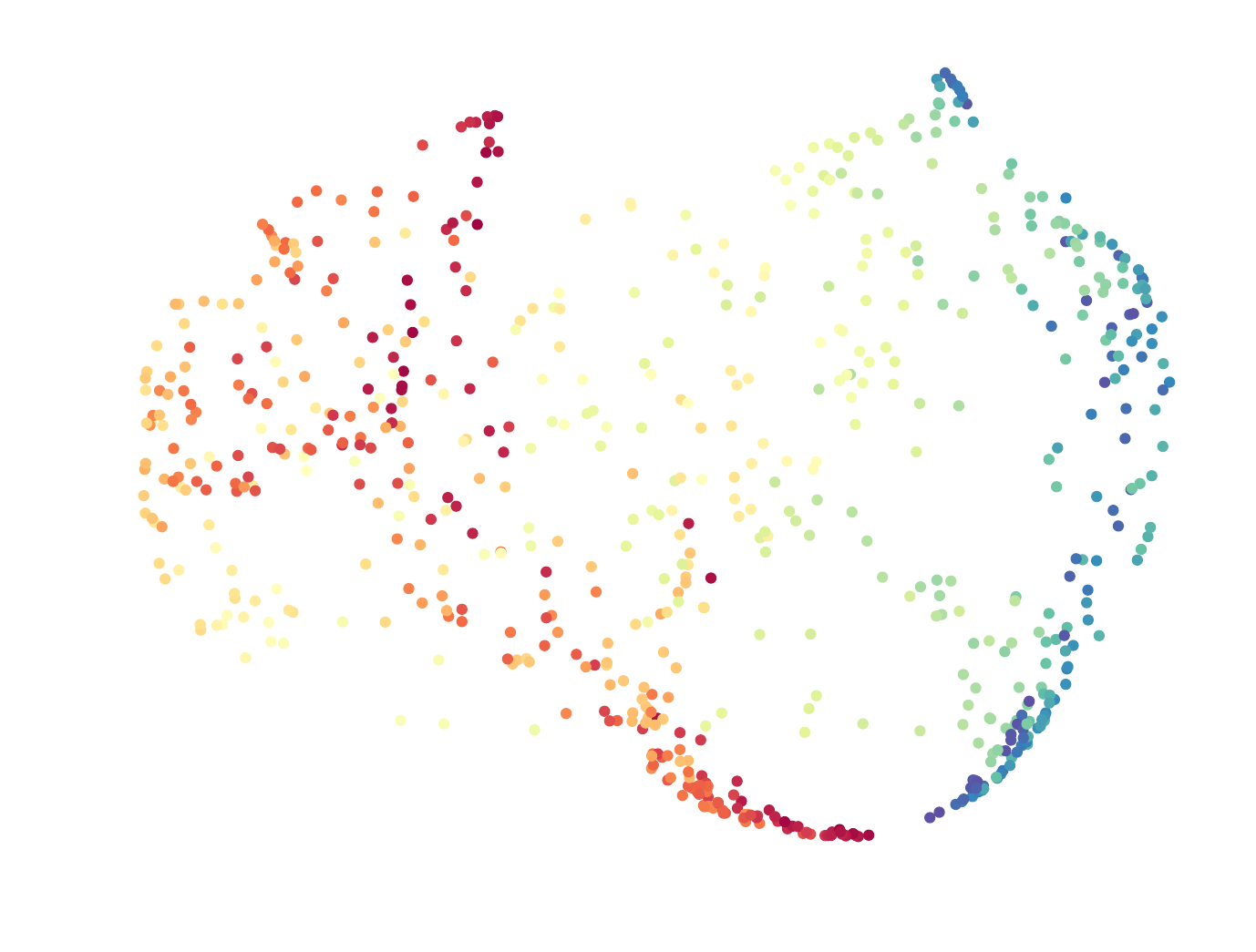}
        \vspace{-0.75cm}
        \subcaption*{HL-SNE}
        \end{minipage}
        \begin{minipage}{0.195\textwidth}
        \includegraphics[width=\linewidth]{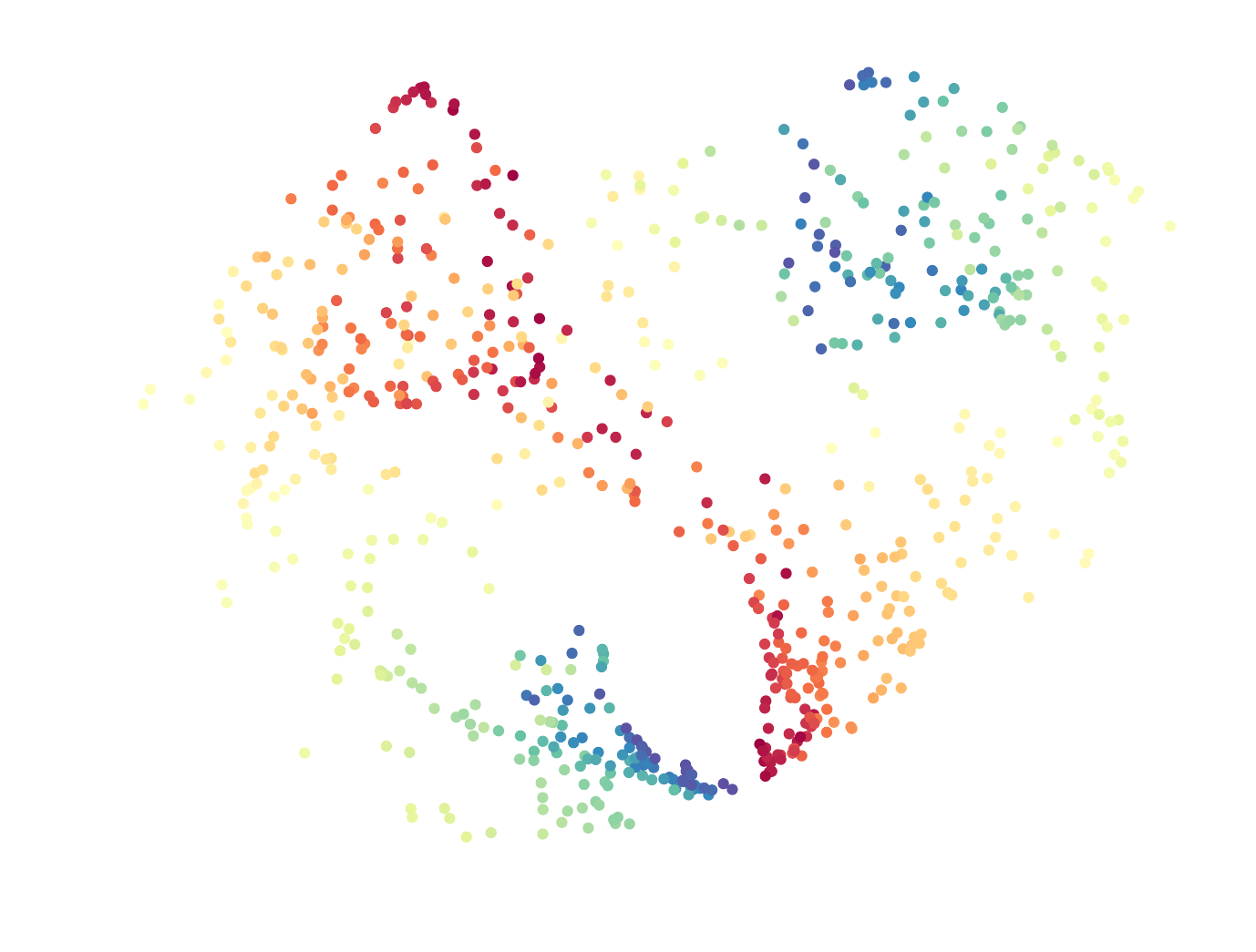}
        \vspace{-0.7cm}
        \subcaption*{CH-SNE}
        \end{minipage}
        \subcaption{Coloured based on pose 2}
    \end{minipage}
    \caption{Face Embeddings using $ft$-SNE. Perplexity=300}
    \label{fig:fSNE_face}
\end{figure}

\begin{figure}[htp]
    \captionsetup{justification=centering}
    \begin{minipage}{\textwidth}
 
        \begin{minipage}{0.195\textwidth}
        \includegraphics[width=\linewidth]{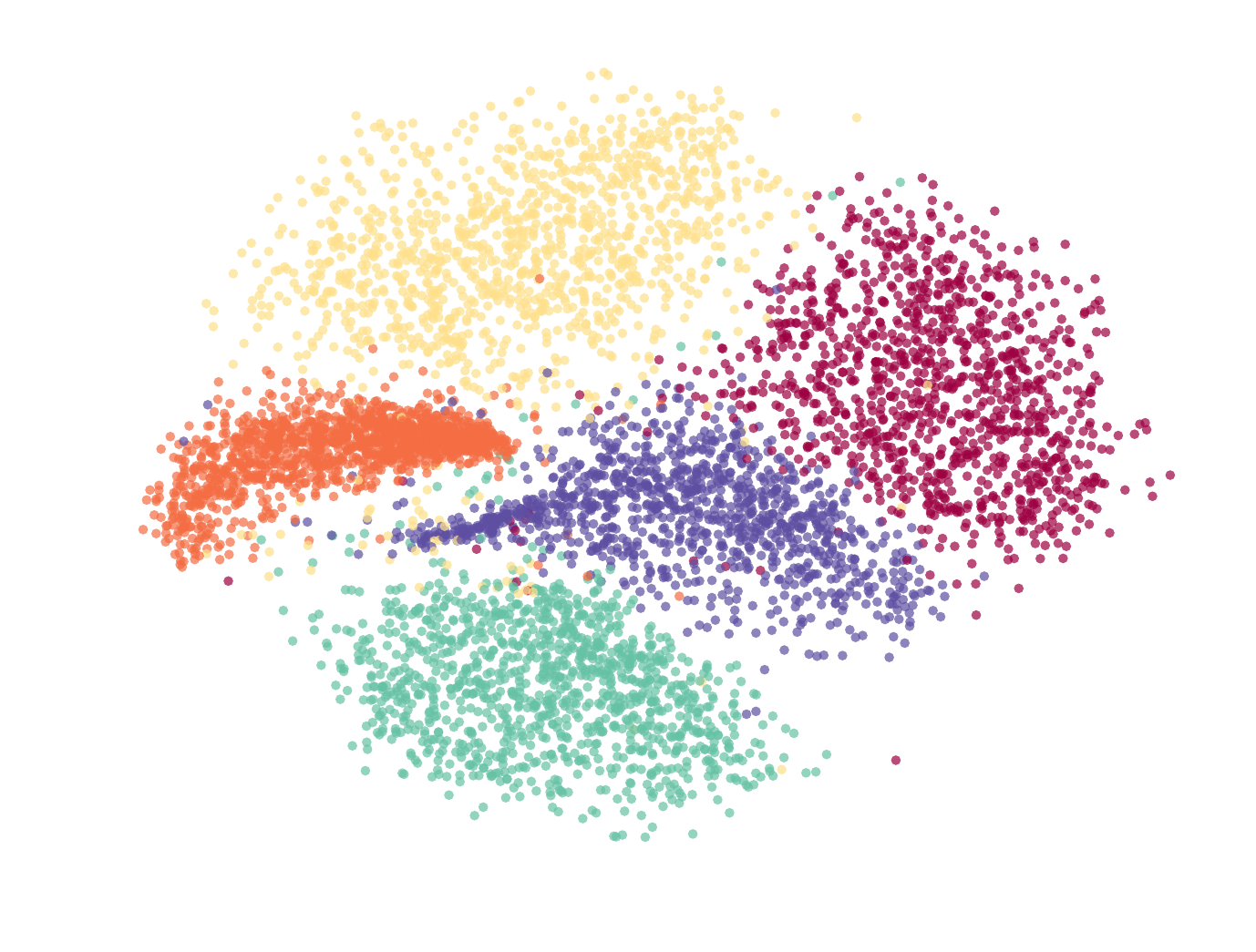}
        \vspace{-0.7cm}
        \subcaption*{KL-SNE}
        \end{minipage}
        \begin{minipage}{0.195\textwidth}
        \includegraphics[width=\linewidth]{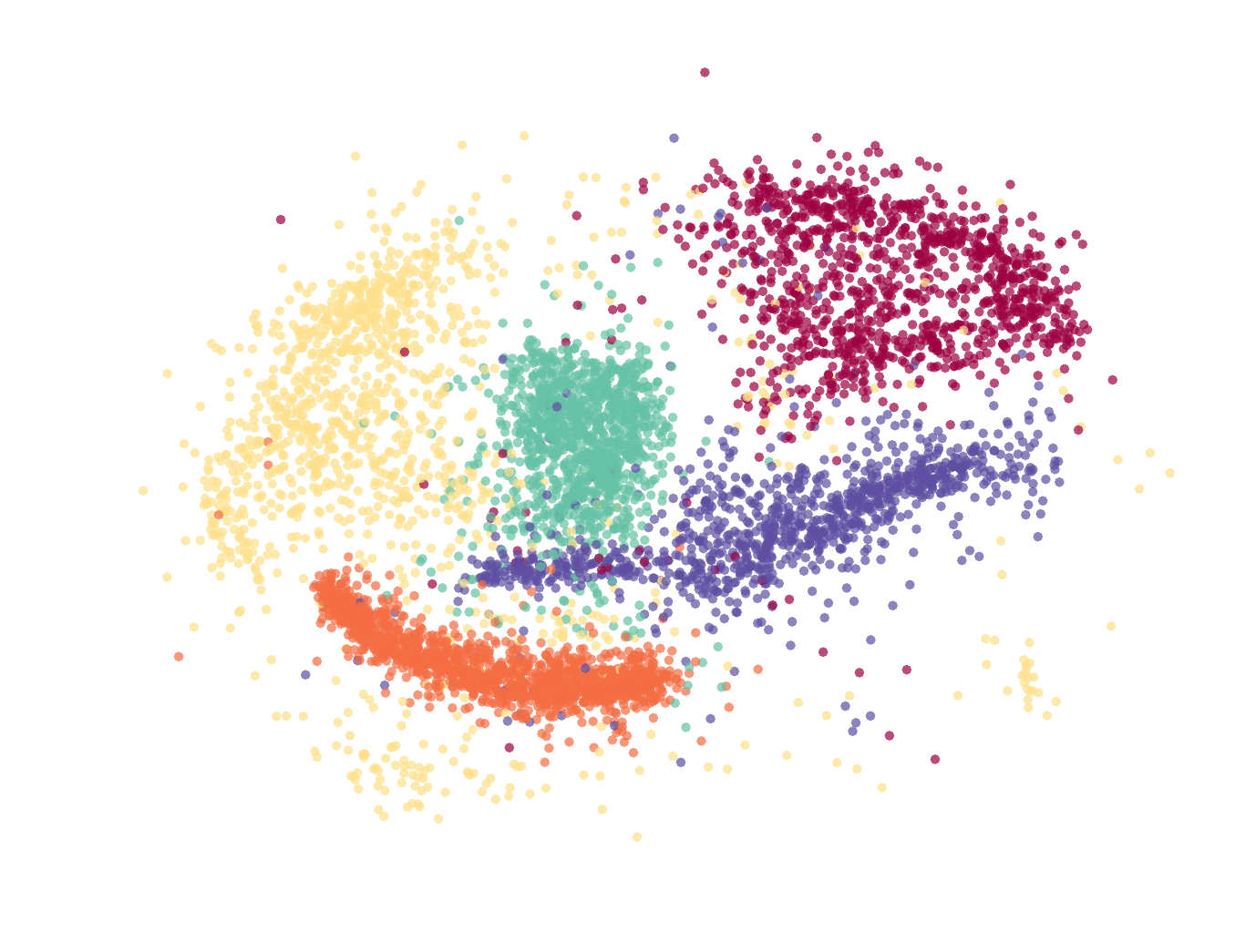}
        \vspace{-0.7cm}
        \subcaption*{RKL-SNE}
        \end{minipage}
        \begin{minipage}{0.195\textwidth}
        \includegraphics[width=\linewidth]{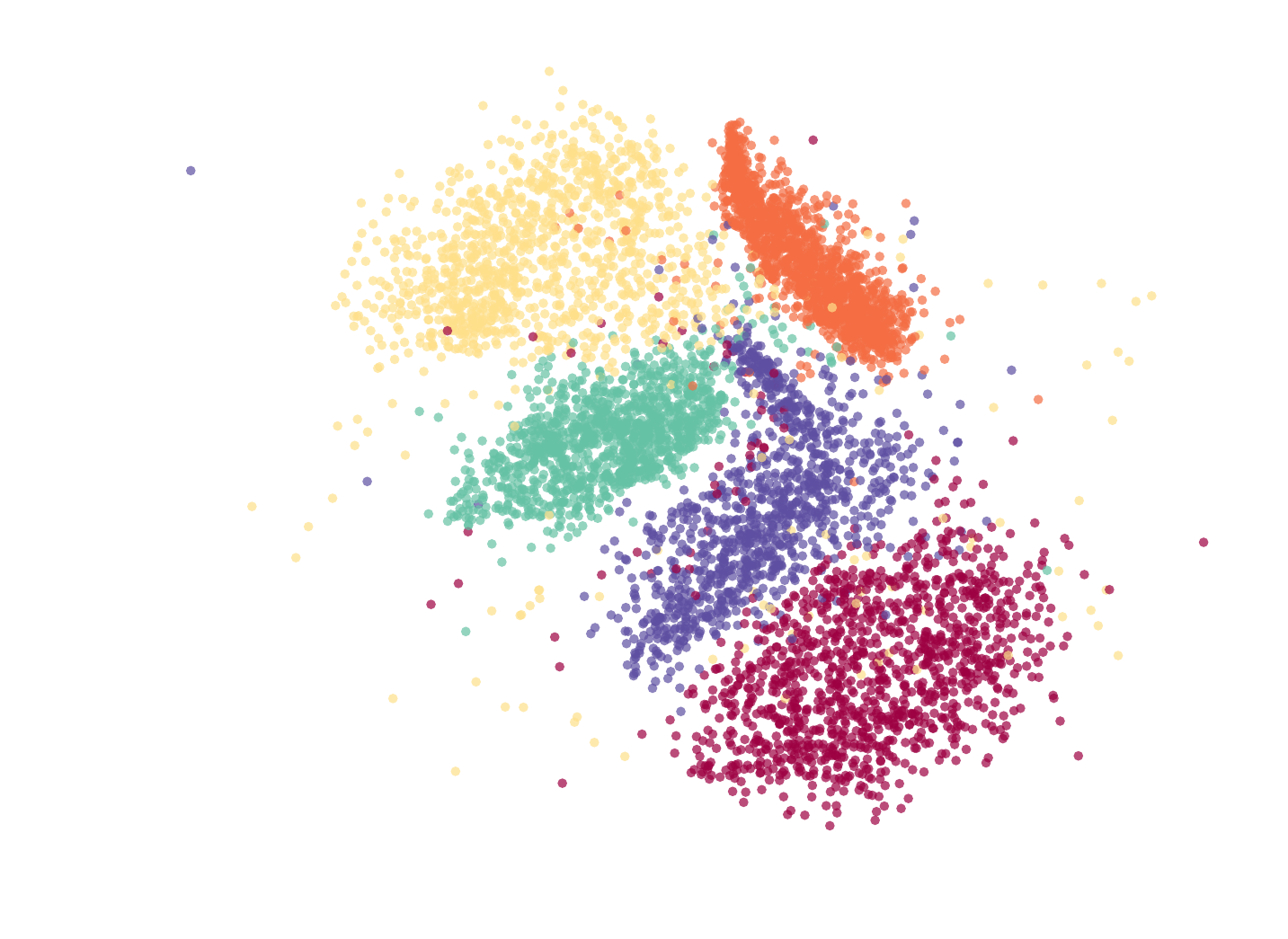}
        \vspace{-0.75cm}
        \subcaption*{JS-SNE}
        \end{minipage}
        \begin{minipage}{0.195\textwidth}
        \includegraphics[width=\linewidth]{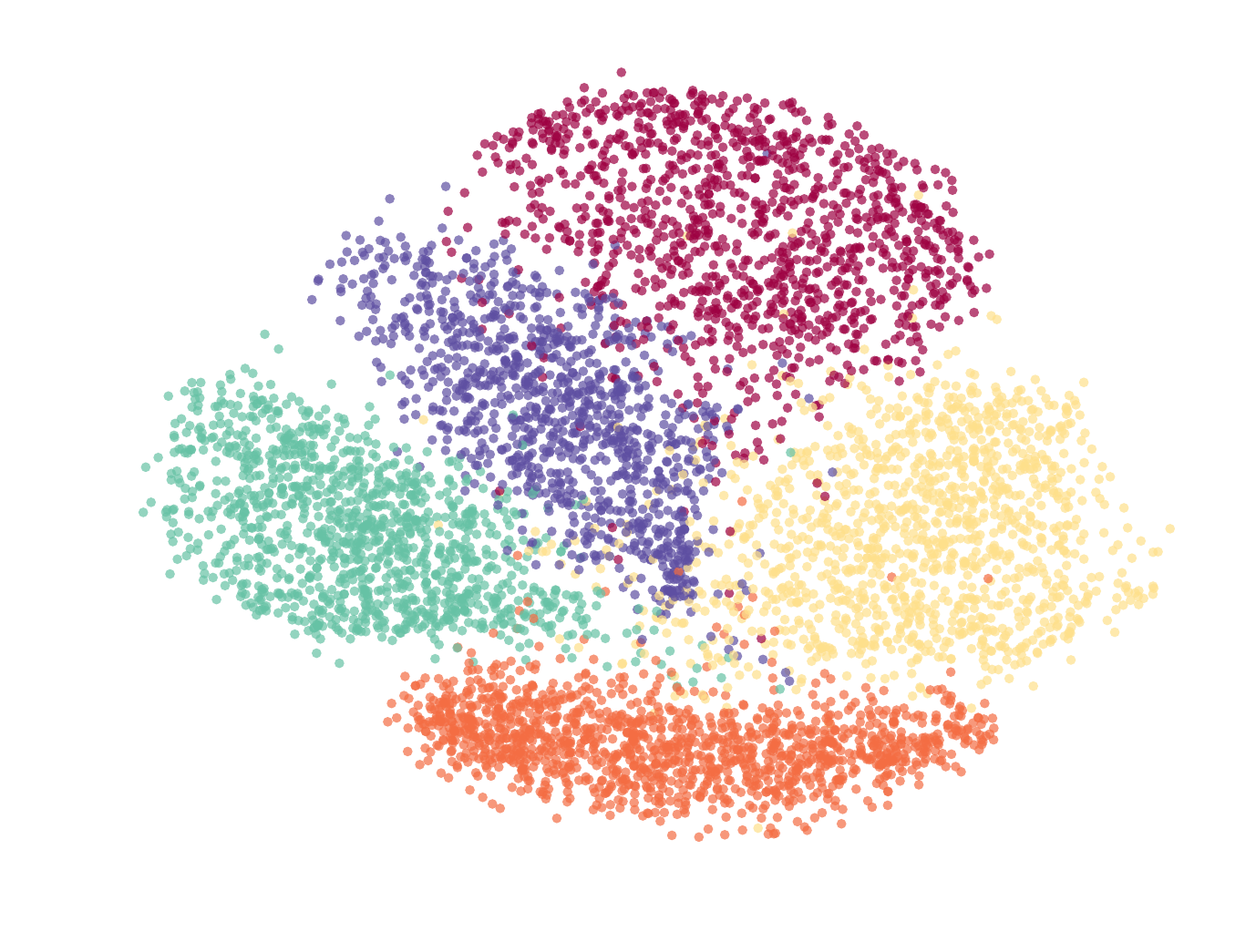}
        \vspace{-0.75cm}
        \subcaption*{HL-SNE}
        \end{minipage}
        \begin{minipage}{0.195\textwidth}
        \includegraphics[width=\linewidth]{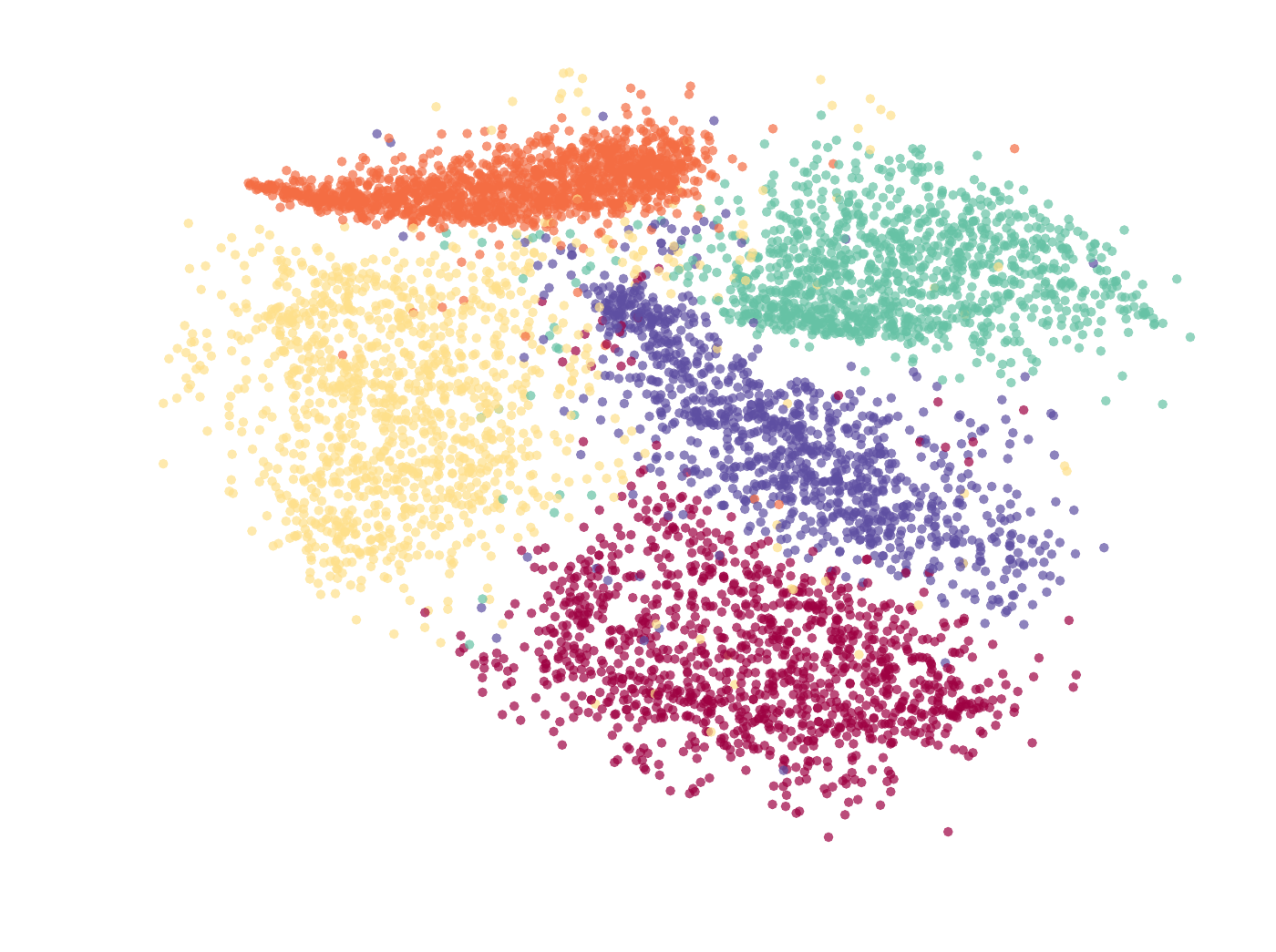}
        \vspace{-0.7cm}
        \subcaption*{CH-SNE}
        \end{minipage}
    \end{minipage}
    \caption{MNIST Embeddings using $f$-SNE. Perplexity: 2000 \& Step: 2000}
    \label{fig:fSNE_mnist}
\end{figure}

\end{appendices}

\end{document}